\documentclass[acmtog,authorversion]{acmart}
\usepackage{multirow}
\usepackage{graphicx}
\usepackage{subfigure}
\usepackage{amsmath}
\usepackage[ruled,vlined]{algorithm2e}
\usepackage{wrapfig}
\usepackage{makecell}
\AtBeginDocument{%
  \providecommand\BibTeX{{%
    \normalfont B\kern-0.5em{\scshape i\kern-0.25em b}\kern-0.8em\TeX}}}

\setcopyright{rightsretained}
\acmJournal{TOG}
\acmYear{2023}\acmVolume{42}\acmNumber{6}\acmArticle{245}\acmMonth{12} \acmDOI{10.1145/3618314}

\begin{document}

\title{Robust Zero Level-Set Extraction from Unsigned Distance Fields Based on Double Covering}


\author{Fei Hou}
\authornote{Both authors contributed equally to this research.}
\email{houfei@ios.ac.cn}
\orcid{0000-0001-8226-6635}
\author{Xuhui Chen}
\authornotemark[1]
\email{chenxh@ios.ac.cn}
\orcid{0009-0005-3112-2857}
\author{Wencheng Wang}
\authornote{Corresponding authors.}
\email{whn@ios.ac.cn}
\orcid{0000-0001-5094-4606}
\affiliation{
  \institution{State Key Laboratory of Computer Science, Institute of Software, Chinese Academy of Sciences \& University of Chinese Academy of Sciences}
  \city{Beijing}
  \country{China}
}
\author{Hong Qin}
\email{qin@cs.stonybrook.edu}
\orcid{0000-0001-7699-1355}
\affiliation{
\institution{Department of Computer Science, Stony Brook University}
\city{New York}
\country{USA}
}
\author{Ying He}
\authornotemark[2]
\email{yhe@ntu.edu.sg}
\orcid{0000-0002-6749-4485}
\affiliation{
\institution{School of Computer Science and Engineering, Nanyang Technological University}
\country{Singapore}
}


\begin{teaserfigure}
\centering
    \includegraphics[width=7.0in]{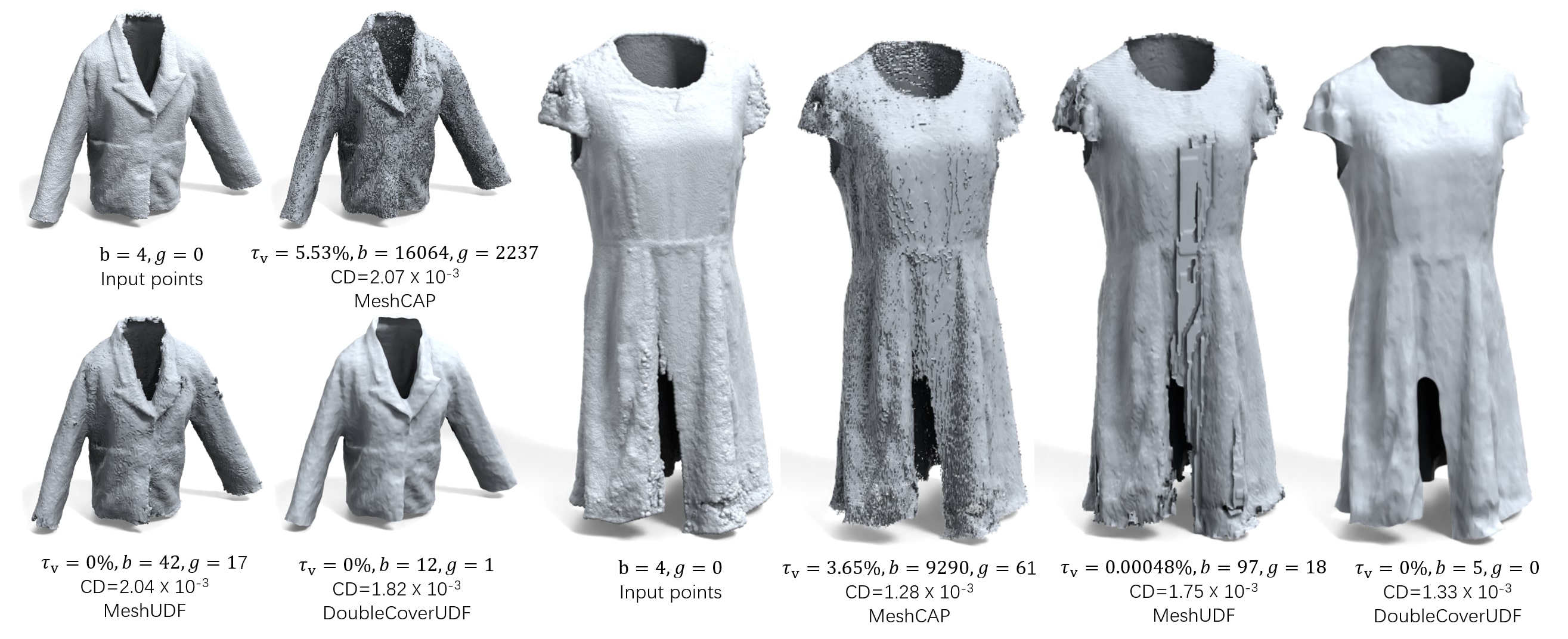}
\caption{DoubleCoverUDF is a robust method for extracting the zero level-set from unsigned distance fields, enabling the representation of both open and closed 3D surfaces. Comparison with MeshUDF~\cite{Guillard2022} and MeshCAP~\cite{Zhou2022} on the open models from the Deep Fashion3D dataset~\cite{Zhu2020} shows that the extracted meshes using DoubleCoverUDF exhibit several favorable characteristics, such as the absence of non-manifold vertices, visually appealing results, and the ability to accurately capture the correct topological features of the target surfaces. Here, $\tau_v$, $b$, and $g$ denote the percentage of non-manifold vertices, the number of boundaries, and the genus of the meshes, respectively. Additionally, we compute the Chamfer distance (CD) between the extracted meshes and the given point clouds to assess their geometric accuracy. A marching cubes resolution of $256^3$ is used in these examples.}
\label{fig:teaser}
\end{teaserfigure}

%
\begin{abstract}
In this paper, we propose a new method, called DoubleCoverUDF, for extracting the zero level-set from unsigned distance fields (UDFs). DoubleCoverUDF takes a learned UDF and a user-specified parameter $r$ (a small positive real number) as input and extracts an iso-surface with an iso-value $r$ using the conventional marching cubes algorithm. We show that the computed iso-surface is the boundary of the $r$-offset volume of the target zero level-set $S$, which is an orientable manifold, regardless of the topology of $S$. Next, the algorithm computes a covering map to project the boundary mesh onto $S$, preserving the mesh's topology and avoiding folding. If $S$ is an orientable manifold surface, our algorithm separates the double-layered mesh into a single layer using a robust minimum-cut post-processing step. Otherwise, it keeps the double-layered mesh as the output. We validate our algorithm by reconstructing 3D surfaces of open models and demonstrate its efficacy and effectiveness on synthetic models and benchmark datasets. Our experimental results confirm that our method is robust and produces meshes with better quality in terms of both visual evaluation and quantitative measures than existing UDF-based methods. The source code is available at \url{https://github.com/jjjkkyz/DCUDF}.

\end{abstract}



\maketitle

\section{Introduction and Motivation}

Marching cubes, originally developed by Lorensen and Cline~\shortcite{Lorensen1987}, is a widely-used algorithm for extracting a polygonal mesh of the zero level-set of a signed distance field (SDF). 
By examining the distance values at the cube vertices, which exhibit opposite signs, the algorithm identifies the cubes that contain a piece of the iso-surface. Triangular faces are generated by looking up in a pre-defined table based on the signs of cube vertices. The marching cubes algorithm is conceptually simple, easy to implement, and highly efficient. Additionally, it guarantees to extract watertight iso-surfaces from SDFs~\cite{Lorensen1987,Ju2002,Chen2021NMC}.

While SDFs are a popular implicit representation of watertight 3D surfaces, they are unable to model open surfaces, non-manifold shapes, or non-orientable surfaces. Some recent works~\cite{Meng2023,Wang2022HSDF,Chen2022} have attempted to learn SDFs to model open surfaces, but they require additional resources to generate open boundaries. On the other hand, unsigned distance fields (UDFs) are capable of representing surfaces of any topology and are becoming an important complement to SDFs. There has been a growing interest in learning UDFs from either point clouds or multiview images, which presents new opportunities~\cite{Zhou2022,Chibane2020NDF,Zhao2021,Liu2023,Long2023}. However, extracting the zero level-set from a UDF is challenging since zero is not a regular value of the distance value. 
In addition, the absence of cubes with vertices having opposite signs based on distance values in UDFs makes it challenging to apply the standard marching cubes algorithm directly to UDFs. Despite the potential of UDFs to become a more powerful and flexible representation of general 3D surfaces, the lack of off-the-shelf toolkits for extracting the zero-crossing surfaces from UDFs severely limits their applicability. This limitation hinders the ability to leverage UDFs in practical applications, e.g., computer graphics and 3D vision.

There have been several research efforts aimed at overcoming the aforementioned challenges of extracting the zero level-set from a UDF. One approach is to extract the iso-surface of a small positive iso-value, as attempted by some works~\cite{Venkatesh2021,Corona2021}, but this method can result in an inflation of the original surface and produce inaccurate results. Another approach is to adapt marching cubes for use with UDFs. Guillard et al.~\shortcite{Guillard2022} and Zhou et al.~\shortcite{Zhou2022} used the gradient directions of a UDF instead of distance signs to determine intersections on cube edges. 
Specifically, the intersection with the zero level-set on an edge exists if the gradient directions on both ends of the edge are opposite. However, the gradient-based methods suffer from a lack of robustness due to their high sensitivity to the accuracy of UDFs. Small errors in distance values may completely change the gradient direction, leading to topological errors in the extracted meshes.
Moreover, these methods often give rise to many intersection configurations that do not exist in the original marching cubes algorithm, and result in a significant number of undesired non-manifold vertices, boundaries, and handles. 

In this paper, we present a new algorithm for robustly extracting the zero level-set from a learned UDF and utilizing it to reconstruct both open and closed surfaces.
Our algorithm takes as input a learned UDF, e.g., encoded in a multilayer perceptron (MLP), and a user-specified parameter $r$ (a small positive real number). Initially, it utilizes the conventional marching cubes algorithm to extract a triangular mesh for the iso-surface with an iso-value $r$. As a ``dilated'' double cover of the target zero level-set $S$, this iso-surface is an orientable closed 2-manifold, irrespective of the topology of $S$. Subsequently, our algorithm computes a covering map to project the double covering onto $S$, preserving the mesh's topology and avoiding issues such as folding and self-intersection. By leveraging the differentiability of the learned UDF, we formulate the projection as an optimization problem and efficiently solve it using the vector Adam solver~\cite{Ling2022}. In cases where the target $S$ is orientable manifold, our algorithm incorporates a post-processing step to separate the double-layered mesh into a single layer. However, for non-manifold or non-orientable surfaces, the algorithm directly outputs the double-layered mesh, which is of different topology to the target surface, without any further modifications. 

We refer to our method as DoubleCoverUDF or DCUDF for brevity. Through extensive validation on synthetic models and benchmark datasets, we demonstrate the robustness and effectiveness of DCUDF. The resulting meshes extracted by DCUDF exhibit visual appeal, characterized by the absence of non-manifold vertices, thereby ensuring topological cleanliness. Moreover, quantitative evaluation highlights the superior robustness and capabilities of our method compared to other UDF-centric approaches.

We make the following contributions in the paper:
\begin{enumerate}
    \item We propose a novel and robust method for extracting the zero level-set from learned UDFs, enabling the handling of both open and closed surfaces.
    \item We demonstrate the effectiveness of our method in generating high-quality triangulated meshes of the zero level-set, characterized by small geometric errors and the absence of non-manifold structures.
    \item We conduct a comprehensive evaluation of our method on various test cases, comparing it with state-of-the-art approaches for extracting the zero level-set from UDFs. The results highlight the superior accuracy and robustness of our method.    
\end{enumerate}

\section{Related Work}

\subsection{Surface Reconstruction}
Surface reconstruction is an important area of study in computer graphics that aims to create a 3D surface from a point cloud. There are two main classes of surface reconstruction algorithms: combinatorial and implicit methods. Combinatorial techniques~\cite{Bernardini1999,Amenta2001,Dey2003} are efficient and can produce both watertight and open surfaces. However, they can result in self-intersecting or discontinuous triangular faces given points with noise and outliers.

Implicit methods, on the other hand, compute a distance field and extract the zero level-set, e.g., by the standard marching cubes~\cite{Lorensen1987} algorithm, which guarantees the generation of watertight models. These methods are more robust than combinatorial methods, as they are less sensitive to noise and outliers. 
Many of these methods require oriented normals as input, e.g., MPU~\cite{Ohtake2003}, Poisson surface reconstruction (PSR)~\cite{Kazhdan2006,Kazhdan2013} and Gaussian formula based reconstruction~\cite{Lu2018}.
These methods are computationally efficient, but only generate watertight models.
A number of normal orientation methods~\cite{Hoppe1992,Alliez2007,Huang2009,Huang2019,Metzer2021,Peng2021,Hou2022,Lin2023,Xu2023} are proposed for surface reconstruction.
Peng et al.~\shortcite{Peng2021} developed a differentiable Poisson solver that can globally update point positions and normals for PSR.
Hou et al.~\shortcite{Hou2022} proposed iPSR that updates the normals iteratively taking advantage of the PSR models.
Xu et al.~\shortcite{Xu2023} developed a global approach using regularized winding-number field for consistent point orientation. However, these methods also only work for watertight models.

\subsection{Extracting Zero Level-Sets from UDFs}
Although UDFs are ideal for reconstructing open models, extracting the zero level-set from a given UDF is a challenging task. Early approaches~\cite{Koo2005,Wang2005} tackled this problem via a two-step framework, where they first constructed a set of voxels containing the zero level-set, and then shrunk the boundary surface of these voxels to obtain the target surface. However, to produce a clean output mesh, these methods required the initial mesh to be a manifold, which cannot be guaranteed in practice.

Due to the increasing interest in the representation capability of UDFs in 3D deep learning, there have been recent developments aimed at enhancing the use of UDFs in this field. Guillard et al.~\shortcite{Guillard2022} proposed a modified version of the  standard marching cubes algorithm for extracting the zero level-set using the gradient direction of UDFs. To address the vulnerability of gradient directions to UDF noise, they introduced a novel concept called pseudo-signed distance and proposed a voting strategy for computing the sign. They also applied a post-processing step to smooth the extracted meshes and remove small artifacts. Their method is computationally efficient and can produce smooth non-watertight surfaces. However, their method may produce small topological artifacts such as non-manifold vertices, boundaries, and handles. 
Zhou et al.~\shortcite{Zhou2022} also adopted a modified marching cubes strategy for the zero-level set extraction from UDFs based on gradient directions, but it tends to generate a large amount of non-manifold vertices.

Recently, Chen et al.~\shortcite{Chen2022NDC} proposed neural dual contouring, which is a unified approach for mesh reconstruction from various input sources, including SDFs, UDFs, binary voxels, unoriented points and noisy raw scans. Their method is data driven, offering high reconstruction accuracy, sharp feature preservation, and excellent runtime performance. When trained with appropriate data, their method can generate excellent results. However, their approach has a relatively high GPU memory consumption and often produces meshes with a large number of non-manifold edges.

\subsection{Neural Distance Fields}
There are many methods for generating continuous and differentiable distance fields. For instance, DeepSDF~\cite{Park2019} uses an MLP trained on a dataset to learn latent codes that represent the SDF of models with SDF ground truth. On the other hand, IGR~\cite{Gropp2020} learns the SDF from surface points input only. To reconstruct more details, some methods~\cite{Chabra2020,Jiang2020} divide the space and use multiple local MLPs to represent the entire model. IDF~\cite{Wang2022IDF} employs two hierarchical SIRENs~ \cite{Sitzmann2020} to generate a smooth base surface and high-frequency details.

In contrast, only a few methods have been proposed to model open surfaces. Chibane et al.'s method~\shortcite{Chibane2020NDF} learns a UDF by IF-Net~\cite{Chibane2020IFNET} and projects sample points to the zero iso-surface using the UDF gradient, resulting in a dense set of surface points. However, this projection is sensitive to noise in the UDF and mesh generation from dense points is often unstable, leading to artifacts in the reconstructed surface. CAPUDF~\cite{Zhou2022} uses an MLP to fit the UDF, which is simple and efficient. However, since it infers the UDF from input points directly without a discrete UDF as input, the learned UDFs are usually inaccurate for models with complex geometry. GeoUDF~\cite{Ren2023} learns a quadratic polynomial for each input point, and formulates the unsigned distance of a query point as the learnable affine averaging of its distances to the tangent planes of neighboring points on the surface. While the aforementioned works primarily concentrate on learning UDFs from point clouds, our research aims at targeting the extraction of zero level-sets from pre-existing, learned UDFs. NeAT~\cite{Meng2023} learns SDFs from multi-view images, subsequently utilizing the standard MC algorithm to extract zero level-sets. This is followed by a postprocessing step to cut the surface open. DeepCurrent~\cite{Palmer2022} adopts a hybrid representation for modeling surface boundaries explicitly. Instead of UDF, these works generate open models with explicit masks, which are different from our target. This sets our work apart from current studies in the field.

To address the challenges in extracting surfaces from learned UDFs, various new representations have been proposed instead of UDFs.
For example, Chen et al.~\shortcite{Chen2022} proposed a three-pole signed distance function to generate open surfaces.
Wang et al.~\shortcite{Wang2022HSDF} proposed a hybrid sign and distance function to model open surfaces.
Ye et al.~\shortcite{Ye2022} proposed a general implicit function for 3D shapes to encode the presence of a surface between two points. Although these methods can use marching cubes or its variant to extract the surface, their representations are more complicated and difficult to learn than UDF.
Given that DCUDF serves a general-purpose tool for extracting zero level-sets from UDFs, it can readily accept UDFs learned through methods such as~\cite{Chibane2020NDF,Zhou2022} as input.

\begin{figure}[!htbp]
    \centering
    \includegraphics[width=3.3in]{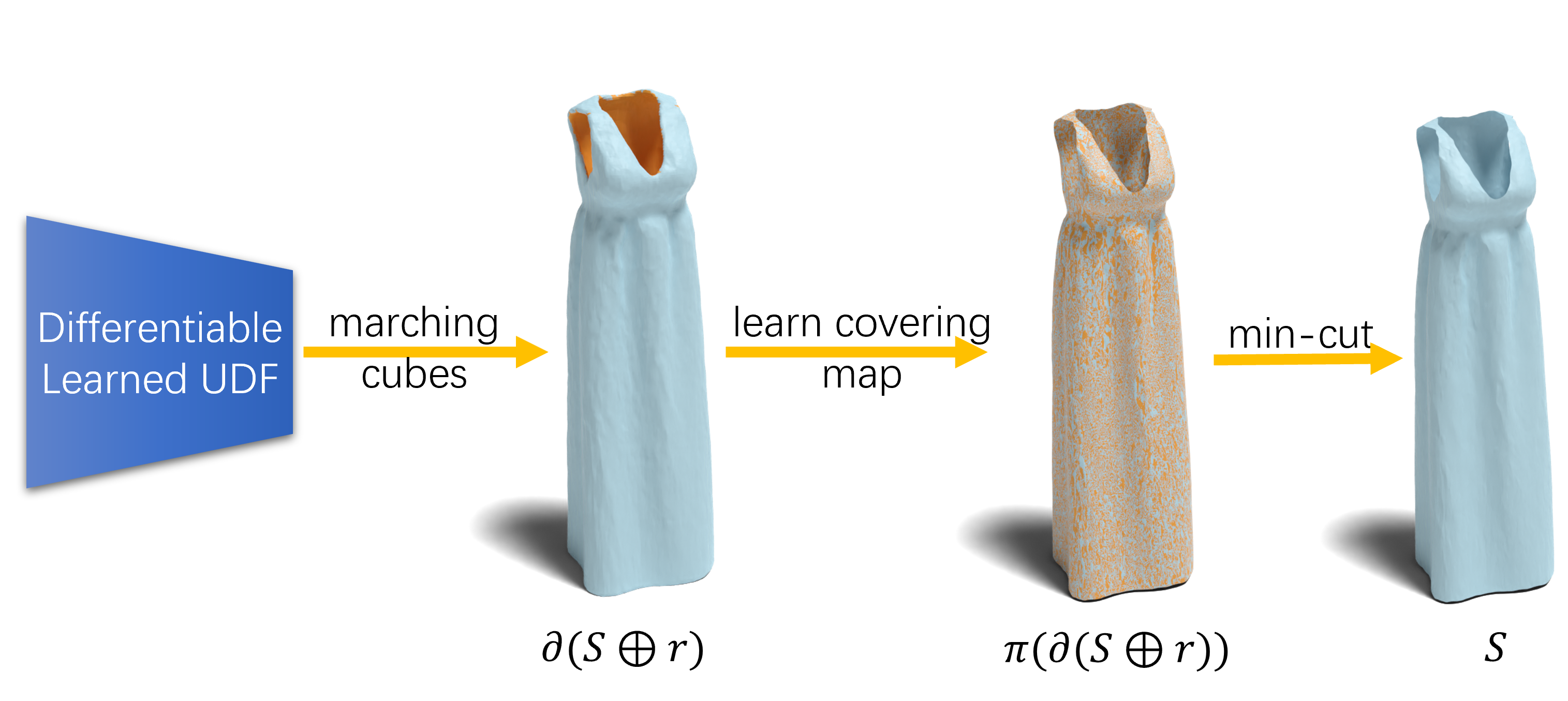}
    \caption{Algorithmic pipeline. Given a learned UDF of the target surface $S$ as input, the algorithm first extracts an iso-surface with iso-value $r$ using the marching cubes algorithm. 
    The iso-surface, denoted by $\partial(S\oplus r)$, is a closed and orientable 2-manifold and represents the boundary of a $r$-offset volume. Next, the algorithm computes a covering map $\pi:\partial(S\oplus r)\rightarrow S$ that projects the double covering of the iso-surface back to $S$, yielding a mesh with the same topology as $\partial(S\oplus r)$ and with two identical layers overlapping each other. If $S$ is an orientable manifold, the algorithm applies a min-cut algorithm to separate the two layers into a single-layer mesh. Otherwise, it retains the double-layered mesh as the output. For visualization purposes, the two layers are distinguished by different colors. }
    \label{fig:pipeline}
\end{figure}

\section{Extracting Zero Level-Set from UDFs}
\label{sec:method}
DoubleCoverUDF extracts the zero level-set from a learned UDF. It makes use of the double covering technique described in Section~\ref{sec:covering} and computes a covering map to project the double covering to the target zero level-set, which is documented in Section~\ref{sec:mapping}. Additionally, our method includes a post-processing step to separate the double-layered mesh into a single layer for orientable and manifold surfaces, as explained in Section~\ref{sec:sep}. We illustrate the algorithmic pipeline in Figure~\ref{fig:pipeline} and show the pseudo code for the method in Algorithm~\ref{alg:DoubleCoverUDF}.

\begin{algorithm}
    \SetAlgoLined
    \SetKwInOut{Input}{input}
    \SetKwInOut{Output}{output}
    \SetKwRepeat{Do}{do}{while}
    \Input{A UDF represented by an MLP $f$ and the offset parameter $r>0$}
    \Output{A triangular mesh representing the zero-crossing iso-surface of the UDF}
    Extract the iso-surface of iso-value $r$ by marching cubes\;
    Compute the covering map by minimizing~(\ref{eqn:step1}) and (\ref{eqn:step2})\;
    \If {the target surface is non-orientable or non-manifold}
    {
    \Return{the projected double-layered mesh $M$}\;
    }
    \ElseIf {the target surface is closed}
    {
    \tcc{the two layers are already separated}
    \Return{the inner or outer layer with more faces}\; 
    }
    \Else{
       \Do{the face count difference $||M_1|-|M_2||<0.15|M|$ }{
        Find a pair of source region and sink region\;
        Compute the $s\text{-}t$ cut\;
        Split $M$ to $M_1$ and $M_2$\;
    }
    \Return {the either $M_1$ or $M_2$ with more faces}\;
    }
    \caption{DoubleCoverUDF}
    \label{alg:DoubleCoverUDF}
\end{algorithm}

\subsection{$r$-Offset Volume and Double Covering}
\label{sec:covering}

Double covering bounds an original surface by attaching two identical copies of it. It has been used in the graphics community for computing global surface parameterization~\cite{Gu2003,Kalberer2007,Nieser2012} and discrete Laplacian for non-manifold meshes~\cite{Sharp2020}. In our method, we apply double covering to get the initial mesh.

\begin{figure}[!htbp]
    \centering
    \includegraphics[height=0.75in]{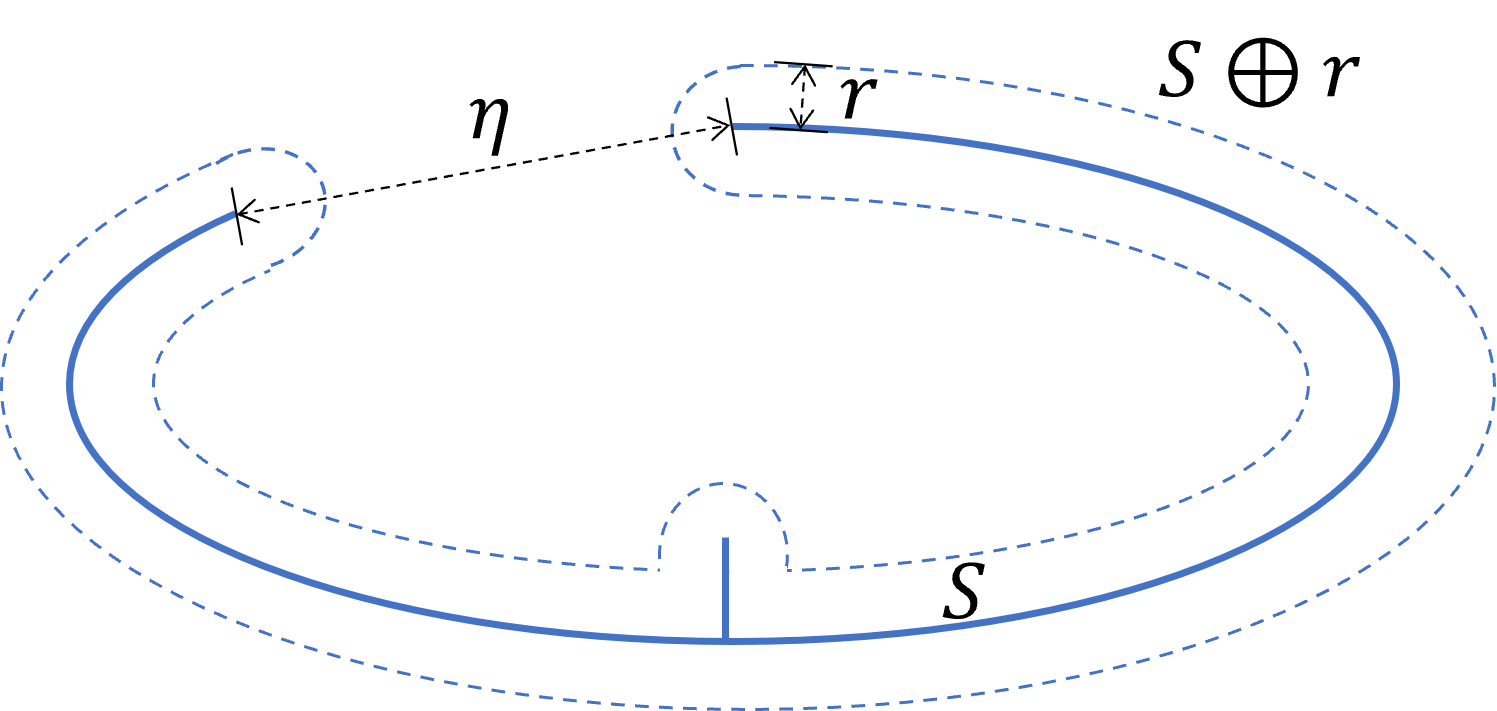}
    \includegraphics[height=0.75in]{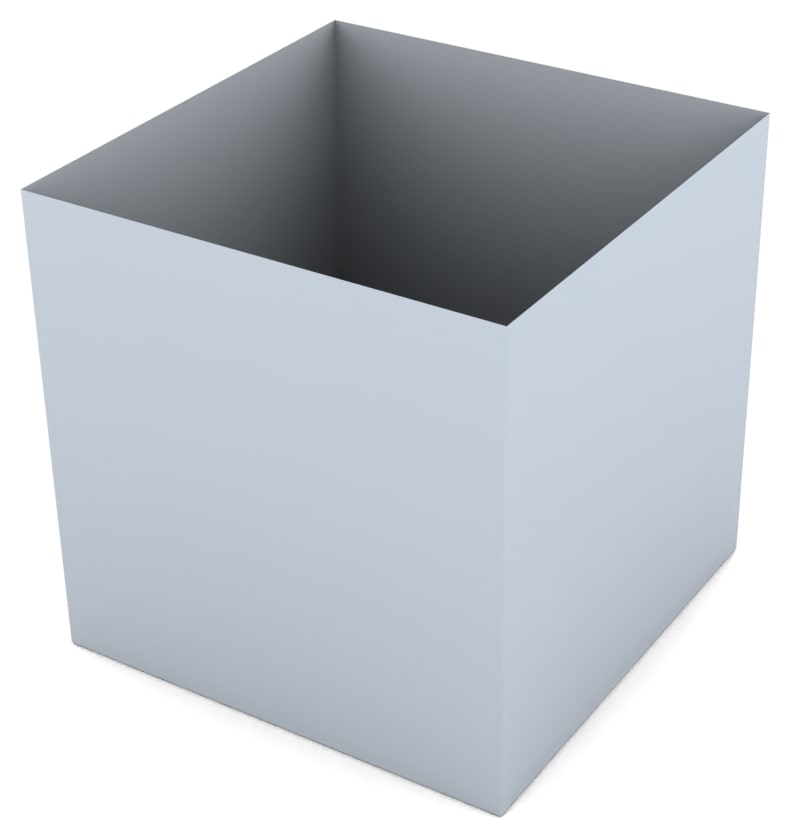}
    \includegraphics[height=0.75in]{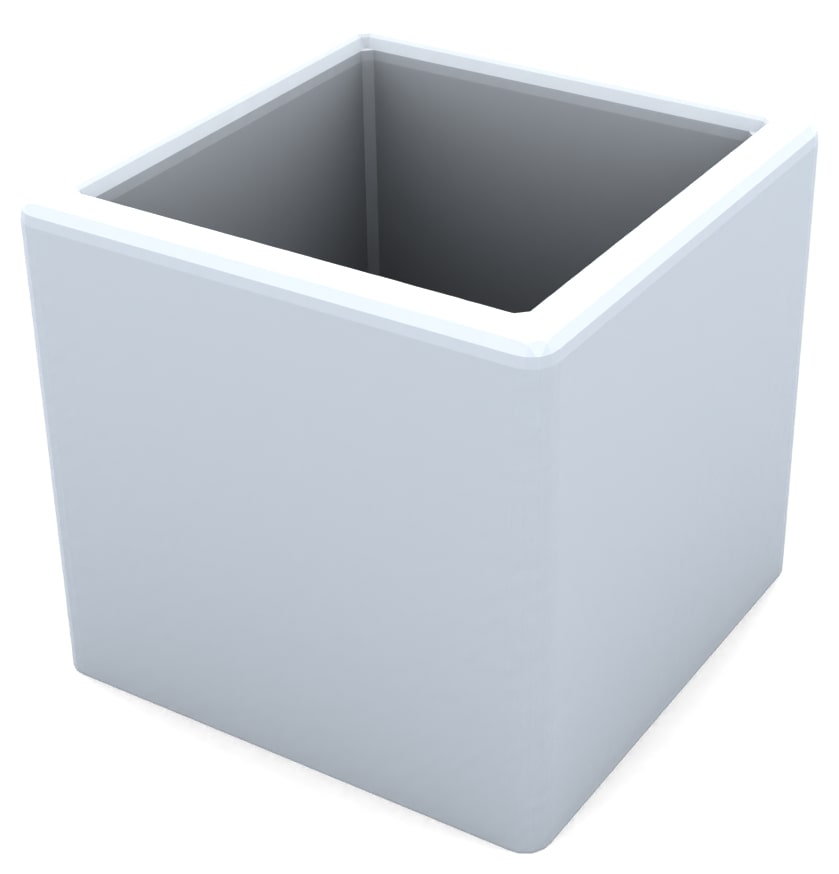}\\
    \makebox[1.8in]{(a) 2D illustration}
    \makebox[1.5in]{(b) $r$-offset volume}\\
    \caption{$r$-offset volume. In the 2D illustration (a), the solid curve represents the surface $S$, and the dotted curve is the boundary of the $r$-offset volume. The thickness parameter $r$ should be chosen to be less than half of the gap size $\eta$ to prevent the filling of the gap between the two ends. Regardless of the type of the input surface, the dilated double covering (i.e., the boundary surface of the $r$-offset volume) is always an orientable, closed 2-manifold. (b) shows an $r$-offset volume example. The model is scaled into a unit box. We use a relatively large offset value of $r=0.05$ for better visualization.}
    \label{fig:doublecover}
\end{figure}

Let $S$ be an arbitrary surface embedded in $\mathbb{R}^3$.
As illustrated in Figure~\ref{fig:doublecover}, for open surfaces, denote by $\eta$ the gap size. Given a positive real number $r<\eta/2$, simialr to~\cite{Lee2009}, we define the $r$-offset volume of $S$ by 
$$S \oplus r=\{p|\exists q\in S, \|p-q\|\leq r\}.$$
The offset, similar to the morphological dilation, assigns a thickness to the target surface and yields a solid. The key insight is that the boundary of the $r$-offset volume $\partial (S\oplus r)$ is always a closed, orientable, and manifold surface regardless the topology of $S$~\cite{Hatcher2002}. This property allows us to develop a unified framework for extracting the zero level-set of both open and closed surfaces. Because of the thickness $r$ assigned to the offset volume, we call $\partial(S\oplus r)$ a \textit{dilated} double cover of surface $S$.

\subsection{Learning the Covering Map $\pi$}
\label{sec:mapping}

Since the dilated double covering $\partial(S\oplus r)$ is closed, orientable 2-manifold, we can apply the conventional marching cubes algorithm~\cite{Lorensen1987} to extract the iso-surface. Denote by $\mathcal{M}$ the triangular mesh extracted by marching cubes. Next, we learn a continuous covering map $\pi$ to project $\mathcal{M}$ to $S$. 
We call such a mapping $\pi:\partial(S\oplus r)\rightarrow S$ a covering map.

\begin{figure}[!htbp]
\centering
    \includegraphics[width=1.1in]{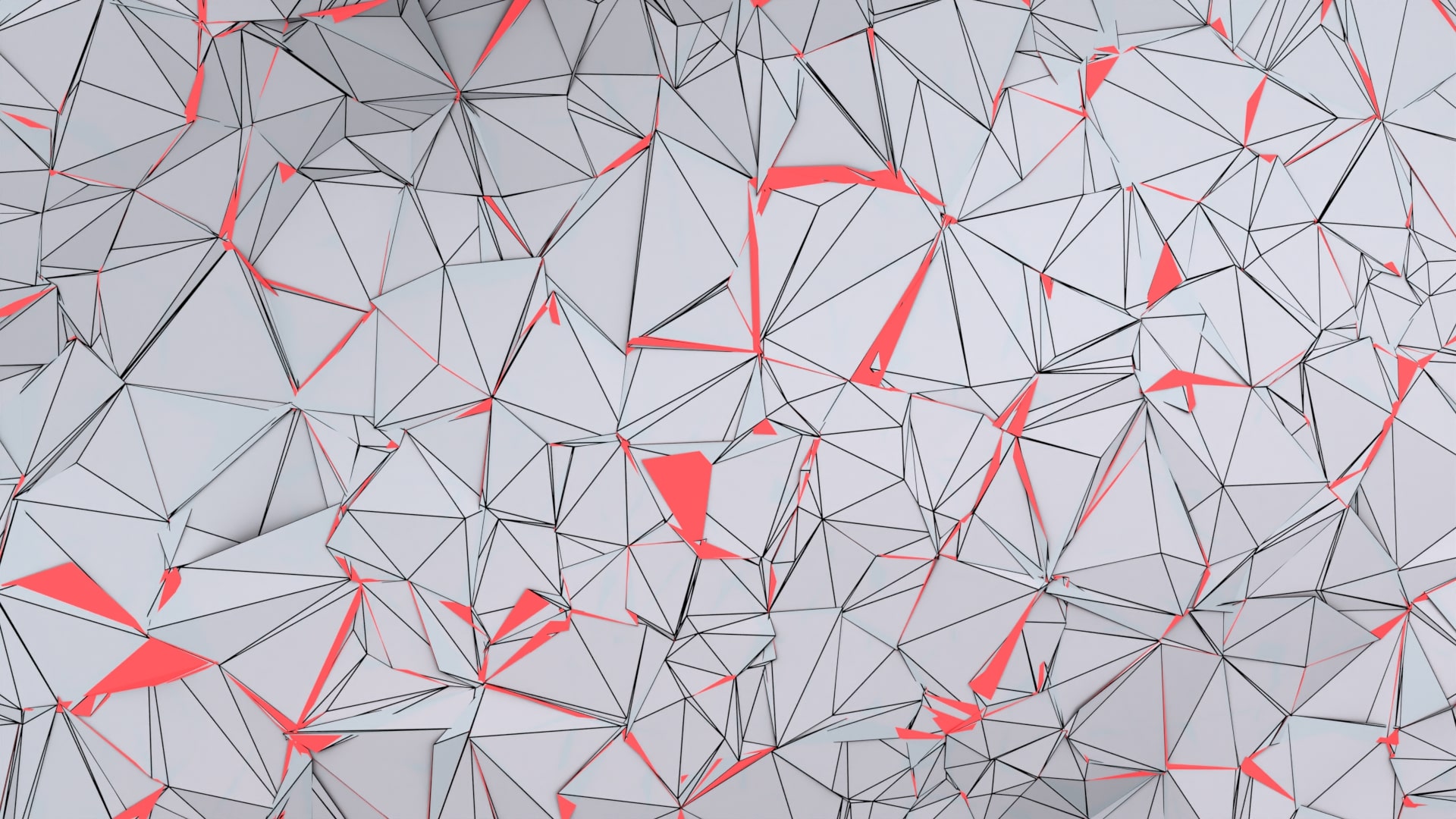}
    \includegraphics[width=1.1in]{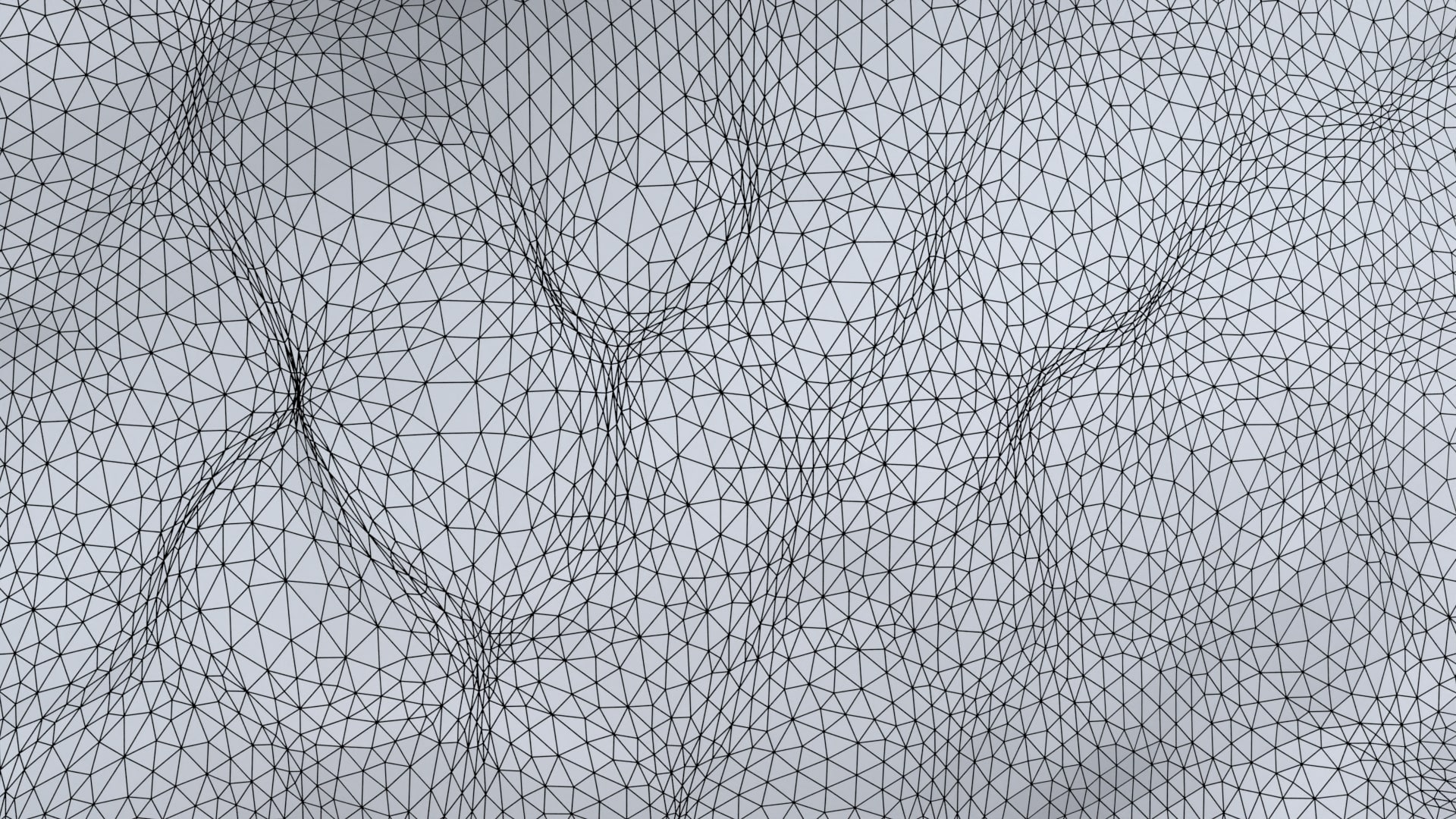}
    \includegraphics[width=1.1in]{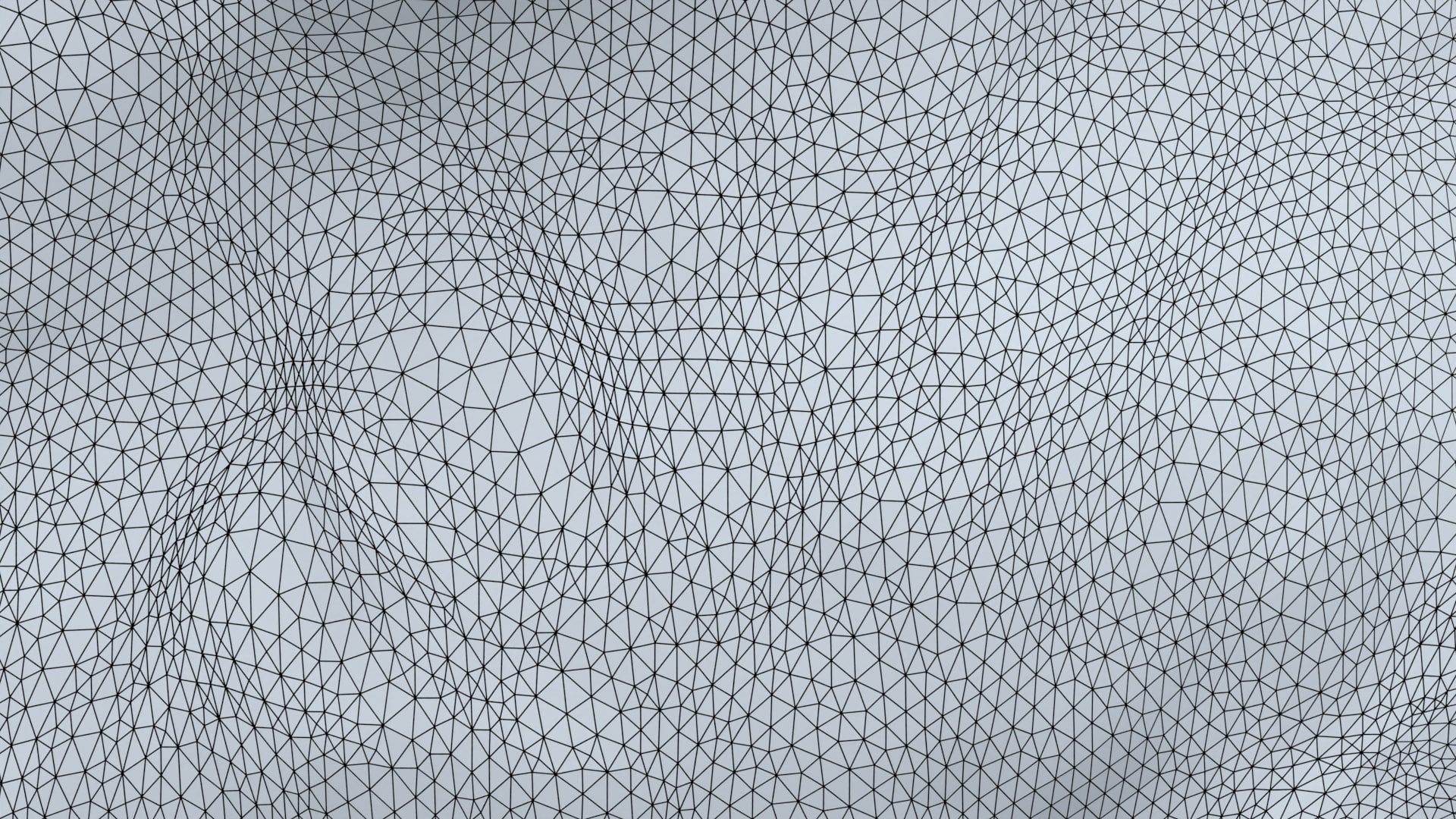}\\
    \begin{small}
    \makebox[1.1in]{(a) w/o Laplacian }
    \makebox[1.1in]{(b) Constant weight}
    \makebox[1.1in]{(c) Adaptive weight}
    \end{small}
    \caption{Effects of the Laplacian term in the objective function for learning the covering map $\pi$. (a) Without the Laplacian term, the resulting mesh is poorly tessellated and contains many flipped and self-intersecting triangles (highlighted in red). (b) When using a constant weight $w(p_i)\equiv 1$ for the Laplacian term,  undesired ``ridges'' or ``valleys'' often appear in certain relatively flat regions. The issue occurs because zero is not a regular value of the UDF, and the local minima of the UDF in the region are smaller than those of the neighboring points. Consequently, the vertices become overly concentrated in the region, resulting in an overcrowded appearance. (c) By employing an adaptive weight $w(p_i)=\sqrt{A_{\mathrm{max}}/A(p_i)}$ for the Laplacian term, the resulting triangles exhibit better quality. A marching cubes resolution $256^3$ is used in all three results.  }
    \label{fig:ablation_laplacian}
\end{figure}

Let $f$ be the input UDF represented by an MLP. Since $f$ is usually approximate, we formulate the projection as a minimization of $f$ for the projected points. Although attaining a solution with exact zero is improbable, for the sake of simplicity, we continue to refer to the extracted surface as the zero level-set. If treating each vertex of $\mathcal{M}$ independently as in~\cite{Chibane2020NDF}, the mesh would be broken after the optimization (see Figure~\ref{fig:ablation_laplacian}(a)). To preserve the mesh structure, we take \textit{all} vertices of $\mathcal{M}$ as the input and optimize their locations together to avoid folding and self-intersection. We also take the centroid of each triangular face of $\mathcal{M}$ into consideration, which serves as additional constraints. To improve the accuracy, we adopt a coarse-to-fine method for learning the covering map $\pi$: we learn an approximate map $\pi_1$ in the coarse step to obtain the normal directions, along which we can fine-tune the position of each vertex to obtain map $\pi_2$.

\paragraph{Learning a coarse map $\pi_1$.} We minimize the following loss function to learn an approximate covering map $\pi_1:\partial(S\oplus r)\rightarrow S$
\begin{small}
\begin{equation}
\label{eqn:step1}
\begin{small}
    \min_{\pi_1}\sum_{p_i\in\mathcal{M}\cup\mathcal{C}} f\big(\pi_1(p_i)\big)
    +\lambda_1\sum_{p_i\in\mathcal{M}} w(p_i)\Bigl\|\pi_1(p_i)-
    \frac{1}{|\mathcal{N}(p_i)|}\sum_{p_j\in\mathcal{N}(p_i)}\pi_1(p_j)\Bigr\|^2,
\end{small}
\end{equation}
\end{small}
where $\mathcal{C}$ is the set of triangle centroids, $\mathcal{N}(p_i)$ denotes the 1-ring neighboring vertices of $p_i\in \mathcal{M}$ and $w(p_i)$ is a position-dependent weight.
The first term $f(\pi_1(p_i))$ induces that the projected point $\pi_1(p_i)$ should be the local minimum of the input UDF. 
The second term serves as a surface Laplacian constraint, ensuring that the 1-ring neighbors of $p_i$ are uniformly distributed. To determine the weight $w(p_i)$ for this term, we consider the area $A(p_i)$ of surrounding triangular faces with vertex $p_i$. 
If $A(p_i)$ is large, $p_i$ is allowed to move freely following the gradient direction of $f$. However, if $A(p_i)$ is small, strong Laplacian constraints are necessary to regularize the points around $p_i$. Therefore, we assign weights $w(p_i)$ to each point $p_i$ based on the inverse correlation with the area $A(p_i)$,
$w(p_i)=\sqrt{A_{\mathrm{max}}/A(p_i)},$
where $A_{\mathrm{max}}=\max_{p\in \mathcal{M}} A(p)$ is the maximum area among all points in the mesh $\mathcal{M}$. It is important to note that $A_{\mathrm{max}}$, $A(p_i)$ and the weight $w(p_i)$ are updated in each epoch of our algorithm. 

Figure~\ref{fig:ablation_laplacian} highlights the effectiveness of the Laplacian term with adaptive weight in our approach. The Laplacian term plays a crucial role in preventing the occurrence of folded and flipped triangles in the extracted meshes. And, the adaptive weight facilitates the generation of triangles with similar sizes.

\paragraph{Learning a fine map $\pi_2$.} 
In the coarse-step optimization, we use the Laplacian term to prevent folding and self-intersection. The resulting  mesh $\pi_1(\mathcal{M})$ is smooth due to the smoothing effect of the Laplacian term.
The goal of the fine-tune step is to further update the position of mesh vertices and triangle centroids so that the learned covering map $\pi(\mathcal{M})=\pi_2\circ\pi_1(\mathcal{M})$ contains fine geometric details. Towards this goal, we compute the face normal $\overrightarrow{n}_i$ of each triangle of $\pi_1(\mathcal{M})$ and encourage each vertex $\pi_1(p_i)$ 
to move along the normal direction by penalizing the tangential displacements.
Specifically, we minimize the following loss function  
\begin{equation}
\label{eqn:step2}
\min_{\pi_2}\sum_{p_i\in\mathcal{M}\cup\mathcal{C}} f\big(\pi_2\circ\pi_1(p_i)\big)+\lambda_2\sum_{p_i\in\mathcal{C}}\left\|\big(\pi_2\circ\pi_1(p_i)-\pi_1(p_i)\big)\times \overrightarrow{n}_i\right\|,
\end{equation}
where $\times$ is the vector cross product.

\begin{figure}[!htbp]
\centering
\includegraphics[width=1.55in]{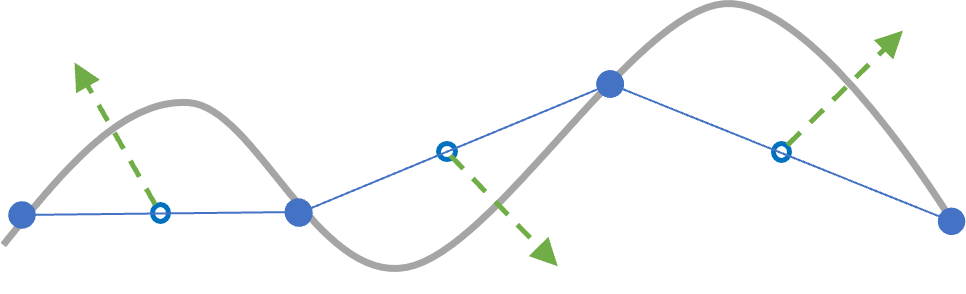}
\hspace{0.1in}
\includegraphics[width=1.55in]{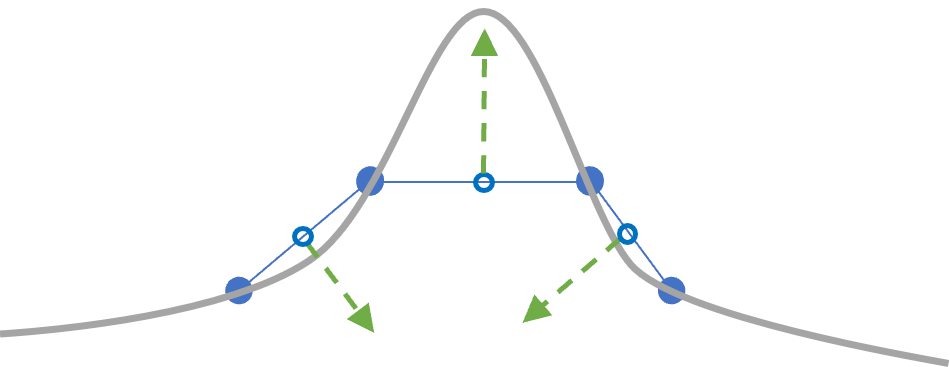}\\
\makebox[1.55in]{(a) Coarse step}
\makebox[0.1in]{}
\makebox[1.55in]{(b) Fine-tune step}\\
\caption{\label{fig:pull} Taking the triangle centroids as additional constraints into the optimization is effective in improving the accuracy in learning the covering map $\pi$. In this 2D illustration, we use the solid dots to indicate mesh vertices and the thin lines for the triangular faces, the hollowed dots for the centroids, and the dashed arrows for the update of the centroids. The centroids are encouraged to move along the normal direction of the corresponding triangle in the fine-tune step, while there is no such requirement in the coarse step.}
\end{figure}

\begin{figure}[!htbp]
    \centering
    \includegraphics[width=0.8in]{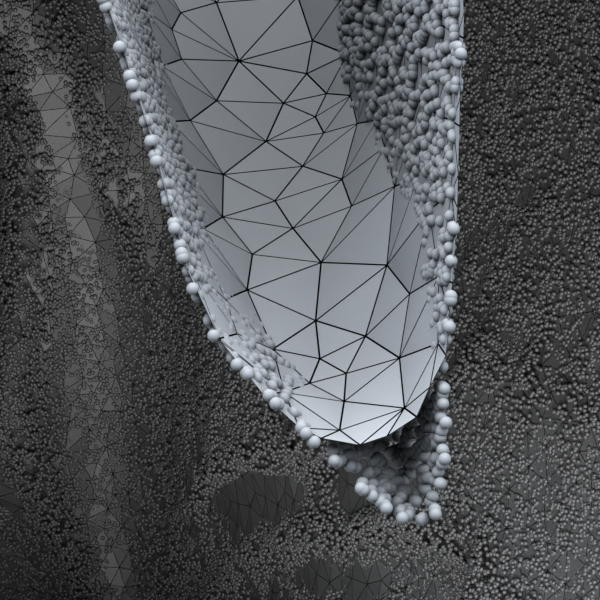 }
    \includegraphics[width=0.8in]{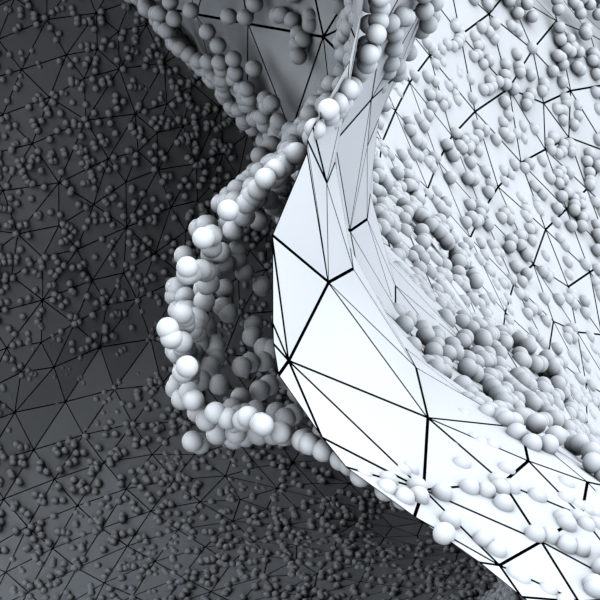 }
    \includegraphics[width=0.8in]{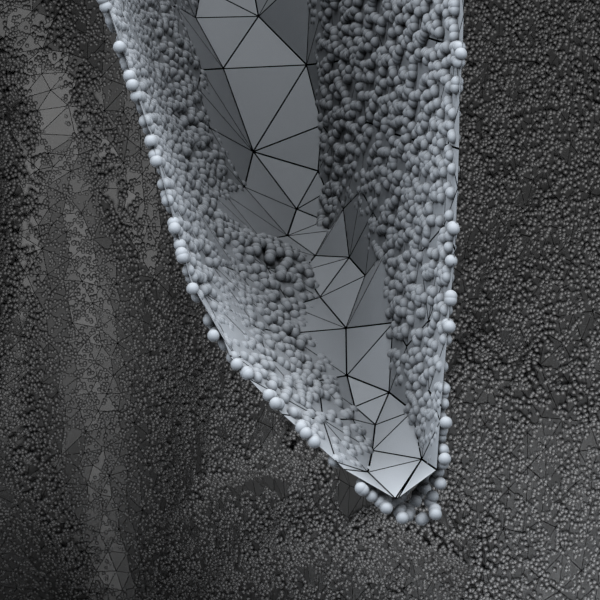 }
    \includegraphics[width=0.8in]{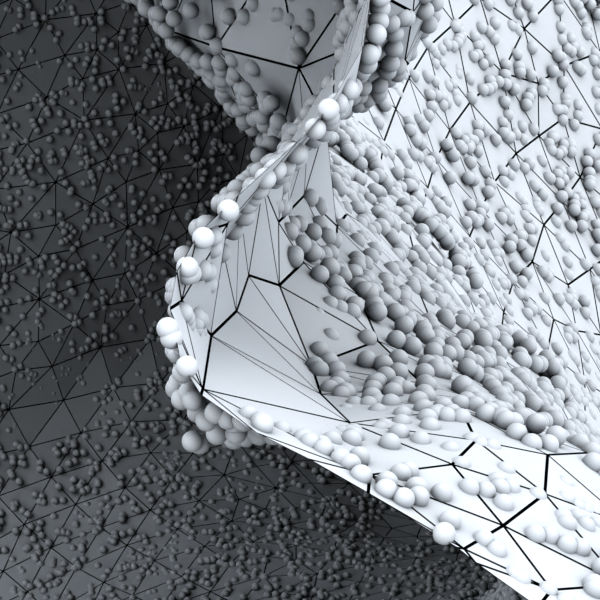 }\\
    \begin{small}
    \makebox[1.6in]{\makecell[c]{CD $0.399\times 10^{-3}$\\ w/o centroids}}
    \makebox[1.6in]{\makecell[c]{CD $0.334\times 10^{-3}$\\ w/ centroids}}\\
    \end{small}
    \caption{Effects of using centroids in learning the covering map $\pi$. Without centroids, the extracted mesh may not fully recover regions with high curvature, such as the tip in this example. This limitation arises because the triangle vertices are already on the zero level set and cannot be further updated. However, by incorporating centroids as additional constraints, we can provide additional forces to guide the triangle vertices to move along the zero level set, leading to further reduction of the objective function. We present the visual results with marching cubes resolutions of $512^3$ and the thickness parameter $r=0.0025$.}
    \label{fig:ablation}
\end{figure}

\textbf{Remark 1.}
It is important to note that although the triangle centroids appear in the objective functions~(\ref{eqn:step1}) and (\ref{eqn:step2}), they are not considered as free variables.
During the optimization process, each centroid is obtained as the average of the three triangle vertices, and only the vertices are optimized.
However, despite not being free variables, the triangle centroids play a \textit{necessary} role in improving the accuracy of the reconstruction. Intuitively, both the triangle vertices and triangle centroids can be seen as particles within the force field generated by the gradient of UDF $f$. As depicted in Figure~\ref{fig:pull}, when the mesh vertices $p_i$ are already located on the zero level-set, they cannot be further improved without additional ``forces''. By introducing the triangle centroids as additional constraints, each centroid is attracted by a force to the surface, leading to a more precise update of the triangle's position and orientation. Figure~\ref{fig:ablation} illustrates the effectiveness of using triangle centroids as additional constraints, demonstrating their crucial role in improving the accuracy of the reconstruction.

\textbf{Remark 2.} In general, computing an injective map between surfaces is a challenging task. Schmidt et al.~\shortcite{Schmidt2019} tackled this problem by formulating an elegant minimization of distortion between surfaces of disk topology that could have widely varying shapes, e.g., a pig and a camel. Note that our setting is different from theirs in that the boundary of the $r$-offset volume typically exhibits non-trivial topology, yet its two ``sides'' are geometrically similar. This unique characteristic of our problem allows us to develop a simple-yet-effective optimization method for computing the covering map $\pi$.

\textbf{Remark 3.} Both our approach and the non-rigid ICP method~\cite{Deng2022} involve a deformation procedure. However, the motivations behind the deformations differ significantly. In non-rigid ICP, the deformation aims to align two shapes, where in our case, the deformation is guided by moving points to reach minimum distance. Consequently, the objective functions for the two approaches are different.

\subsection{Segmenting the Double Layers for Open Models}
\label{sec:sep}
After projection, we obtain a double-layered mesh $\pi(\partial(S\oplus r))$. Since $\partial(S\oplus r)$ is a closed  orientable 2-manifold, 
so is the projected mesh $\pi(\partial(S\oplus r))$.
Apparently, any subset of $\pi(\partial(S\oplus r))$ is also orientable.
If $S$ is non-orientable or non-manifold, we cannot separate the double-layered mesh to reproduce $S$.
Consequently, we simply retain the double-layered mesh as the output, which is of different topology to $S$.

\begin{figure}[!htbp]
    \centering
    \includegraphics[width=1.15in]{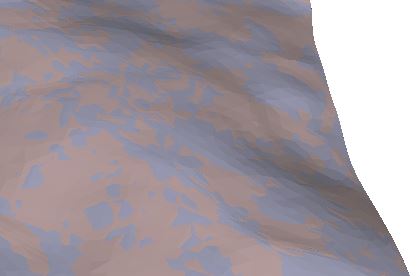}
    \includegraphics[width=1.95in]{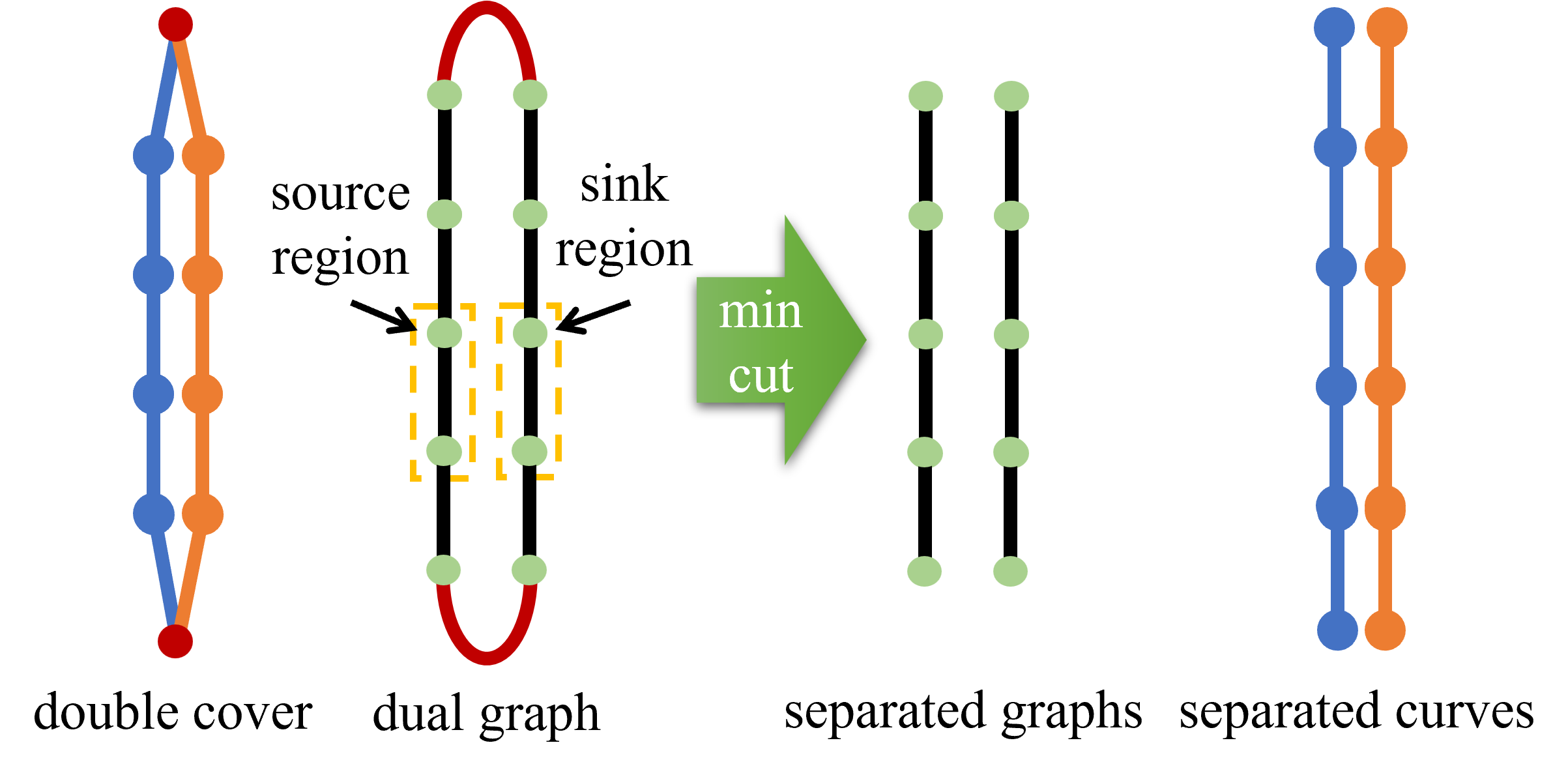}\\
    \makebox[1.3in]{(a)} \makebox[1.8in]{(b)}
    \caption{Segmenting the double layers. (a) For an open, orientable and manifold surface, the covering map $\pi$ generates two overlapping layers (colored in brown and grey, respectively) with nearly identical geometry.
    (b) illustrates the separation of the double layers in 2D. For visualization purpose, the two layers are intentionally depicted with different colors and a gap between them. The minimum cut in the dual graph consists of the two red edges with the lowest weights (corresponding to $0^\circ$). Removing these edges effectively separates the two layers. }
    \label{fig:min-cut}
\end{figure}

When $S$ is an orientable manifold with boundary, it is highly desirable to output a single-layered mesh. Towards this goal, we develop a post-processing step to separate the two layers. Observe that the two layers of $\pi(\partial(S\oplus r))$ are overlapping and almost congruent in geometry. This implies that the cutting curve consists of the edges whose dihedral angles are close to $0^\circ$ (see Figure~\ref{fig:min-cut}(a)). However, it is not robust to na\"{i}vely applying this heuristic to search for the cutting edges, since it is hard to find a good angle threshold in practice. To separate the layers in a robust manner, we adopt a \textit{global} strategy based on minimum cut in graph theory~\cite{Boykov2004}. 

As illustrated in Figure~\ref{fig:min-cut}, we build a dual graph of the double-layered mesh, where every triangular face corresponds to a graph node and two nodes are connected if the two corresponding triangles share a common edge.
Therefore, every graph edge corresponds to a dihedral angle of the double-layered mesh. Let $\alpha_{\min}$ be the minimum angle among all dihedral angles.
Consider a graph edge $(i,j)$ with dihedral angle $\alpha_{ij}$, where $0\leq\alpha_{ij}\leq\pi$. 
We assign the edge a weight $\exp(200(\alpha_{ij}-\alpha_{\min}))$, so that small angles are associated with small weights, and the other way around.

Since our goal is to separate the two layers evenly, we should compute a minimum normalized cut, which, unfortunately, is an NP-hard problem. To tackle the challenge, we use a simple strategy instead. Take a \textit{group} of nearby nodes with large dihedral angles as the source $s$ and its twin counterpart as the sink $t$. We then compute a minimum $s\text{-}t$ cut, which can be solved efficiently in polynomial time.

For simplicity, we only deal with the largest connected component in our implementation. Initially, we select a random node to serve as the seed of the source region. We then employ a breadth-first search to expand this seed until the region contains $5\%$ of the total number of graph nodes, establishing our source region.
Next, we identify its ``twin node'', - a node that is close in Euclidean distance to the source seed but distant in terms of graph distance - as the seed of the sink. We then grow the sink region in a manner similar to the source.
If the two regions do not share any common triangles, we consider them as a valid source-sink pair. Then we apply Boykov and Kologorov's algorithm~\shortcite{Boykov2004} to compute the $s\text{-}t$ cut.

Upon slicing, we obtain two distinct meshes, $M_1$ and $M_2$. We further check whether $M_1$ and $M_2$ are congruent by comparing their respective face counts. We accept a separation if the face difference is within $15\%$ of the total face count of $M$. In such a case, we return the mesh with the higher number of faces. If this condition is not met, we identify a new pair of source and sink regions and repeat the $s\text{-}t$ cut until achieving an acceptable separation. Our experimental findings affirm the efficacy of this approach, as most tests yield successful cuts in no more than 3 attempts. On the rare occasion where a cut cannot be determined after 5 attempts, we decrease the source/sink regions size by half and repeat the above procedure.

\textbf{Remark 4.} It is worth noting that closed model does not necessitate a cut, as its double cover comprises two disconnected components: one representing the inner side and the other the outer side. Consequently, we simply retain the component with the greater number of faces as the final output.

\section{Experiments and Comparisons}

\paragraph{Implementation}
We implemented DoubleCoverUDF in Python and tested it on a workstation equipped with an AMD Ryzen 9 3950X CPU, 64GB of RAM, and an NVIDIA GeForce 3080 GPU with 10GB of memory. The processes for learning the UDF and computing the covering map $\pi$ were performed on the GPU, whereas computations for the dilated double cover $\partial (S\oplus r)$ and the minimum $s\text{-}t$ cut were carried out on the CPU. Given that the UDF $f$ is encoded in an MLP, Equations~(\ref{eqn:step1}) and (\ref{eqn:step2}) are differentiable. We used the vector Adam solver~\cite{Ling2022}, a specialized extension of the standard Adam solver~\cite{Kingma2015} designed for  rotational equivalence geometry optimization, to minimize these equations. For optimizing Equations~(\ref{eqn:step1}) and (\ref{eqn:step2}), we performed 300 and 100 epochs, respectively. In each epoch, we computed all vertex gradients in multiple batches before collectively updating the vertex positions. We uniformly scaled the point cloud models into a unit box, so that the parameters are independent of the models' scale. We empirically assigned the weights $\lambda_1=2000$ in (\ref{eqn:step1}) and $\lambda_2=0.5$ in (\ref{eqn:step2}) in our experiments.

\paragraph{Quantitative measures}
To assess the quality of the extracted meshes, we utilize both geometrical and topological metrics. Specifically, we employ the Chamfer distance (CD) as a geometrical metric and several topological metrics, including the percentage of non-manifold vertices $\tau_v$, the percentage of non-manifold edges $\tau_e$, the genus $g$, the number of boundaries $b$, and the 0-dimensional Betti number $\beta_0$, measuring the number of connected components. Non-manifold vertices are the vertices whose local neighborhood is the union of two or more topological disks or half-disks (see the right inset). This occurs when two or more surfaces touch each other at a
\begin{wrapfigure}{r}{0.12\textwidth}
\centering
\includegraphics[width=0.12\textwidth]{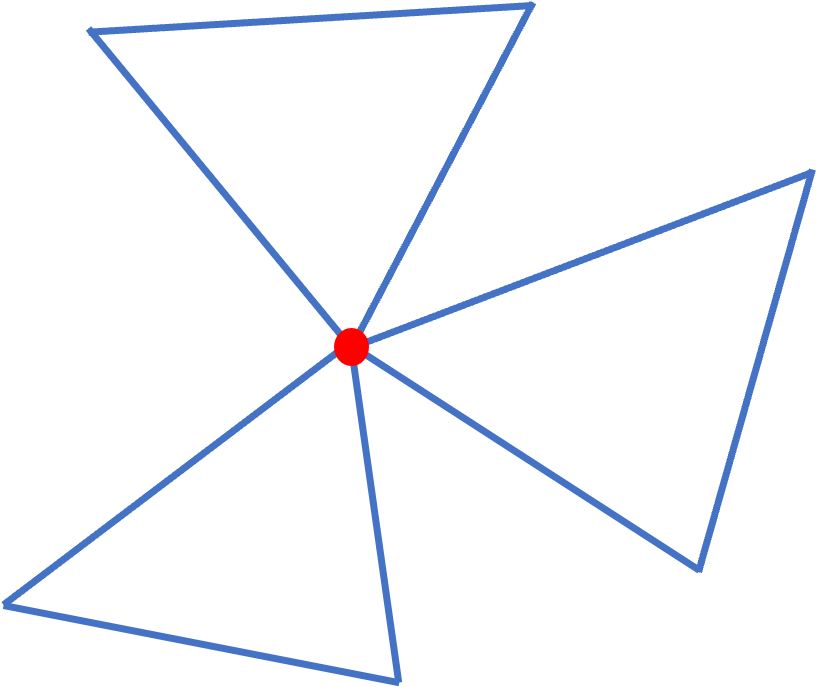}
\end{wrapfigure}
single point. Non-manifold vertices are generally considered undesirable in generated meshes due to their potential impact on the robustness of downstream applications. It is preferable to prevent the generation of non-manifold vertices as removing them often results in the creation of undesired holes in the mesh.
Genus and number of boundaries are the global topological features of the meshes. We compute genus for manifold meshes using Euler characteristic. For non-manifold meshes, we first split non-manifold vertices and remove non-manifold edges before counting the number of boundaries and computing the genus. 

\paragraph{The choice of $r$} The offset thickness parameter $r$ plays a crucial role in our algorithm, and it should be chosen carefully to satisfy certain criteria. On one hand, in theory, the thickness $r$ needs to be less than half of the minimal gap size, i.e., $r\leq \frac{\eta_{\min}}{2}$, to ensure the dilated double cover does not fill in the gap. On the other hand, the value of $r$ cannot be arbitrarily small due to practical considerations. Firstly, considering the potential presence of UDF learning errors, the UDF values of the points on the target surface may not precisely reach zero. Thus, to avoid  unintended handles on the double covered mesh and subsequent creation of undesired holes upon separating the double layers, it is important to select an appropriate value of $r$ that is greater than the maximum UDF value for the points on the zero level-set. Secondly, the choice of $r$ is also influenced by the resolution of the marching cubes algorithm. It is important to ensure that $r$ is greater than half of the edge length of the cubes to preserve the double cover structure. Failure to do so could result in missing intersections at both ends of an edge crossing the $r$ iso-surface. Therefore, the thickness parameter $r$ should satisfy the following condition:
\begin{equation}
\max\left\{\frac{1}{2k}, d_{\max}\right\} 
\leq r \leq \frac{\eta_{\min}}{2},
\label{eqn:range}
\end{equation}
where $d_{\max}$ is the maximum UDF value for points on the zero level-set and $k$ is the marching cubes resolution, typically ranging from 256 to 1024. In general, the maximum UDF value for the points on the target surface is influenced by the accuracy of the learned UDF and the noise level of the given point clouds.

Note that the minimal gap size $\eta_{\min}$ is specific to each model and is often unknown in real-world scenarios. Additionally, for noisy point clouds, the exact value of $d_{\max}$ is not available until the zero level-set is extracted, creating a chicken-egg dilemma. To address these challenges, we provide flexibility to the users by allowing them to specify the parameter $r$ and adapt DoubleCoverUDF to their specific data and desired outcomes.

\begin{table}[!htbp]
\caption{\label{tab:shapenet}Evaluation on the ShapeNet-Car dataset with different $r$ values. The resolution of marching cubes is set to $256^3$ for all methods. We utilize CAPUDF to learn the UDFs and measure the quality of the extracted meshes using the Chamfer distances. MeshCAP achieves the smallest errors, but their extracted meshes contain many non-manifold vertices and are often broken, resulting in many disconnected components. MeshUDF can effective reduce non-manifold vertices, however, it cannot ensure the absence of those vertices. Additionally, the extracted meshes from MeshUDF tend to have higher CD errors and exhibit visual artifacts such as non-smoothness and broken parts. $\overline{\beta_0}$ is the average number of connected components in the output mesh.}
\begin{small}
\setlength\tabcolsep{2pt}
\begin{tabular}{l|c|c|c|c|c}
\hline
                      & \multicolumn{3}{c|}{CD $(10^{-3})$} &\multirow{2}{*}{$\tau_v$} & \multirow{2}{*}{$\overline{\beta_0}$}\\
                      \cline{2-4}
                      & \begin{small}mean\end{small} & \begin{small}gt $\rightarrow$ pred\end{small} & \begin{small}pred $\rightarrow$ gt \end{small}& &\\
                      \hline
                      \hline
MeshCAP  & 3.007  &  2.886  & 3.129 & 4.44\% & 19534.92\\
MeshUDF  &  3.412 &  2.887 &  3.886  & 0.0018\% & 392.09\\
Ours ($r=0.0100$) & 3.556   & 3.450  & 3.661	  &	 0 &	43.70\\
Ours ($r=0.0075$) & 3.403   & 3.246  &  3.559   &	 0 & 54.38\\
Ours ($r=0.0050$) & 3.290   & 3.105  &	3.476   &	 0 & 60.48\\
Ours ($r=0.0025$) & 3.164   & 3.006   & 3.321	  &	 0 & 80.81\\
\hline
\end{tabular}
\end{small}
\end{table}

To understand the impact of different $r$ values on the results, we conducted extensive experiments on the ShapeNet-Car dataset, which comprises 3,514 car models, at a marching cubes resolution of $256^3$. We generated double-layered meshes. Note that the models in this dataset are noise-free, which enables us to obtain highly accurate UDFs. As a result, we established a lower bound of $r$ at $0.00195$ for the resolution of $256^3$. With this theoretical lower bound in mind, we evaluated four different values of $r$ (0.0025, 0.005, 0.0075, and 0.01) for each model. The consistent results presented in Table~\ref{tab:shapenet} demonstrate that $r=0.0025$ produces the most favorable outcomes. This finding confirms that choosing a lower value of $r$ while satisfying the conditions in (\ref{eqn:range}) leads to better extracted meshes.

For noisy point clouds, such as the clothes models in the Deep Fashion3D dataset~\cite{Zhu2020}, the learned UDFs are less accurate. To account for the noise and tolerate the inaccuracies in the input UDFs, it is recommended to use a relatively larger value of $r$. However, due to the lack of ground truth meshes in the Deep Fashion3D dataset, $d_{\max}$ is unavailable. Therefore, we empirically set $r=0.005$ for all the clothes models when using marching cubes of resolution $256^3$. 

\begin{figure}[!htbp]
    \centering
    \includegraphics[width=1.1in]{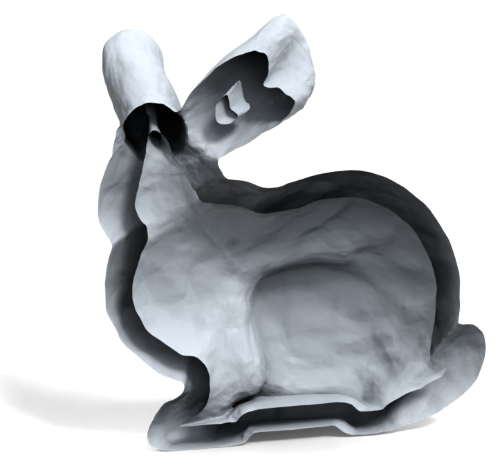 }
    \includegraphics[width=1.1in]{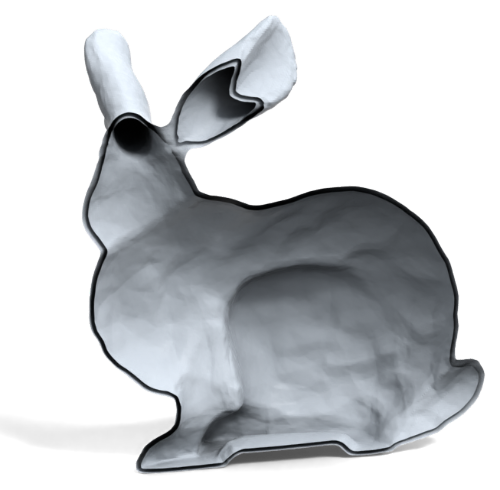 }
    \includegraphics[width=1.1in]{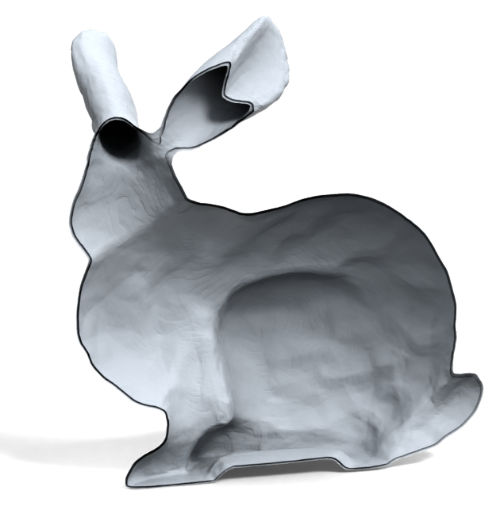 }\\
    \makebox[1.1in]{$r=0.03$}
    \makebox[1.1in]{$r=0.005$}
    \makebox[1.1in]{$r=0.0025$}\\
    \includegraphics[width=1.1in]{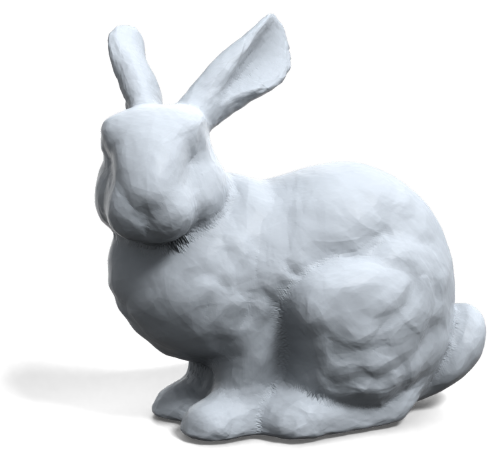 }
    \includegraphics[width=1.1in]{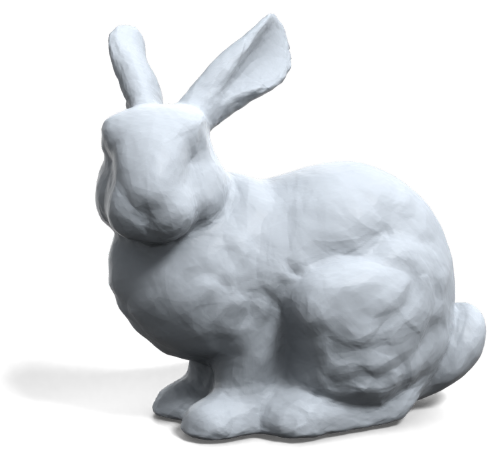 }
    \includegraphics[width=1.1in]{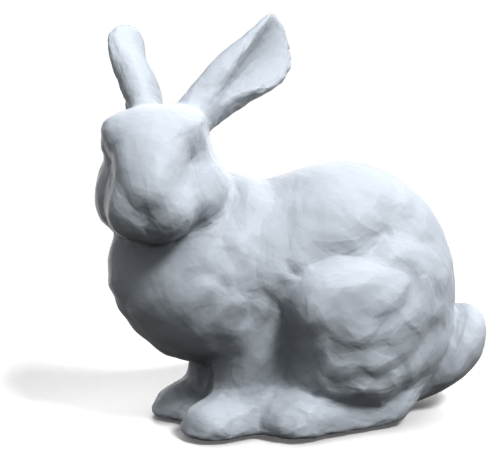 }\\
    \makebox[1.1in]{CD $=2.632\times 10^{-3}$}
    \makebox[1.1in]{CD $=2.580\times 10^{-3}$}
    \makebox[1.1in]{CD $=2.534\times 10^{-3}$}\\
    \includegraphics[width=1.1in]{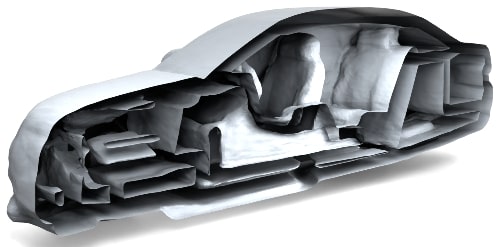 }
    \includegraphics[width=1.1in]{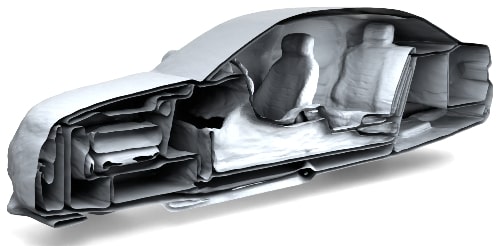 }
    \includegraphics[width=1.1in]{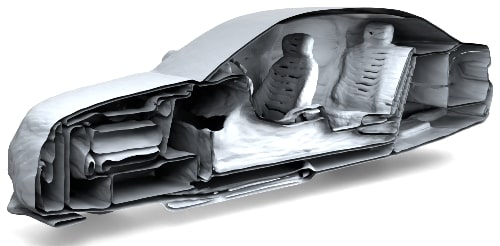 }\\
    \makebox[1.1in]{$r=0.01$}
    \makebox[1.1in]{$r=0.0025$}
    \makebox[1.1in]{$r=0.0019$}\\
    \includegraphics[width=1.1in]{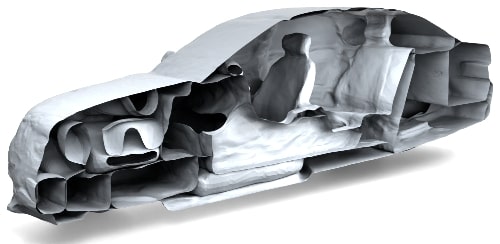}
    \includegraphics[width=1.1in]{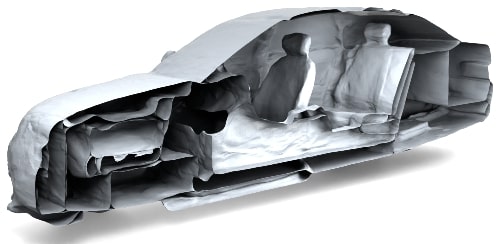 }
    \includegraphics[width=1.1in]{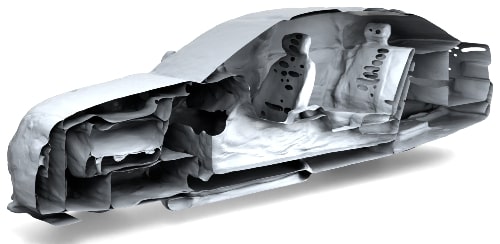 }\\
    \includegraphics[width=1.1in]{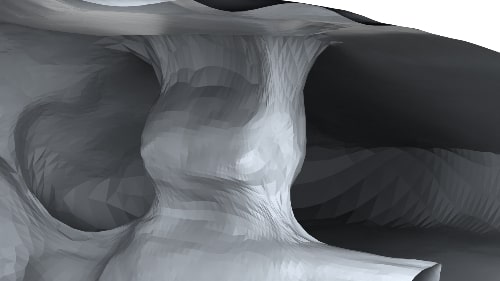}
    \includegraphics[width=1.1in]{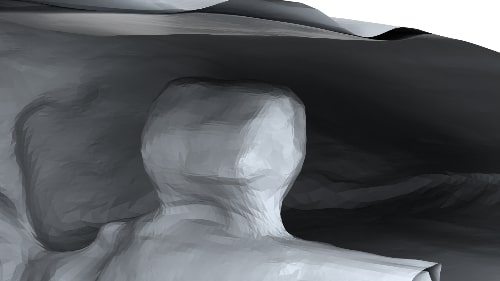 }
    \includegraphics[width=1.1in]{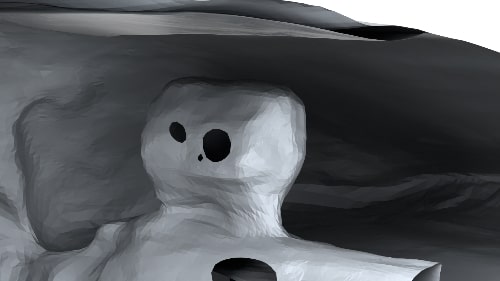 }\\
    \makebox[1.1in]{CD $=3.487\times 10^{-3}$}
    \makebox[1.1in]{CD $=2.045\times 10^{-3}$}
    \makebox[1.1in]{CD $=3.041\times 10^{-3}$}\\
    \includegraphics[width=0.90in]{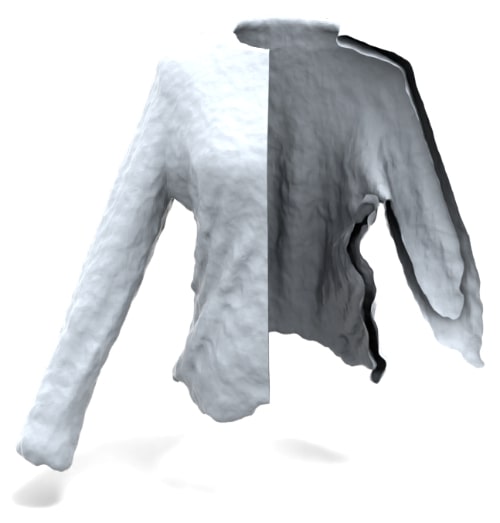 }
    \hspace{0.2in}
    \includegraphics[width=0.90in]{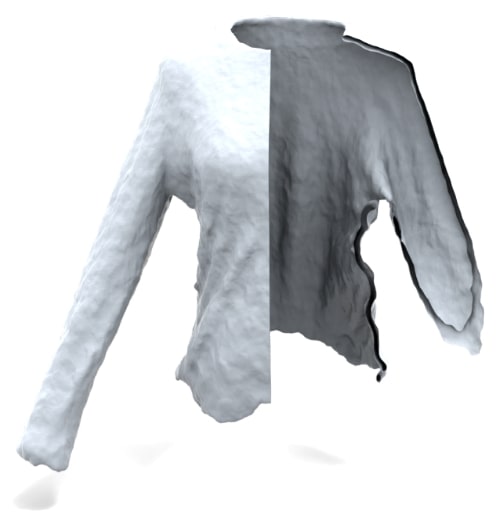 }
    \hspace{0.2in}
    \includegraphics[width=0.90in]{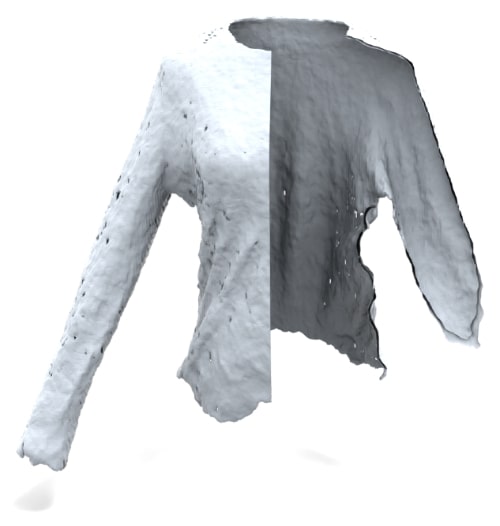 }
    \hspace{0.2in}\\
    \makebox[0.90in]{$r=0.01$}
    \hspace{0.2in}
    \makebox[0.90in]{$r=0.005$}
    \hspace{0.2in}
    \makebox[0.90in]{$r=0.0025$}
    \hspace{0.2in}\\
    \includegraphics[width=0.90in]{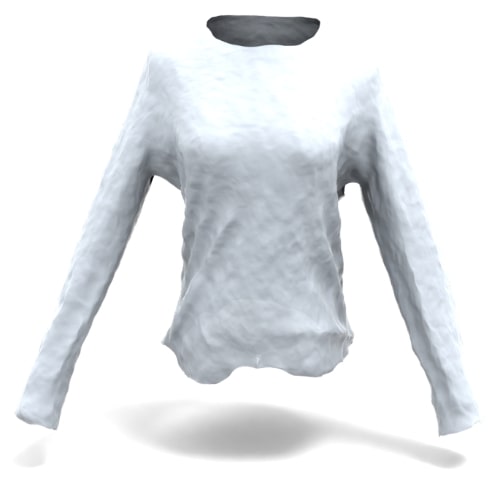}
    \hspace{0.2in}
    \includegraphics[width=0.90in]{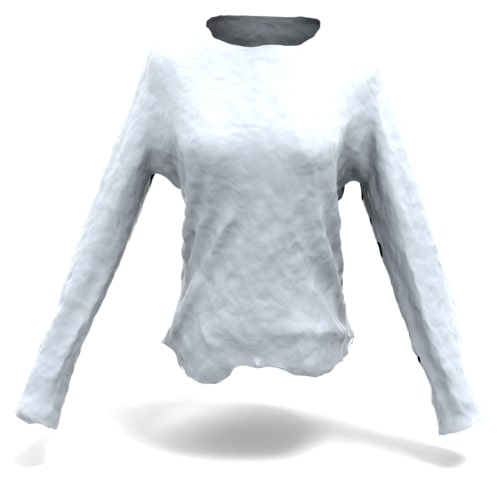 }
    \hspace{0.2in}
    \includegraphics[width=0.90in]{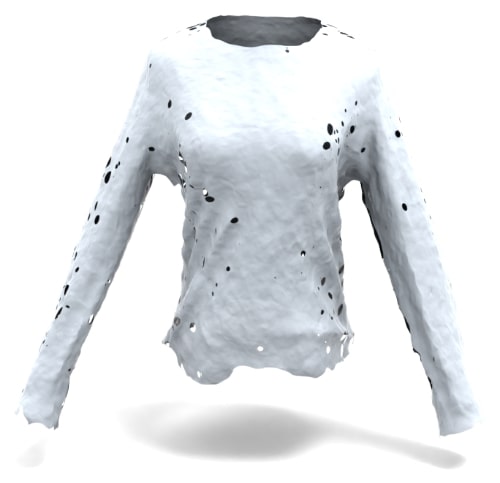 }
    \hspace{0.2in}\\
    \makebox[0.90in]{CD $=1.857\times 10^{-3}$}
    \hspace{0.2in}
    \makebox[0.90in]{CD $=1.850\times 10^{-3}$}
    \hspace{0.2in}
    \makebox[0.90in]{CD $=1.873\times 10^{-3}$}
    \hspace{0.2in}\\
    \caption{Impact of the thickness parameter $r$. When applied to the Bunny model, which is a clean model without gaps and thin structures, the choice of $r$ has little impact on the extracted meshes. However, for the car model with  thin structures, it is necessary to choose a relatively small value of $r$ in order to effectively separate the nearby sheets. For the noisy clothes model, it is recommended to use a larger value of $r$ to tolerate the noise and potential inaccuracies present in the input UDFs. A marching cubes resolution of $256^3$ is used for all three models. For each model, we present a cut view of the double covered mesh and the extracted meshes.}
    \label{fig:MC_threshold}
\end{figure}

Figure~\ref{fig:MC_threshold} provides a visual illustration of the impact of different $r$ values on the quality of the extracted meshes for the Bunny model, a car model from the ShapeNet-Car dataset, and a garment model from the Deep Fashion3D dataset. The evaluation is conducted at a marching cubes resolution of $256^3$. Both the Bunny model and the car model are noise-free and have ground truth meshes, allowing us to measure the accuracy using the Chamfer distance. While the Bunny model lacks thin structures, the car model exhibits thin structures. On the other hand, the clothes model from the Deep Fashion3D dataset is noisy and does not have a ground truth mesh. Therefore, we measured the Chamfer distances between the extracted meshes and the input noisy points. 

For the Bunny model, our evaluation showed that the choice of $r$ within a wide range of $[0.0025, 0.030]$ has little impact on the quality of the extracted mesh. Specifically, we observed that setting a smaller value of $r$, such as $0.0025$, resulted in a marginal improvement in accuracy of approximately $3.7\%$ compared to using a larger value of $r$, such as $0.03$. This finding suggests that for clean models lacking thin structures, the choice of $r$ can be flexible, and even a small change in $r$ has limited impact on the overall quality of the extracted mesh.

For the car model, we extracted double-layered meshes. We observed that its complex inner structures inside the car body make it more sensitive to the choice of $r$. When using a larger value of $r=0.01$, the top of the seats becomes connected to the roof of the car, which is not desired. On the other hand, a smaller value of $r$, such as $0.0025$, correctly separates the seats and the roof. We also conducted experiments with $r=0.0019$, which is even lower than the theoretical lower bound of $0.00195$ (half of the cube edge length). This choice of $r$ results in a slightly higher CD error compared to a valid $r$ value of $0.0025$ and introduces a few undesired holes. This example highlights the importance of selecting a valid $r$ that satisfies both the lower and upper constraints to achieve optimal results.

The clothes model from Deep Fashion3D is a noisy point cloud that contains a few holes due to the limitations in the image-based 3D reconstruction algorithm. In addition, the accuracy of the input UDF for the clothes model is not as high as that for the Bunny and car models. When using a small value of $r$, such as $0.0025$, more undesired holes appear in the extracted mesh. Increasing the value of $r$ to $0.01$ can effectively fill most of the undesired holes.

The above discussions offer a general guideline for users to set the thickness parameter $r$. In our experiments, we used the default values $r=0.0025$ for the clean models and $r=0.005$ for the noisy models and observed satisfactory results. In more challenging scenarios, we recommend users to determine the appropriate $r$ through a trial-and-error process.

\begin{figure*}[!htbp]
\setlength\tabcolsep{1pt}
\begin{small}
\begin{tabular}{cccccccc}
    MeshUDF & MeshCAP & DCUDF &  GT/Input & MeshUDF & MeshCAP & DCUDF &  GT/Input \\
    \includegraphics[width=0.545in]{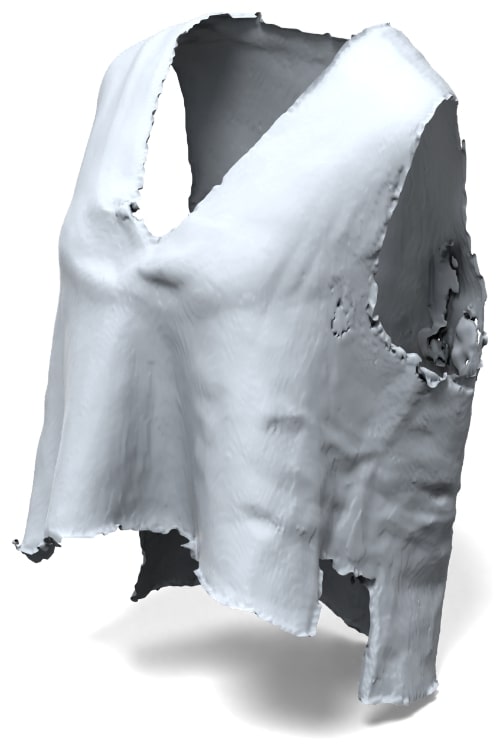} & 
    \includegraphics[width=0.545in]{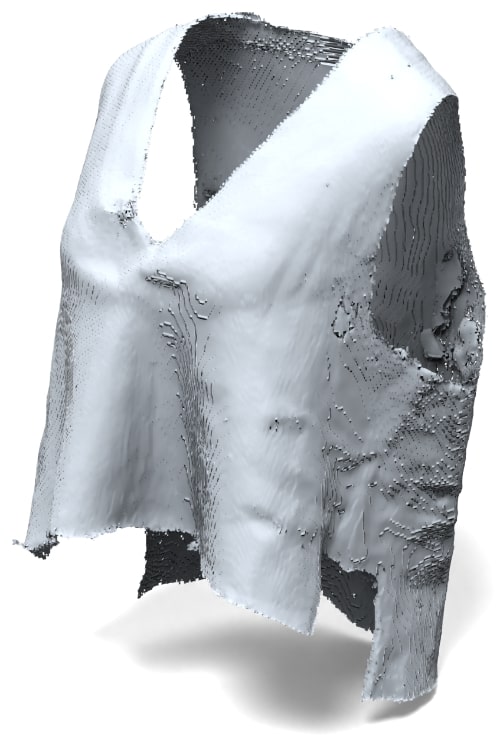} & 
    \includegraphics[width=0.545in]{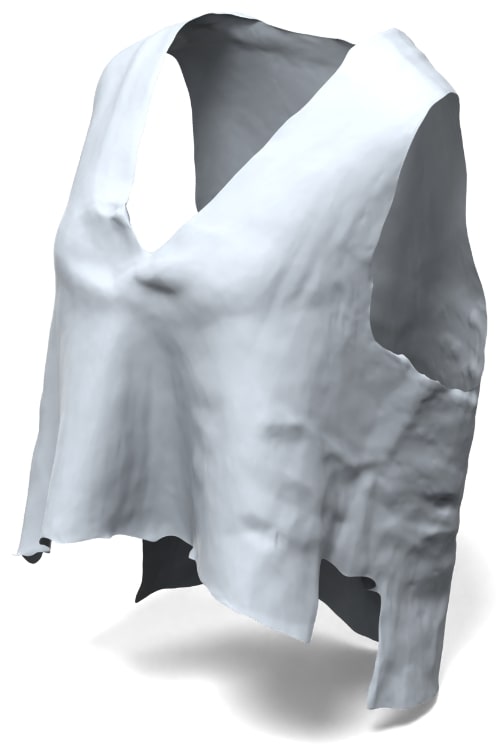} & 
    \includegraphics[width=0.545in]{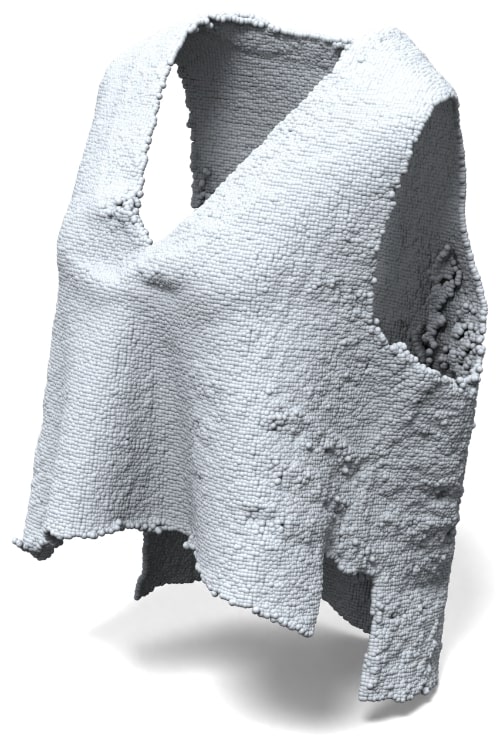} & 
    \includegraphics[width=0.845in]{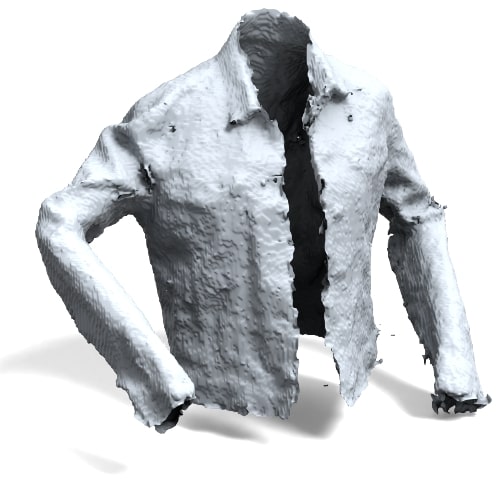} & 
    \includegraphics[width=0.845in]{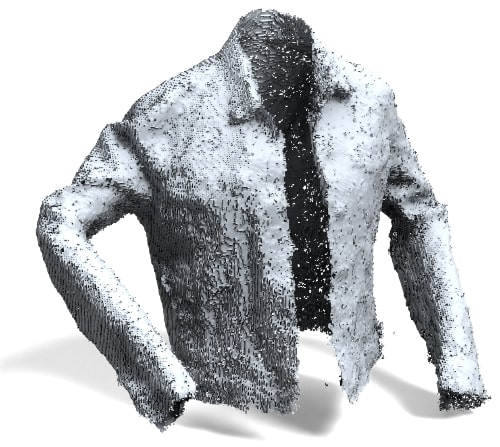} & 
    \includegraphics[width=0.845in]{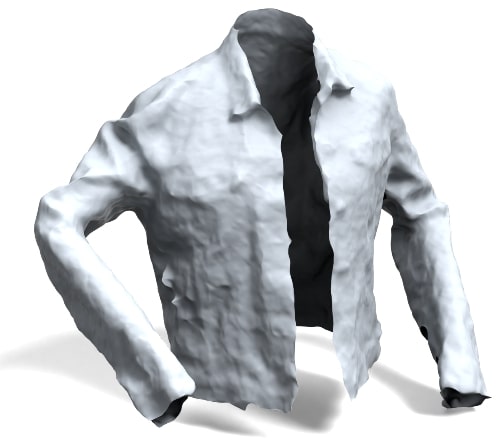} &
    \includegraphics[width=0.845in]{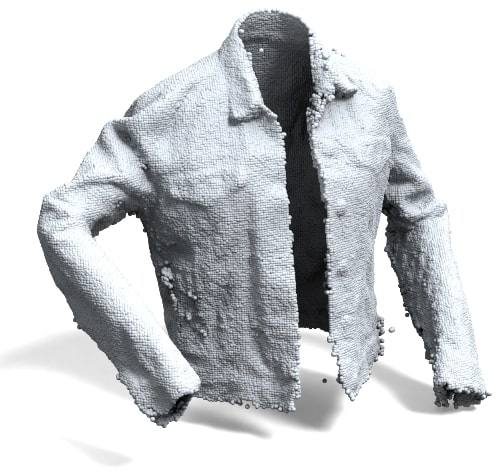 }\\
    \includegraphics[width=0.7in]{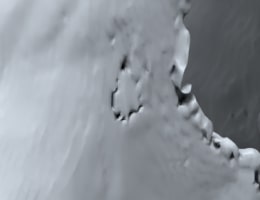} &
    \includegraphics[width=0.7in]{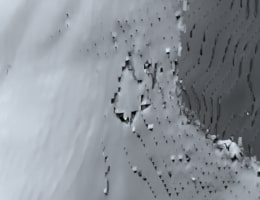} &
    \includegraphics[width=0.7in]{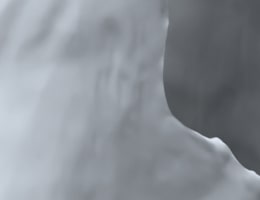} &
    \includegraphics[width=0.7in]{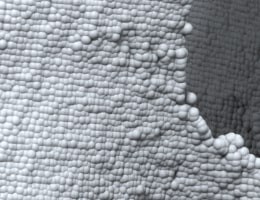} & 
    \includegraphics[width=0.7in] {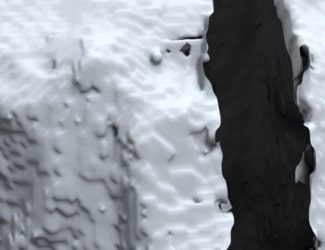} & 
    \includegraphics[width=0.7in]{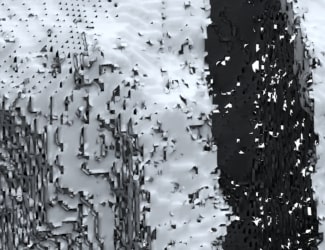} & 
    \includegraphics[width=0.7in]{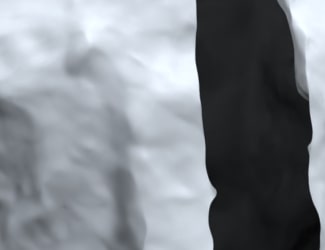} &
    \includegraphics[width=0.7in]{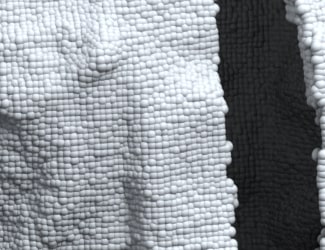 }\\
 $\tau_v=0\%$ & $\tau_v=1.209\%$ & $\tau_v=0\%$ & --- & $\tau_v=0.00053\%$ & $\tau_v=6.791\%$ &  $\tau_v=0\%$ & ---\\
    $b=50,g=61$ & $b=2461,g=129$ & $b=6,g=0$ & $b=4,g=0$ & $b=66,g=30$ & $b=16763,g=51$ & $b=14,g=1$ & $b=3,g=0$\\
    CD = 2.113$\times 10^{-3}$ & CD = 1.981$\times 10^{-3}$ & CD = 2.109$\times 10^{-3}$ &--- & CD = 2.291$\times 10^{-3}$ & CD = 2.177$\times 10^{-3}$ & CD = 2.193 $\times 10^{-3}$&---\\

    \includegraphics[width=0.8in]{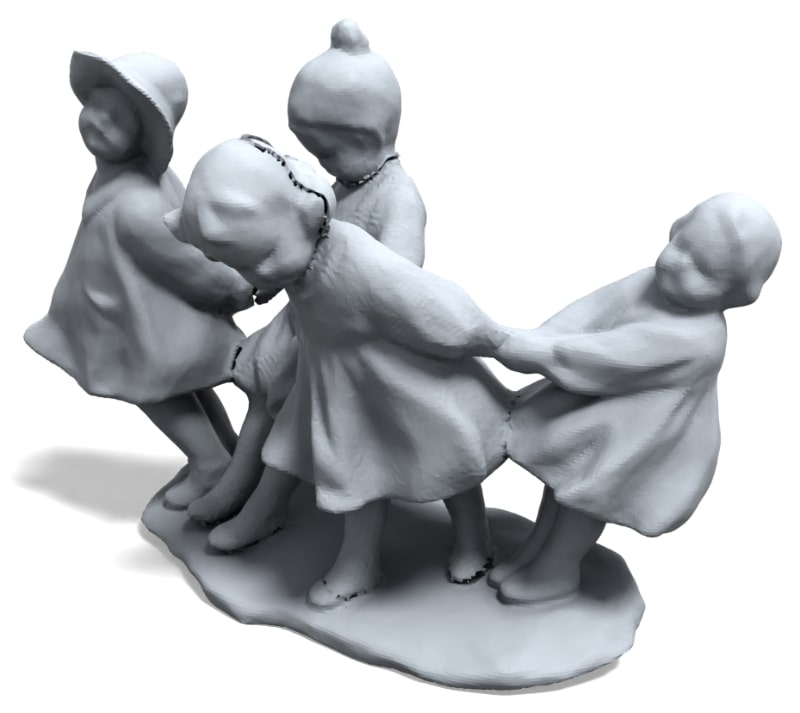 } & 
    \includegraphics[width=0.8in]{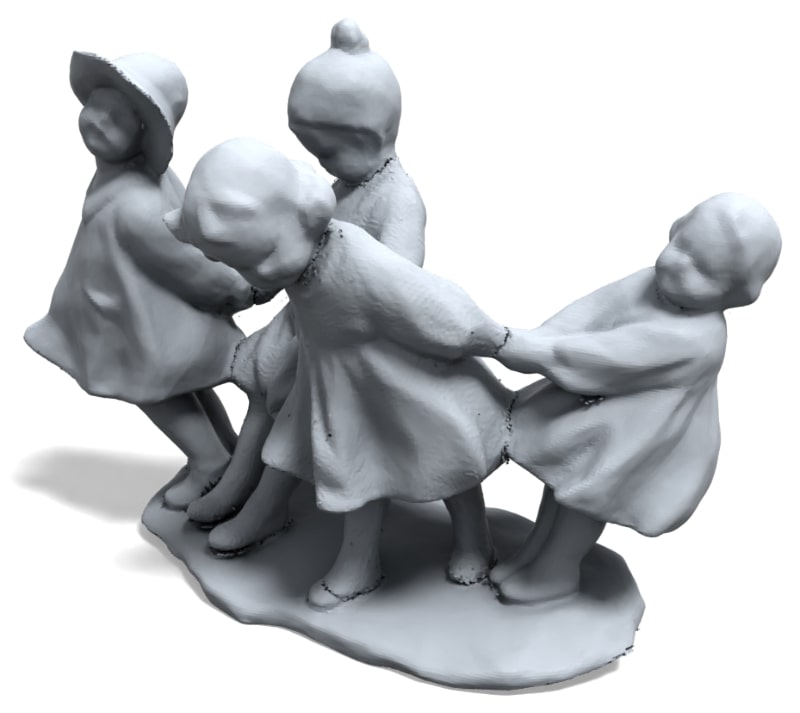} &
    \includegraphics[width=0.8in]{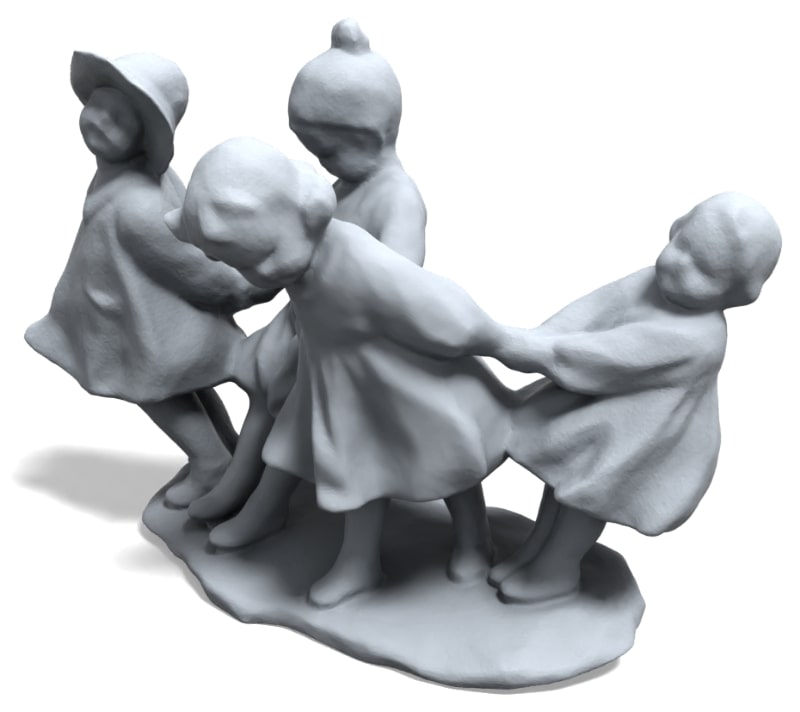} &
    \includegraphics[width=0.8in]{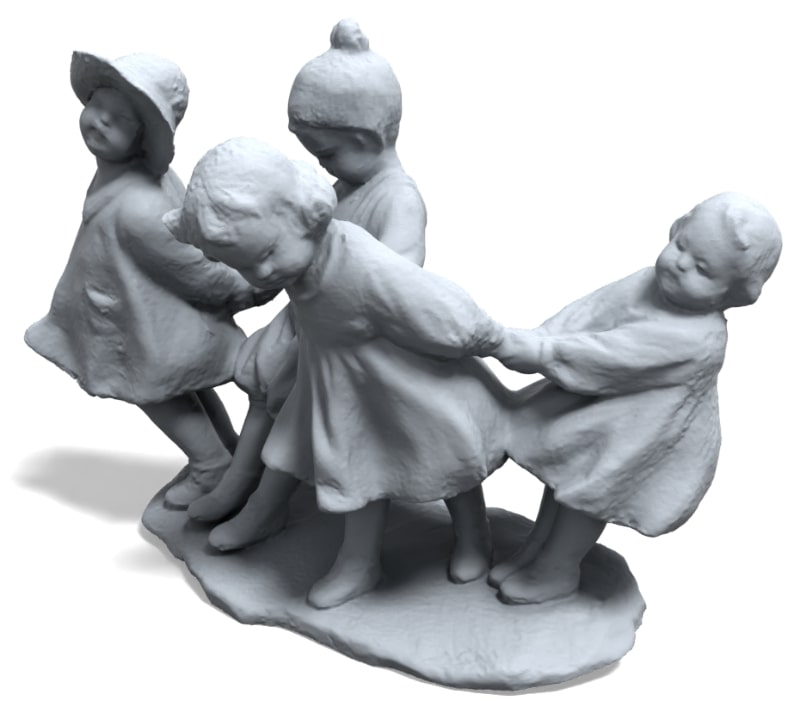} &
    \includegraphics[width=0.7in]{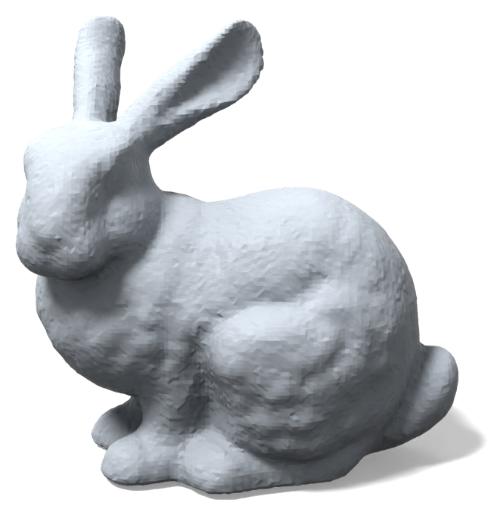} &
    \includegraphics[width=0.7in]{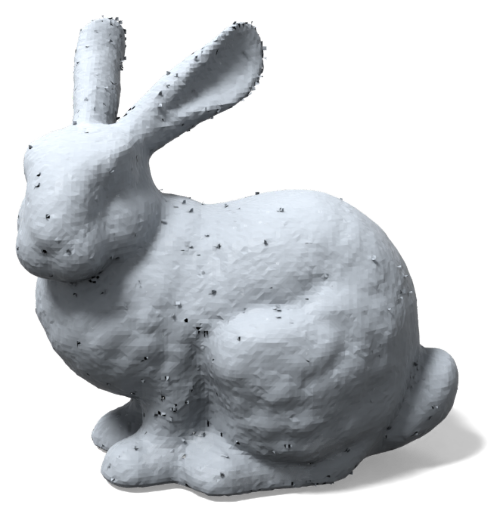} &
    \includegraphics[width=0.7in]{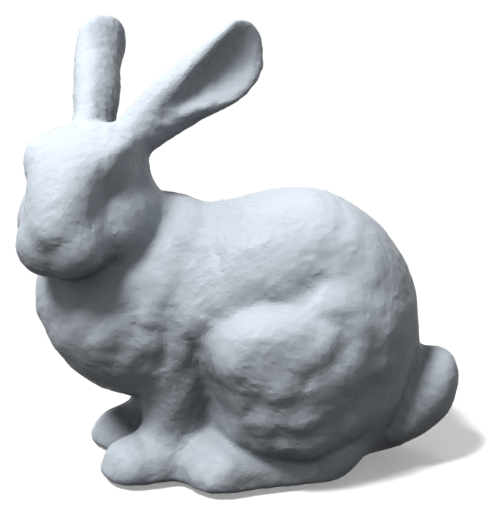} &
    \includegraphics[width=0.7in]{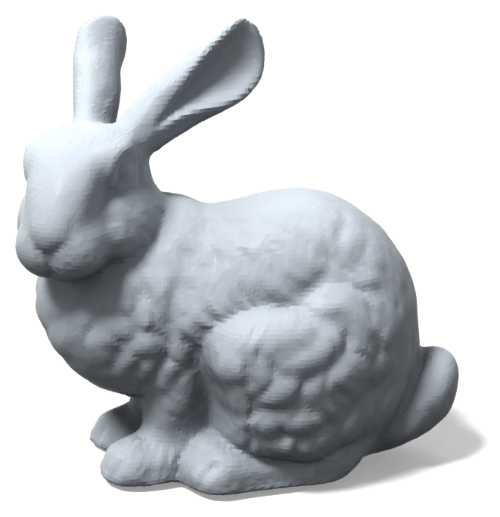}\\
    \includegraphics[width=0.7in]{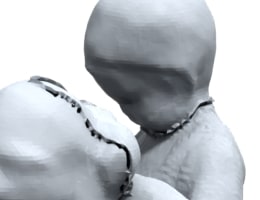 } &
    \includegraphics[width=0.7in]{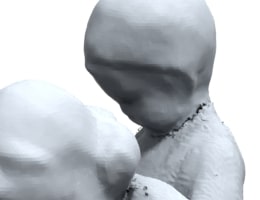} &
    \includegraphics[width=0.7in]{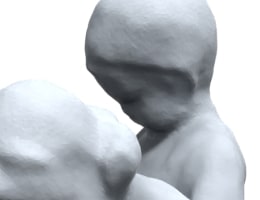} &
    \includegraphics[width=0.7in]{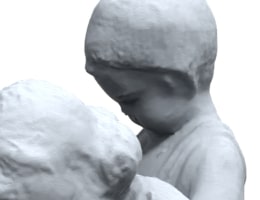} &
    \includegraphics[width=0.8in]{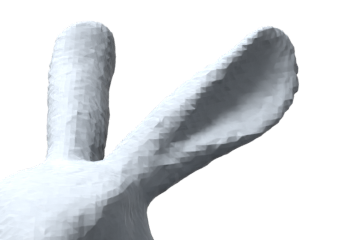} &
    \includegraphics[width=0.8in]{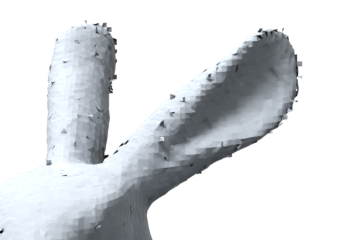} &
    \includegraphics[width=0.8in]{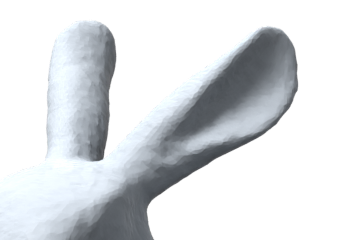} &
    \includegraphics[width=0.8in]{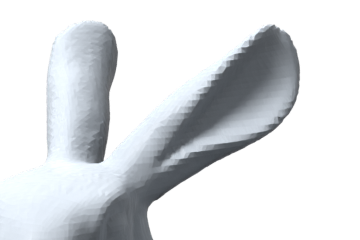}\\
    $\tau_v=0\%$ &  $\tau_v=0.457\%$ &   $\tau_v=0\%$ &   --- &    $\tau_v=0\%$ &  $\tau_v=2.136\%$ &    $\tau_v=0\%$ &    --- \\
    $b=12,g=99$ &    $b=1916,g=194$ &    $b=0,g=8$&    $b=0,g=8$&    $b=4,g=0$&    $b= 7569,g=22$&    $b=0,g=0$&    $b=0,g=0$\\
     CD = 2.279$\times 10^{-3}$ & CD = 2.243$\times 10^{-3}$ & CD = 2.127$\times 10^{-3}$ & --- & CD = 2.302$\times 10^{-3}$ & CD = 2.264$\times 10^{-3}$ & CD = 2.225$\times 10^{-3}$ & ---\\
\end{tabular}

\vspace{0.2in}
    \makebox[3.2in]{CAPUDF}
    \makebox[3.2in]{Simple MLP}\\
\end{small}
    \caption{Experimental results on open and closed surfaces using a resolution of $256^3$. We consider two types of UDFs as input: CAPUDF (left) and Simple MLP (right). For DCUDF, we set the thickness parameter $r$ to $0.005$.}
    \label{fig:models}
\end{figure*}

\begin{figure}[!htbp]
    \centering
    \makebox[1.1in]{MeshUDF}
    \makebox[1.1in]{MeshCAP}
    \makebox[1.1in]{DCUDF}\\
    \includegraphics[width=1.1in]{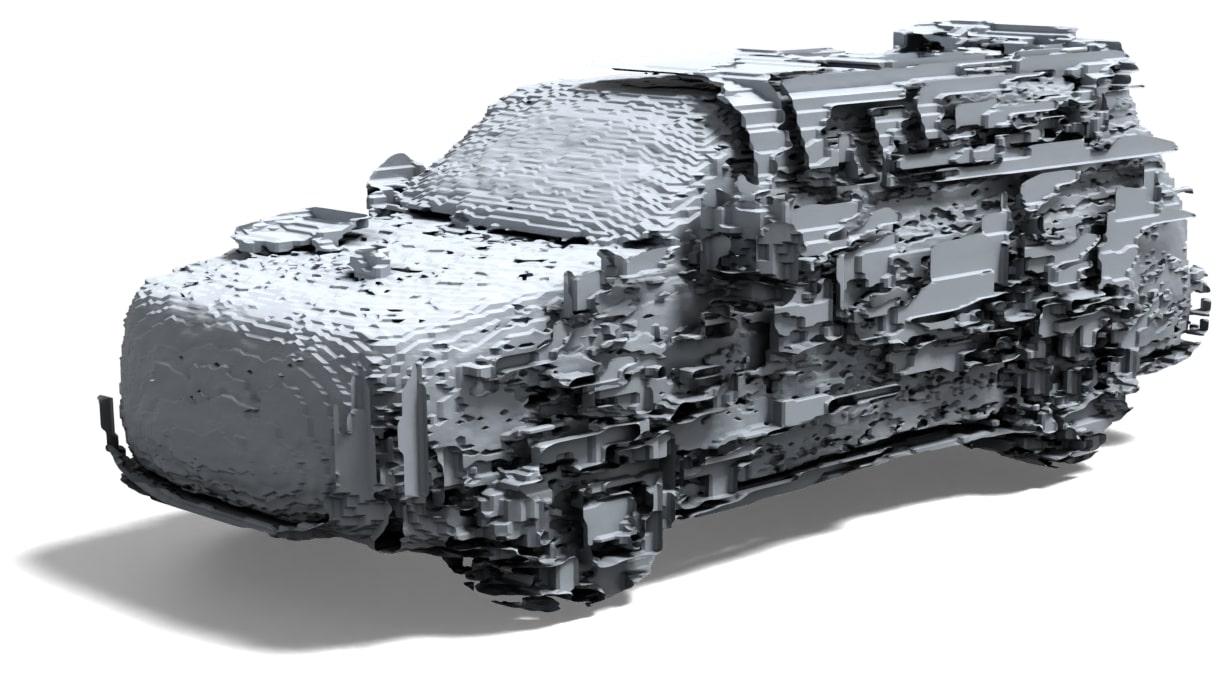 }
    \includegraphics[width=1.1in]{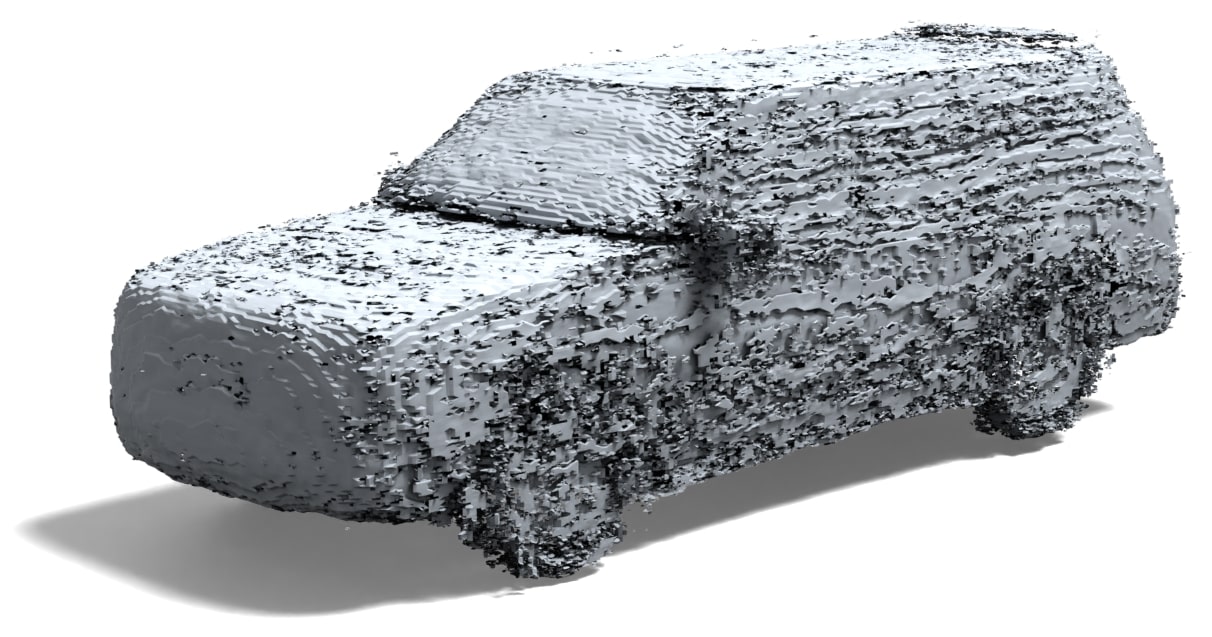}
    \includegraphics[width=1.1in]{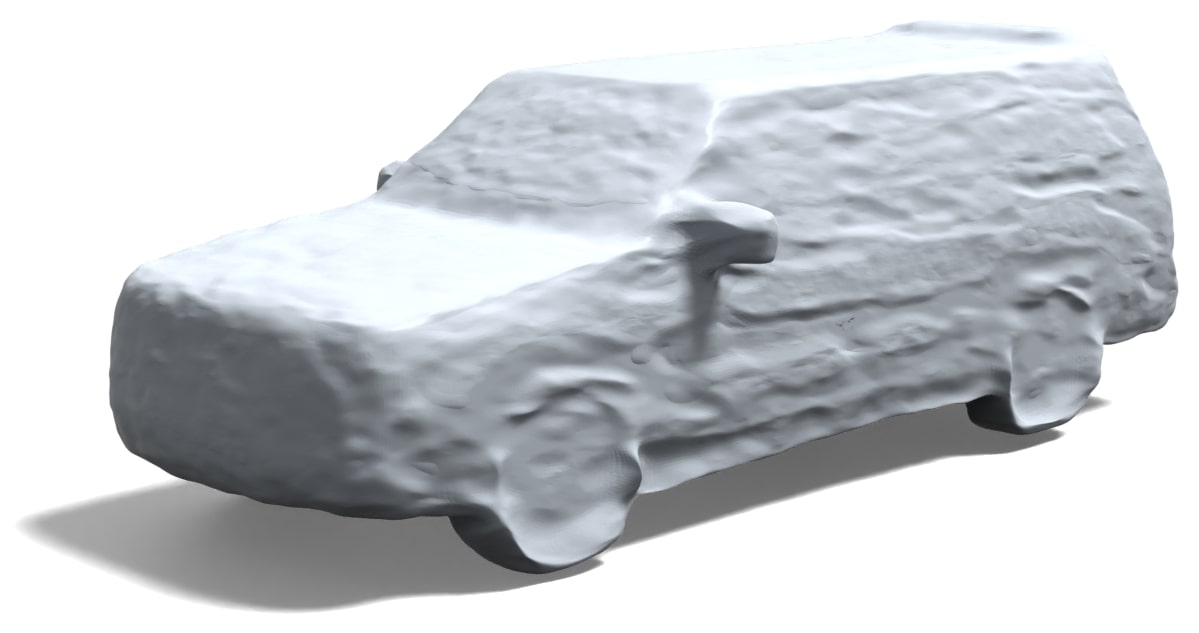}\\
    \makebox[1,1in]{$\tau_v=0.004\%$}
    \makebox[1,1in]{$\tau_v=10.581\%$}
    \makebox[1,1in]{$\tau_v=0\%$}\\
    \makebox[1,1in]{CD = $6.293\times 10^{-3}$}
    \makebox[1,1in]{CD = $4.221\times 10^{-3}$}
    \makebox[1,1in]{CD = $4.102\times 10^{-3}$}\\
    \makebox{(a) NDF}\\
    \includegraphics[width=1.1in]{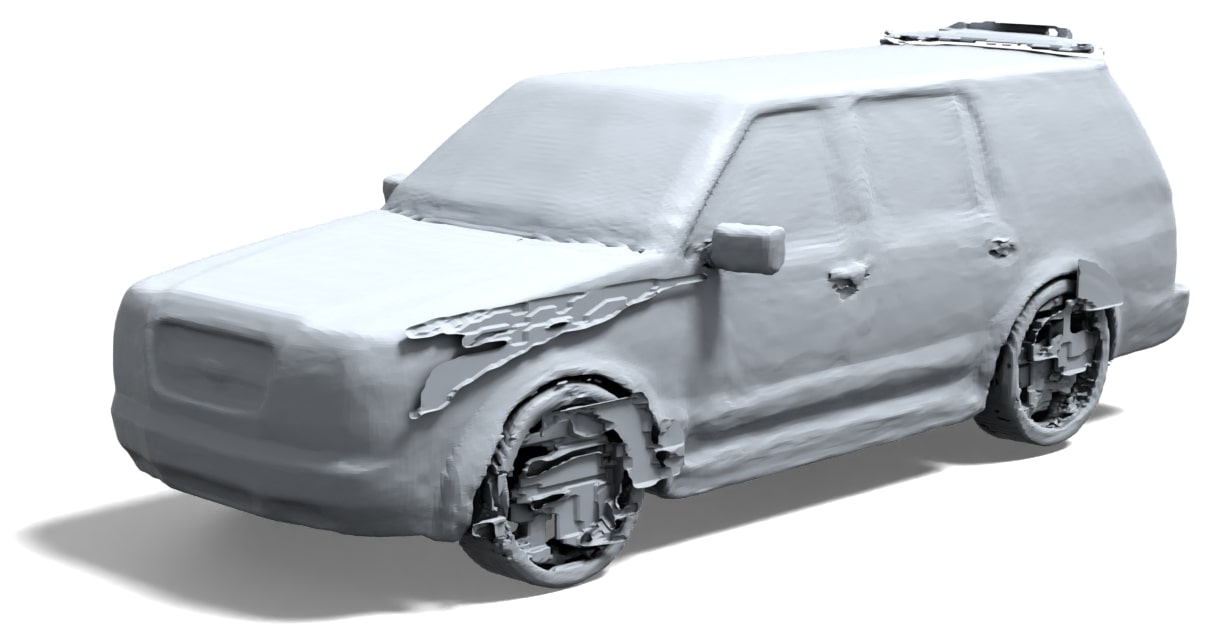}
    \includegraphics[width=1.1in]{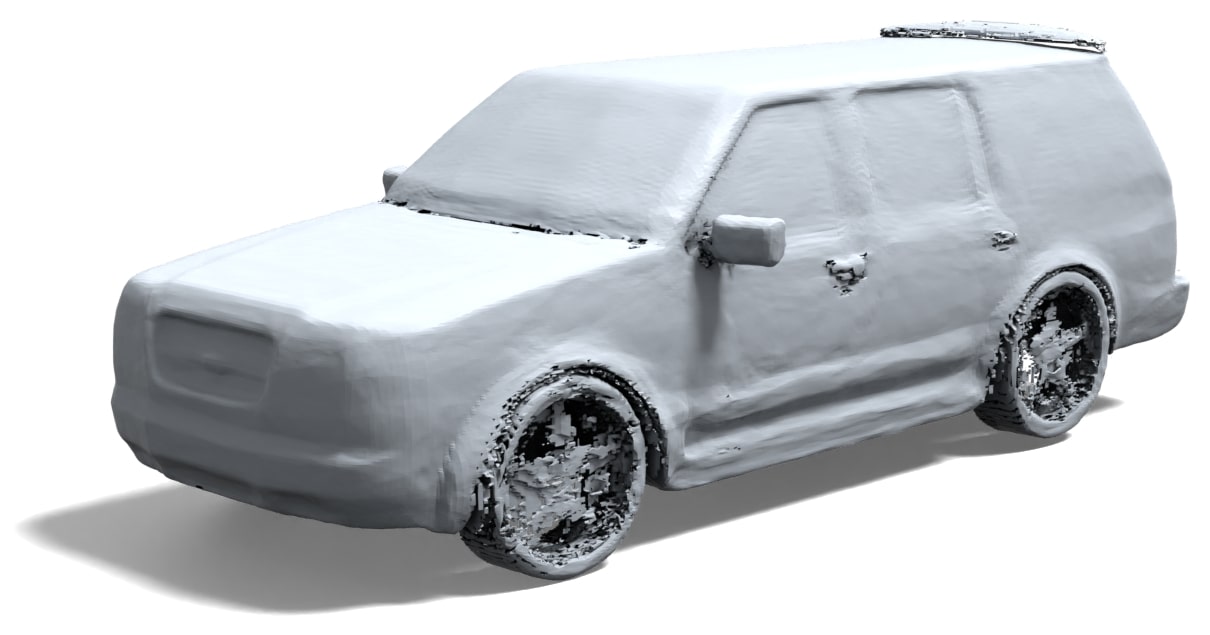}
    \includegraphics[width=1.1in]{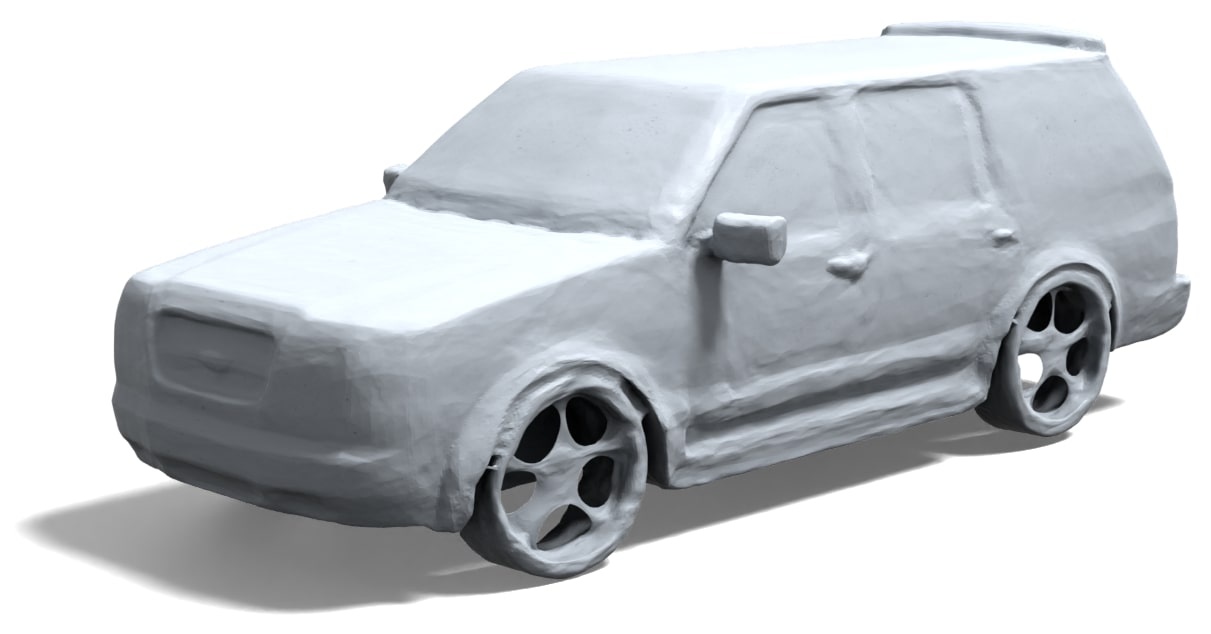}\\
    \makebox[1,1in]{$\tau_v=0.002\%$}
    \makebox[1,1in]{$\tau_v=3.526\%$}
    \makebox[1,1in]{$\tau_v=0\%$}\\
    \makebox[1,1in]{CD = $2.941\times 10^{-3}$}
    \makebox[1,1in]{CD = $2.564\times 10^{-3}$}
    \makebox[1,1in]{CD = $2.550\times 10^{-3}$}\\
    \makebox{(b) CAPUDF}\\
    \includegraphics[width=1.1in]{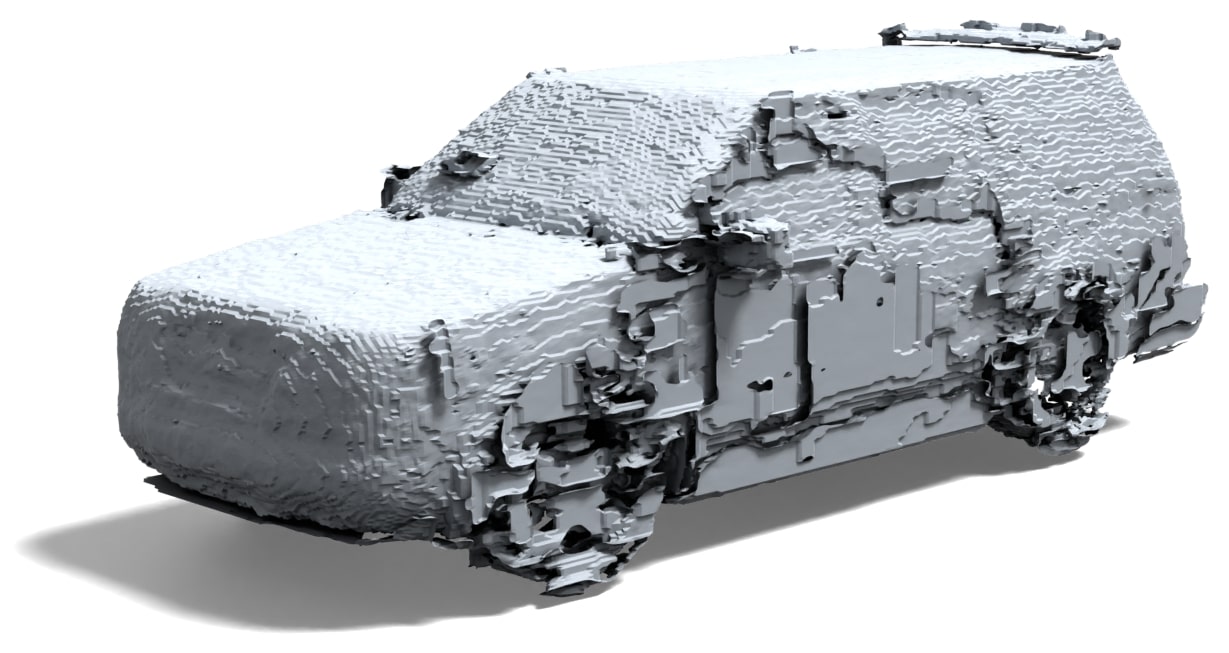}
    \includegraphics[width=1.1in]{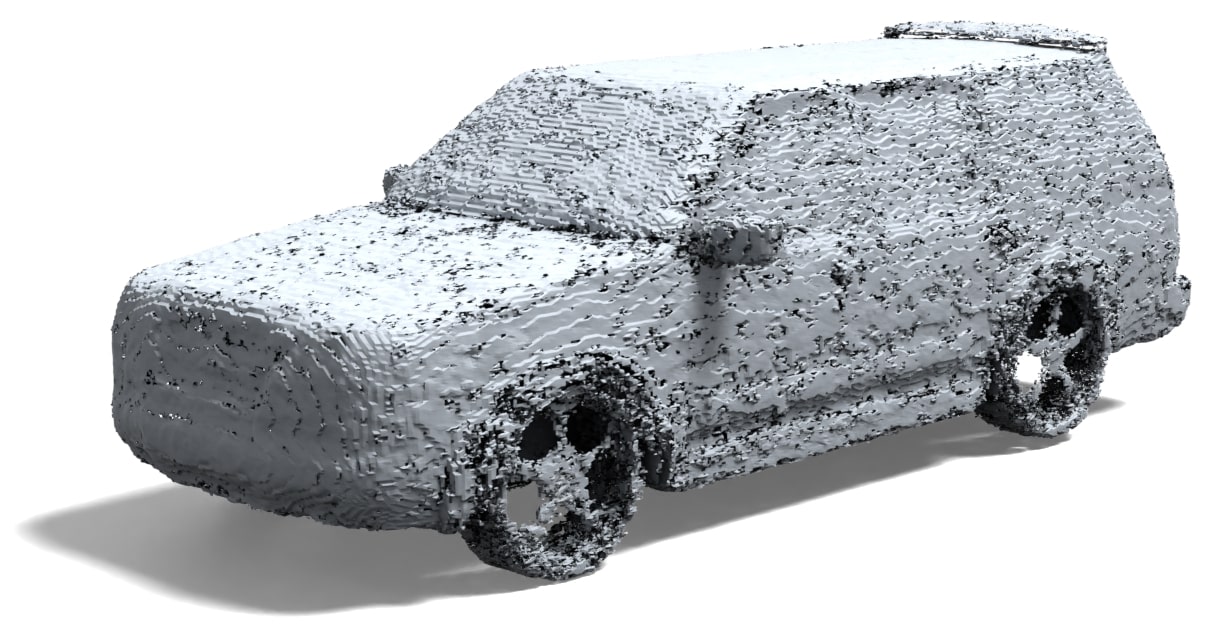}
    \includegraphics[width=1.1in]{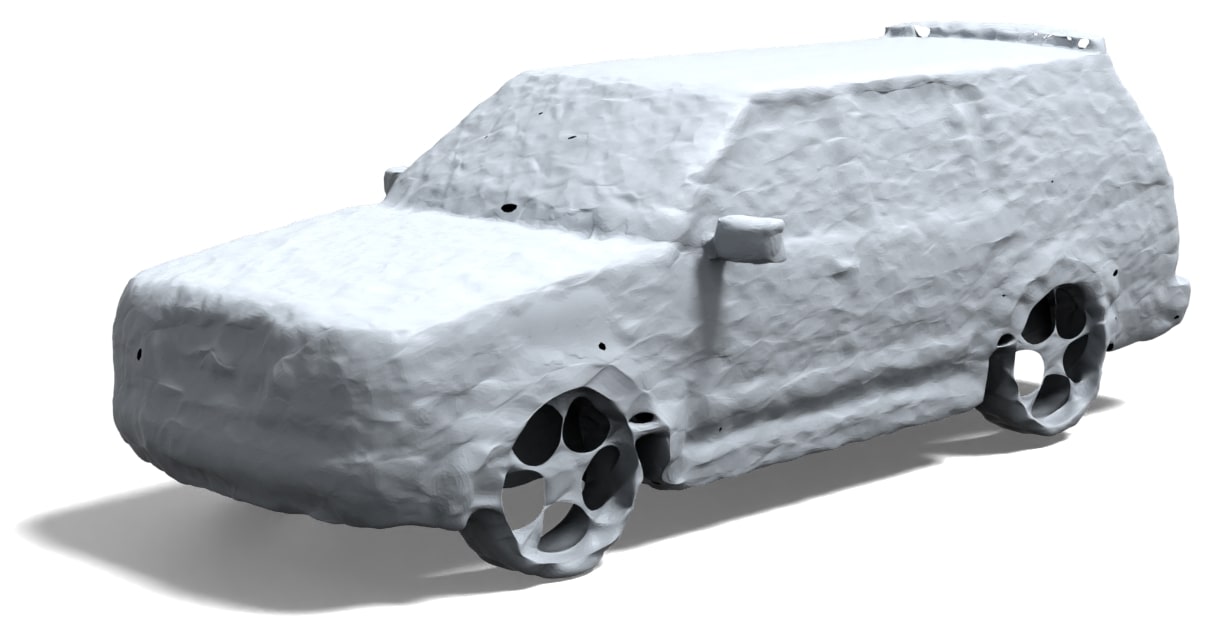}\\
    \makebox[1,1in]{$\tau_v=0.0014\%$}
    \makebox[1,1in]{$\tau_v=12.41\%$}
    \makebox[1,1in]{$\tau_v=0\%$}\\
    \makebox[1,1in]{CD = $3.337\times 10^{-3}$}
    \makebox[1,1in]{CD = $2.638\times 10^{-3}$}
    \makebox[1,1in]{CD = $2.653\times 10^{-3}$}\\
    \makebox{(c) Simple MLP}
    \caption{Comparison of UDFs learned from different methods at a MC resolution of $256^3$. MeshCAP yields a significant number of non-manifold vertices, and MeshUDF exhibits noticeable artifacts in the extracted meshes. In contrast, our method, with a thickness parameter $r=0.0025$ consistently adapts well to all input data and produces visually appealing results without any non-manifold vertices or edges. }
    \label{fig:mix_comparison}
\end{figure}

\paragraph{Input UDFs}
It is worth noting that DCUDF is compatible with any continuous and differentiable UDF representations that are capable of providing gradient calculations for the optimizations in Equations~(\ref{eqn:step1}) and (\ref{eqn:step2}). In this paper, we generate UDFs from unoriented points using two state-of-the-art methods: NDF~\cite{Chibane2020NDF} and CAPUDF~\cite{Zhou2022}, where the UDFs are encoded in a 3D CNN and an MLP, respectively. Additionally, we demonstrate that DoubleCoverUDF is not limited to any particular UDF learning method by employing a simple approach. Given a set of sample points $\mathcal{P}$ on or close to the target surface, we compute the unsigned distance $\mathrm{UDF}(x)=\|p-x\|$ for each query $x\in\mathbb{R}^3$, where $p\in\mathcal{P}$ is the closest sample point. We then train an MLP $f$ by minimizing 
$\min_{\theta}\sum_{x\in\mathcal{P}} \left\|f(x;\theta)-\mathrm{UDF}(x)\right\|^2$,
where $\theta$ is the MLP weights. Following~\cite{Zhou2022}, the MLP consists of 8 layers, with $256^2$ neurons in each layer, and we use an $8$-dimensional positional encoding~\cite{Mildenhall2020} to enable the network to learn high-frequency details. Due to its straightforward design, we refer to this learning method as ``Simple MLP''.

Note that there are alternative methods for computing unsigned distance fields. For instance, GeoUDF employs a more nuanced approach by considering not only the closest sample point to a query point but also its surrounding neighbors~\cite{Ren2023}. This method then computes the distance as a weighted sum of the shortest distances from these various sample points, with the weights learned by the network. Generally speaking, GeoUDF offers greater accuracy than our Simple MLP learning method. However, this comes at the cost of increased complexity in gradient calculations. Since our focus is not primarily on UDF learning, we opt to use NDF, CAPUDF and Simple MLP to learn UDF from input point clouds.

\paragraph{Comparison with MeshUDF and MeshCAP}
MeshUDF~\cite{Guillard2022} and MeshCAP~\cite{Zhou2022} are two state-of-the-art methods for extracting the zero level sets from UDFs. They both deviate from the standard marching cubes algorithm by utilizing gradient directions instead of vertex signs to determine cube intersections. As marching cubes variants, they maintain high computational efficiency and guarantee that the extracted meshes free of self-intersections. However, these modifications introduce undefined cube configurations. While there are $2^8$ configurations of vertex signs, there are $2^{12}$ edge intersection configurations. Consequently, not every edge intersection configuration corresponds to a valid vertex sign configuration. To address the artifacts resulting from these undefined configurations, MeshUDF incorporates post-processing strategies to smooth boundary curves and eliminate certain redundant faces. However, as a consequence of this process, small holes may appear in the final mesh.

The reliance on gradient directions in MeshCAP and MeshUDF make them sensitive to the accuracy of the UDF learning process, as even a small perturbation can lead to flipped directions, resulting in inappropriate mesh connections and incorrect topology.
Figure~\ref{fig:models} illustrates the extracted meshes of open and closed models on UDFs learned by CAPUDF and Simple MLP. MeshUDF and MeshCAP generate lots of boundaries and handles. In contrast, the number of boundaries and handles of DoubleCoverUDF are much closer to the ground truth.
In Figure~\ref{fig:mix_comparison}, we assess the sensitivity of the iso-surface extraction algorithm to the learned UDF inputs on a car model, which is a clean model but exhibits complex geometry and includes non-manifold structures. MeshCAP produces a large number of boundaries and non-manifold vertices in the extracted meshes. Although MeshUDF significantly reduces the number of non-manifold vertices, it tends to introduce redundant structures.
In contrast, our DoubleCoverUDF consistently produces satisfactory double-layered mesh results, highlighting its versatility and effectiveness across different UDF learning methods.

\begin{figure}[!htbp]
    \centering
    \makebox[0.8in]{MeshUDF}
    \makebox[0.8in]{MeshCAP}
    \makebox[0.8in]{DCUDF}
    \makebox[0.8in]{Reference}\\
    \includegraphics[width=0.8in]{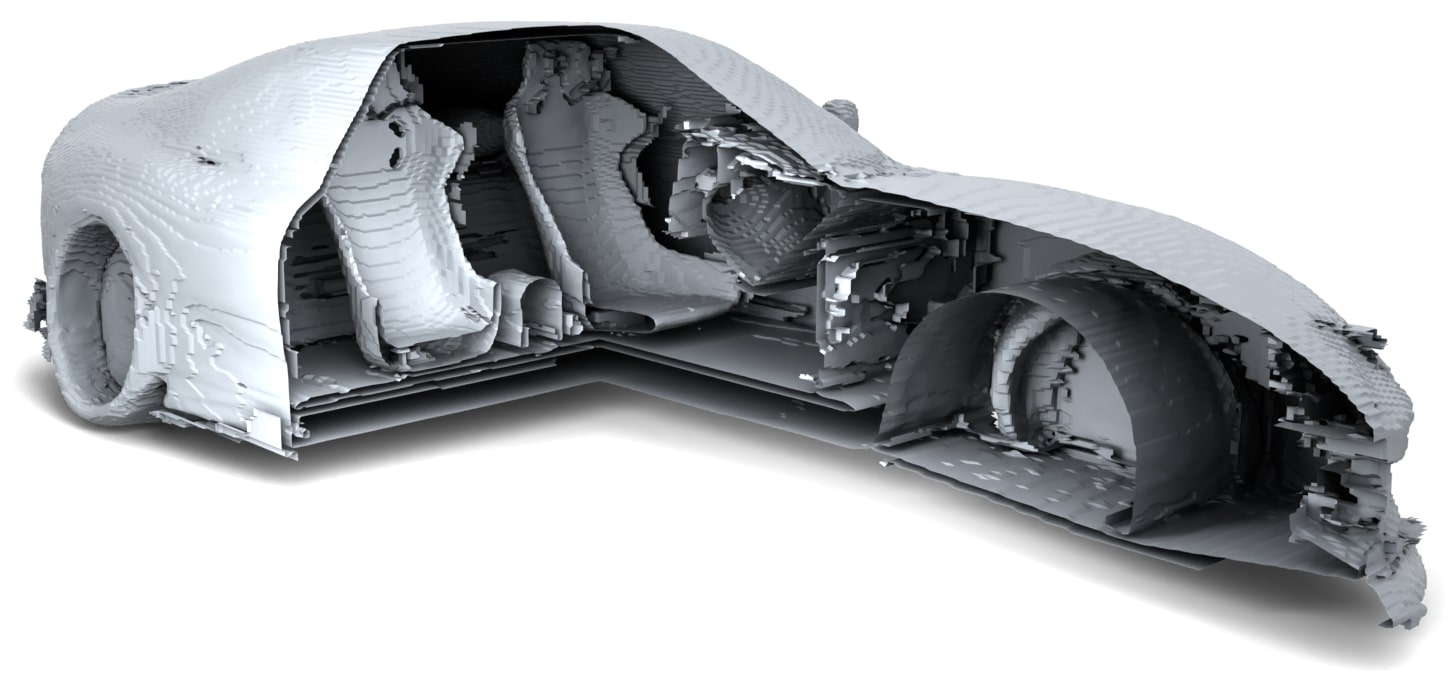 }
    \includegraphics[width=0.8in]{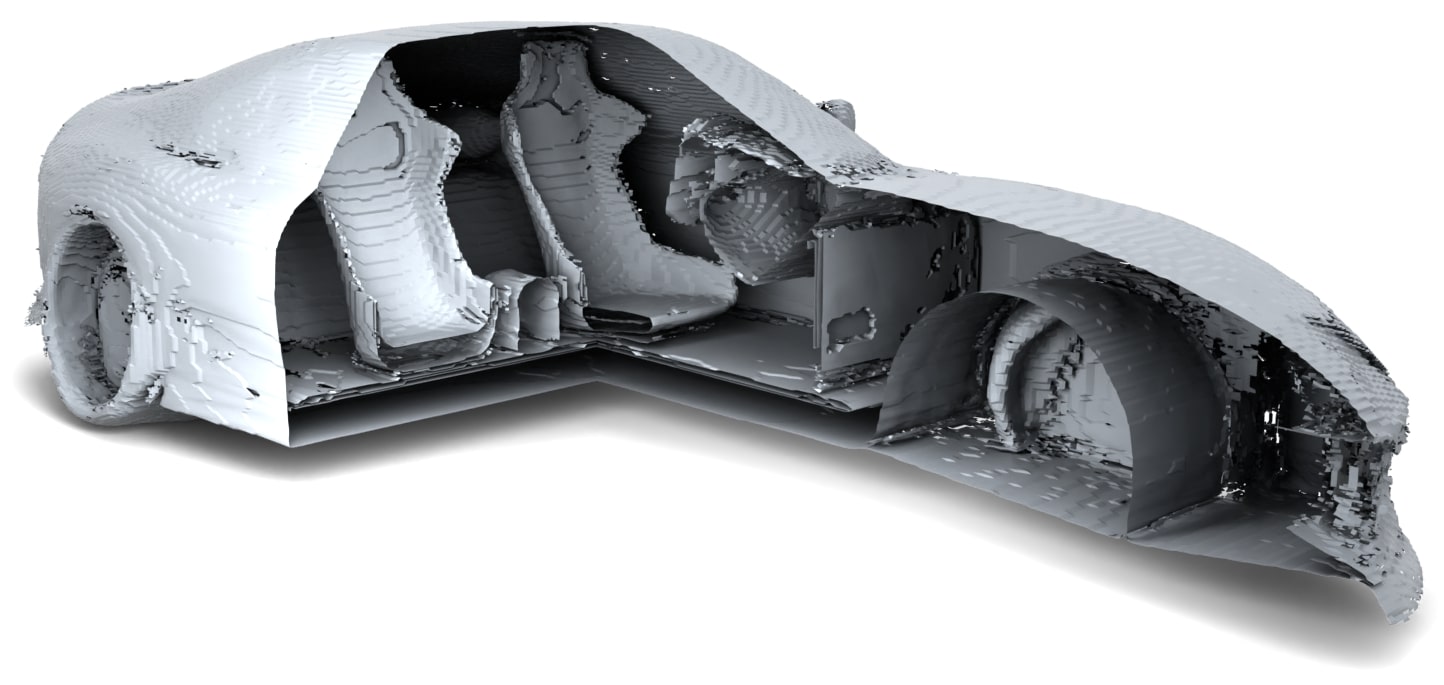 }
    \includegraphics[width=0.8in]{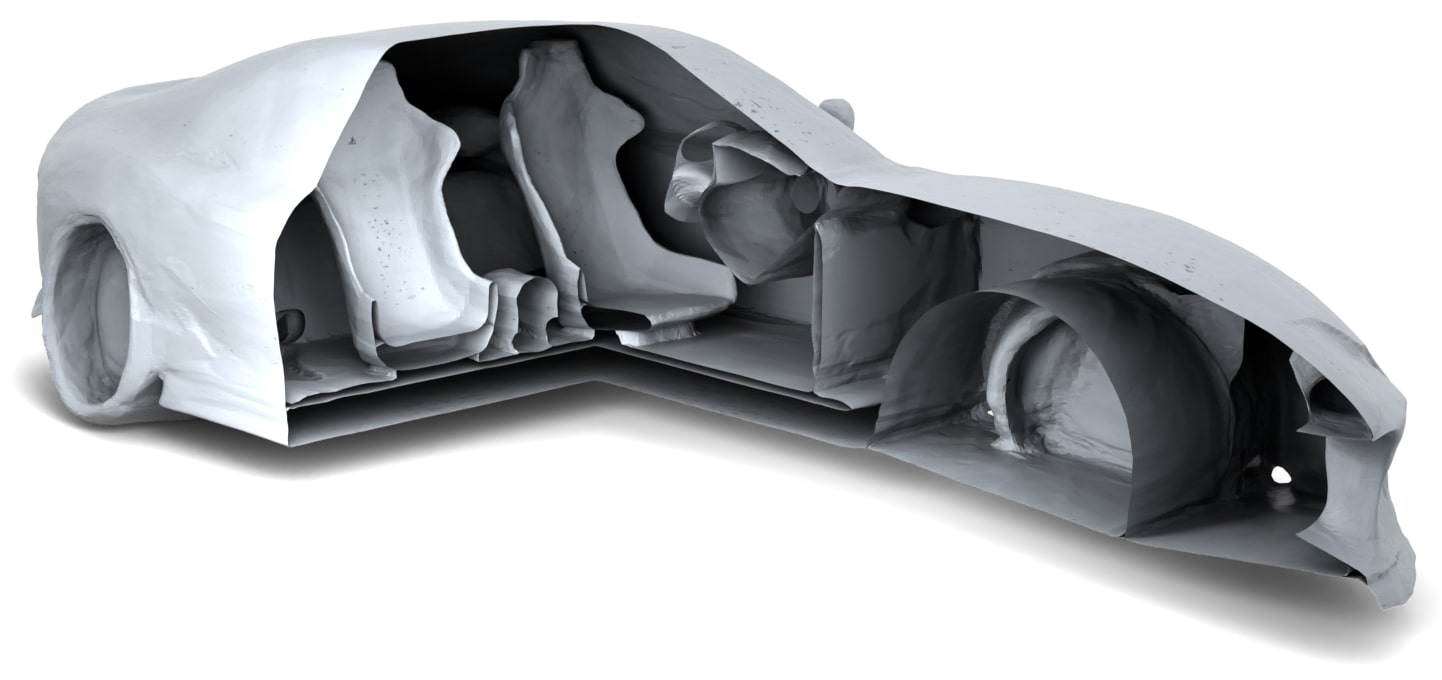 }
    \includegraphics[width=0.8in]{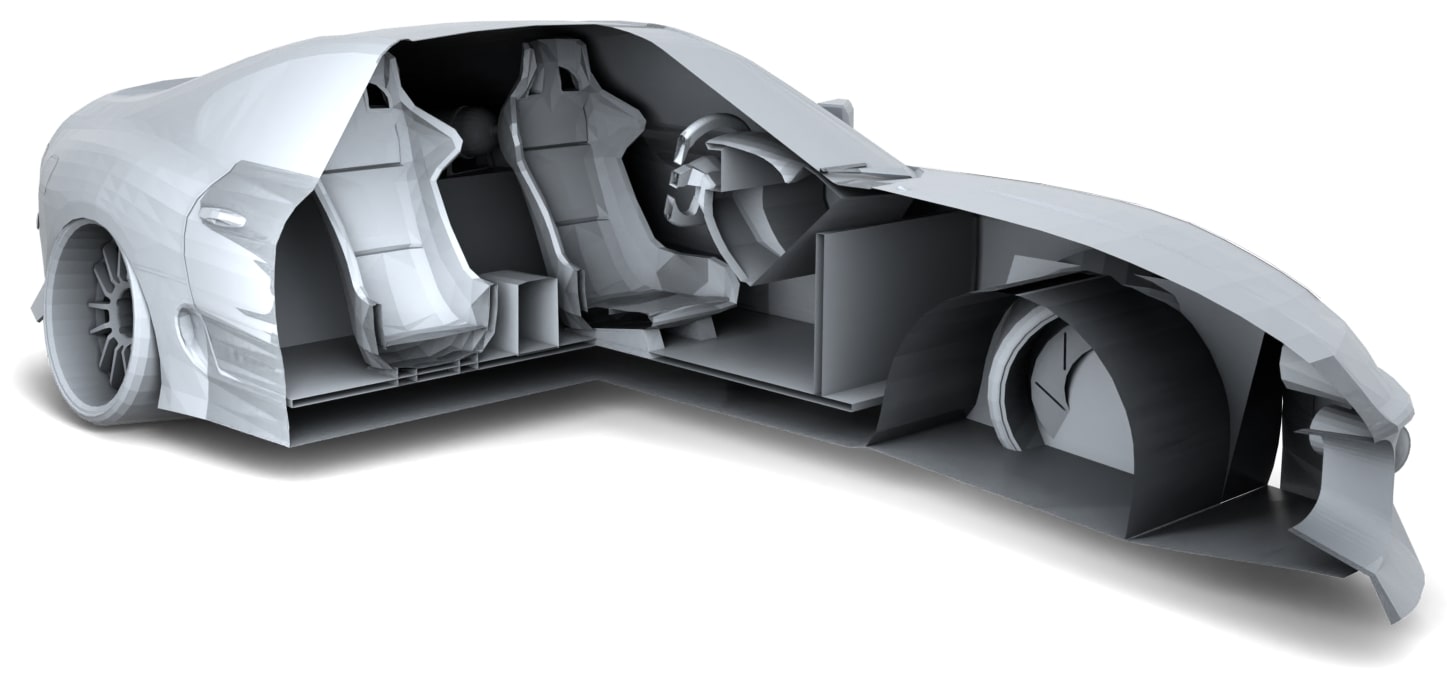 }\\
    \includegraphics[width=0.8in]{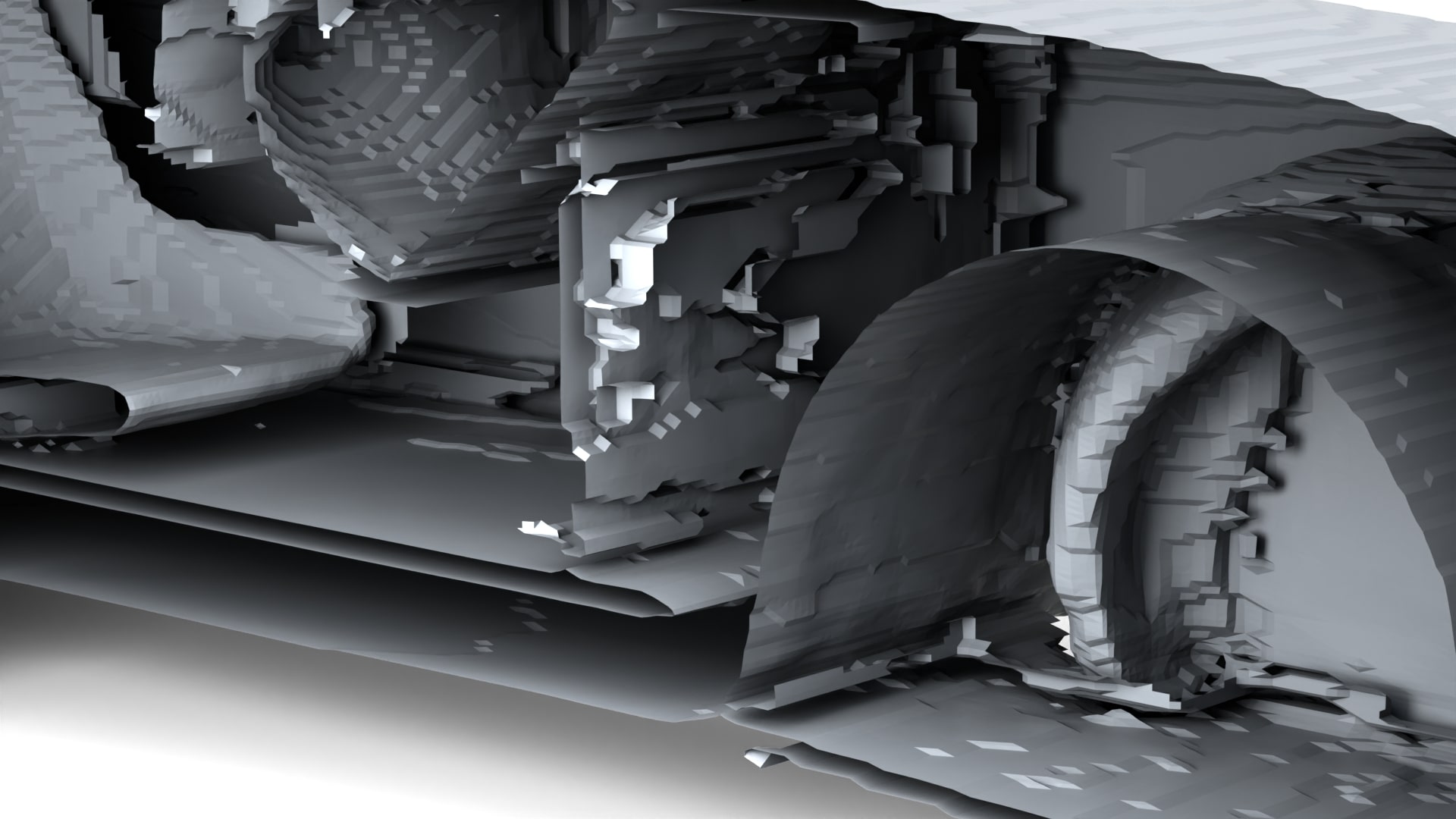 }
    \includegraphics[width=0.8in]{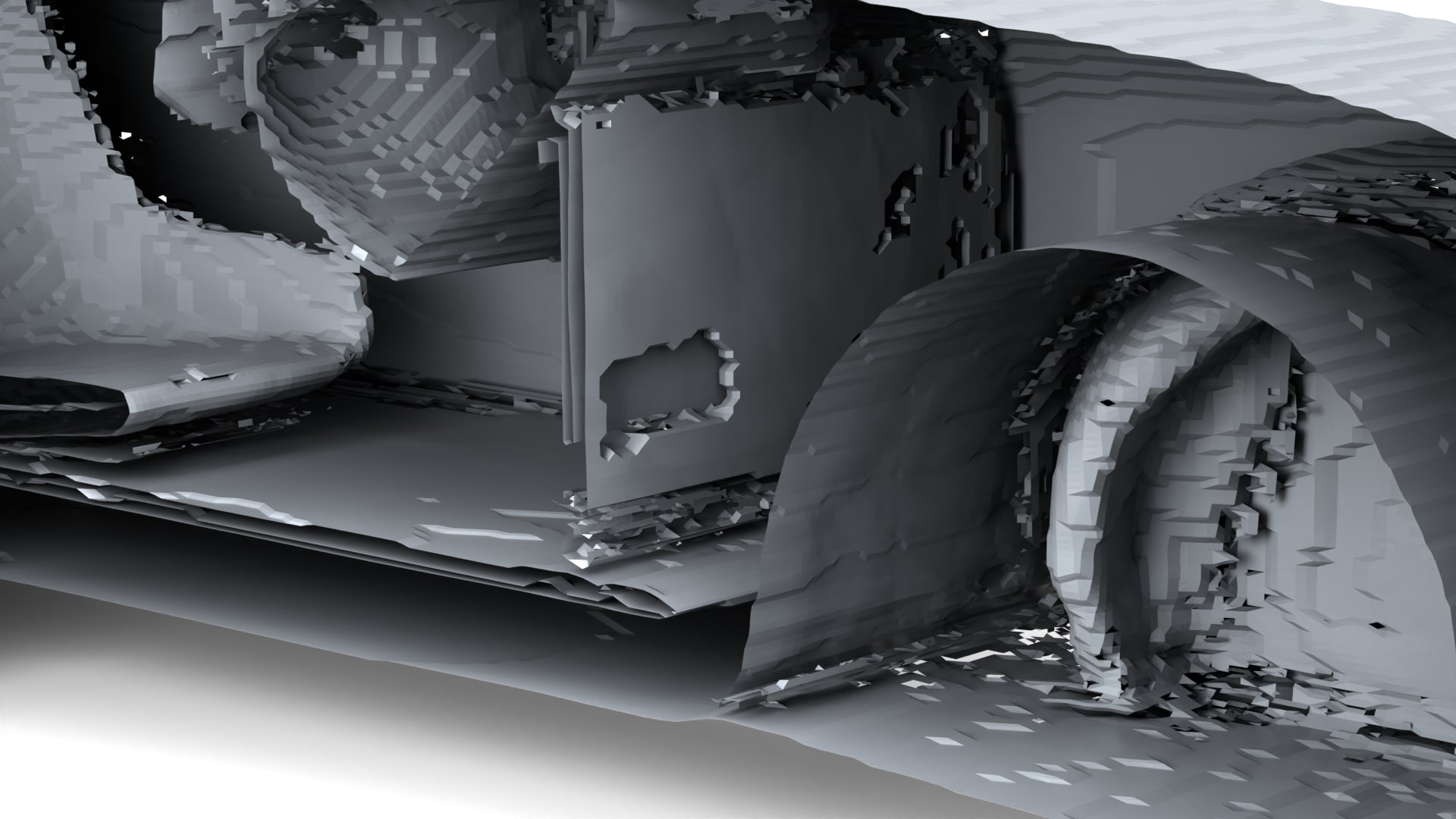 }
    \includegraphics[width=0.8in]{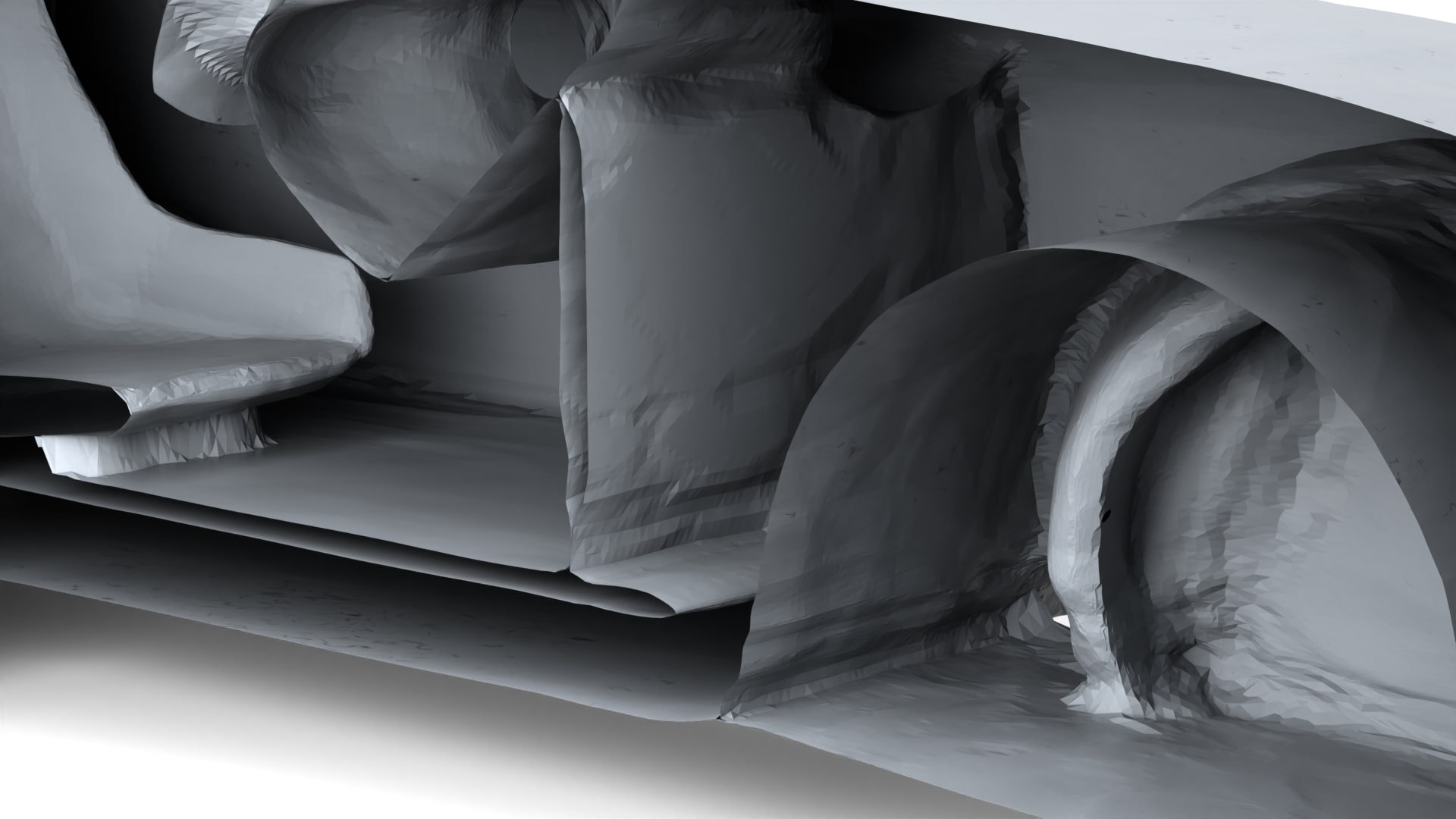 }
    \includegraphics[width=0.8in]{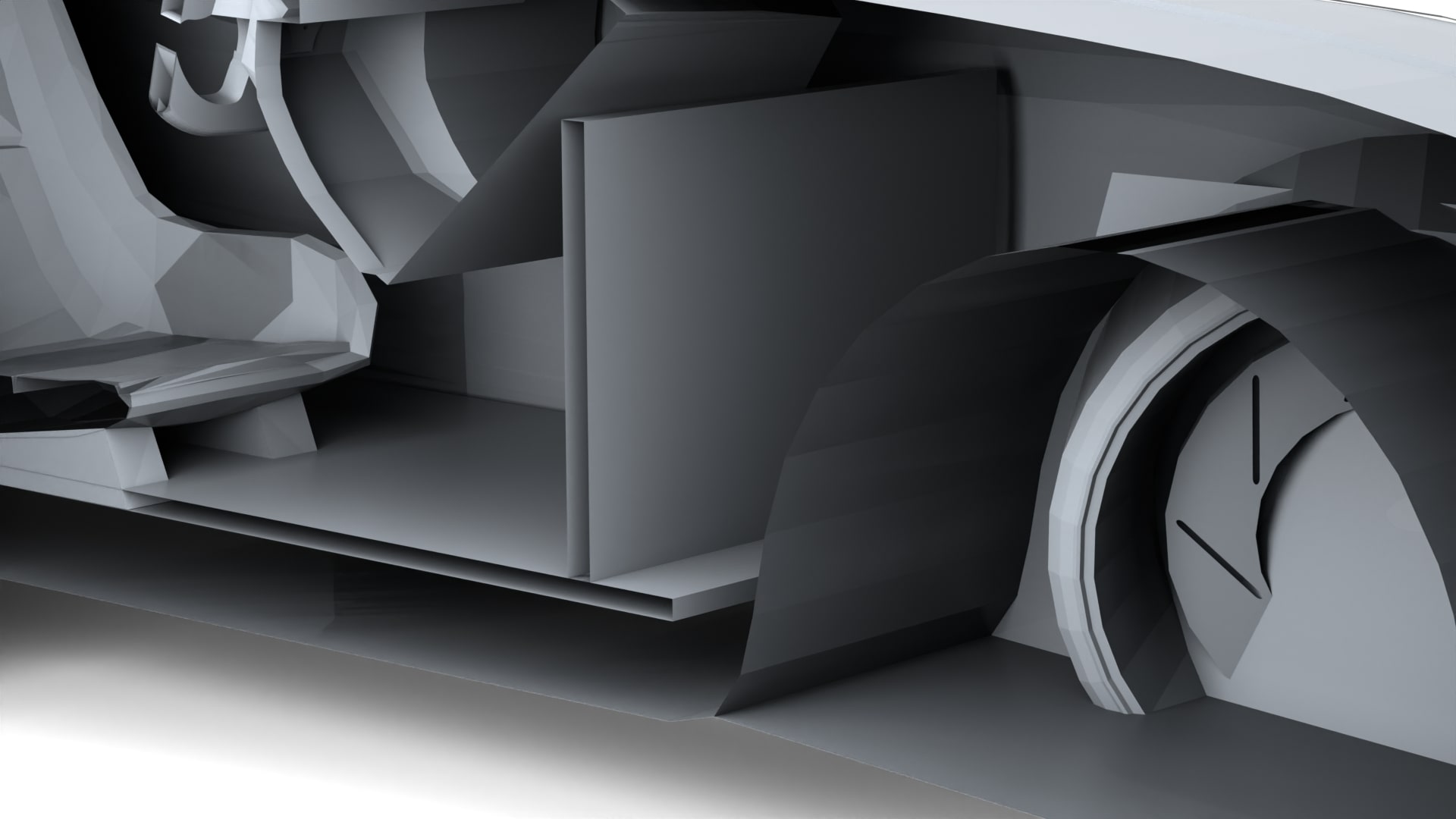 }\\
    \makebox[0.8in]{$\tau_v=0.0011\%$}
    \makebox[0.8in]{$\tau_v=2.77\%$}
    \makebox[0.8in]{$\tau_v=0\%$}
    \makebox[0.8in]{$\tau_v=0.11\%$}\\
    \makebox[0.8in]{$\beta_0=297$}
    \makebox[0.8in]{$\beta_0=15184$}
    \makebox[0.8in]{$\beta_0=36$}
    \makebox[0.8in]{$\beta_0=86$}
    \includegraphics[width=0.8in]{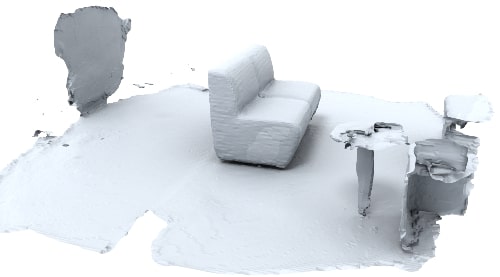 }
    \includegraphics[width=0.8in]{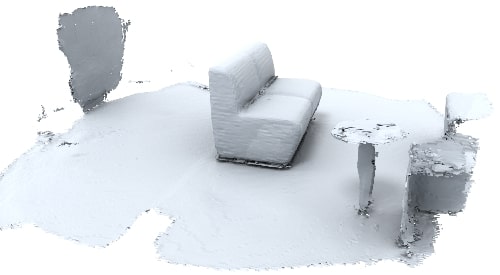 }
    \includegraphics[width=0.8in]{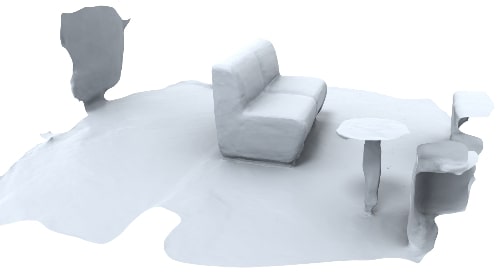 }
    \includegraphics[width=0.8in]{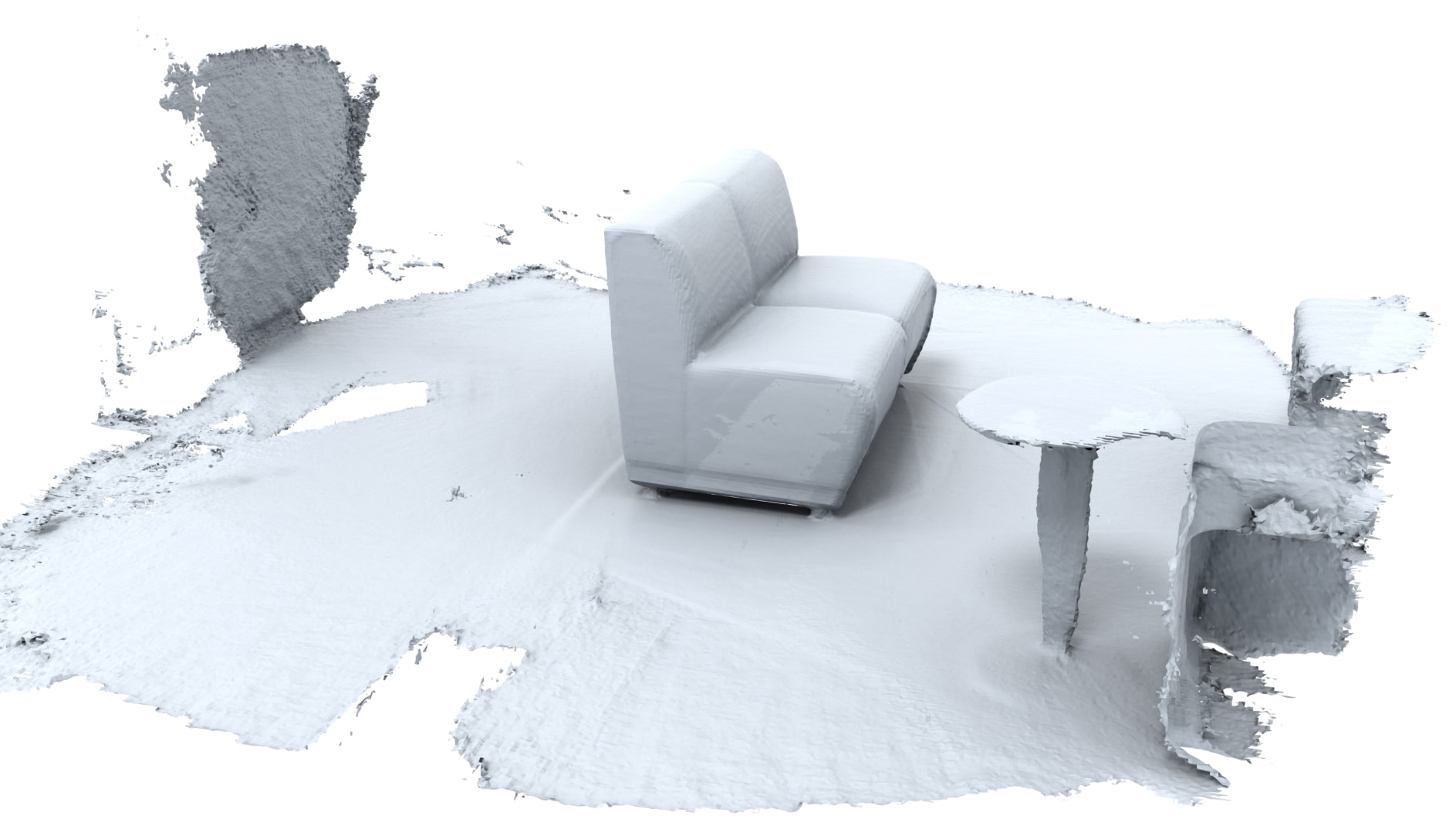}\\
    \includegraphics[width=0.8in]{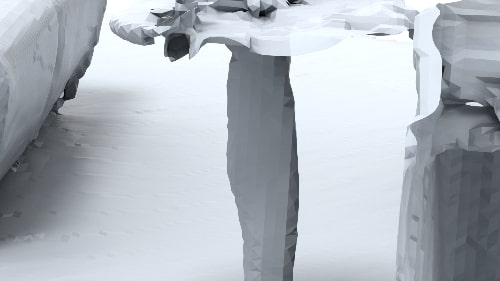 }
    \includegraphics[width=0.8in]{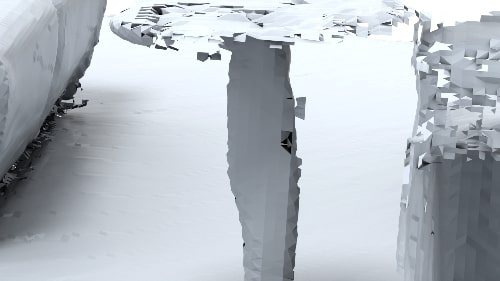 }
    \includegraphics[width=0.8in]{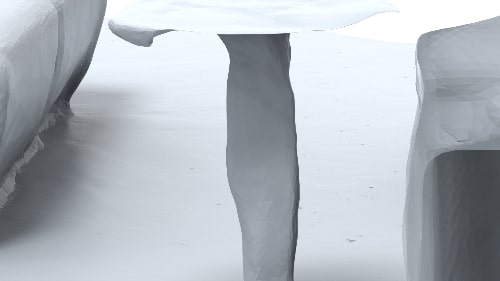 }
    \includegraphics[width=0.8in]{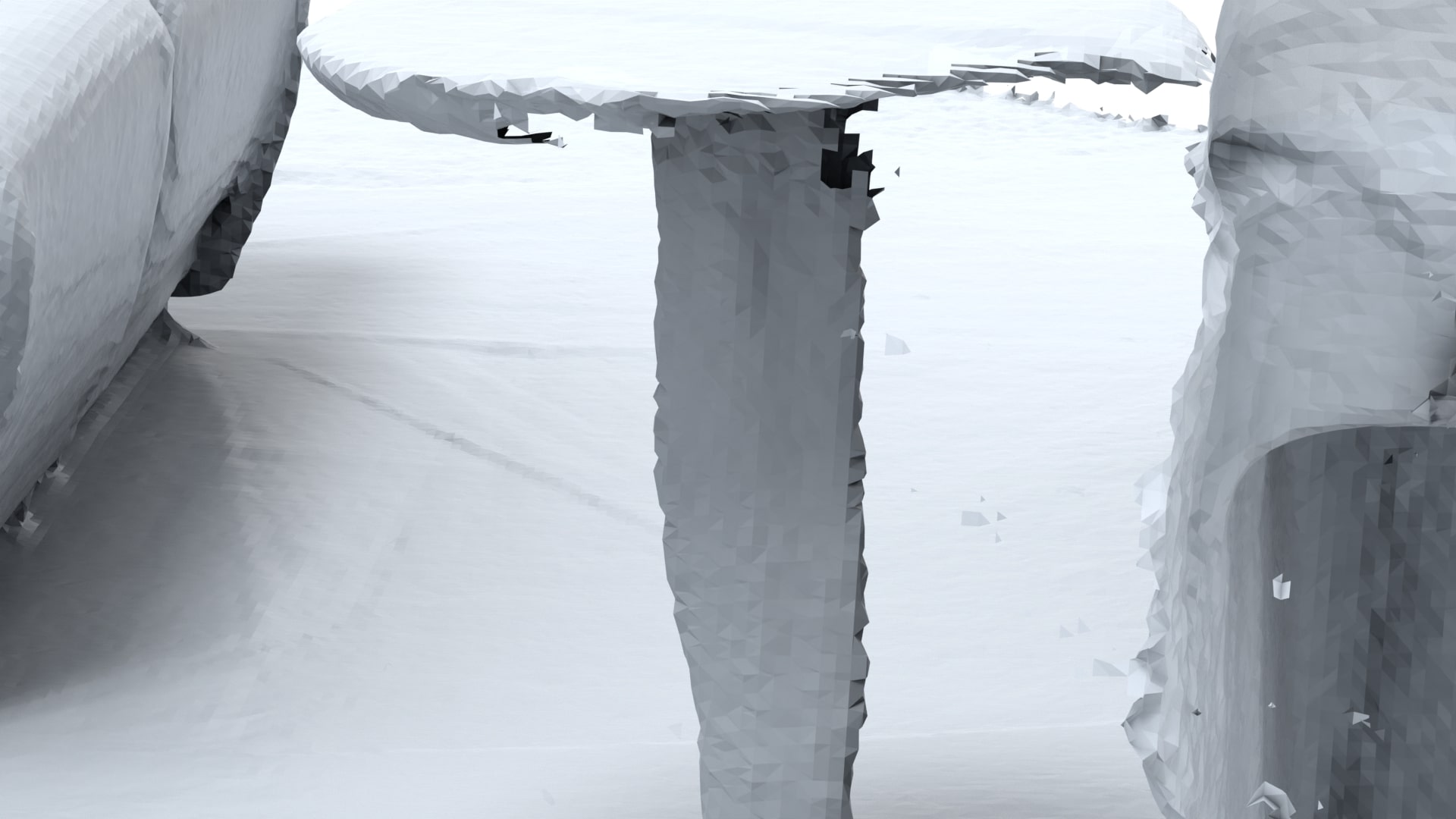}\\
    \makebox[0.8in]{$\tau_v=0\%$}
    \makebox[0.8in]{$\tau_v=1.93\%$}
    \makebox[0.8in]{$\tau_v=0\%$}
    \makebox[0.8in]{$\tau_v=0.0310\%$}\\
    \makebox[0.8in]{$\beta_0=43$}
    \makebox[0.8in]{$\beta_0=2112$}
    \makebox[0.8in]{$\beta_0=22$}
    \makebox[0.8in]{$\beta_0=2004$}\\
    \caption{Complex examples. Top: models with inner structures. Bottom: scene data. Our results are visually pleasing, free of non-manifold vertices and smaller number of connected components.
    We set the marching cubes resolution of $256^{3}$ for all methods and the thickness parameter $r=0.0025$ for DCUDF. The bottom scene data and the reference model are from~\cite{Choi2016}.}
    \label{fig:inner_structures}
\end{figure}

\paragraph{Complex examples}
Figure~\ref{fig:inner_structures} (top) illustrates the cutaway view of a car model from the ShapeNet-Car dataset~\cite{Chang2015}. The model has complicated inner structures and many non-manifold edges. Our method effectively preserves the interior structures and avoids generating non-manifold vertices in the extracted double-layered mesh. In contrast, MeshUDF and MeshCAP struggle to preserve the interior structures as effectively as our method. We quantify the completeness of the model by counting the number of connected components $\beta_0$, and our method achieves a satisfactory level of completeness. In contrast, both MeshUDF and MeshCAP produce broken meshes with an excessive number of connected components due to the presence of non-manifold edges of the input.
Additionally, when dealing with scene data, which often suffer from issues such as uneven sampling, missing parts and noise, our method produces smoother and visually pleasing result of double-layered mesh, as demonstrated in Figure~\ref{fig:inner_structures} (bottom), outperforming MeshCAP and MeshUDF in terms of quality. 

\begin{table}[!htbp]
\caption{\label{tab:df3d}Quantitative comparison on the Deep Fashion3D dataset with 598 models. The UDFs are learned by our Simple MLP. The marching cubes resolution is set to $256^3$ for all methods. Due to the lack of the ground-truth meshes, the Chamfer distances were measured using the input point clouds.}
\begin{small}
\begin{tabular}{l|c|c|c}
\hline
                      & MeshCAP         & MeshUDF &Ours   \\
                      \hline
                      \hline
    CD (input$\rightarrow$pred, $10^{-3}$)    &   2.016       &    2.161   &    2.250  \\
   CD (pred$\rightarrow$input, $10^{-3}$)    &   1.525       &    1.931   &   1.401 \\
  CD (average, $10^{-3}$)                    &   1.770       &    2.046   &    1.825  \\
   avg. non-manifold vertices                 &      18548.610 &  1.459    &  0\\
    avg. non-manifold edges                  &       0.168   &   0.157    &  0 \\
    avg. genus                              &      69.848     &   93.361   &  3.334  \\
    avg. boundaries                           &       18428.280   &   104.732    &  15.895 \\
\hline
\end{tabular}
\end{small}
\end{table}

\paragraph{Evaluation on Deep Fashion3D}
The Deep Fashion3D dataset~\cite{Zhu2020} contains point cloud models of clothes. There are 598 garments, each with multiple poses. The point clouds are generated using computer vision algorithms from the multi-view real images. We observed that the point clouds are not smooth and often have small missing parts.
In our evaluation, we used the first poses of all the 598 garments as a benchmark.
Figures~\ref{fig:teaser},~\ref{fig:MC_threshold} (garment) and~\ref{fig:models} (row 1) showcase a few typical reconstruction examples and Table~\ref{tab:df3d} reports the statistics of geometry and topology quality measures.
For a fair comparison, we retain only the largest connected component as the final result for all the Deep Fashion3D models.
Our comprehensive evaluation demonstrates that DoubleCoverUDF outperforms MeshUDF and MeshCAP in terms of several key aspects.
Firstly, it achieves similar CD errors to MeshCAP and much smaller than MeshUDF, indicating superior accuracy in reconstructing the target surfaces.
Secondly, our extracted meshes are free of non-manifold vertices and non-manifold edges, distinguishing it from MeshUDF and MeshCAP which tend to produce such artifacts. Thirdly, DoubleCoverUDF excels in accurately capturing topological features, including number of boundaries and genus \footnote{
Since the models of the Deep Fashion3D dataset consists of raw point clouds, the ground truth values of average genus and average boundaries are not available. However, it is reasonable to expect that the clothes models would generally exhibit a low genus and a minimal number of boundaries, based on common sense.}, while 
MeshCAP and MeshUDF often generate an excessive number of unnecessary boundaries and handles, compromising the quality and the integrity of the reconstructed meshes. These findings highlight the robustness and high-quality output of DoubleCoverUDF compared to its counterparts.

\begin{table}[!htbp]
\caption{\label{tab:accurate_udf}Quantitative comparison with UNDC, MeshUDF, MeshCAP and the standard Marching Cubes on accurate UDFs, which are converted from SDFs using ground truth meshes. $k$ is the resolution of marching cubes and the Chamfer distances are measured in the units of $10^{-3}$. We set the thickness parameter $r$ of DCUDF to $0.0025$ for all resolutions.}
\setlength\tabcolsep{2pt}
\begin{small}
\begin{tabular}{c|c|c|c|c|c|c|c|c}
\hline
                     Model &  Method      & $k$ & CD & $\tau_e$  &$\tau_v$ & $g$ & $b$ & $\beta_0$ \\
                     \hline
                      \hline
\multirow{11}{*}{\makecell[c]{Dragon\\$|V|=3.6$m\\ $g=1$\\$b=0$\\$\beta_0$=1}}
                        & MC  & $256^3$ & 0.571 & 0 & 0 & 4 &0 &  1 \\
                       & UNDC & $256^3$     &   0.729        &  11.24\%      &    0  & 357 & 2177 & 1 \\
                       & MeshCAP & $256^3$   &  0.715 & 0 & 1.95\% & 199 & 2096  & 490\\
                       & MeshUDF & $256^3$  &      0.579     &     0  &   0 & 6 & 226 & 7\\
                       & DCUDF & $256^3$     &   0.785      &     0   &      0   & 8 & 0 & 1\\
                        \cline{2-9}
                       & MC  & $512^3$ &  0.358 & 0 & 0 & 3 & 0 & 1\\
                       & MeshCAP & $512^3$   &  0.415 & 0 & 0.60\% & 202 & 3126 & 1171\\
                       & MeshUDF & $512^3$     &     0.364      &     0   &      0  & 7 & 104 & 4\\
                       & DCUDF & $512^3$     &     0.334      &     0   &      0   & 4 & 0 & 1 \\
                        \cline{2-9}
                       & MC  & $1024^3$ & 0.313 &  0& 0 & 2 &0 & 1\\
                       & DCUDF & $1024^3$  & 0.317 & 0 &0 & 1 &0 & 1\\
                       \hline
                        \hline
\multirow{11}{*}{\makecell[c]{Dancing\\Children\\$|V|=724$k \\$g=8$\\$b=0$\\$\beta_0$=1}}
                        & MC  & $256^3$ & 0.518 & 0 & 0 & 8 &0 & 1\\
                       & UNDC & $256^3$       &   0.567        &     2.84\%  &     0 & 104 & 1125 & 1 \\
                       & MeshCAP & $256^3$   &  0.556 & 0 & 0.19\% & 50 & 433 & 178\\
                       & MeshUDF & $256^3$  &      0.530     &     0  &   0 & 9 & 30 & 2 \\
                       & DCUDF & $256^3$     &     0.507      &     0   &      0  & 8 &0 & 1 \\
                       \cline{2-9}
                       & MC  & $512^3$ & 0.460 & 0 &0  & 8 &0  & 1 \\
                        & MeshCAP & $512^3$   &  0.479 & 0 & 0.03\% & 17 & 3427 & 182 \\
                       & MeshUDF & $512^3$     &     0.473      &     0   &      0  & 8 & 2  & 1\\
                       & DCUDF & $512^3$     &     0.455      &     0   &      0 & 8 & 0 & 1 \\
                       \cline{2-9}
                       & MC  & $1024^3$ & 0.445 & 0 & 0 & 8 &0 & 1\\
                       & DCUDF & $1024^3$     &  0.458         & 0       &  0   & 8 & 0 & 1\\
                       \hline
                       \hline
  \multirow{11}{*}{\makecell[c]{Armadillo\\$|V|=172$k \\ $g=0$\\$b=0$\\$\beta_0$=1}}
                        & MC  & $256^3$ & 0.485 & 0 & 0 & 0 &0 & 1\\
                        & UNDC & $256^3$   &   0.522  &    3.19$\%$  &  0.002$\%$   & 75 & 868 & 1 \\
                        & MeshCAP & $256^3$   &  0.524 & 0 & 0.20\% & 22 & 300 & 111 \\
                       & MeshUDF & $256^3$  &      0.504     &     0  &   0  & 0 & 14 & 1\\
                       & DCUDF & $256^3$     &     0.433      &     0   &      0 & 1 & 0 & 1 \\
                       \cline{2-9}
                       & MC  & $512^3$ & 0.385 & 0 & 0 & 0 &0 & 1\\
                        & MeshCAP & $512^3$   &  0.409 & 0 & 0.02\% & 0 & 244 & 150\\
                       & MeshUDF & $512^3$     &     0.405      &     0   &      0  & 0 & 1 & 1\\
                       & DCUDF & $512^3$     &     0.358      &     0   &      0  & 0 & 0  & 1\\
                       \cline{2-9}
                       & MC  & $1024^3$ & 0.367 &0  & 0 & 0 &0 & 1\\
                       & DCUDF & $1024^3$     &   0.354       &    0    &     0  &   0  &  0 & 1 \\
                       \hline

\end{tabular}
\end{small}
\end{table}

\begin{figure*}[!htbp]
    \centering
    \includegraphics[width=1.6in]{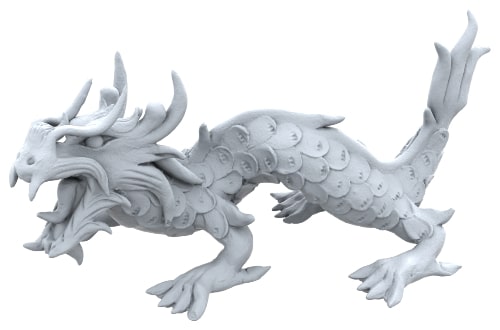}
    \includegraphics[width=1.6in]{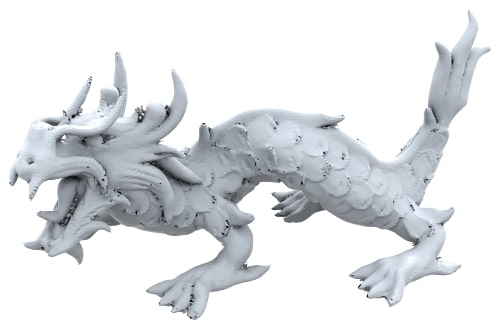 }
    \includegraphics[width=1.6in]{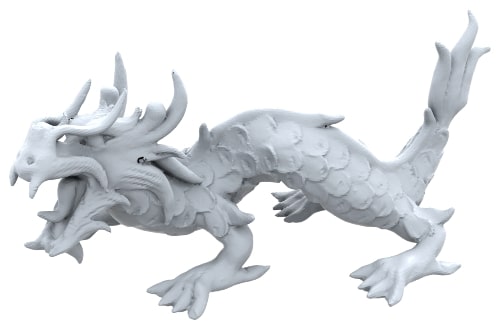 }
    \includegraphics[width=1.6in]{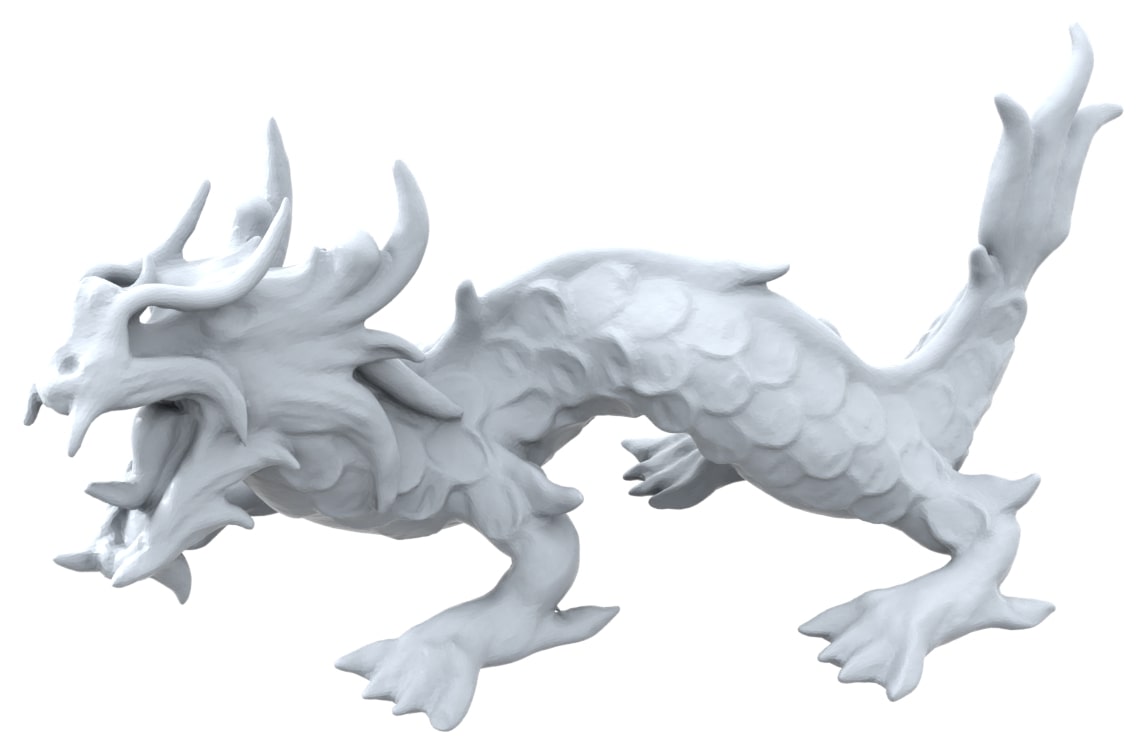}\\
    \includegraphics[width=0.8in]{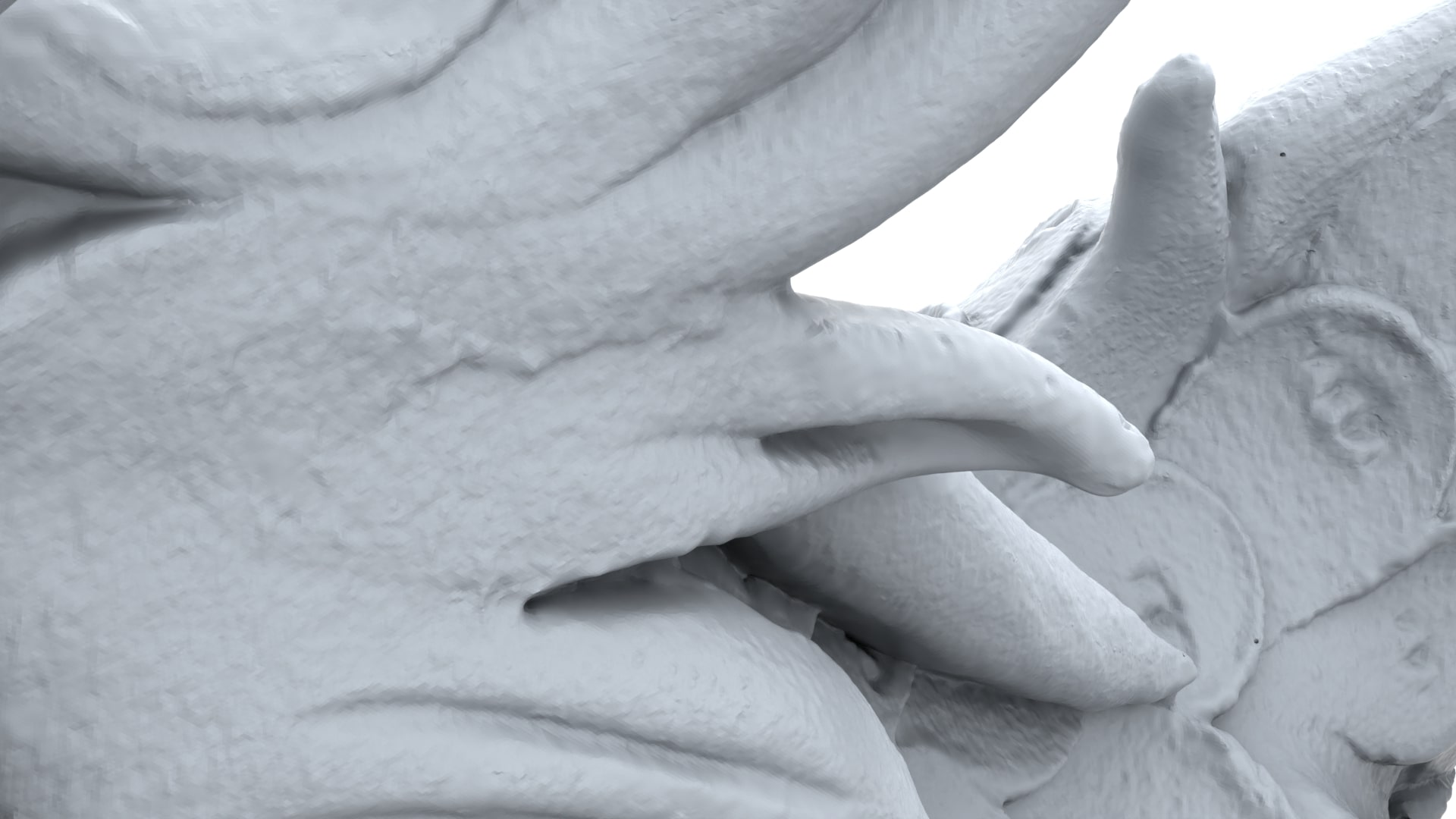}
    \includegraphics[width=0.8in]{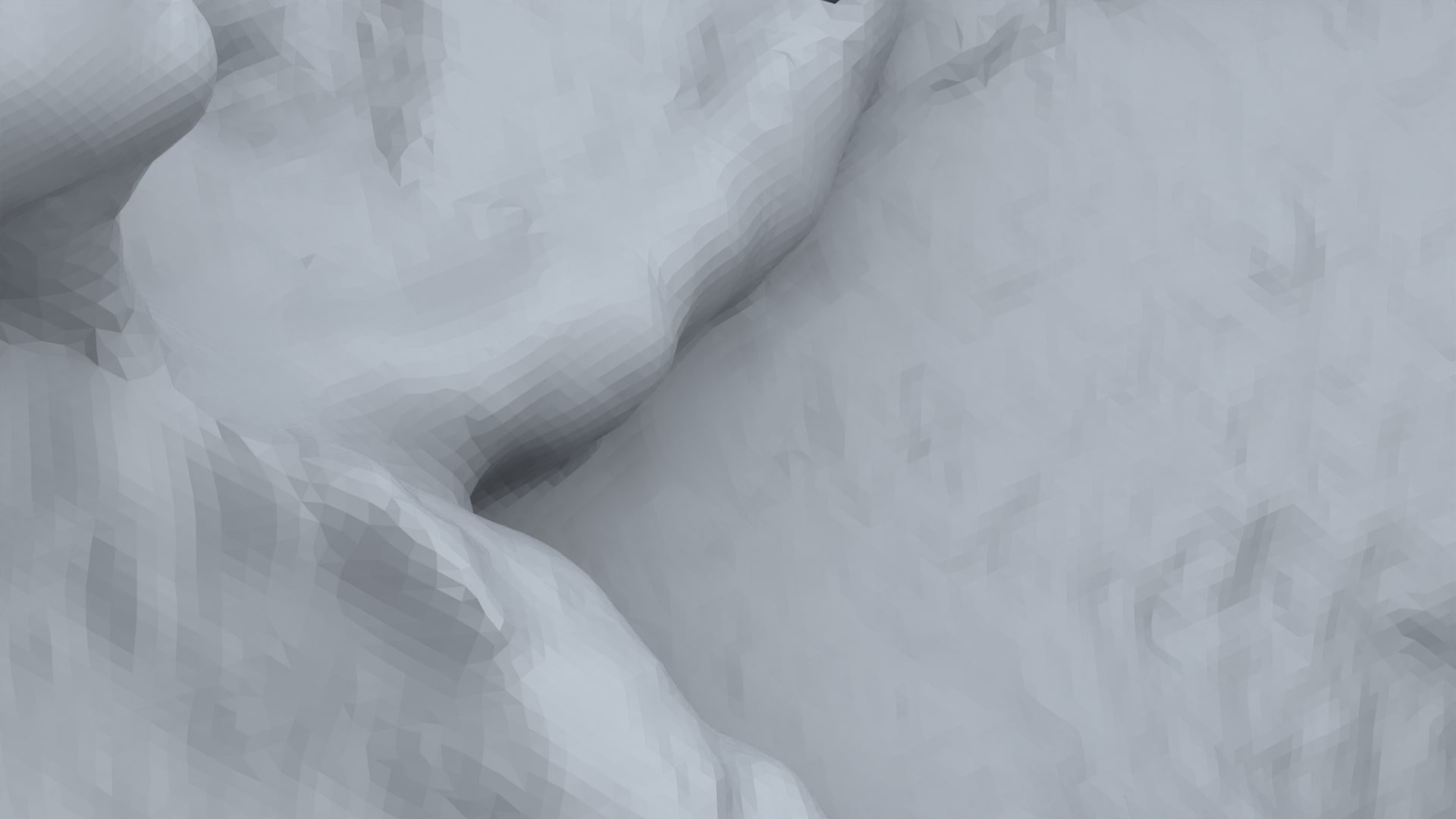}
    \includegraphics[width=0.8in]{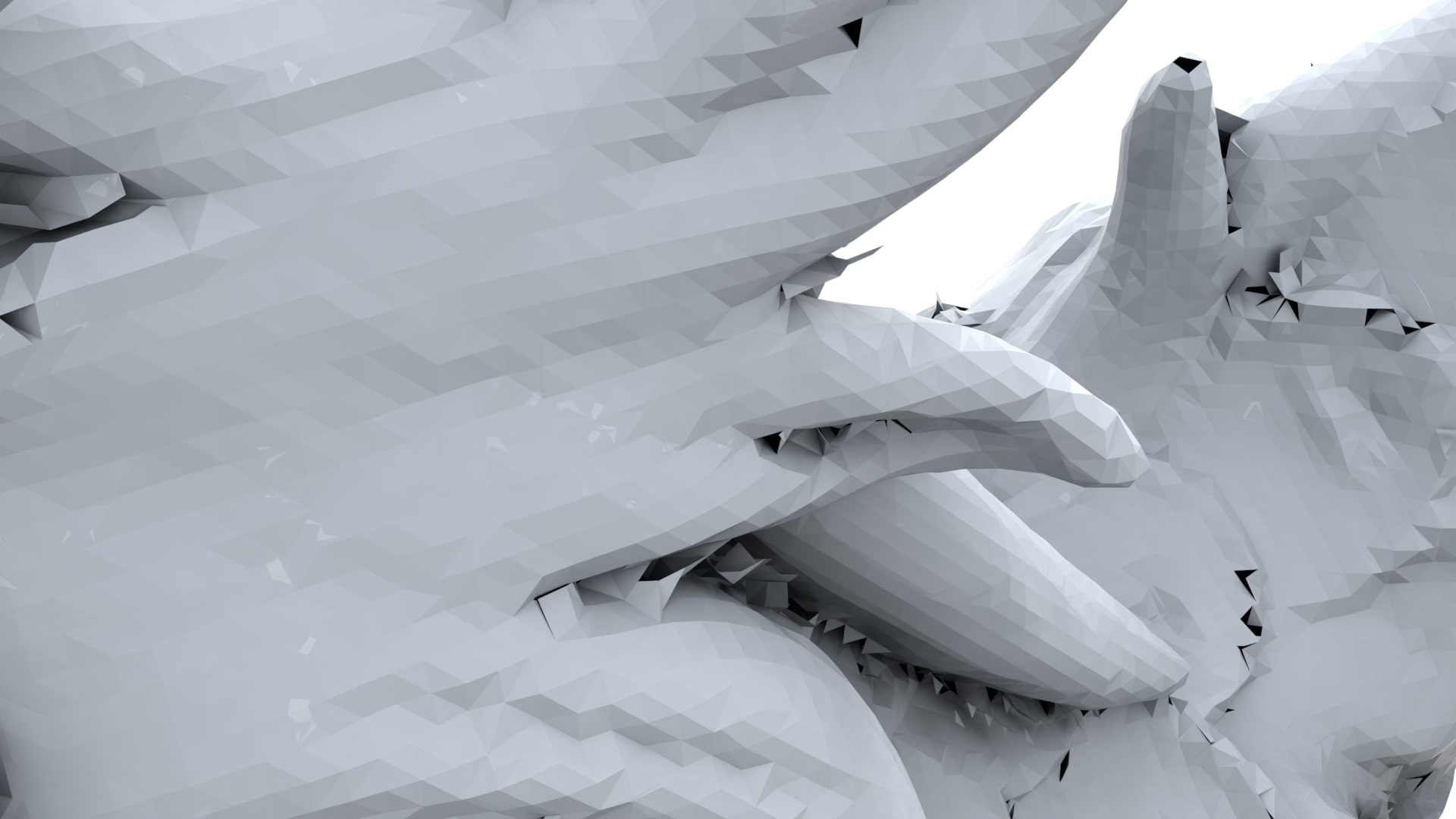 }
    \includegraphics[width=0.8in]{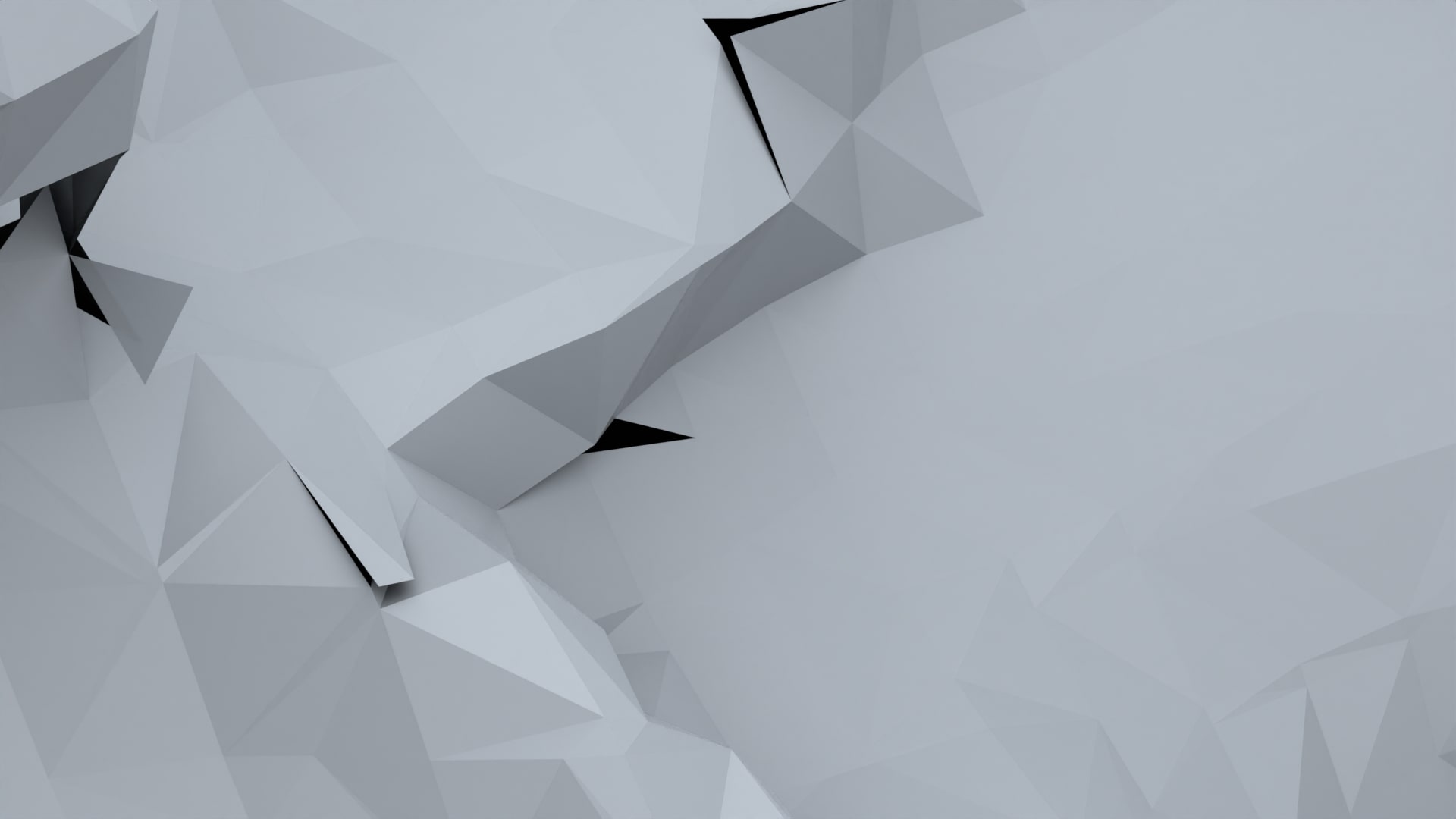 }
    \includegraphics[width=0.8in]{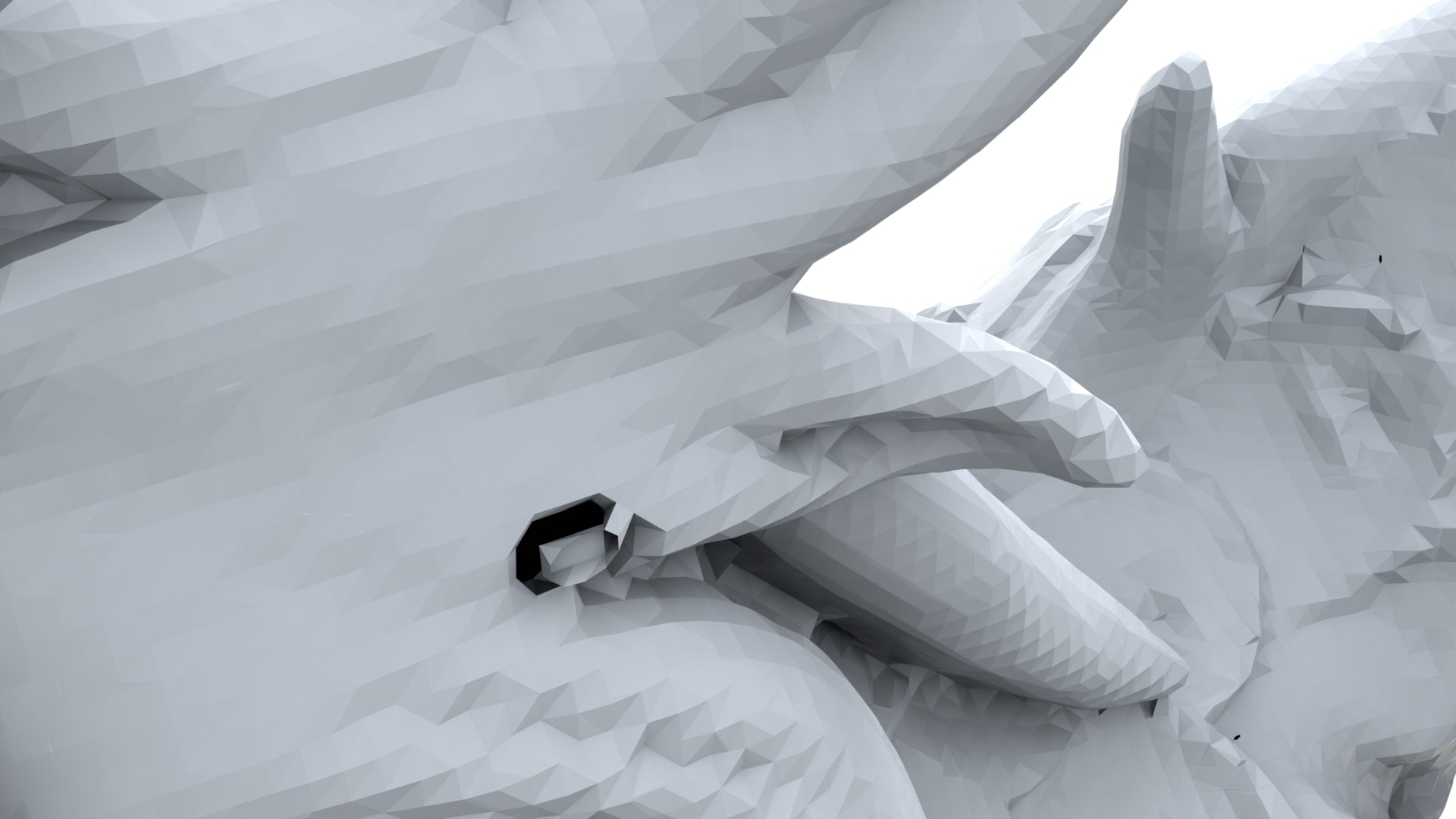 }
    \includegraphics[width=0.8in]{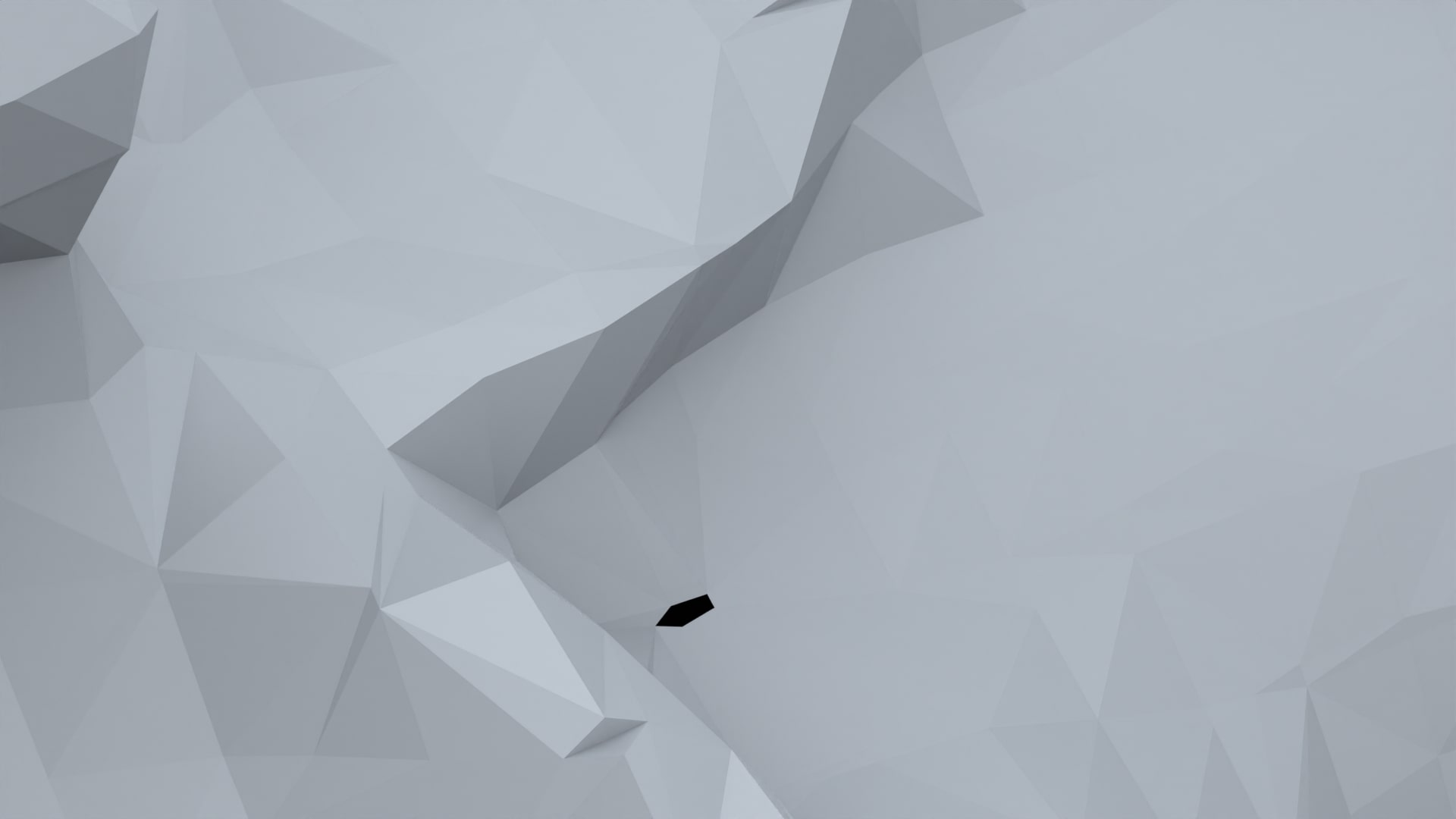 }
    \includegraphics[width=0.8in]{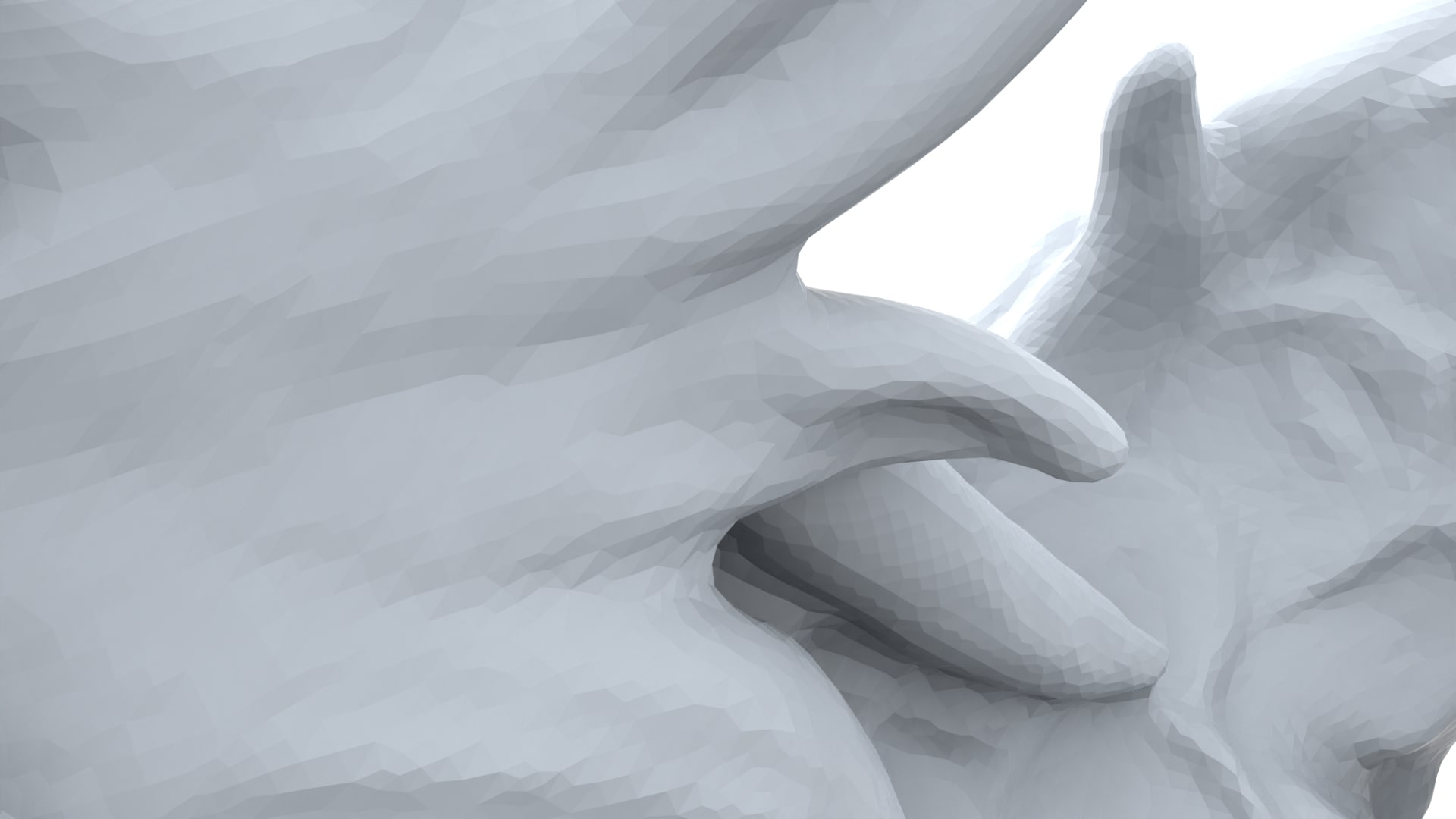}
    \includegraphics[width=0.8in]{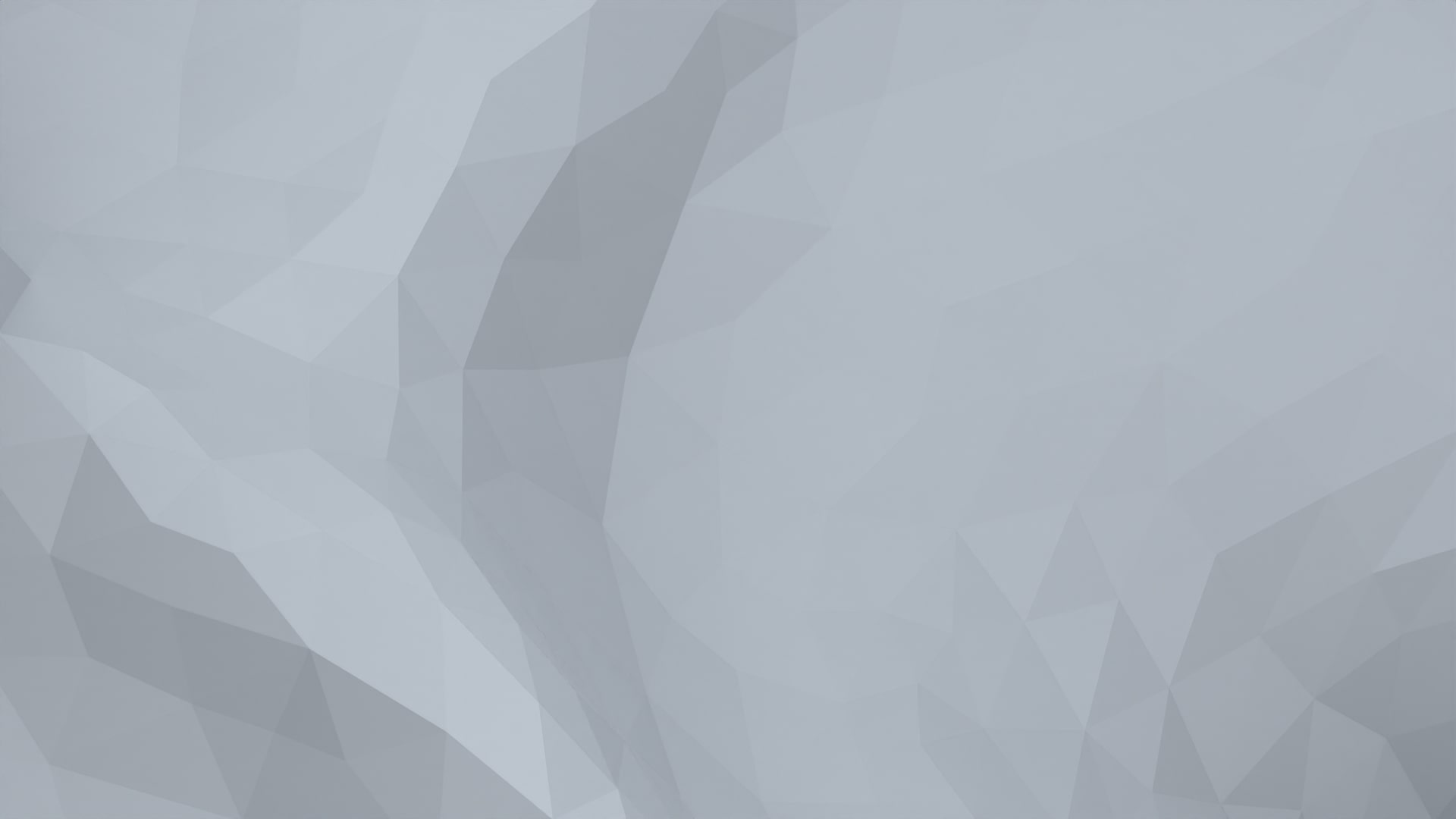}\\
    \makebox[1.6in]{GT mesh (3.6m vertices)}
    \makebox[1.6in]{MeshCAP $256^3$}
    \makebox[1.6in]{MeshUDF $256^3$}
    \makebox[1.6in]{DCUDF $256^3$}\\
    \includegraphics[width=1.6in]{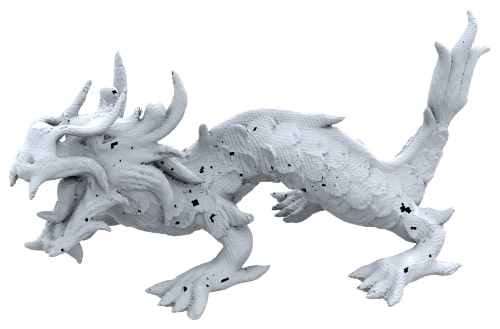}
    \includegraphics[width=1.6in]{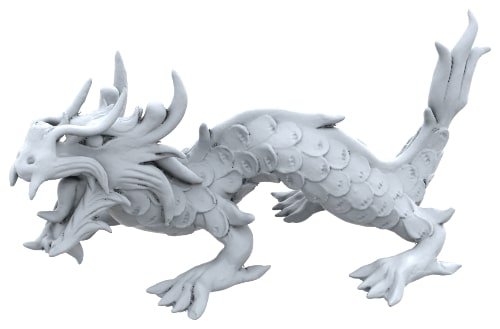 }
    \includegraphics[width=1.6in]{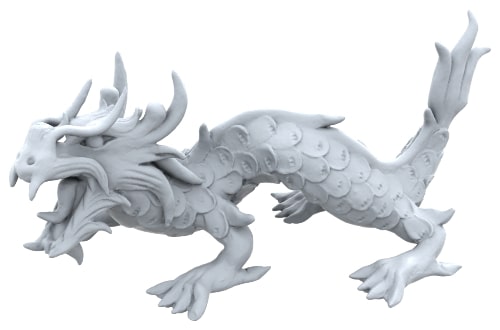 }
    \includegraphics[width=1.6in]{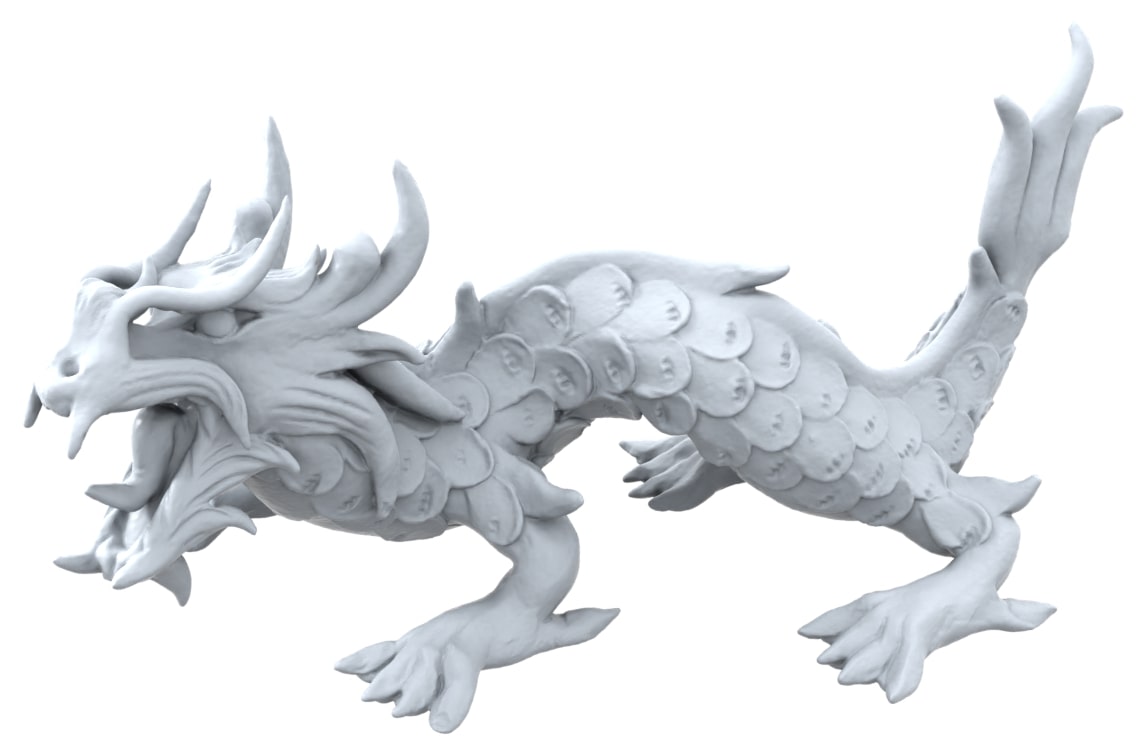}\\
    \includegraphics[width=0.8in]{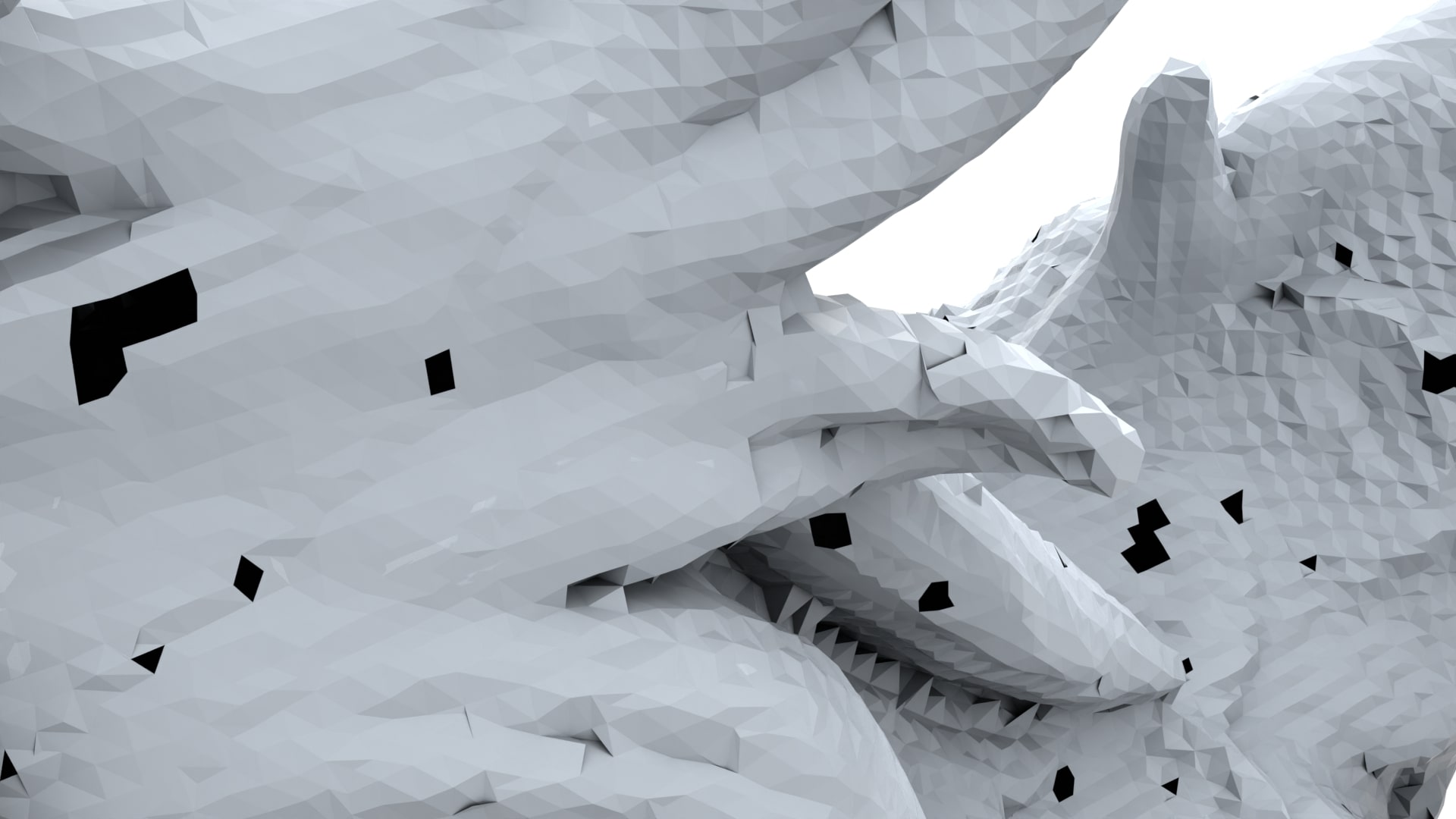}
    \includegraphics[width=0.8in]{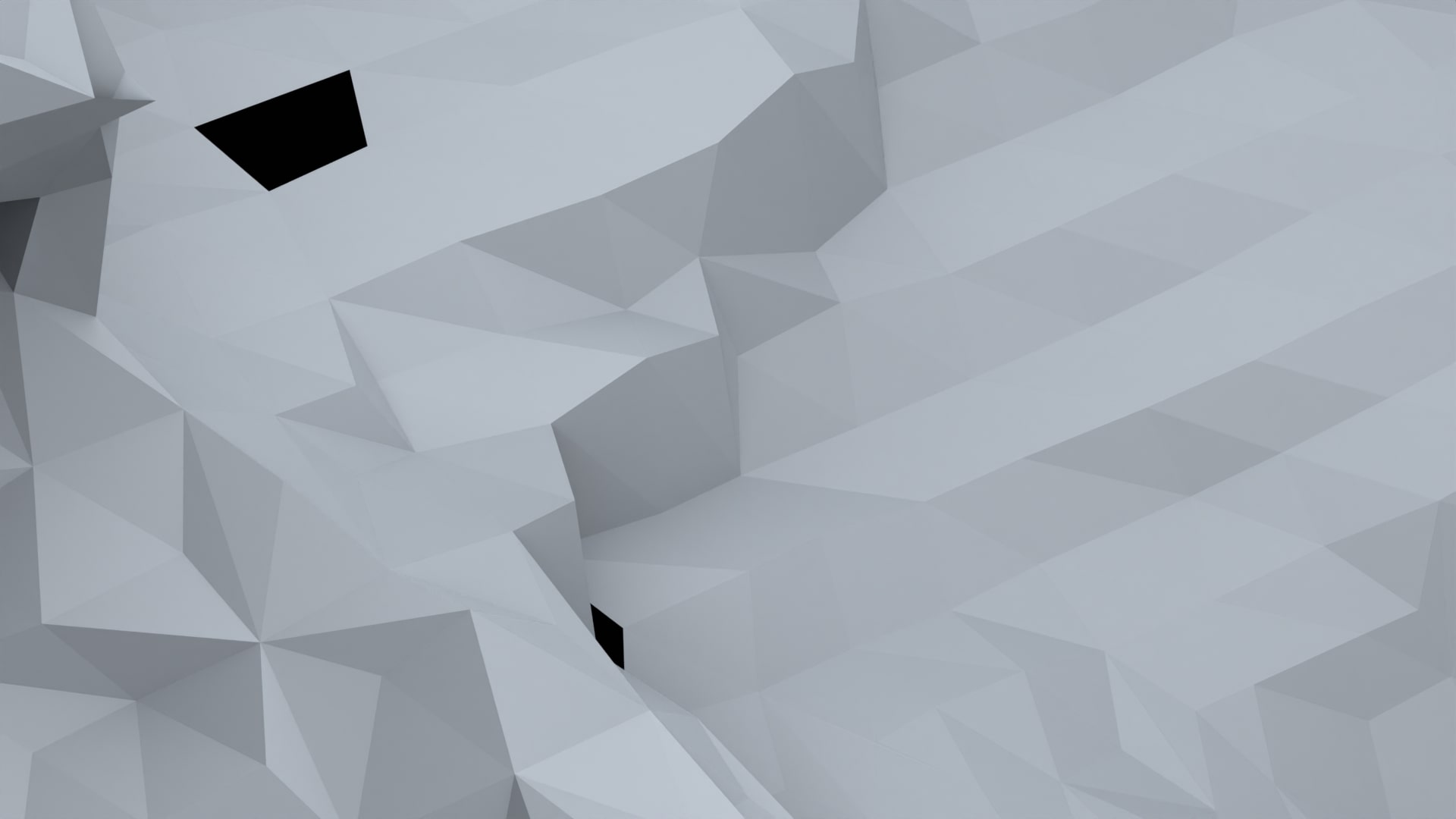}
    \includegraphics[width=0.8in]{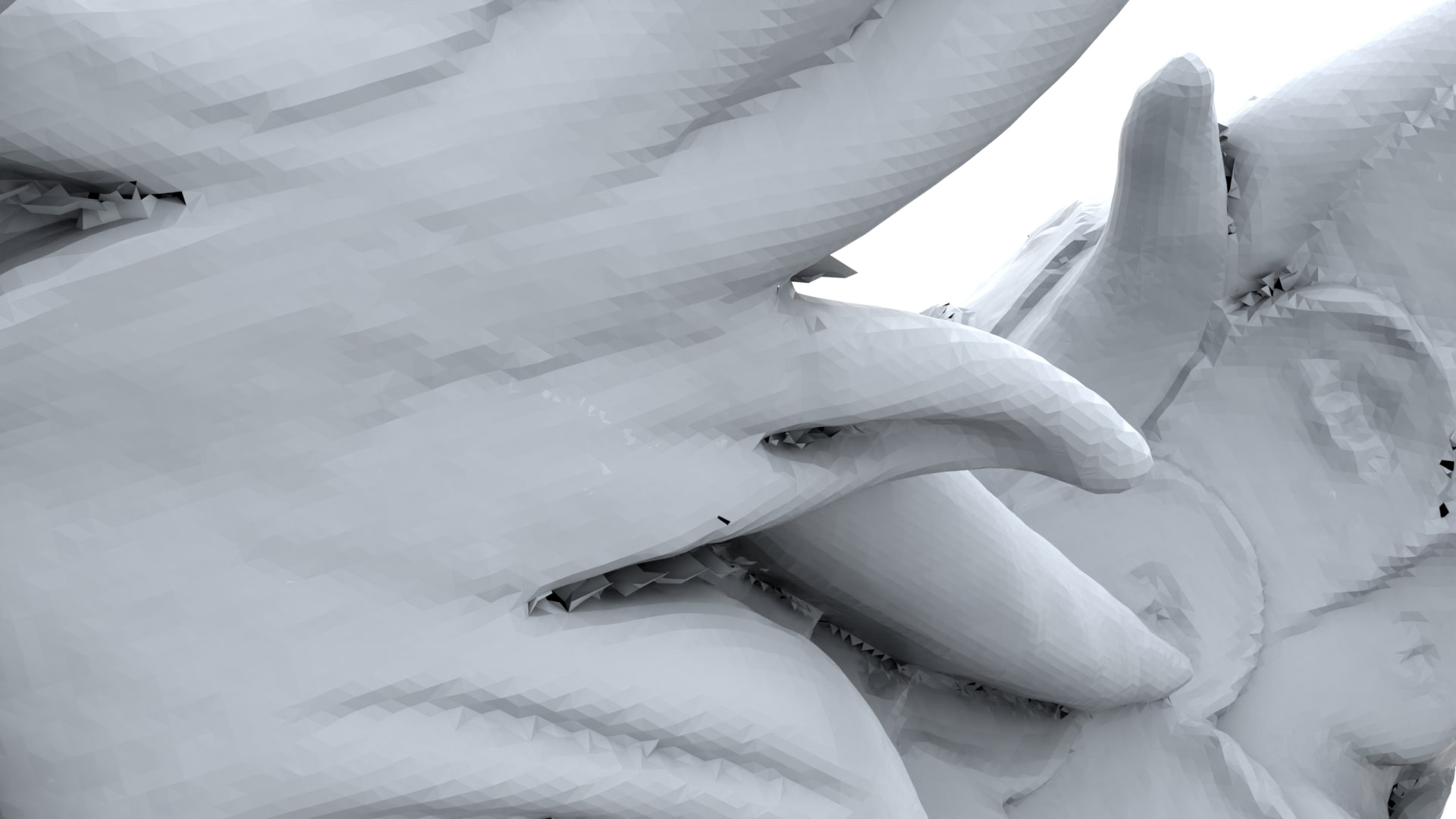 }
    \includegraphics[width=0.8in]{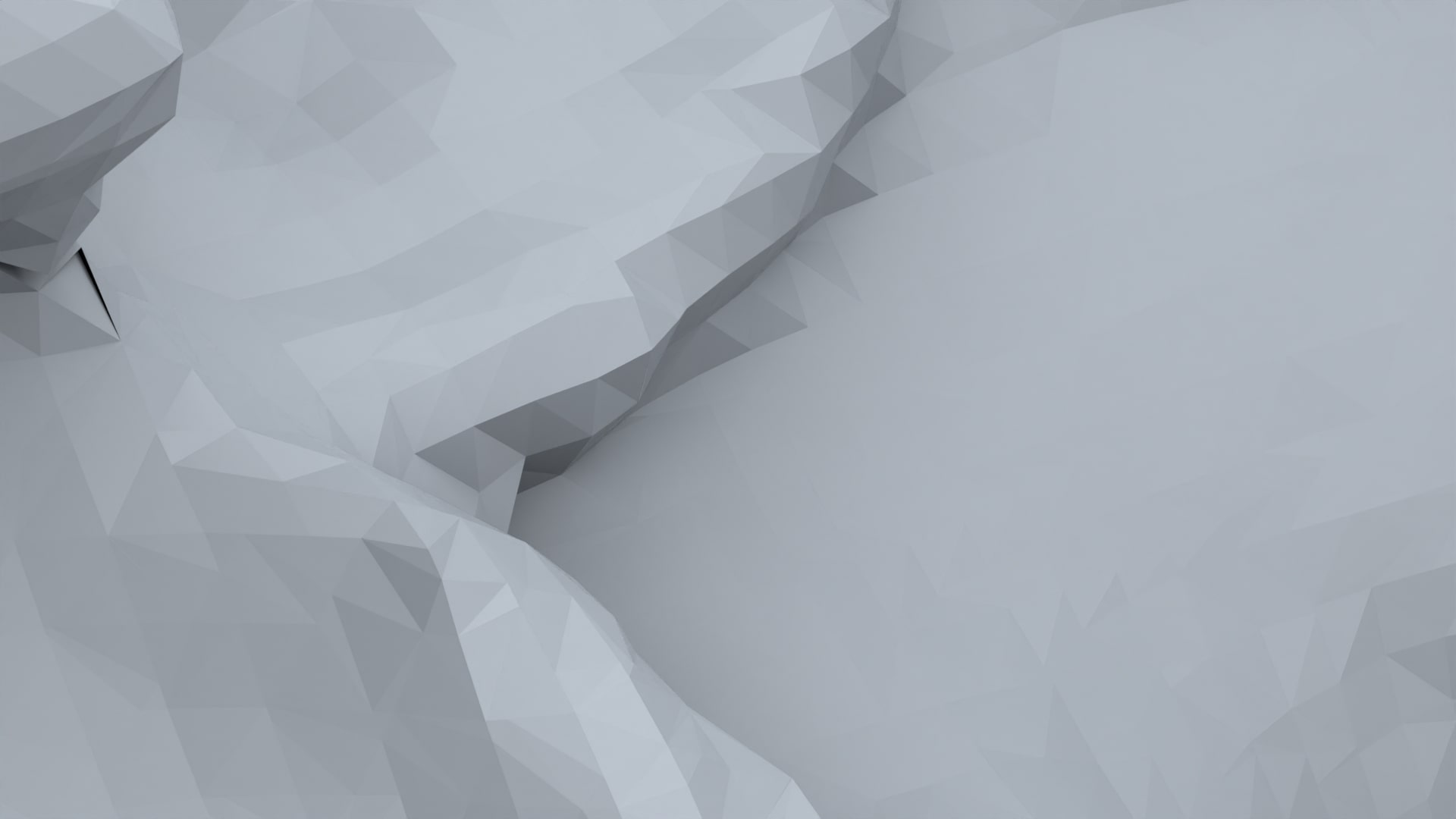 }
    \includegraphics[width=0.8in]{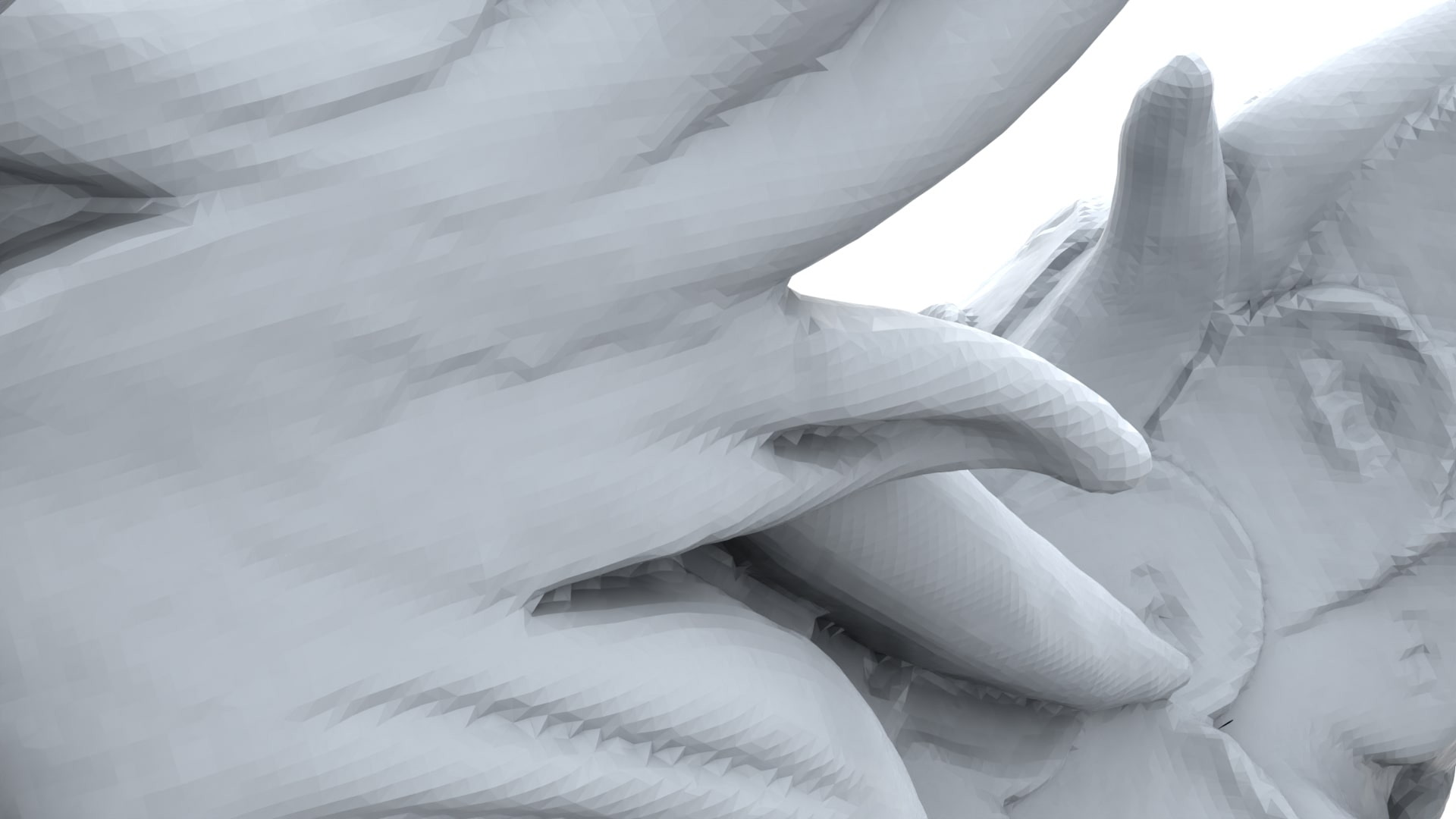 }
    \includegraphics[width=0.8in]{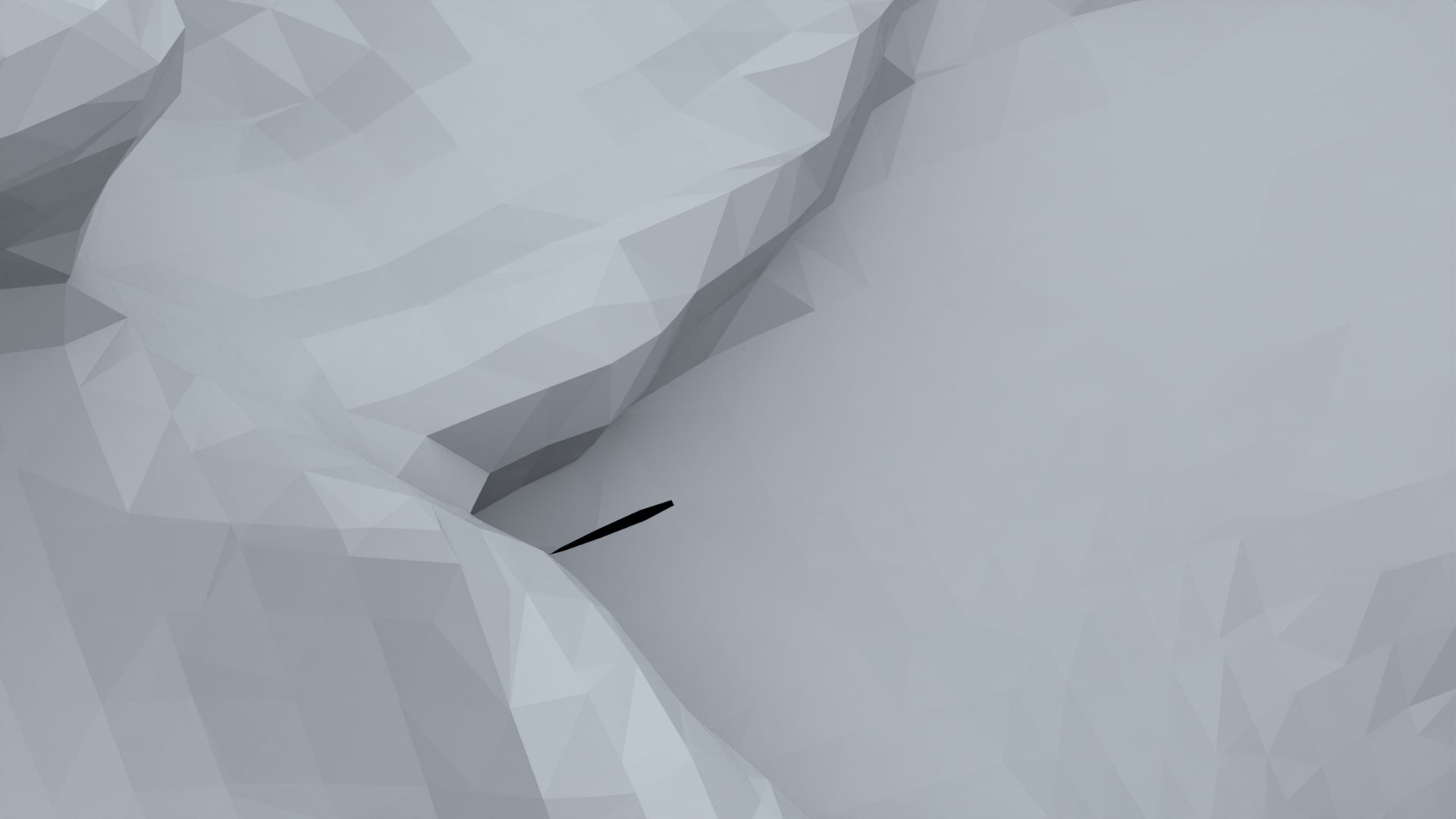 }
    \includegraphics[width=0.8in]{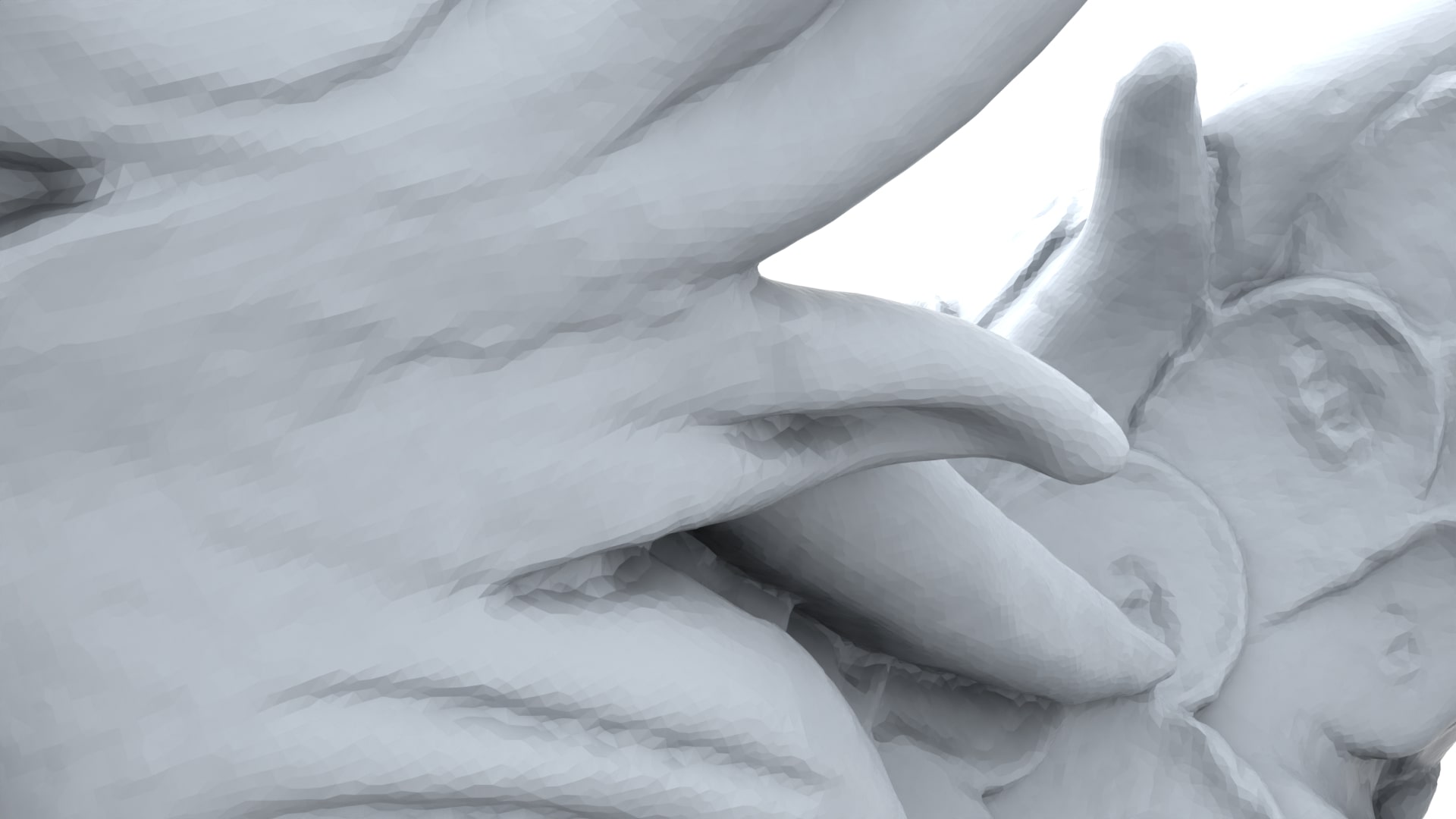}
    \includegraphics[width=0.8in]{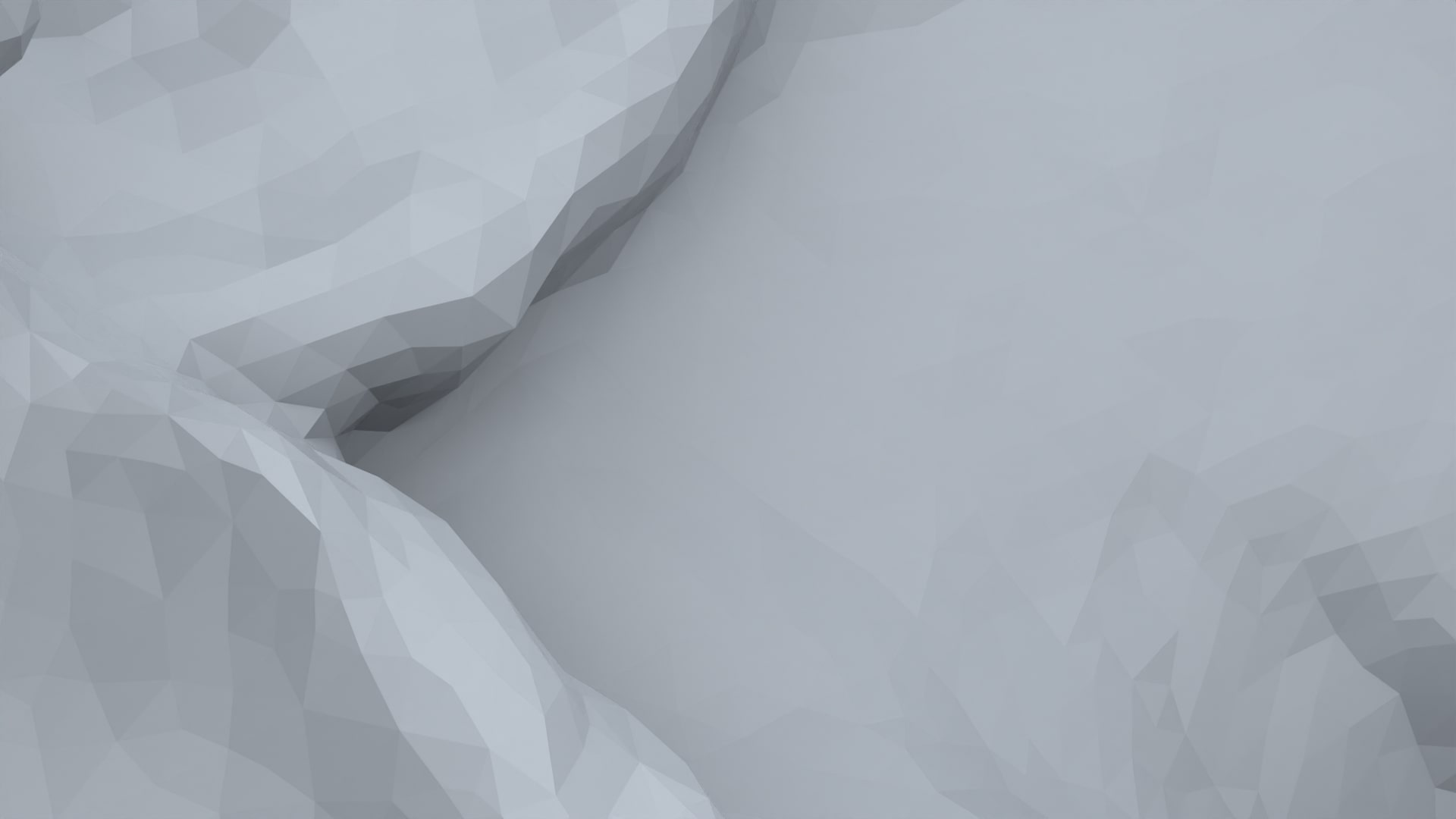}\\
    \makebox[1.6in]{UNDC $256^3$}
    \makebox[1.6in]{MeshCAP $512^3$}
    \makebox[1.6in]{MeshUDF $512^3$}
    \makebox[1.6in]{DCUDF $512^3$} \\
    \includegraphics[width=1.35in]{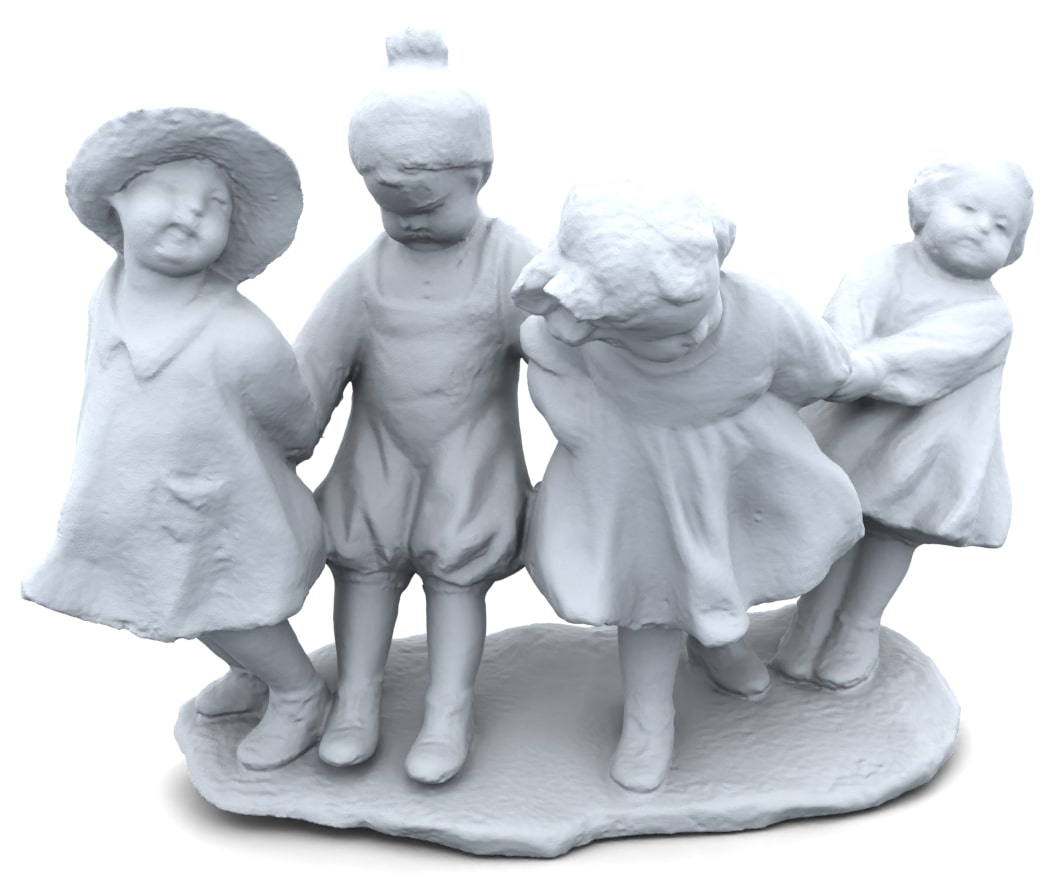 }
    \includegraphics[width=1.35in]{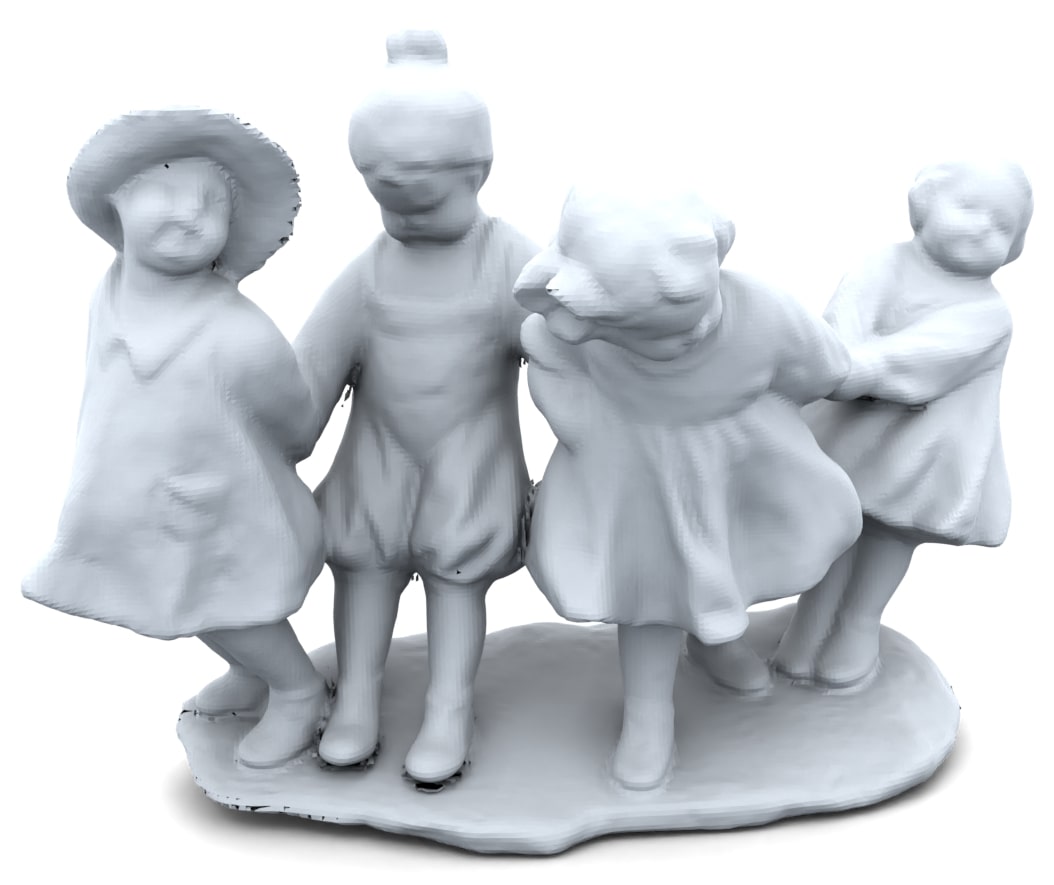 }
    \includegraphics[width=1.35in]{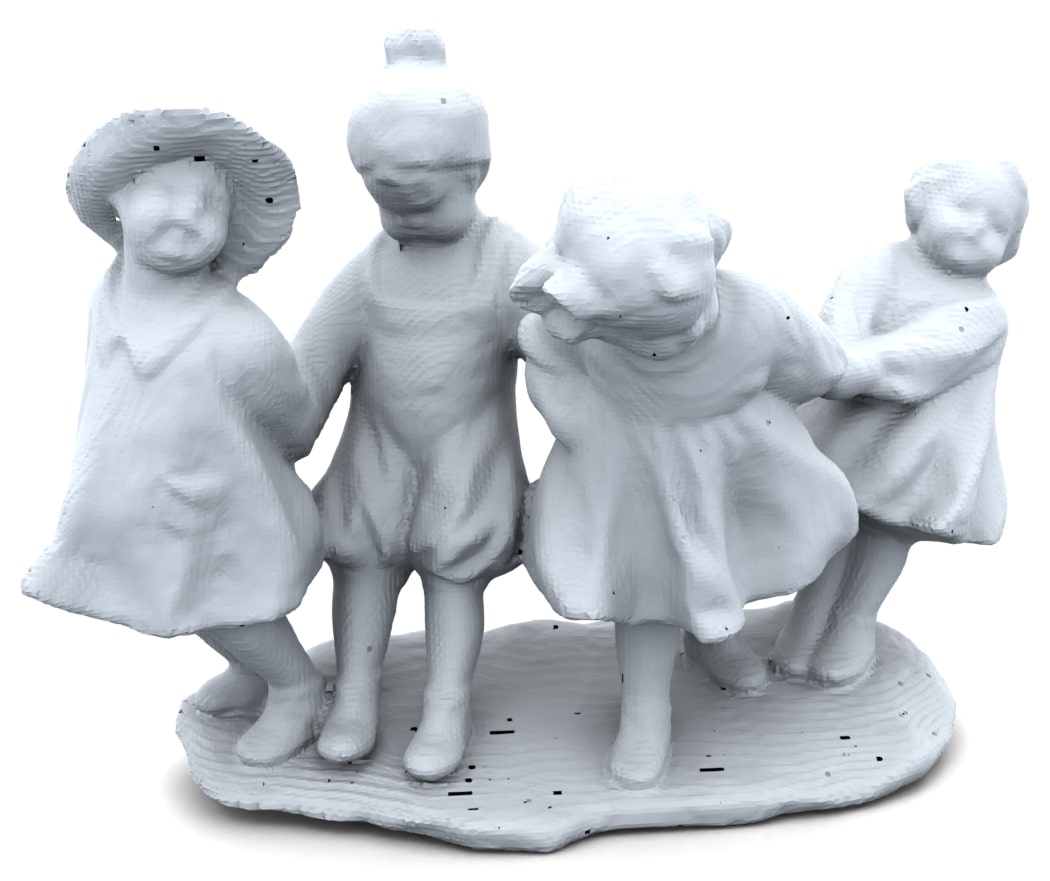 }
    \includegraphics[width=1.35in]{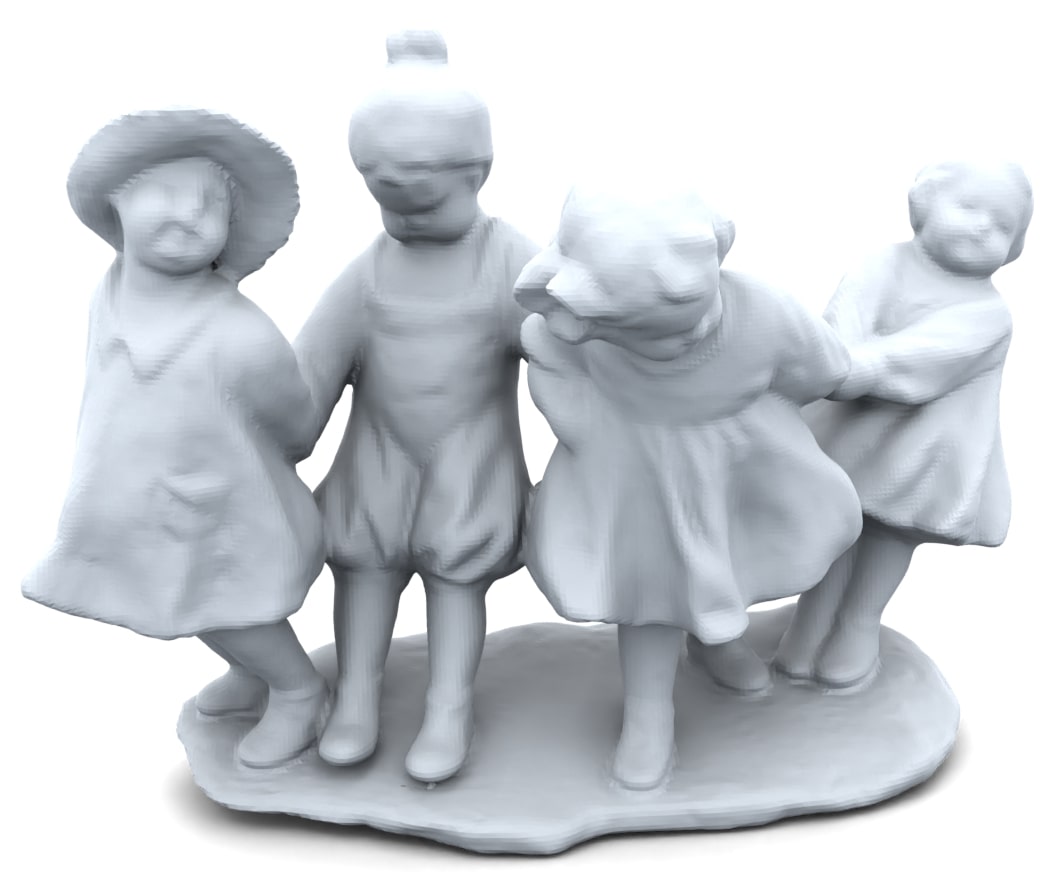 }
    \includegraphics[width=1.35in]{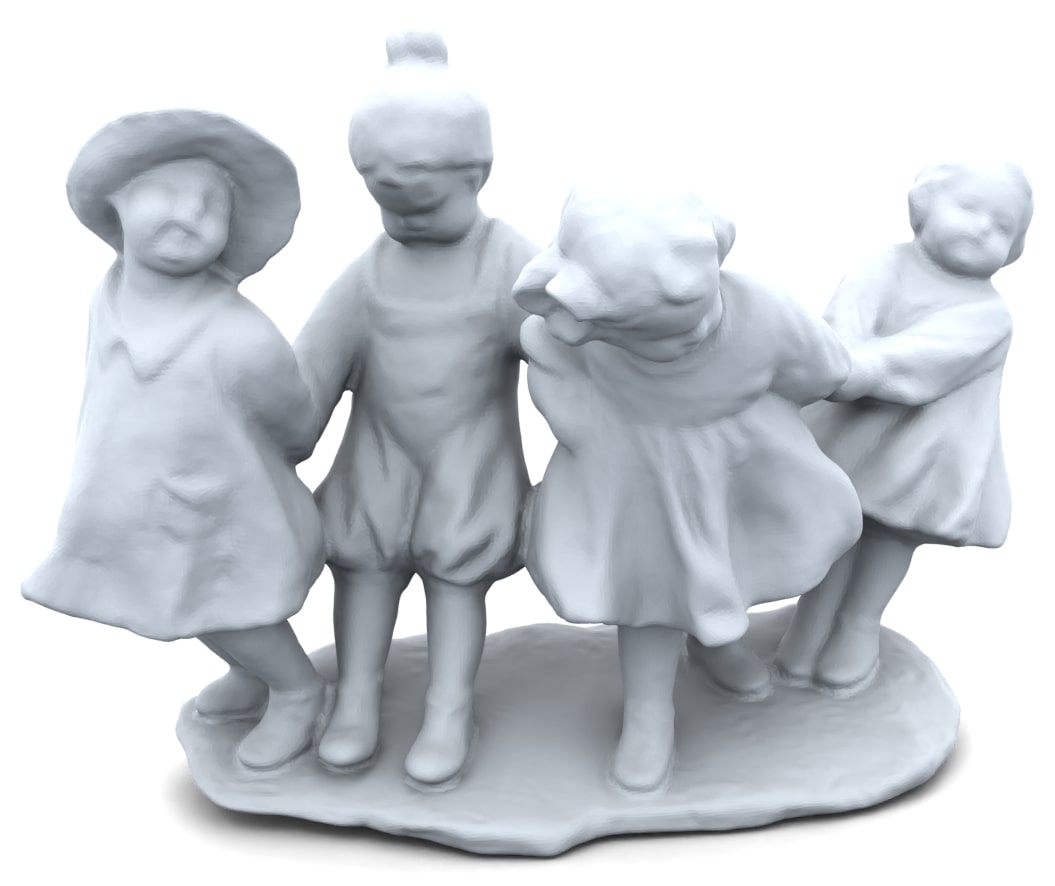 }\\
    \includegraphics[width=1.35in]{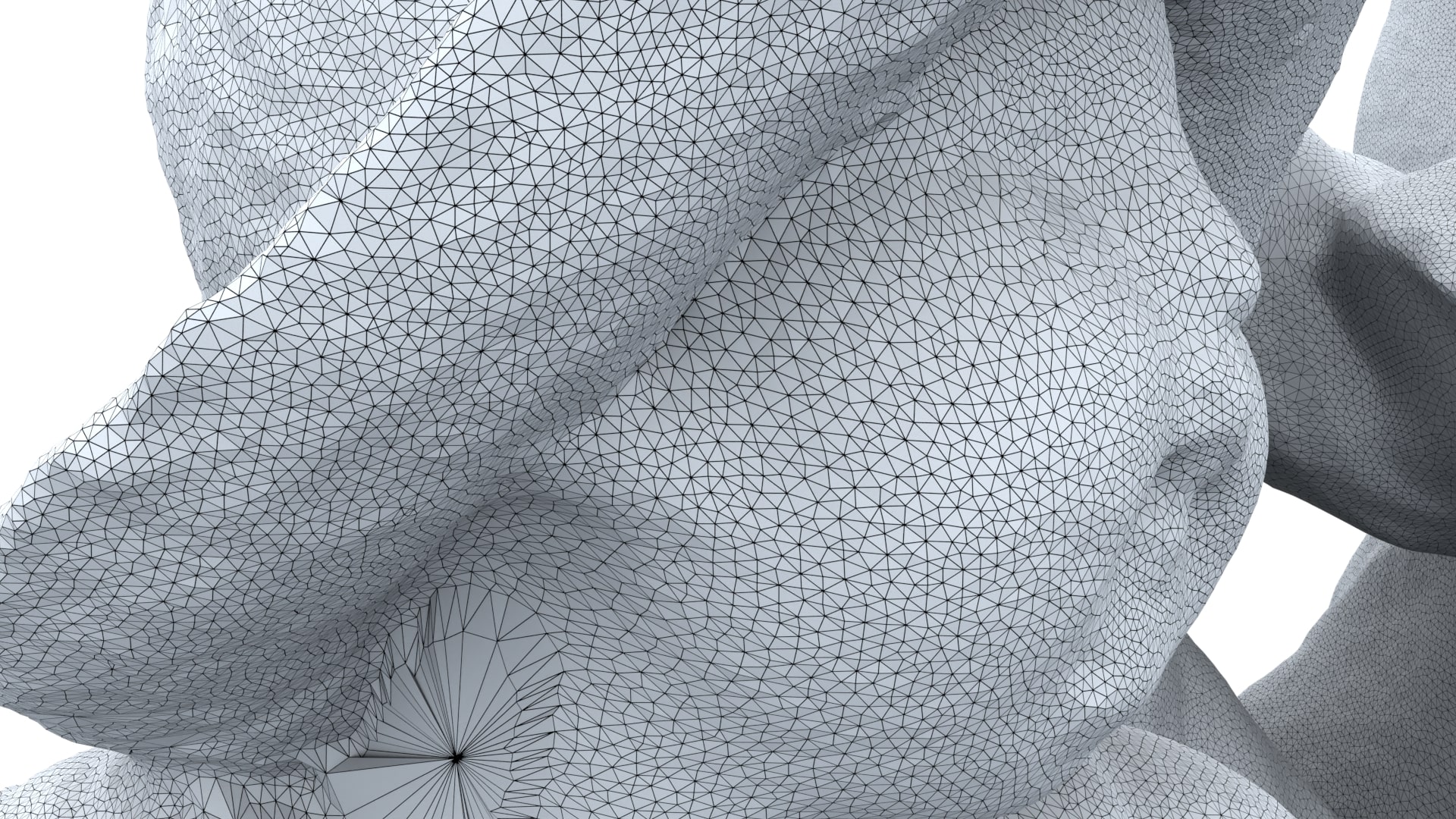 }
    \includegraphics[width=1.35in]{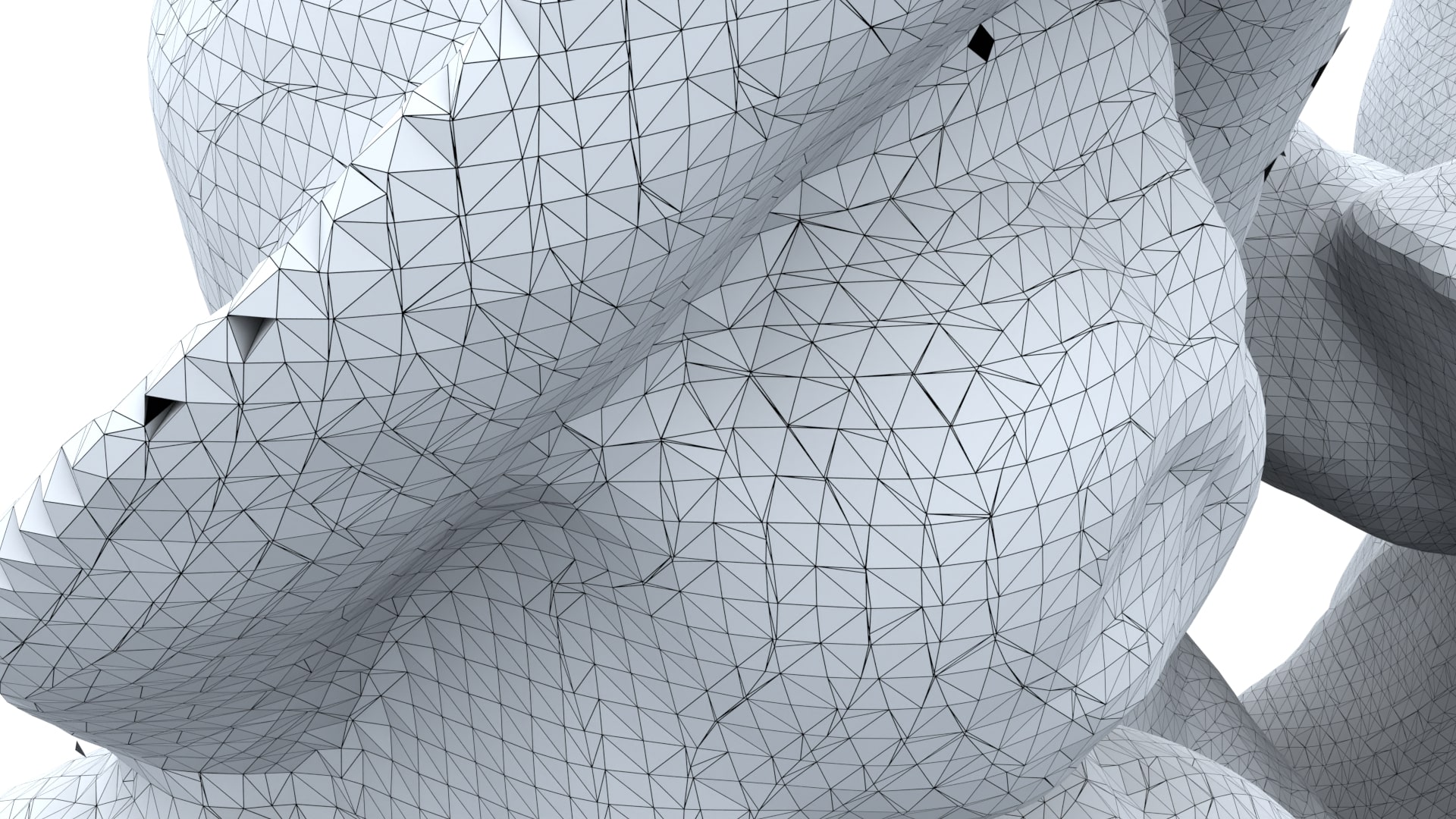 }
    \includegraphics[width=1.35in]{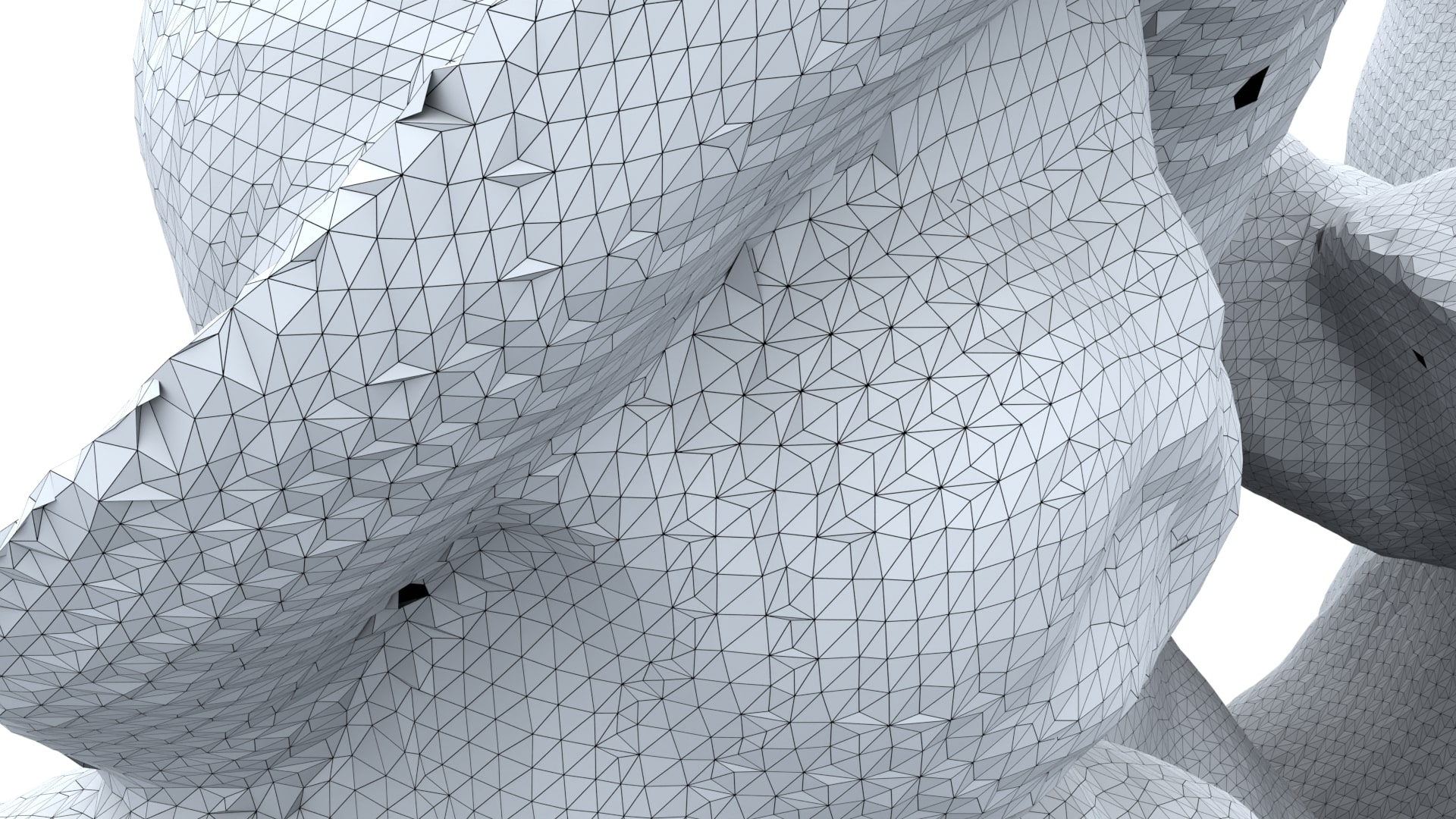 }
    \includegraphics[width=1.35in]{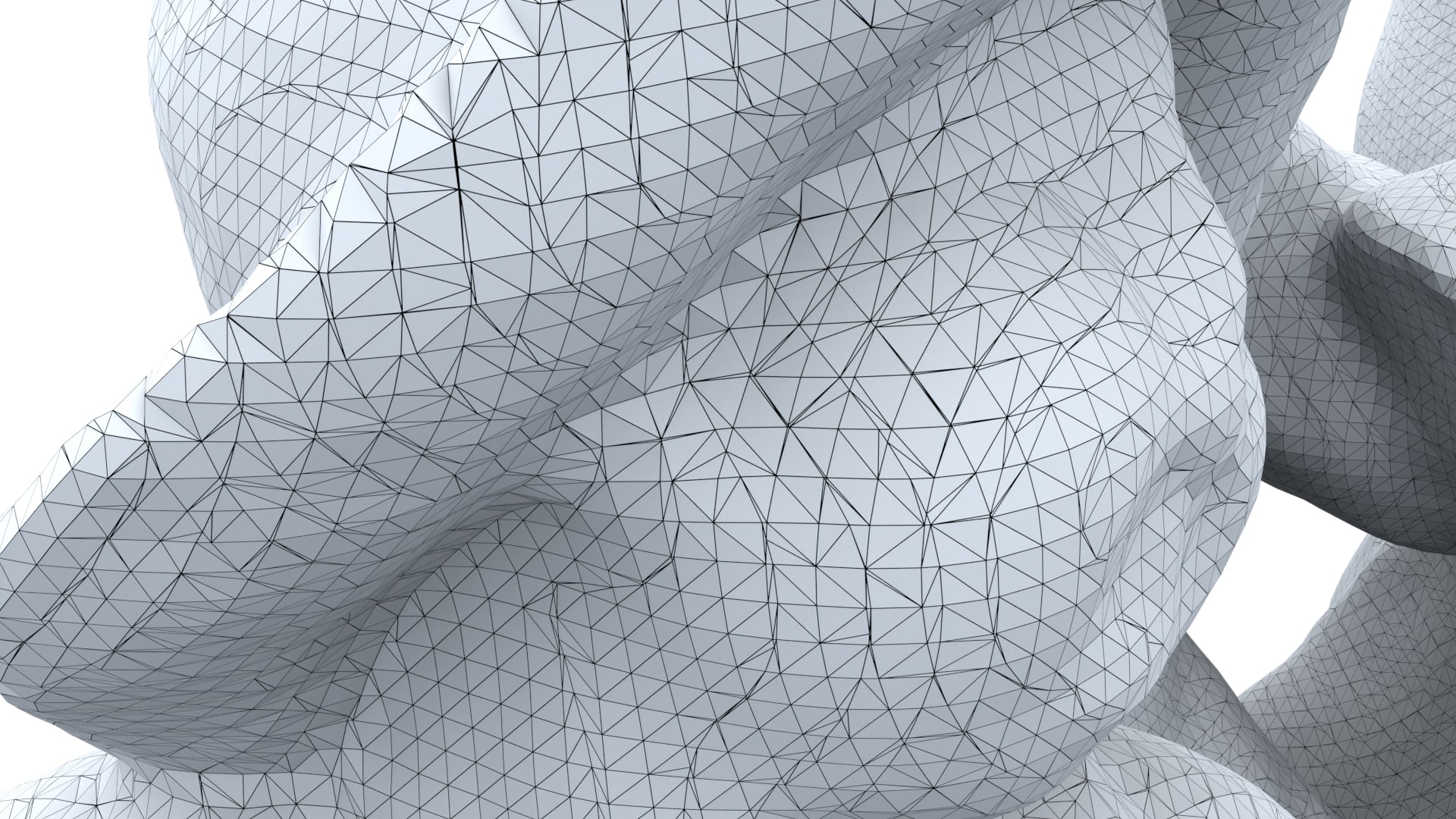}
    \includegraphics[width=1.35in]{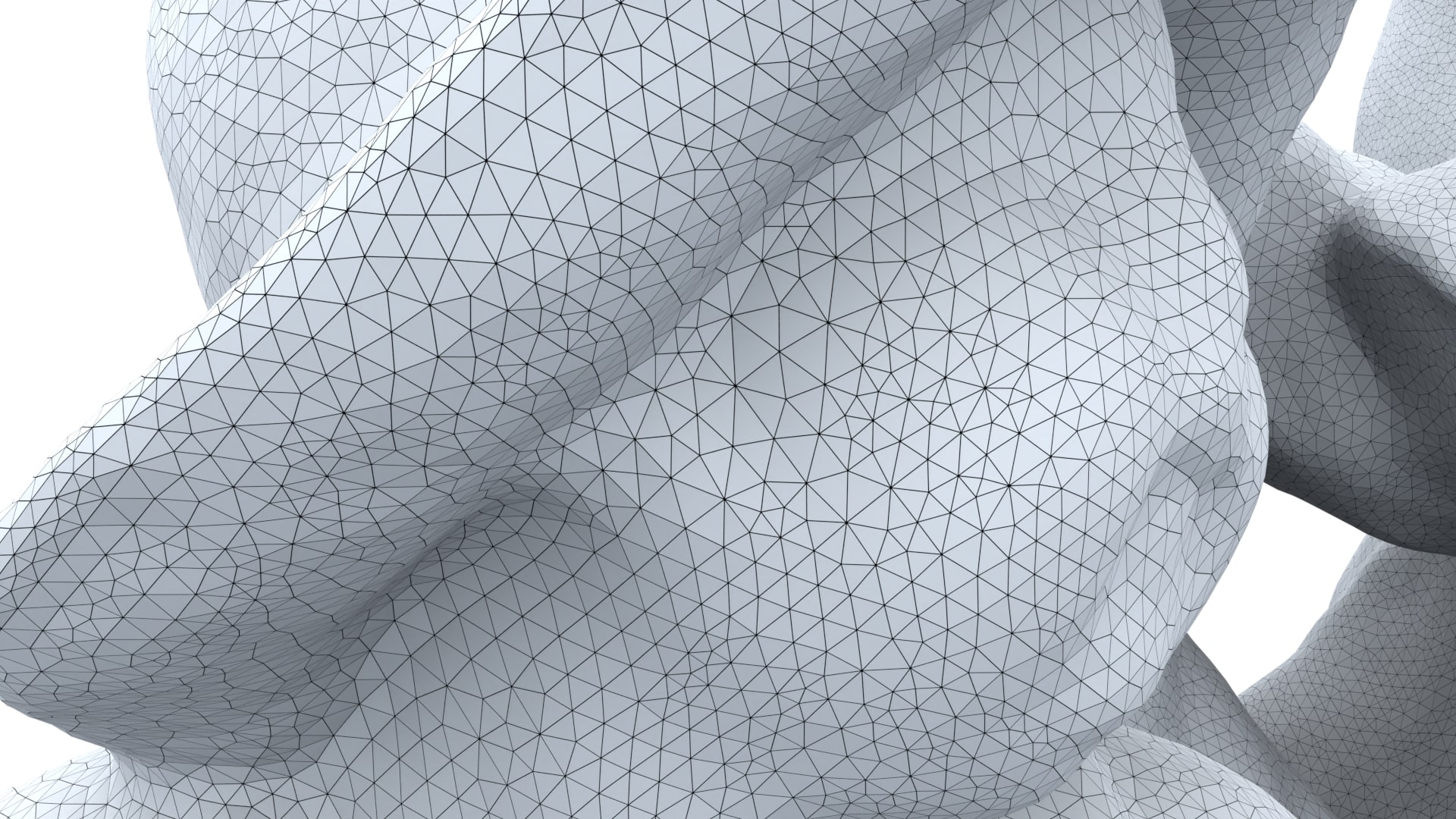 }\\
    \makebox[1.35in]{GT mesh (724k vertices)}
    \makebox[1.35in]{MeshCAP $256^3$}
    \makebox[1.35in]{UNDC $256^3$}
    \makebox[1.35in]{MeshUDF $256^3$}
    \makebox[1.35in]{DCUDF $256^3$}\\    
       \includegraphics[width=1.35in]{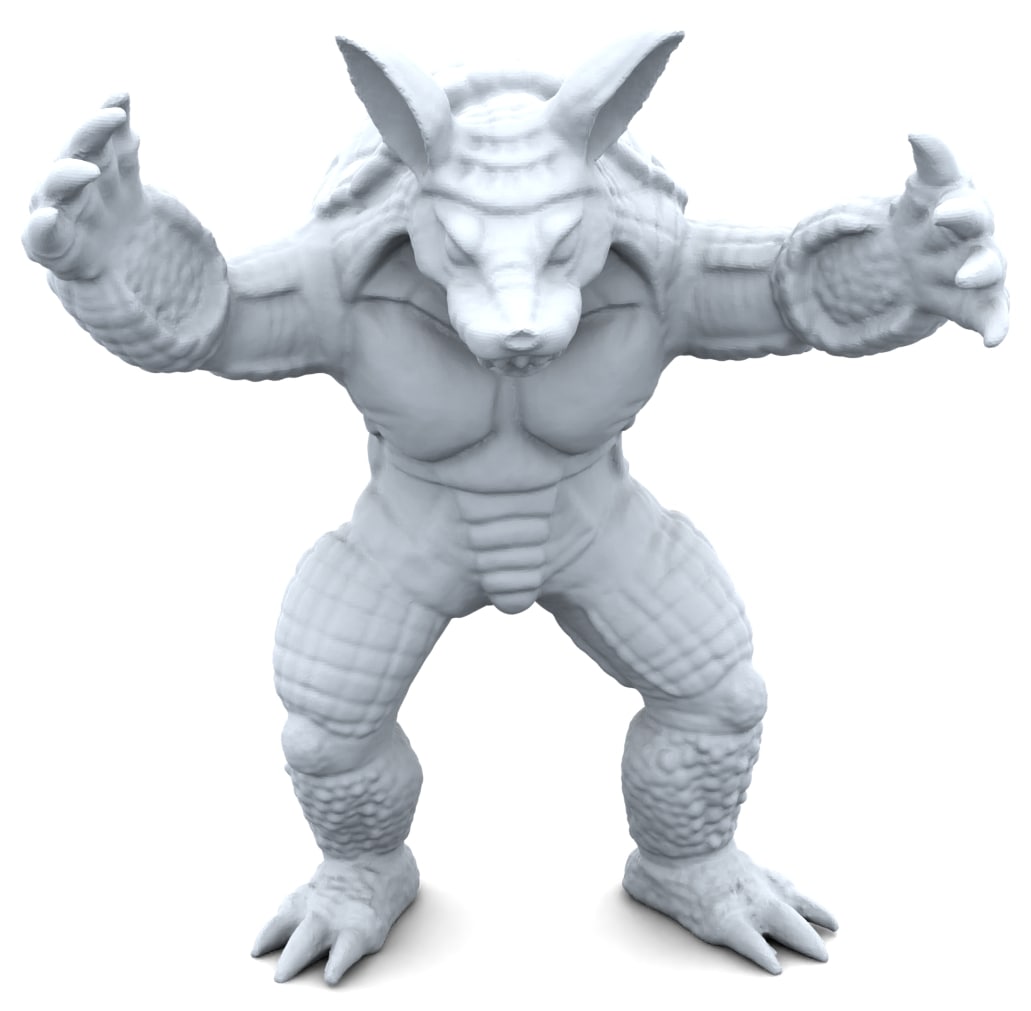 }
    \includegraphics[width=1.35in]{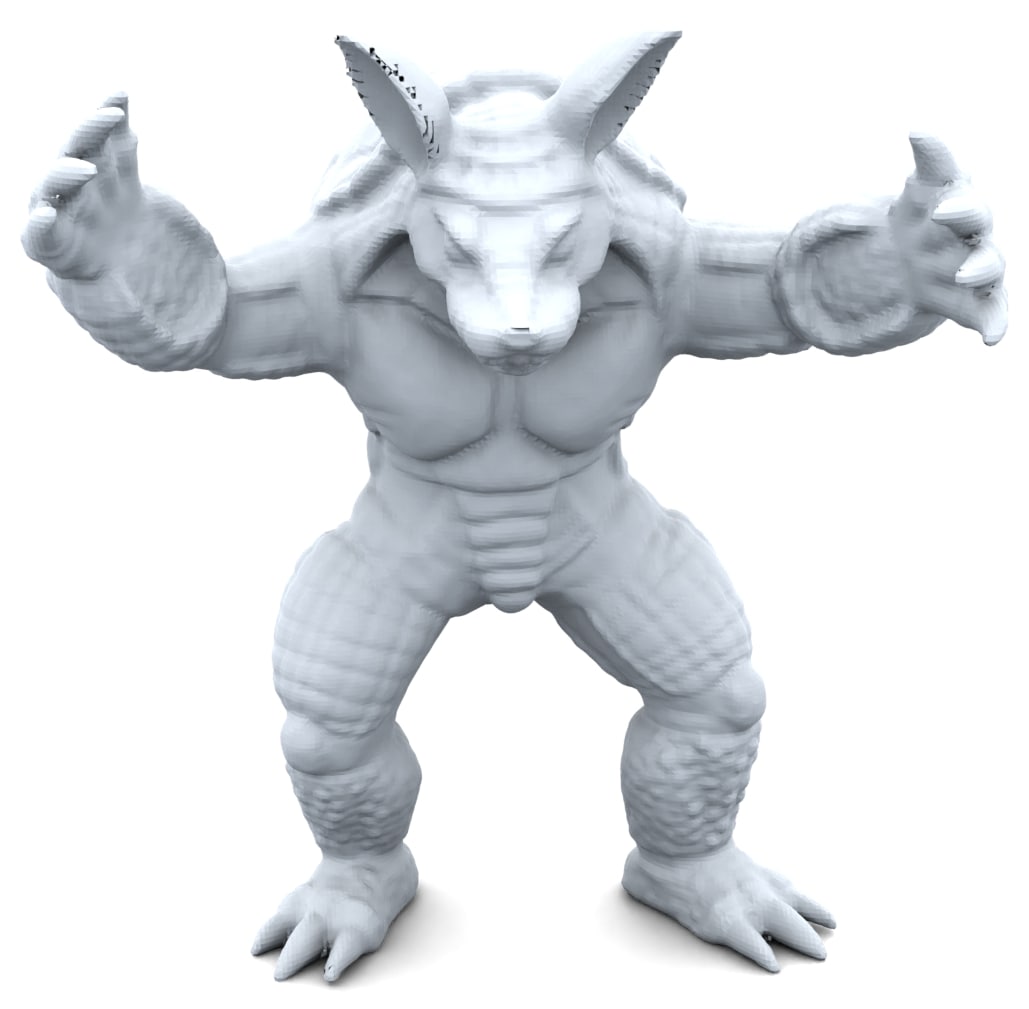 }
    \includegraphics[width=1.35in]{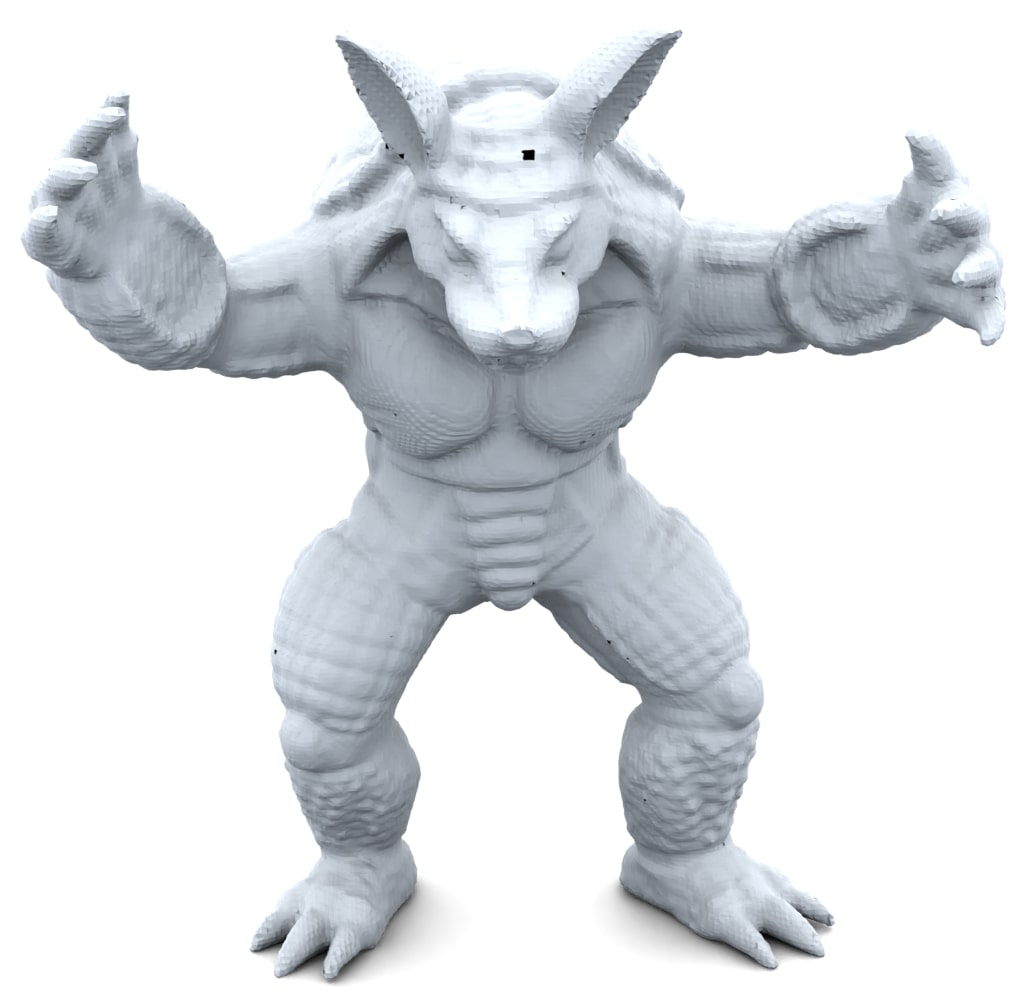 }
    \includegraphics[width=1.35in]{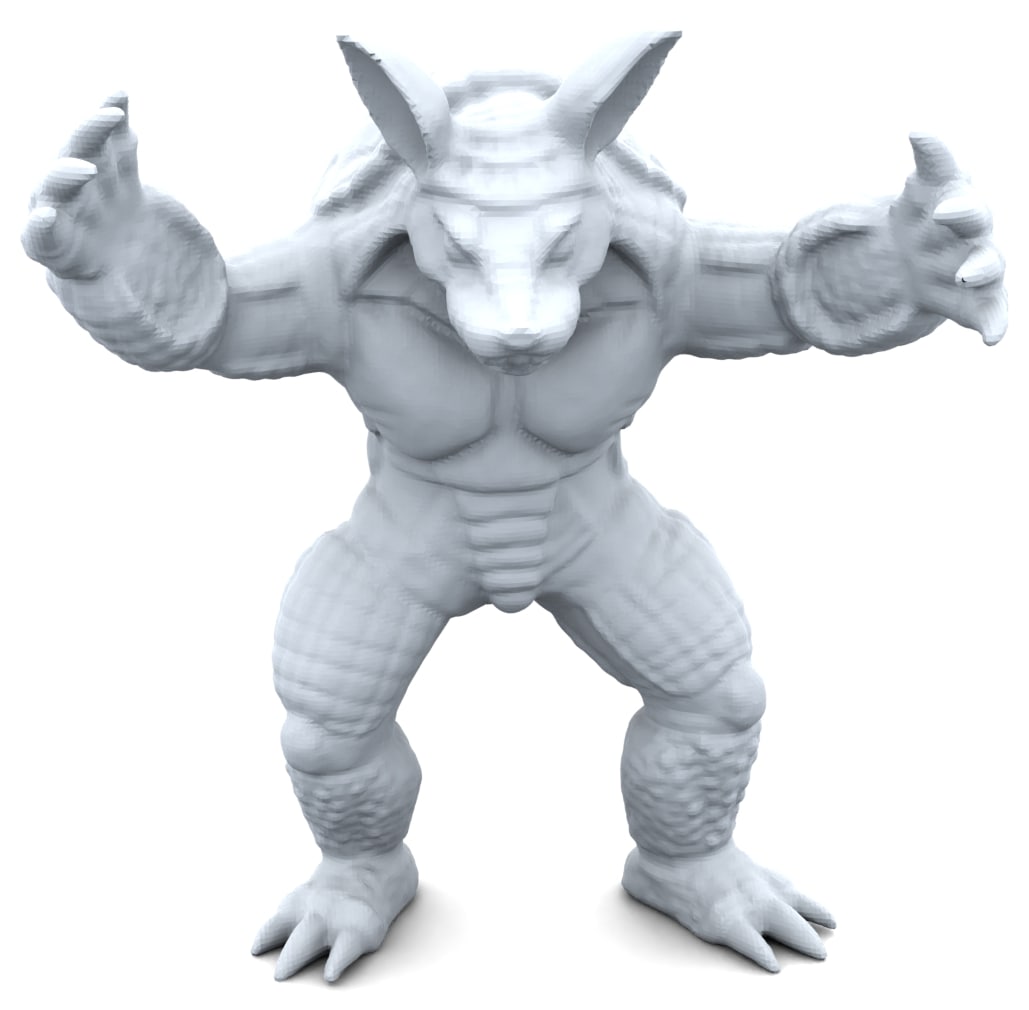 }
    \includegraphics[width=1.35in]{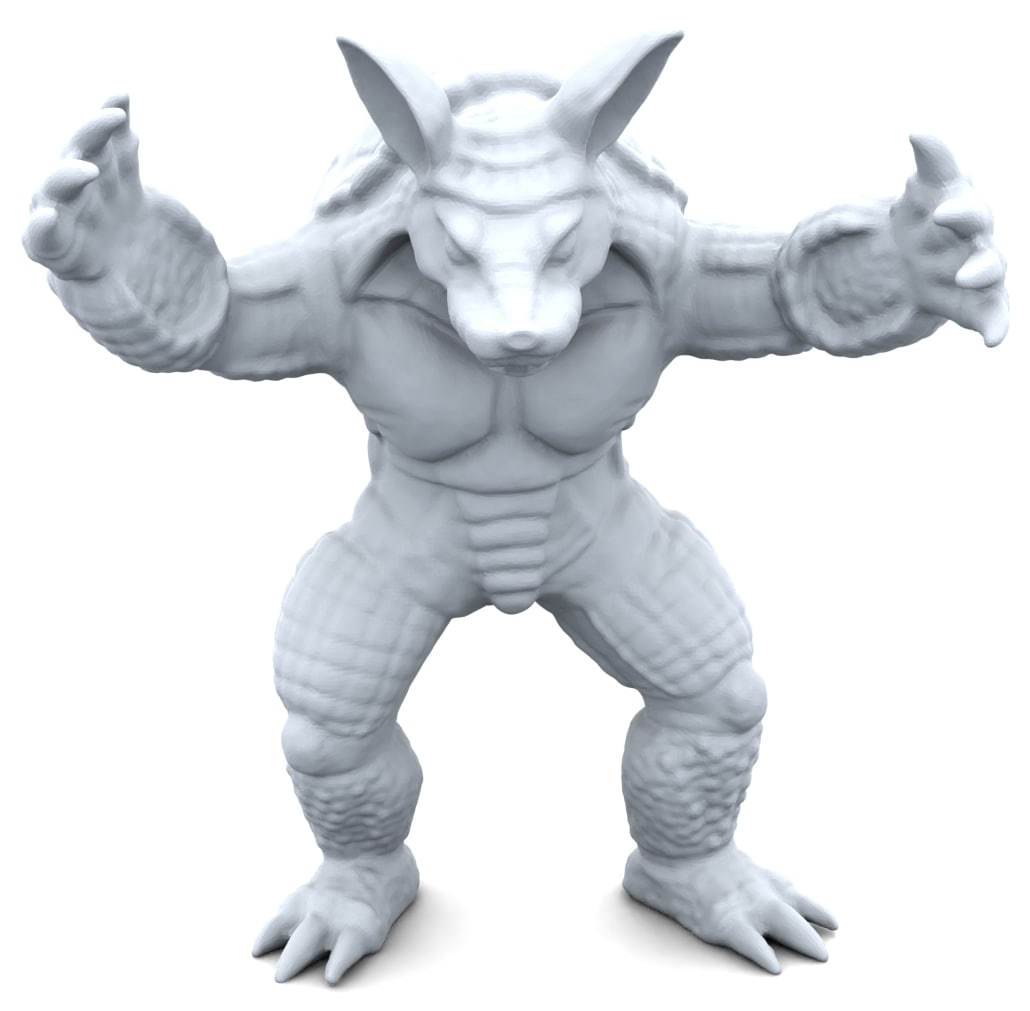}\\
    \includegraphics[width=1.35in]{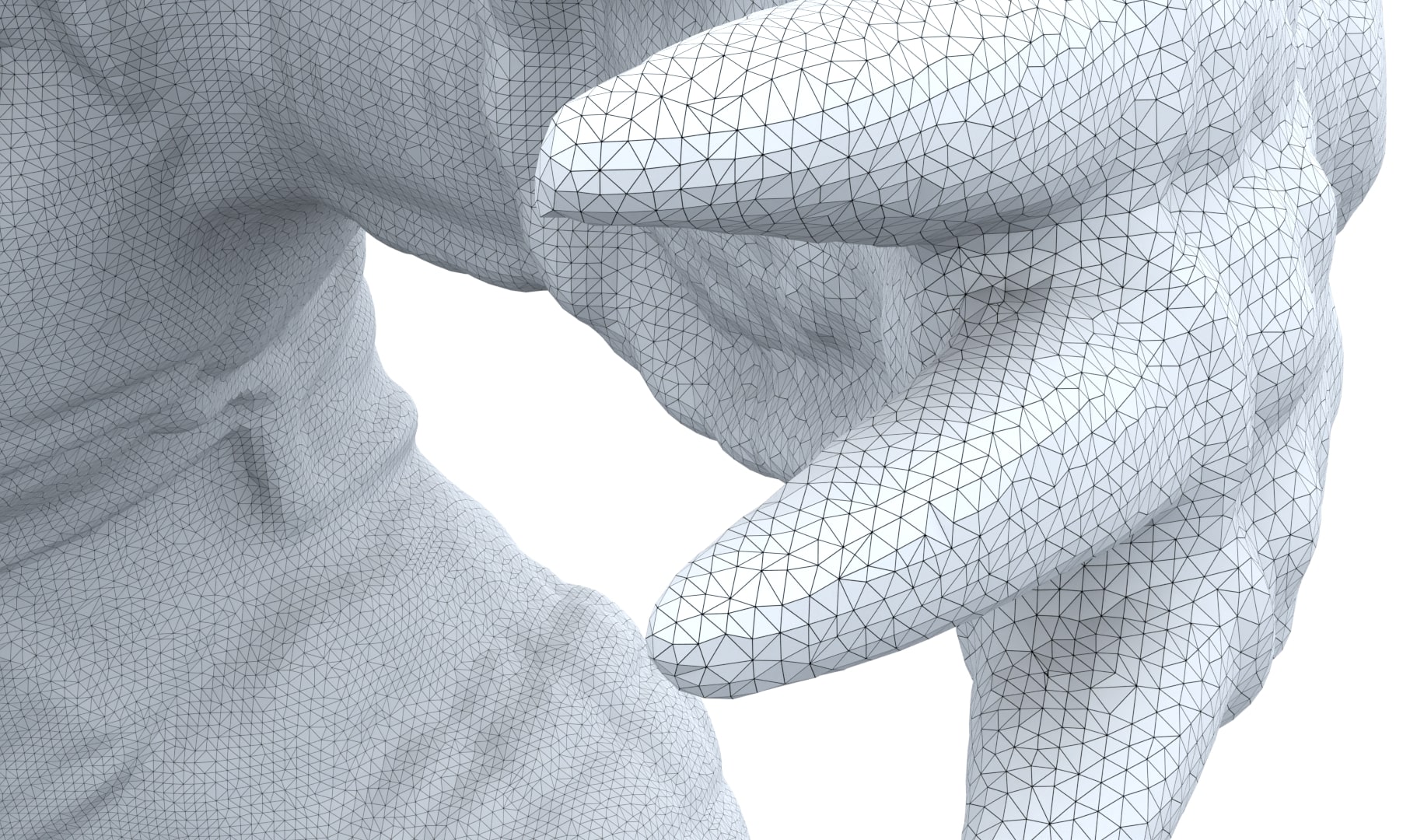 }
    \includegraphics[width=1.35in]{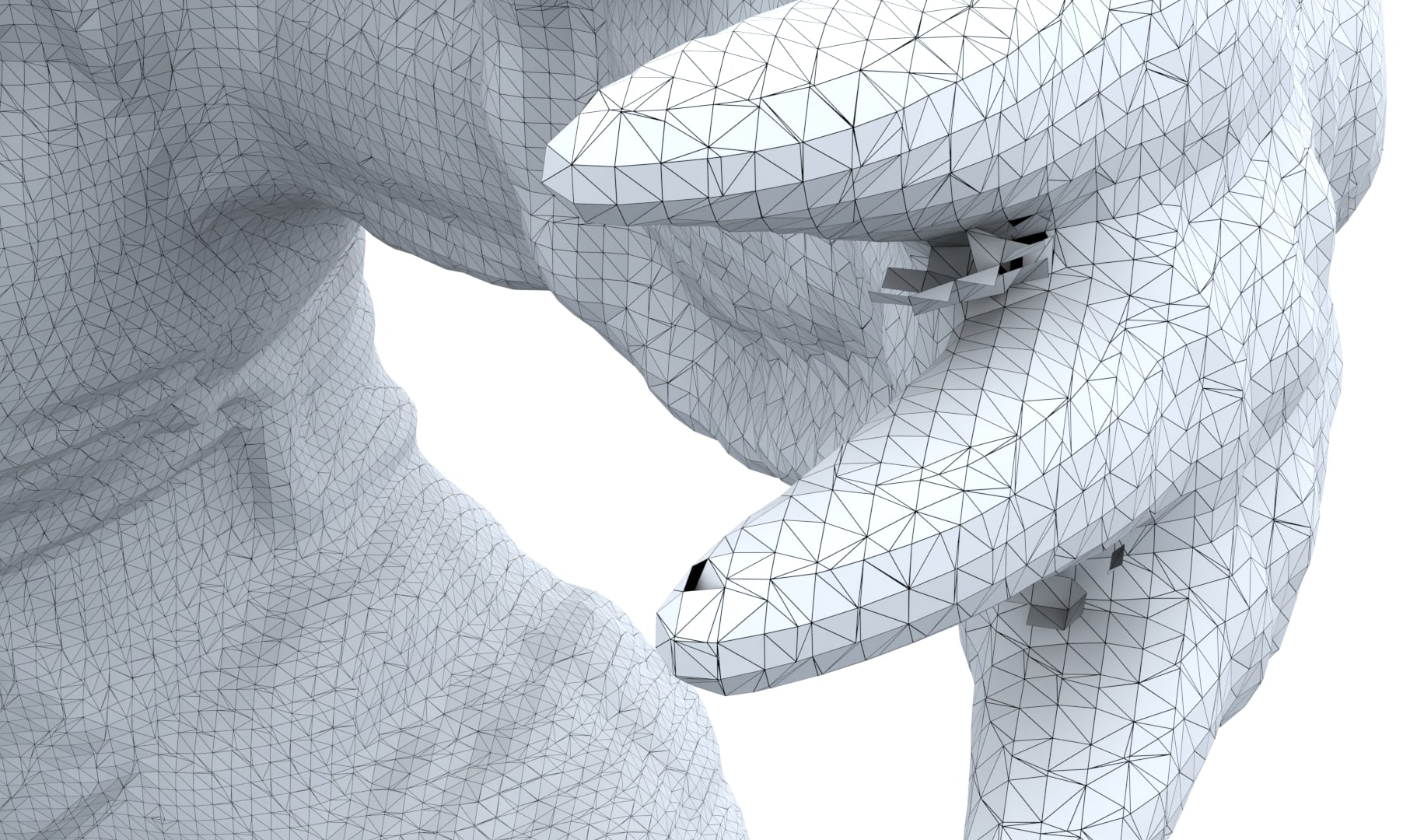 }
    \includegraphics[width=1.35in]{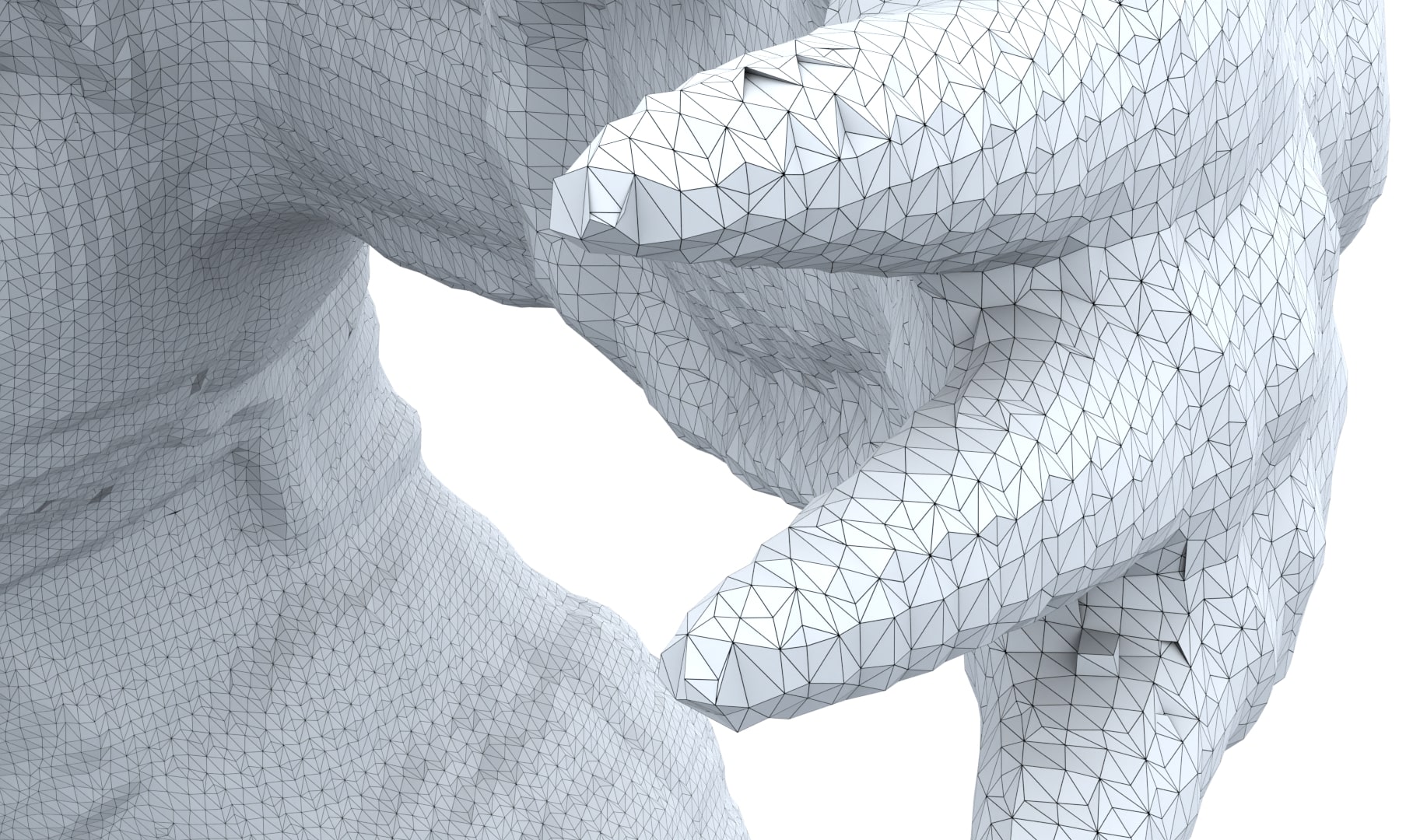 }
    \includegraphics[width=1.35in]{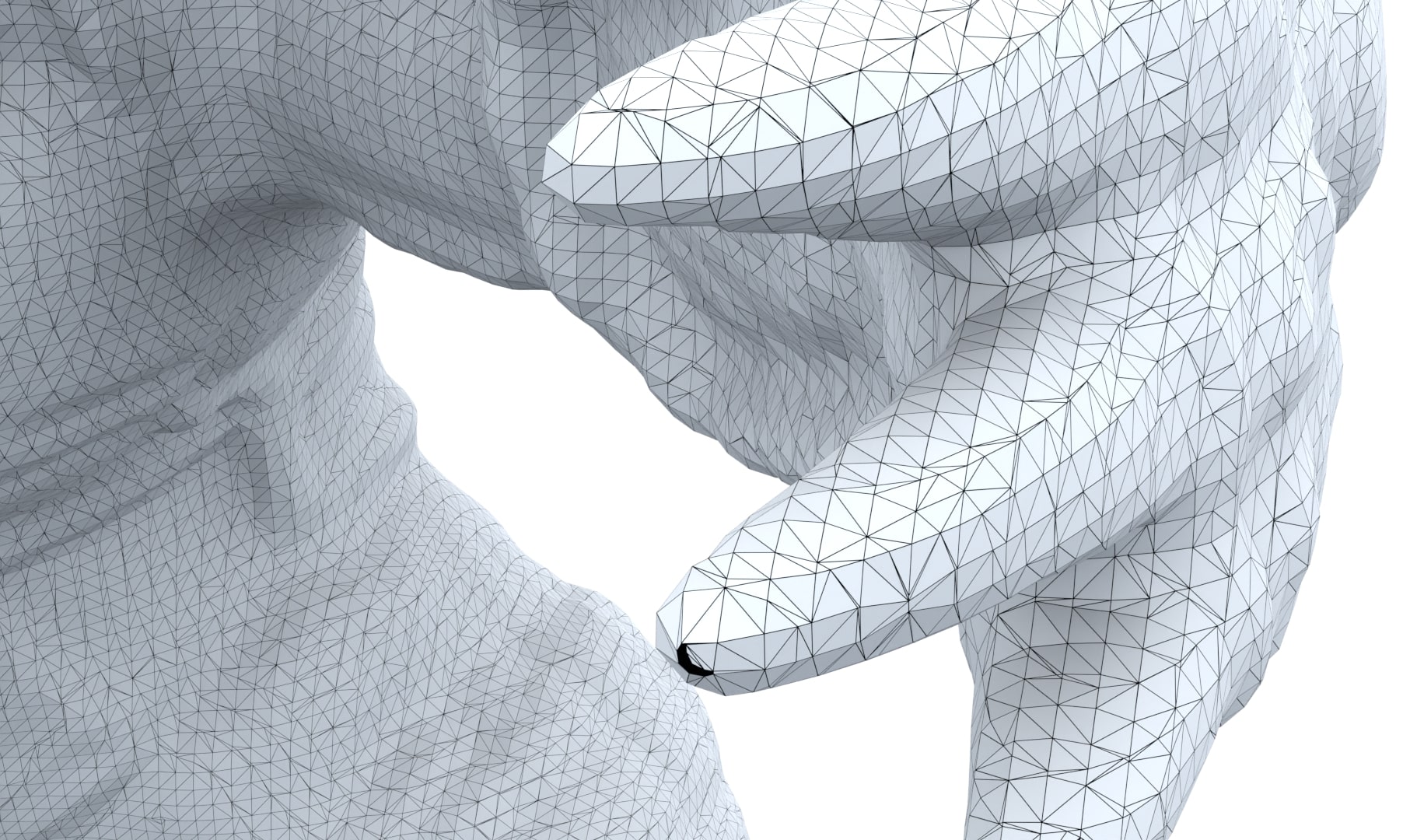 }
    \includegraphics[width=1.35in]{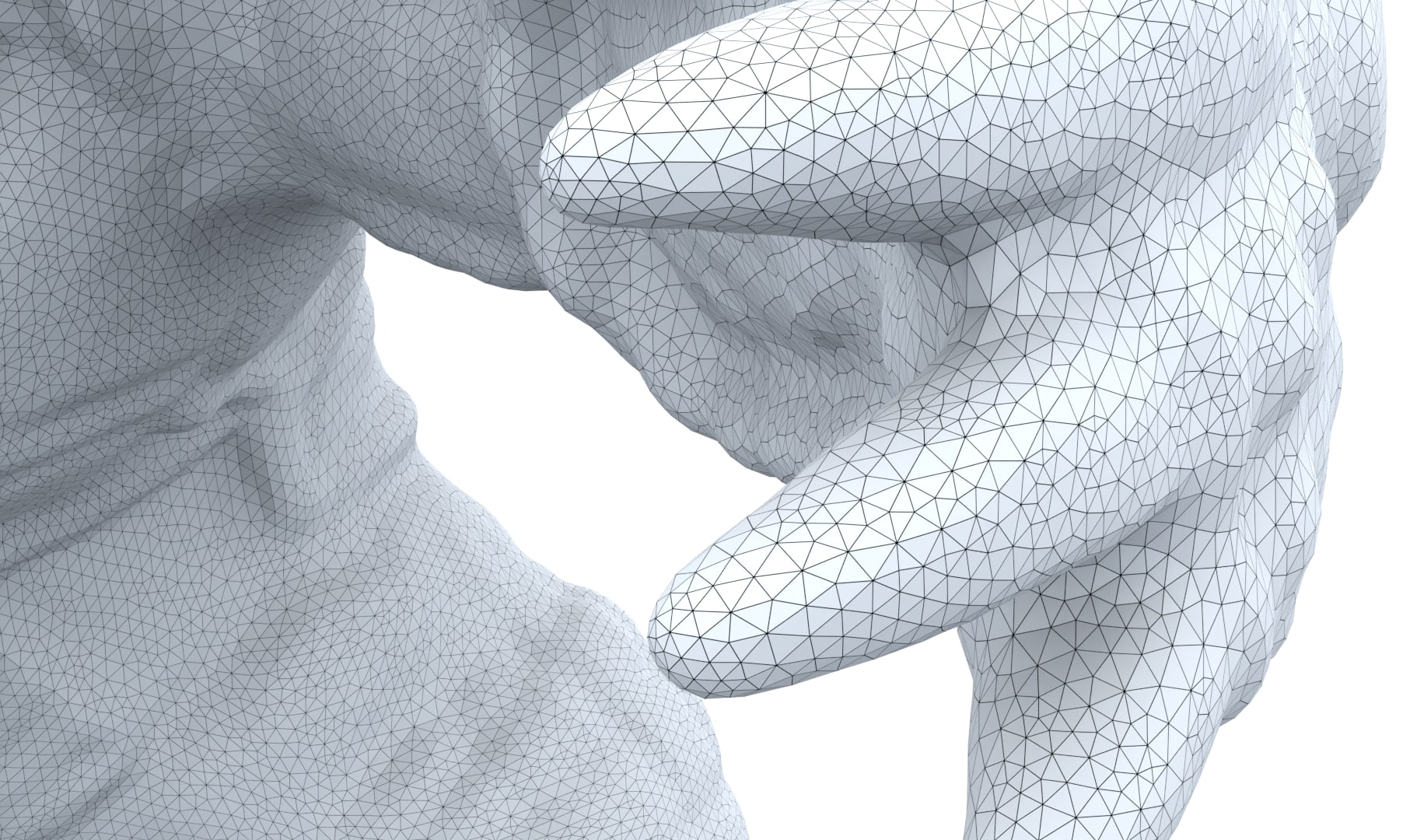}\\
    \makebox[1.35in]{GT mesh (172k vertices)}
    \makebox[1.35in]{MeshCAP $256^3$}
    \makebox[1.35in]{UNDC $256^3$}
    \makebox[1.35in]{MeshUDF $256^3$}
    \makebox[1.35in]{DCUDF $256^3$}\\
    \caption{Comparisons with UNDC, MeshUDF and MeshCAP on accurate UDFs, which are converted from SDFs of high-resolution watertight meshes. Due to its  high GPU memory consumption, UNDC can only work at a resolution $256^3$.}
    \label{fig:high_res}
\end{figure*}

\paragraph{Comparison with UNDC on highly accurate UDFs}
UNDC~\cite{Chen2022NDC} is a highly efficient data-driven approach that utilizes a pre-trained neural network to predict surface gradients and intersection points from UDF, enabling the generation of triangular faces through dual contouring~\cite{Ju2002}.
Due to its supervised nature, the performance of UNDC heavily relies on the training data used. We observed that the pre-trained UDF model faces challenges in  generating satisfactory results on the Deep Fashion3D dataset. Retraining the model is a non-trivial task due to the lack of ground-truth meshes, which results in the absence of labelled data for dual contouring. Therefore, for quantitative comparisons, we employ highly accurate distance fields that allow UNDC to produce more reasonable outcomes.
Specifically, we use IDF~\cite{Wang2022IDF} to learn accurate SDFs of watertight models and convert them to UDFs using the absolute function, resulting in precise UDFs. As shown in Table~\ref{tab:accurate_udf}, the Chamfer distances of the meshes generated by all methods are similar to the standard marching cubes algorithm, indicating similar geometric qualities given accurate UDFs as input. However, there are significant differences in terms of topological measures. Their extracted meshes often exhibit many non-manifold vertices and edges, while our method ensures the absence of such structures. Moreover, our method excels in capturing global topological features, e.g., the genus and the number of boundaries, outperforming the other methods. The results are illustrated in Figure~\ref{fig:high_res}.

\paragraph{Comparison with boundary voxel methods}
Boundary voxel methods~\cite{Koo2005}\cite{Wang2005} first identify a set of voxels that contain the zero level-set, and then shrink the boundary surface of these voxels to obtain the target surface. The quality of the extracted mesh highly depends on the initial boundary voxels. Koo et al.'s method~\shortcite{Koo2005} divides the 3D space into cells, using the faces of cells that contain at least one input point to form the initial mesh. The method then iteratively moves each corner of a boundary voxel toward the closest point in the input cloud and applies Laplacian smoothing. While this projection-smoothing strategy is easy to implement, it is problematic for non-uniform point clouds, as uneven points often distort the shrinking directions, yielding many flipped and self-intersecting triangles. Moreover, the contraction process may not yield a single-layered mesh for noise points, leaving a gap between the two contracting sides of the initial mesh. Their initial meshes may contain numerous non-manifold structures, where two adjacent voxels share only a common edge. Once those non-manifold structures are present, they cannot be eliminated using their method. 
Wang et al.~\shortcite{Wang2005} proposed several heuristics (such as topological thinning) to mitigate such artifacts. However, their method requires the boundary voxels must be single-layered, which cannot be guaranteed in practice. Furthermore, their evaluation was primarily focused on simple synthetic models, leaving uncertainty regarding the effectiveness of their thinning strategy for real-world data that often exhibit diverse types of defects.

Our method is fundamentally different from boundary voxel methods in both the generation of the initial mesh and the subsequent shrinking process. Unlike these methods, our initial mesh is generated using double cover, which is guaranteed to be an orientable 2-manifold. This removes the issue of non-manifold structures that often plague other techniques. Additionally, we do not rely on the initial surface having a unit thickness. Instead, we simply control it by specifying a global parameter $r$, giving us greater flexibility and control. We shrink the mesh in an optimization process taking into account the centroids of the faces. This additional constraint has proven to be effective in reducing reconstruction error, further distinguishing our method from existing boundary voxel approaches.

\begin{figure*}[!htbp]
    \centering
    \includegraphics[width=1.6in]{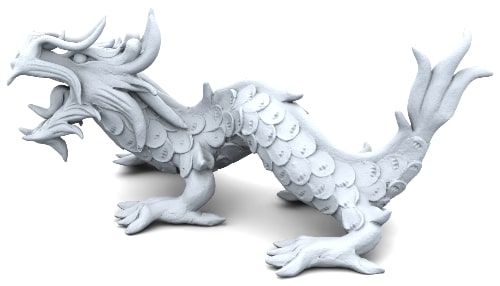 }
    \includegraphics[width=1.6in]{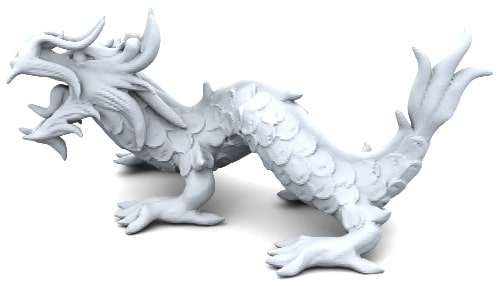 }
    \includegraphics[width=1.6in]{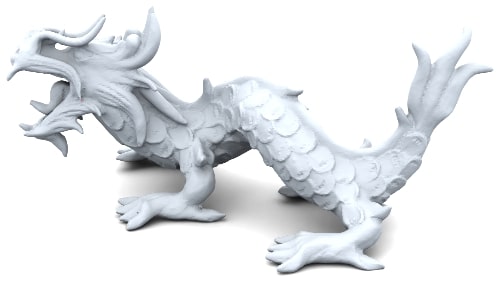 }
    \includegraphics[width=1.6in]{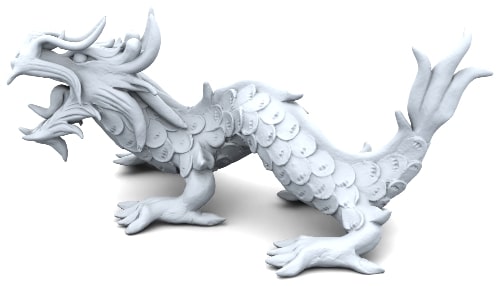 }\\
    \includegraphics[width=1.6in]{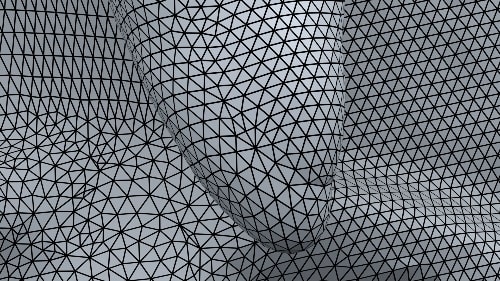 }
    \includegraphics[width=1.6in]{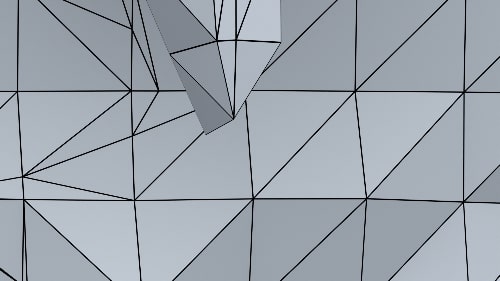 }
    \includegraphics[width=1.6in]{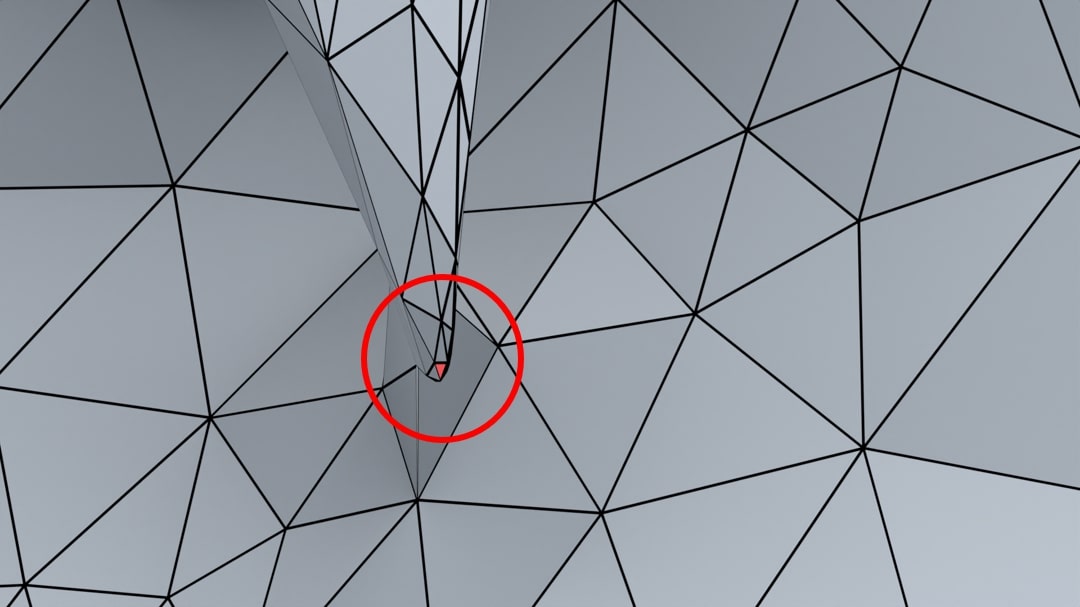 }
    \includegraphics[width=1.6in]{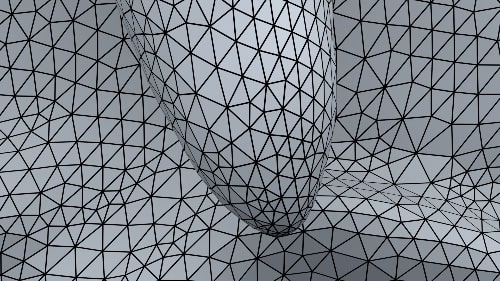 }\\
    \begin{scriptsize}
    \makebox[1.6in]{\makecell[c]{(a) GT \\ 0 self-intersections}} 
    \makebox[1.6in]{\makecell[c]{(b) MC $256^3$\\ 0 self-intersections}} 
    \makebox[1.6in]{\makecell[c]{(c) DCUDF $256^3$\\ 4 self-intersections}}
    \makebox[1.6in]{\makecell[c]{(d) DCUDF $1024^3$\\0 self-intersections}}
\end{scriptsize}
\caption{Limitation. Our method cannot guarantee the absence of self-intersections, which are likely to occur in models featuring sharp tips when rendered at low resolutions and low accuracy. Typically, increasing the MC resolution can reduce the occurrence of self-intersections. The self-intersecting triangles are highlighted in red.}
\label{fig:selfintersection}
\end{figure*}

\section{Limitations}

Firstly, our approach is optimization-driven, inherently demanding more computational time compared to non-optimization or non-iterative methods such as UNDC, MeshUDF and MeshCAP. As reported in Table~\ref{tab:df3d_time}, our method runs 5 times slower than MeshUDF on the Deep Fashion3D dataset. Nonetheless, this increased computational burden is offset by the improved quality of the extracted meshes.

\begin{table}[!htbp]
\caption{\label{tab:df3d_time}Running time (seconds) breakdown on the Deep Fashion3D dataset with 598 models. We set the marching cubes resolution of $256^3$ across all methods. Both MeshCAP and MeshUDF require computing the distance and gradient for each grid point. Our mesh generation process is optimization-driven, which results in longer computational times compared to MeshCAP and MeshUDF. Following the mesh extraction, our method necessitates a min-cut postprocessing step to transition from a double-layered to a single-layered output.}
\begin{small}
\begin{tabular}{l|c|c|c}
\hline
                      & MeshCAP         & MeshUDF & Ours   \\
                      \hline
                      \hline
    Grid extraction    &    34.36     &   34.36 & / \\
   
   Marching cubes & /  & / & 5.94  \\
   Mesh generation    &    155.91    &   10.51 &   201.18\\
 Min-cut           &   /          &   /     &   17.60\\
\hline
\end{tabular}
\end{small}
\end{table}

Secondly, our method cannot guarantee that the extracted meshes are free of self-intersections, a characteristic inherently found in variants of the marching cubes method like MeshCAP and MeshUDF. Such self-intersections are typically observed in models featuring sharp tips or thin handles, especially when processed at lower resolution. Figure~\ref{fig:selfintersection}(a)-(c) illustrates such a situation where intersections appear at a sharp tip due to a relatively low resolution $256^3$ of matching cubes and low accuracy. This sharp tip collapses into self-intersecting triangles when shrinking the double covered mesh into a single layer. Despite of their occurrence, these self-intersecting triangles are typically rare and isolated, thereby having little impact on the overall quality of the extracted meshes. Additionally, increasing the resolution of the marching cubes has proven effective in reducing the frequency of self-intersections. For example, at an MC resolution of $1024^3$, the reconstruction is more accurate, with sharp tips becoming rounded; consequently, the shrinking avoids the creation of self-intersecting triangles, as shown in  Figure~\ref{fig:selfintersection}(d).

Thirdly, when dealing with target surfaces that contain very thin structures, it is necessary to choose both a sufficiently small $r$ value and a sufficiently high marching cubes resolution to accurately capture the correct topological features. This, however, can significantly increase the computation resources required. A possible way is to adopt adaptive resolution in marching cubes, such as sparse voxel octrees~\cite{Laine2010}.

\begin{figure*}[!htbp]
    \centering
    \makebox[1.32in]{GT}
    \makebox[1.32in]{NDC $256^3$}
    \makebox[1.32in]{DCUDF $256^3$}
    \makebox[1.32in]{DCUDF $512^3$}
    \makebox[1.32in]{DCUDF $1024^3$}\\
    \includegraphics[width=1.32in]{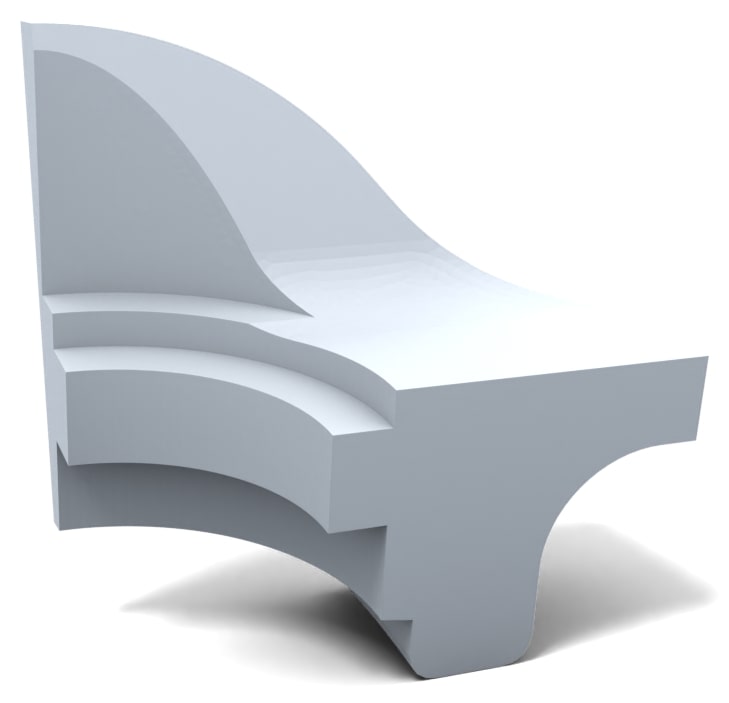 }
    \includegraphics[width=1.32in]{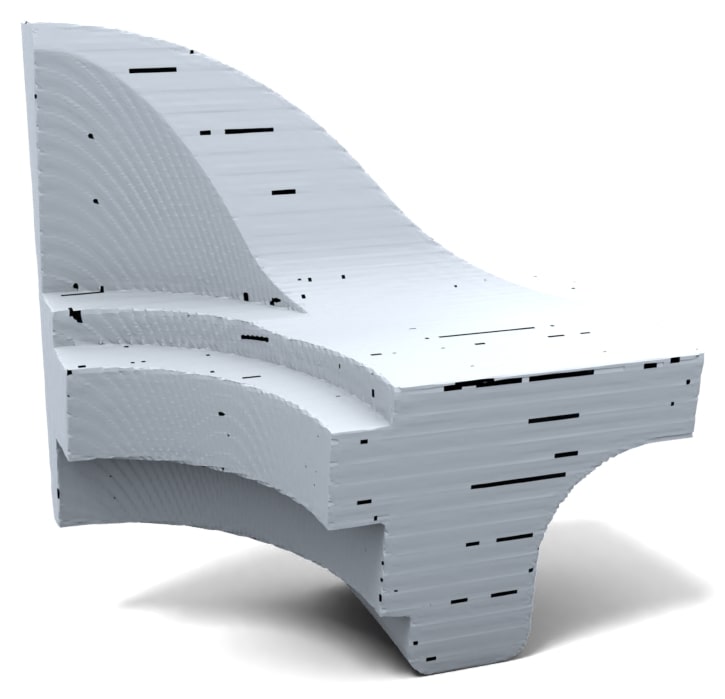 }
    \includegraphics[width=1.32in]{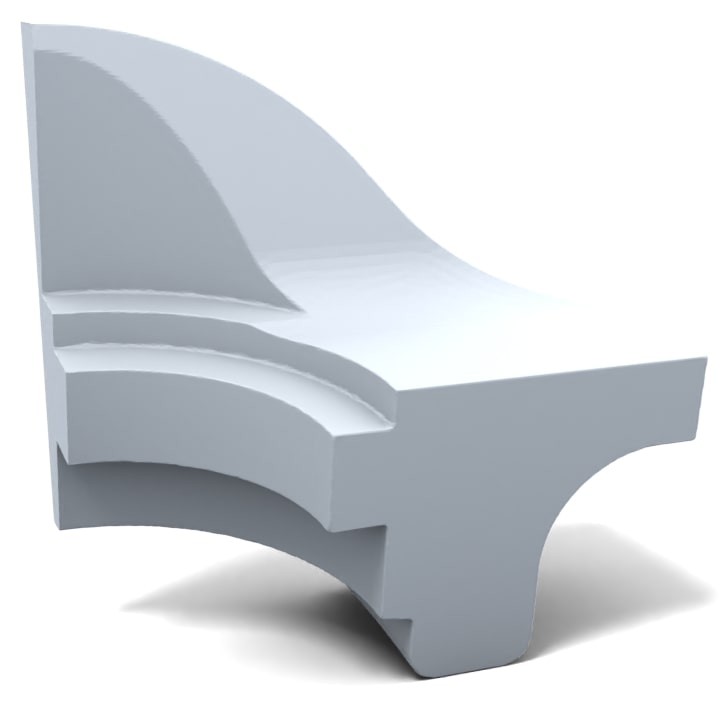 }
    \includegraphics[width=1.32in]{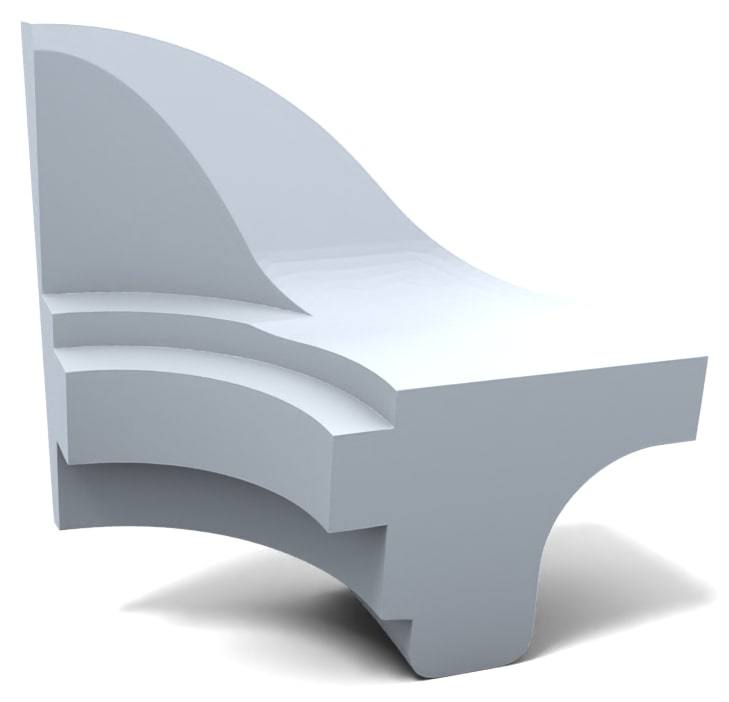 }
    \includegraphics[width=1.32in]{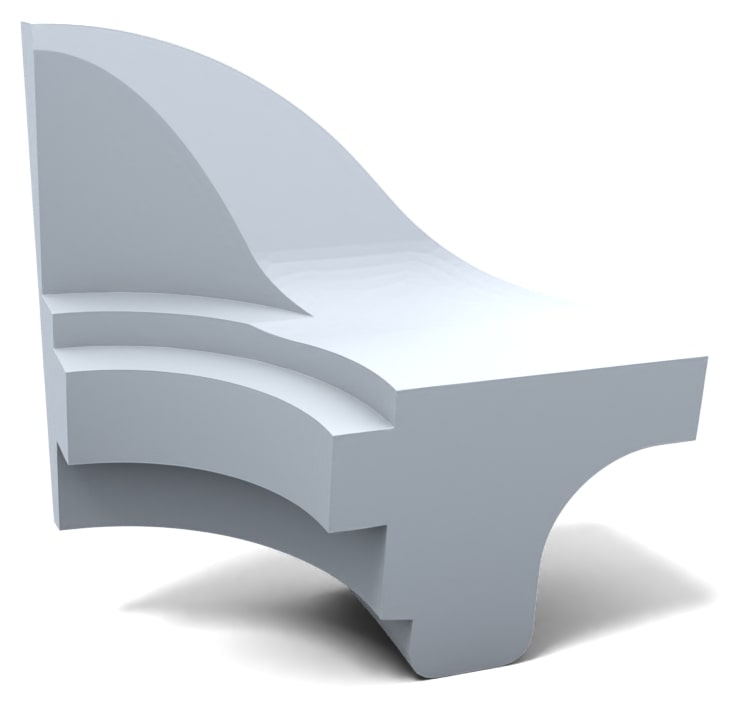 }\\
    \includegraphics[width=1.32in]{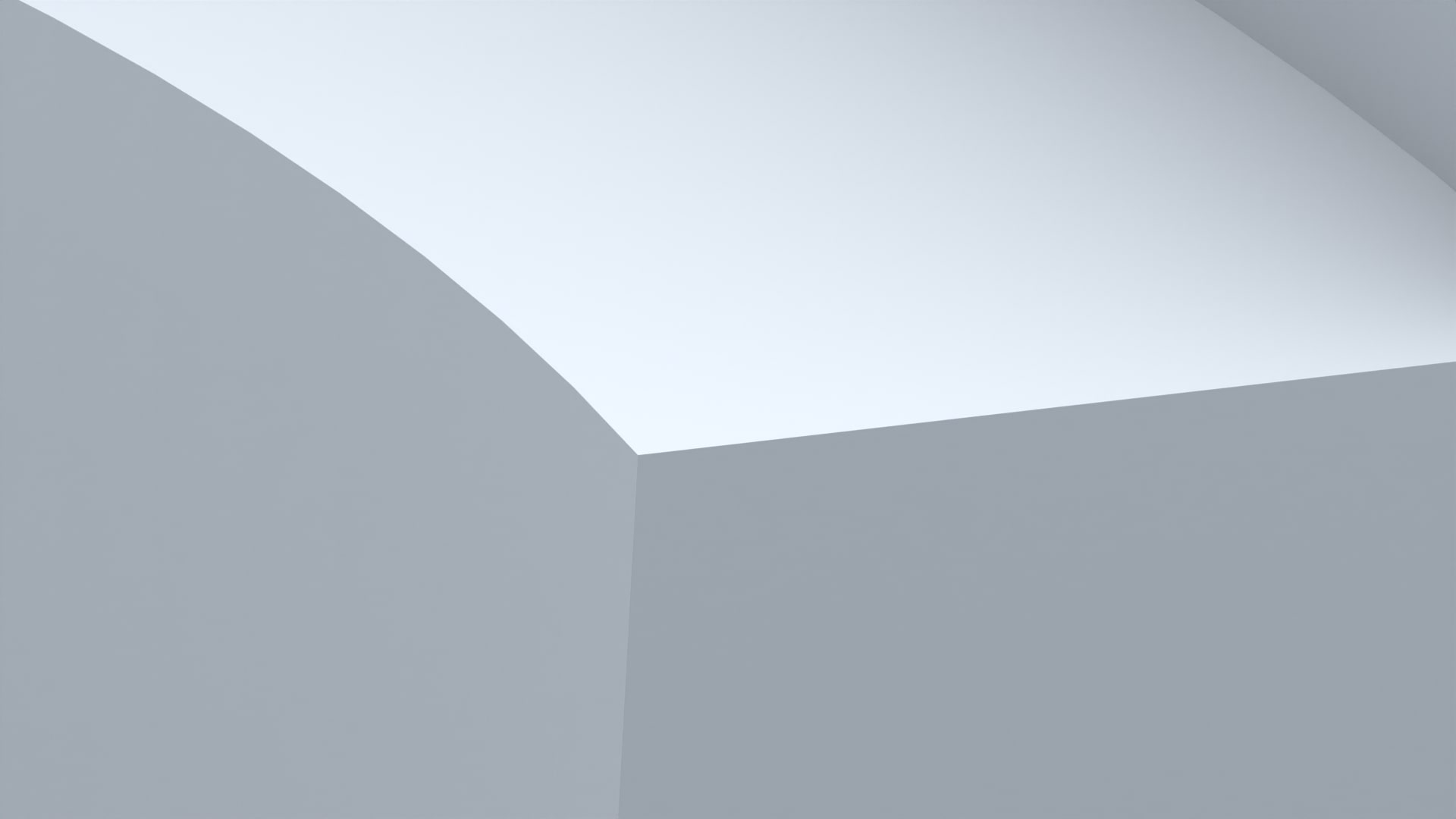 }
    \includegraphics[width=1.32in]{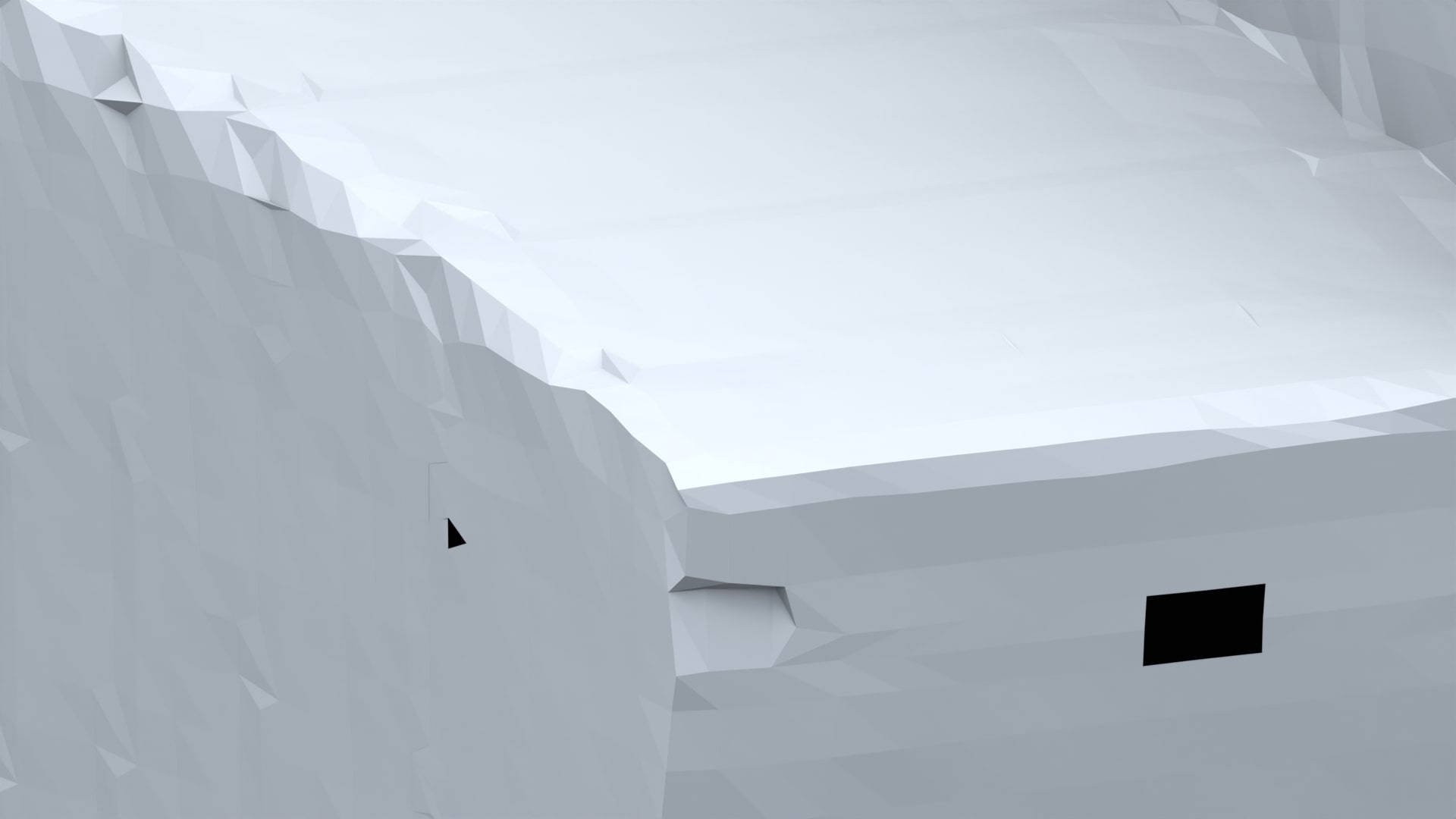 }
    \includegraphics[width=1.32in]{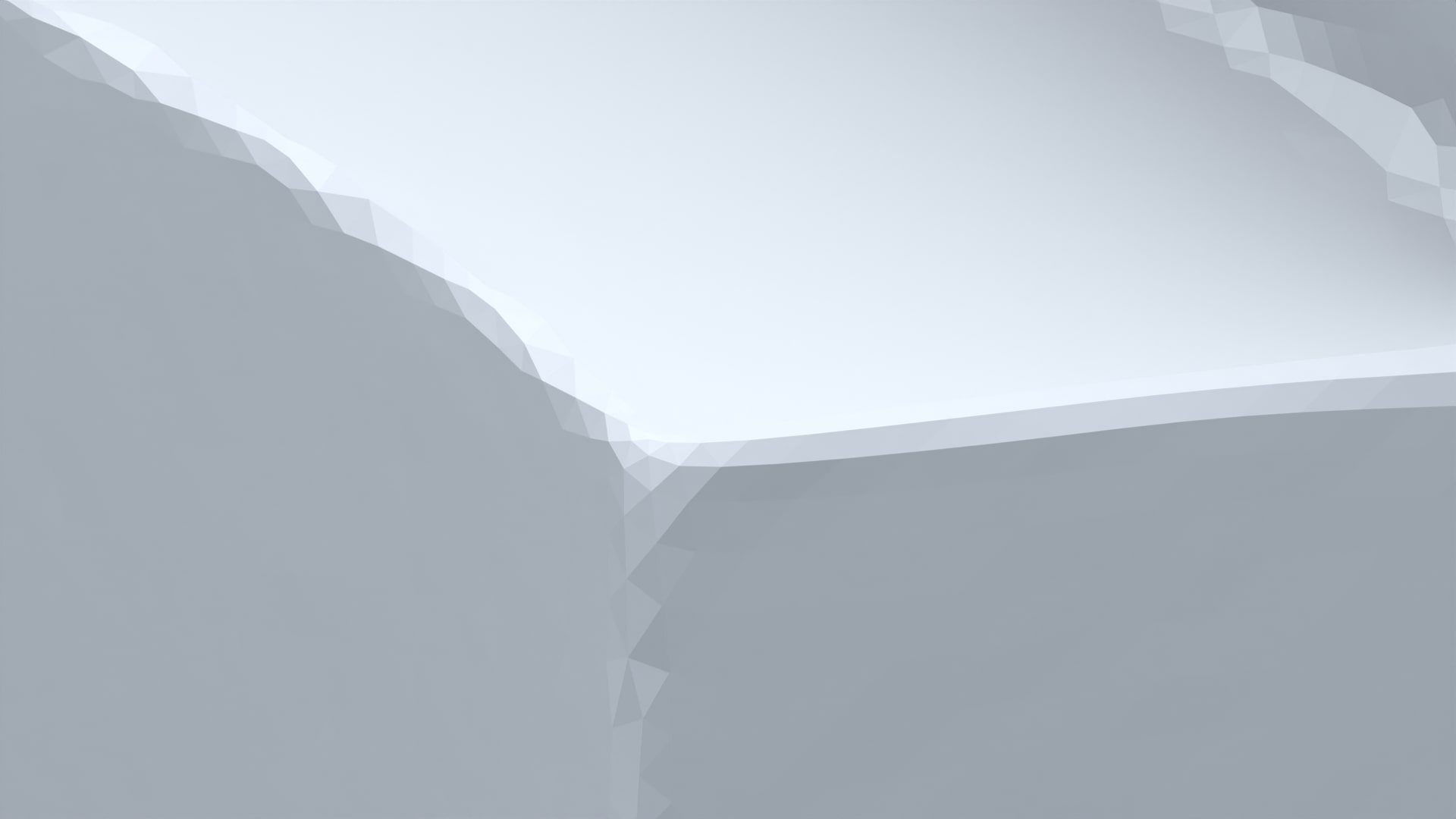 }
    \includegraphics[width=1.32in]{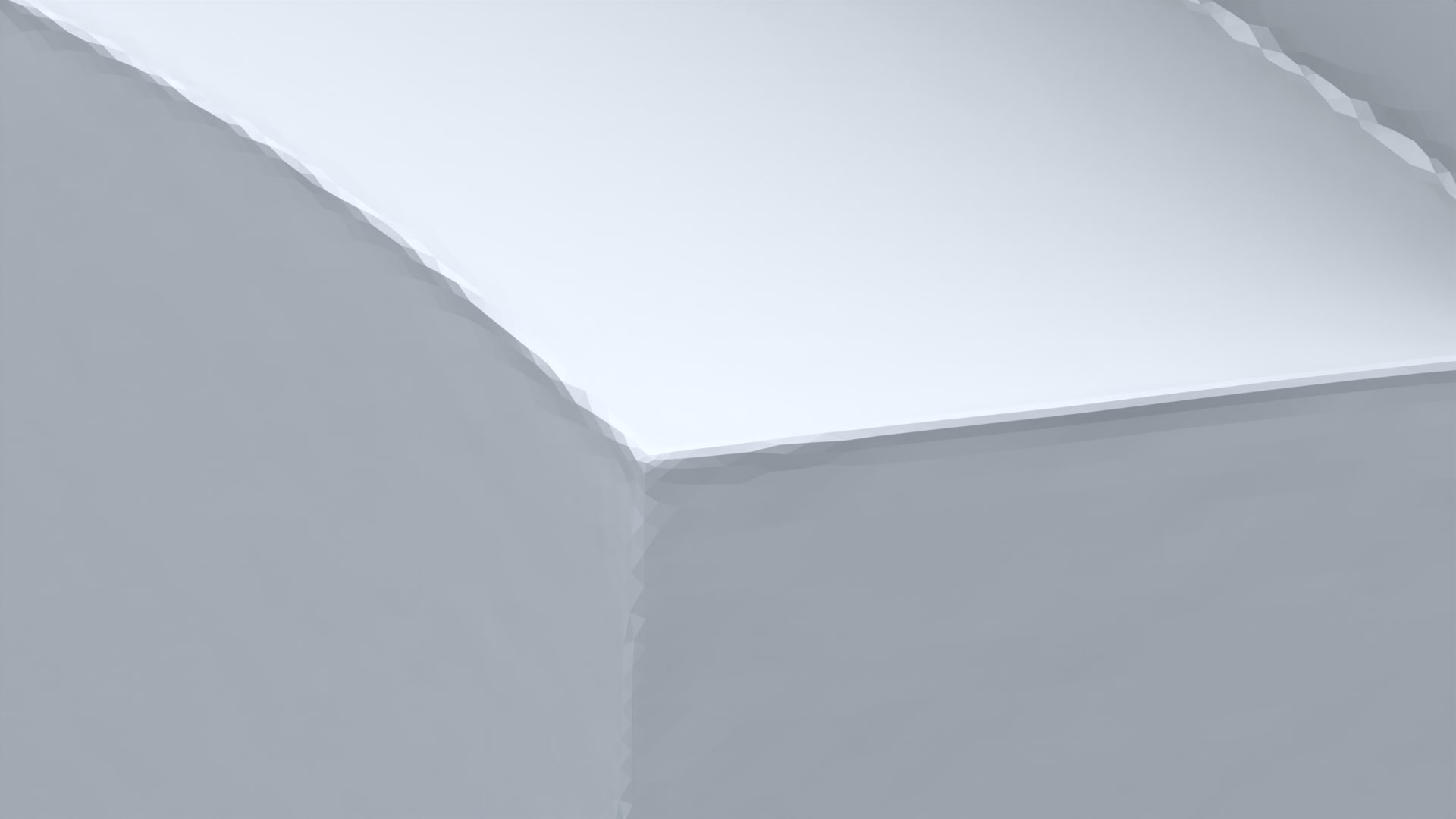 }
    \includegraphics[width=1.32in]{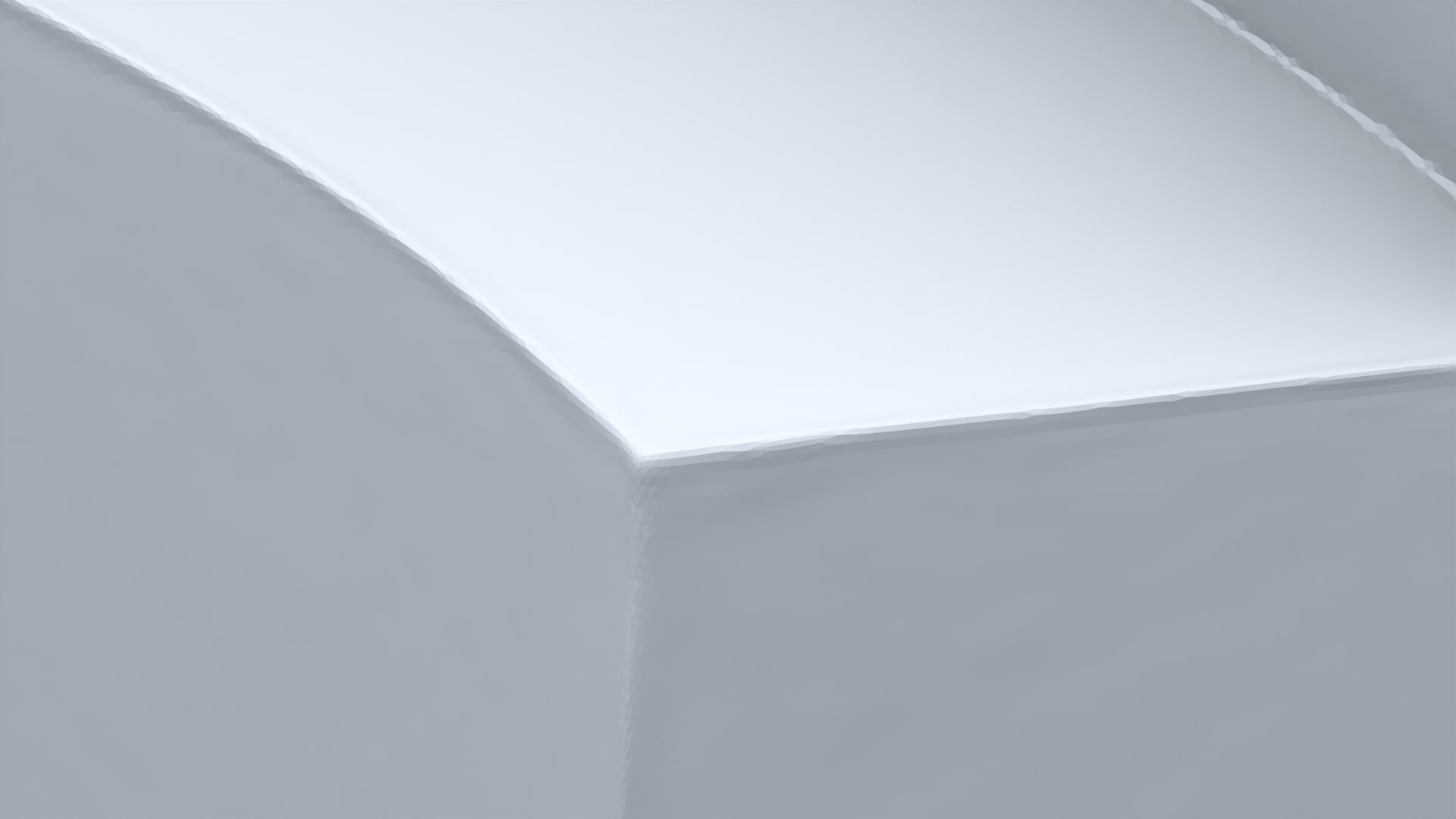 }\\
    \makebox[1.32in]{}
    \makebox[1.32in]{CD = 0.433$\times 10^{-3}$}
    \makebox[1.32in]{CD = 0.330$\times 10^{-3}$}
    \makebox[1.32in]{CD = 0.311$\times 10^{-3}$}
    \makebox[1.32in]{CD = 0.306$\times 10^{-3}$}
    \caption{Although our method does not inherently preserve sharp features, increasing the MC resolution generally improves DCUDF's capability to recover sharp edges in the extracted meshes.}
    \label{fig:sharp}
\end{figure*}

Fourthly, our method is not tailored to preserve sharp features explicitly. Consequently, sharp edges and corners may appear smoothed, particularly when the input UDF is of low accuracy. This smoothing artifact can be reduced by increasing the UDF accuracy and the marching cubes resolution. See Figure~\ref{fig:sharp}.

Lastly, DCUDF requires users to explicitly indicate the type of the model - whether it is closed, open, or non-manifold/non-orientable - in order to select the appropriate post-processing step. Future research could explore automating this aspect of the algorithmic pipeline.

\begin{figure*}[!htbp]
\includegraphics[width=3.3in]{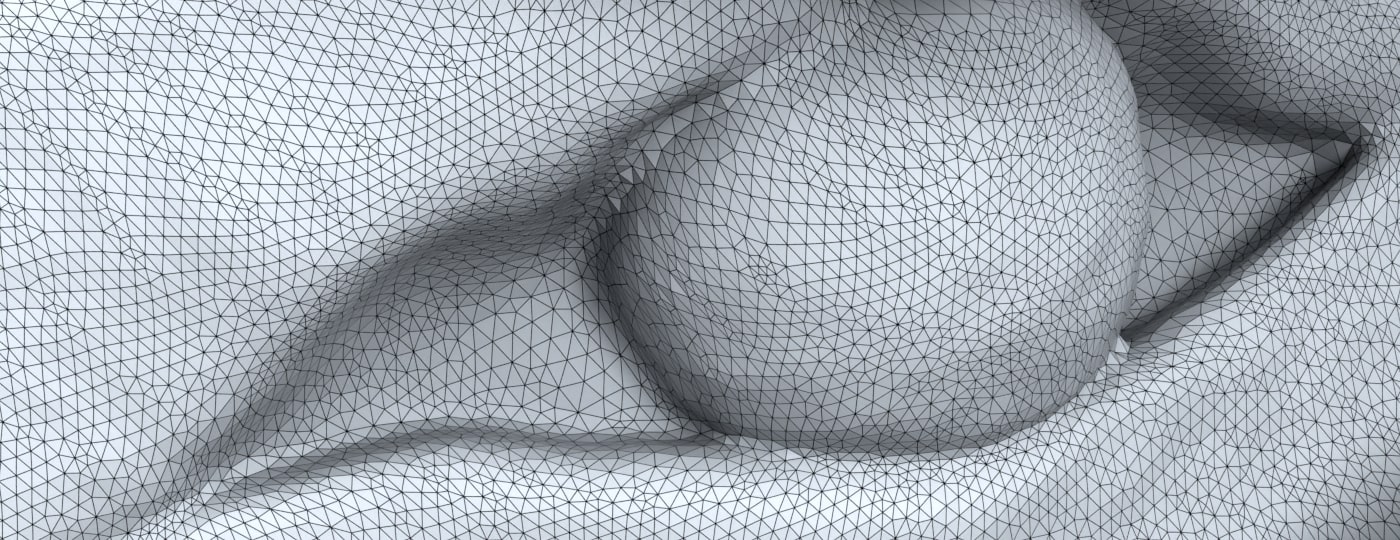}
\hspace{0.1in}
\includegraphics[width=3.3in]{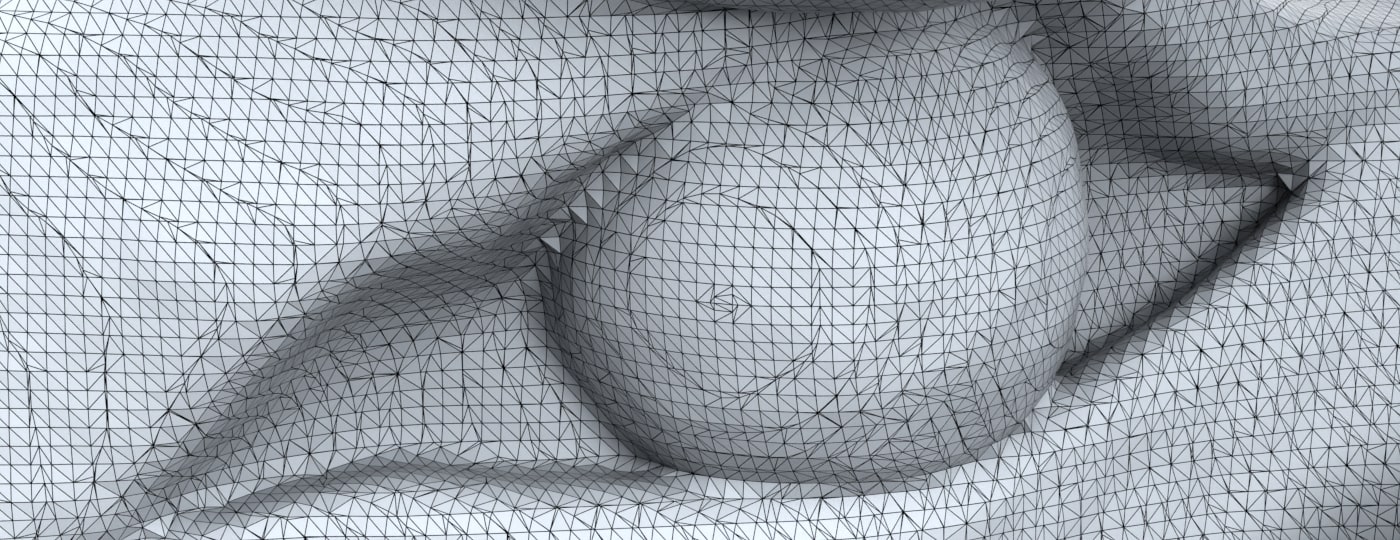}
\\
\caption{Our method is memory efficient and can extract meshes with a resolution of $1024^3$ on a standard graphics card. 
Our extracted meshes (left) are free from the typical presence of obtuse triangles often found in marching cubes results (right), making them well-suited for direct use in downstream applications.}
\label{fig:dragonhead}
\end{figure*}

\begin{figure}[!htbp]
    \centering
    \small
    \makebox[1.1in]{Input points}
    \makebox[1.1in]{Simple MLP + DCUDF}
    \makebox[1.1in]{CAPUDF + DCUDF}\\
    \includegraphics[width=1.1in]{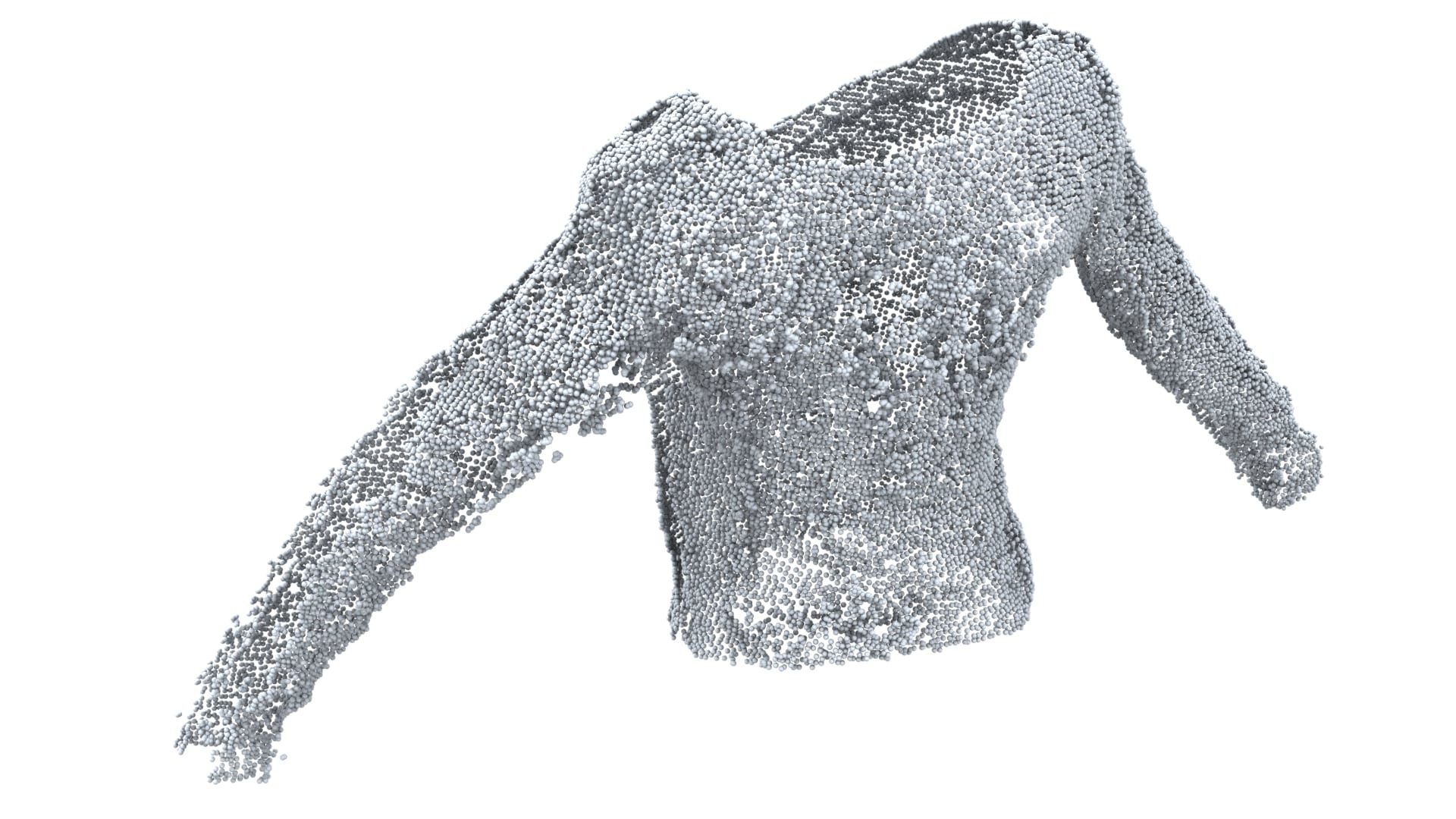 }
    \includegraphics[width=1.1in]{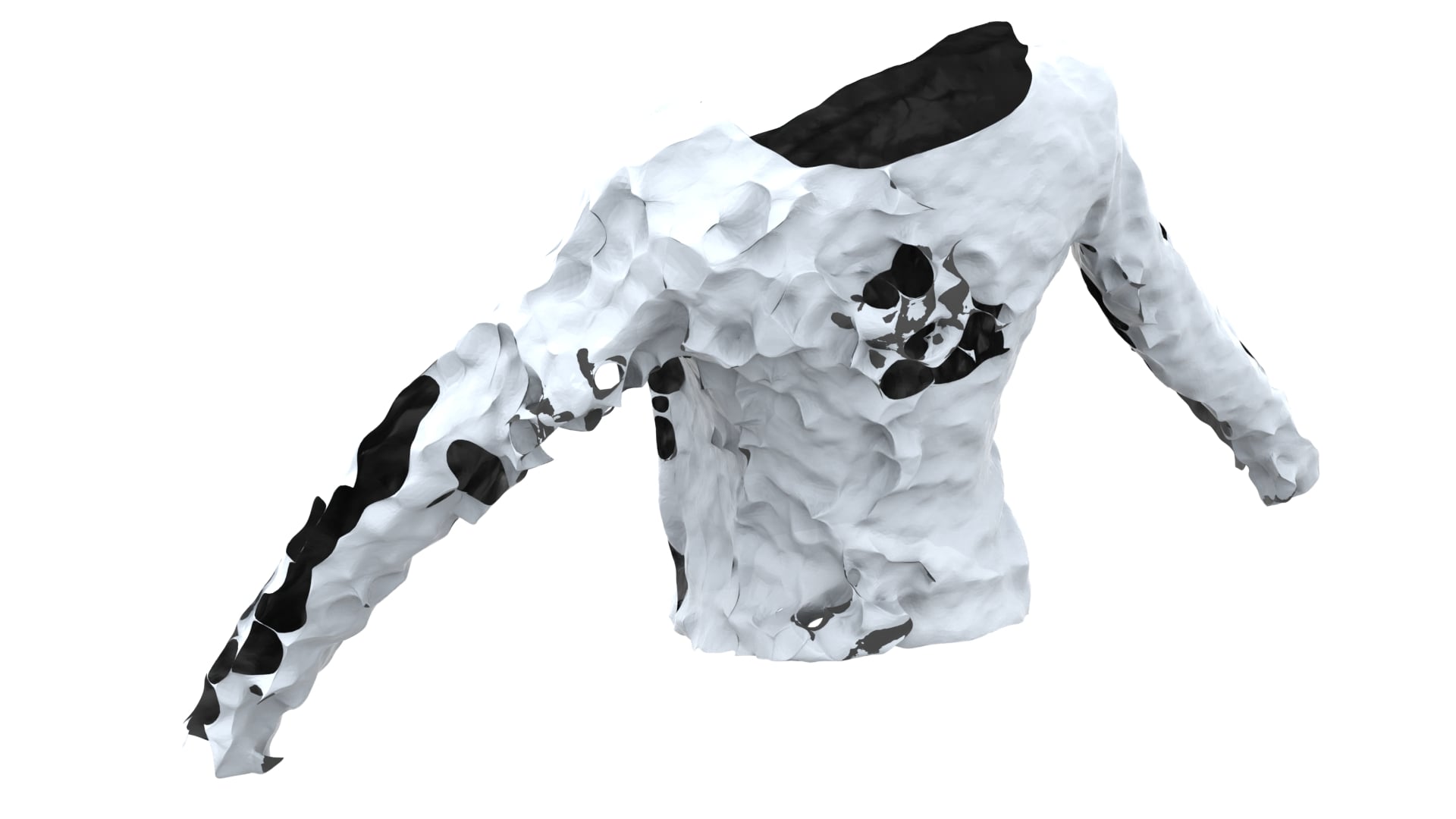 }
    \includegraphics[width=1.1in]{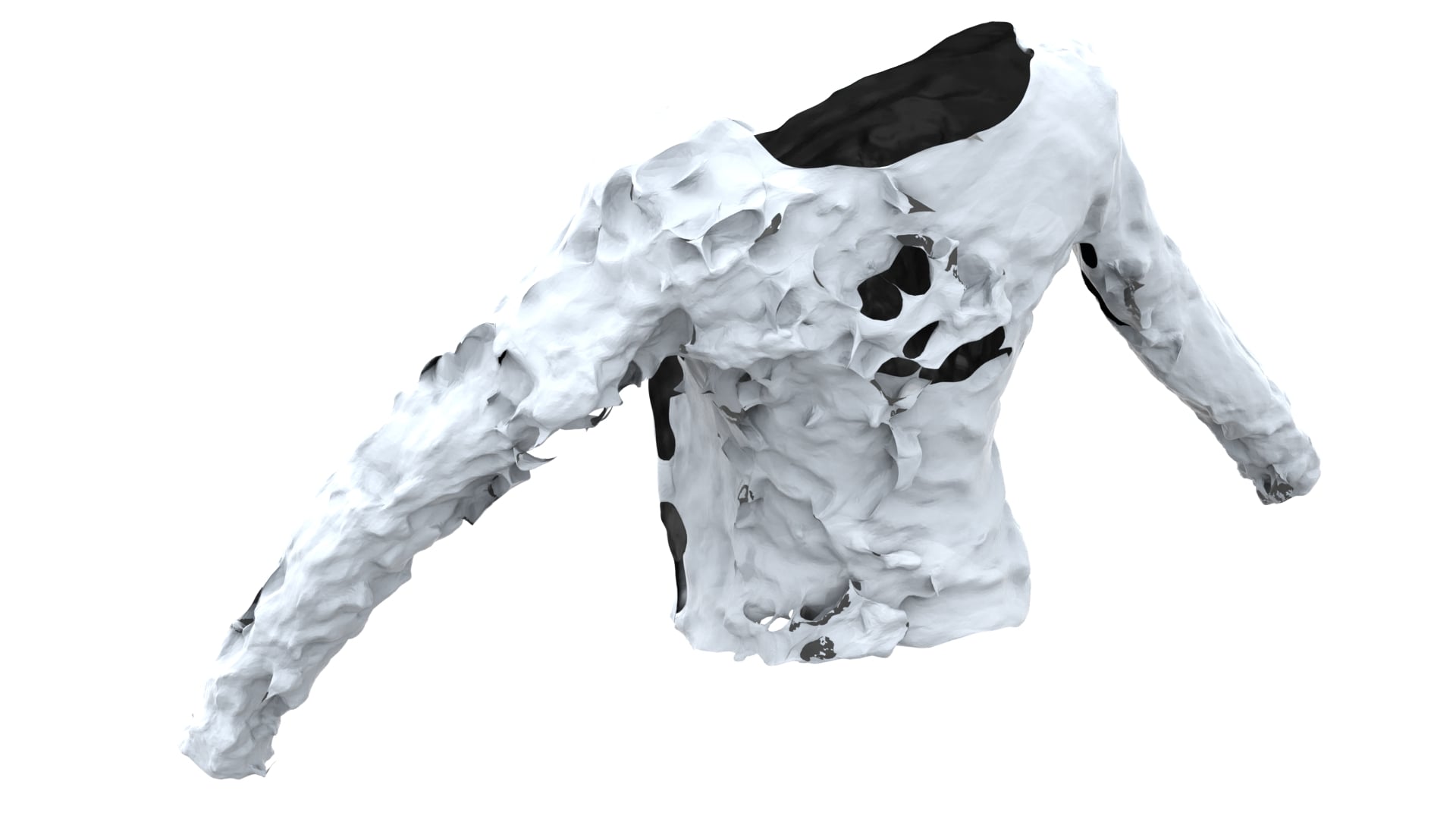 }\\
    \includegraphics[width=1.1in]{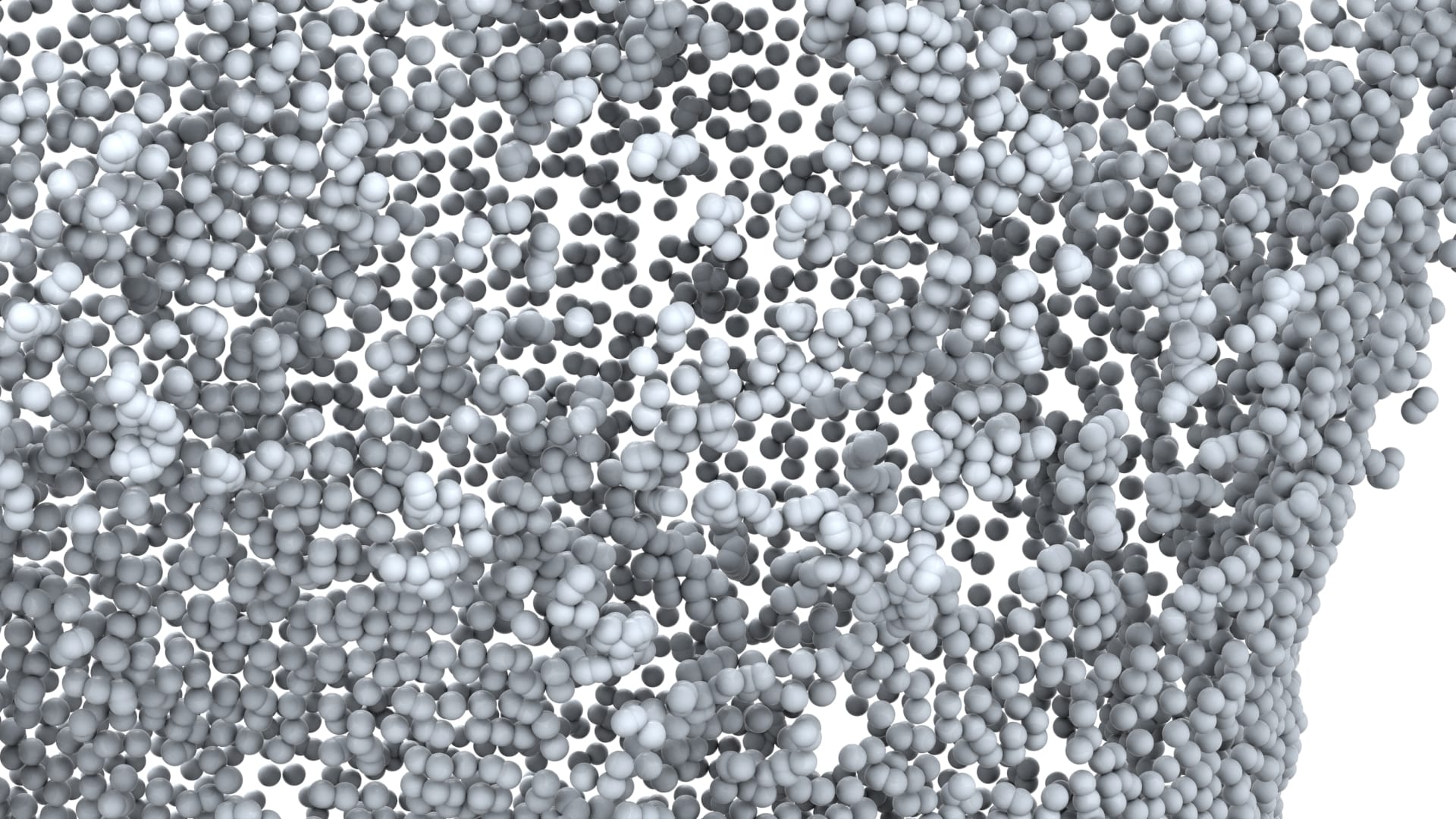 }
    \includegraphics[width=1.1in]{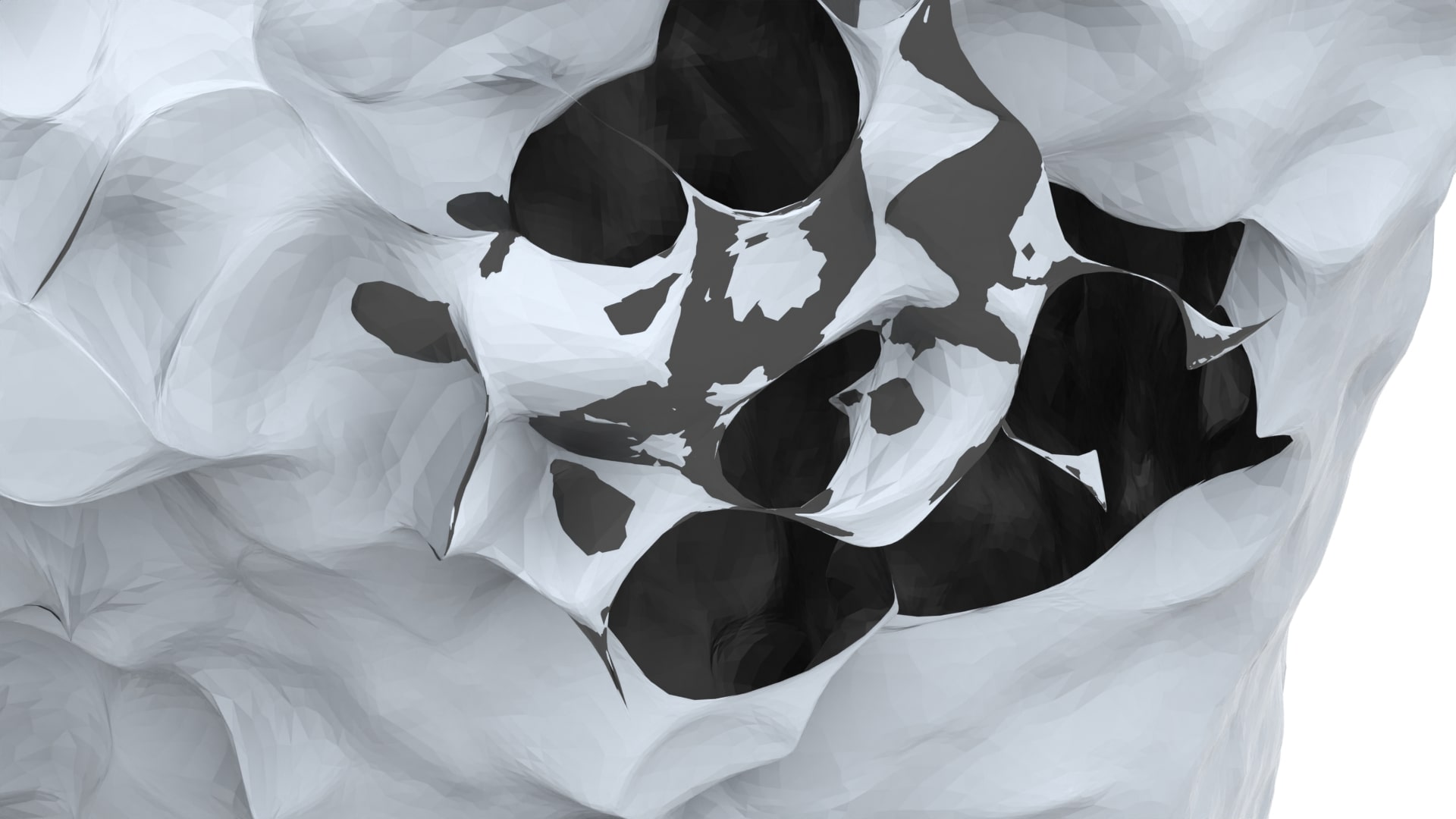 }
    \includegraphics[width=1.1in]{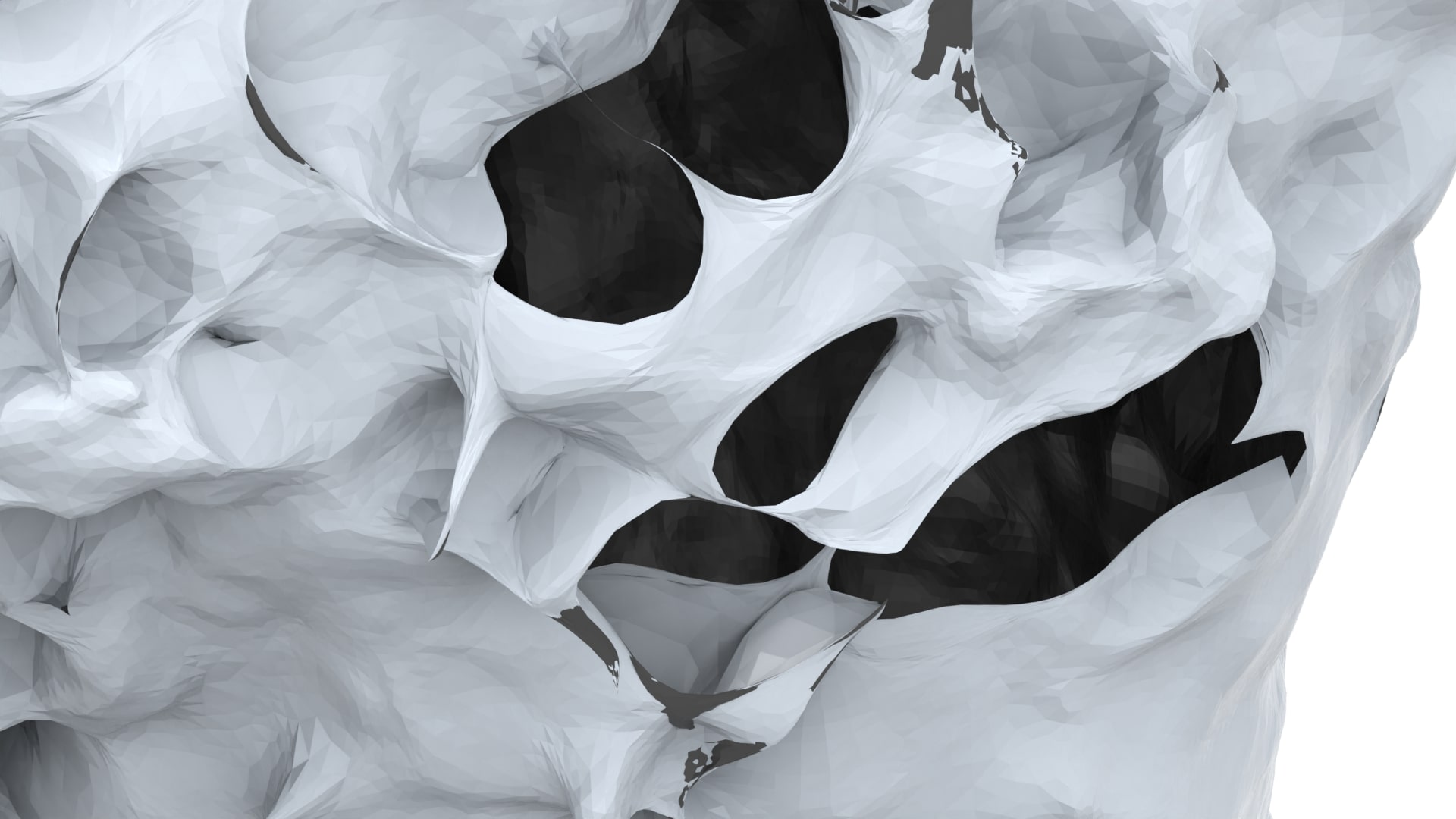 }\\
    \caption{A failed example. The input model has a high degree of noise, presenting a significant challenge for existing UDF learning methods. Due to the inaccuracies in the learned UDF, the resulting meshes are not only lacking in smoothness, but are also plagued with cutting artifacts. The dark regions are holes and the light gray regions are with two layers. The MC resolution is set to $256^3$.}
    \label{fig:bad}
\end{figure}

\section{Conclusion \& Future Work}
In this paper, we presented DoubleCoverUDF, a robust method for extracting the zero level-set from unsigned distance fields. Our approach involves computing the offset volume of the target surface and extracting its boundary mesh, which represents the dilated double covering of the surface. By solving an optimization problem, we project the double covering back to the original surface while preserving its topology and avoiding issues such as folding and self-intersection. Additionally, we introduced a post-processing step to segment the double-layered mesh into a single layer for orientable and manifold surfaces. By conducting various experiments on synthetic models and benchmark datasets, we have demonstrated DCUDF outperforms the state-of-the-art methods, such as UNDC, MeshUDF and MeshCAP, in terms of robustness and better quality of the extracted meshes. Additionally, our method demonstrates excellent memory efficiency, enabling the extraction of meshes with marching cubes resolutions up to $1024^3$ using a standard graphics card with 10GB of memory. As Figure~\ref{fig:dragonhead} shows, the extracted meshes exhibit high-quality triangulations, which can be directly used in downstream applications. 

An interesting direction for future work is to improve the runtime performance of DCUDF. As discussed in Section~\ref{sec:method} in the paper, the inclusion of centroids, which provide additional guidance for accurate projection in regions with high curvature, significantly enhances the quality of the covering map. However, this enhancement comes at the cost of increased computational complexity, as we use a centroid for each triangle. One possible approach to address this is to selectively employ centroids only in highly curved regions and at a later stage in the optimization process, focusing computational resources on areas that require finer details. Additionally, exploring an adaptive marching cubes approach, where higher resolutions are selectively applied to regions with intricate geometric features and lower resolutions are used for relatively flat areas, could potentially reduce the number of mesh vertices. This reduction in vertex count can contribute to improved runtime performance of DCUDF. 

We notice that existing UDF learning methods are sensitive to noise levels. As illustrated in Figure~\ref{fig:bad}, input models with high degree of noise result in imprecise UDFs. Consequently, the extracted meshes lack smoothness and contain various artifacts, e.g., holes and two layers mesh. There is a pressing need to develop robust and efficient UDF learning methods that can handle noisy input.



\begin{acks}
This project was partially supported by the National Natural Science Foundation of China Grants (61872347, 62072446), the Ministry of Education, Singapore, under its Academic Research Fund Grants (MOE-T2EP20220-0005, RG20/20 \& RT19/22), USA NSF (IIS-1715985, IIS-1812606, awarded to Hong QIN), and a gift fund from OPPO US Research Center.
\end{acks}

\bibliographystyle{ACM-Reference-Format}
\bibliography{reference}


\begin{thebibliography}{59}


\ifx \showCODEN    \undefined \def \showCODEN     #1{\unskip}     \fi
\ifx \showDOI      \undefined \def \showDOI       #1{#1}\fi
\ifx \showISBNx    \undefined \def \showISBNx     #1{\unskip}     \fi
\ifx \showISBNxiii \undefined \def \showISBNxiii  #1{\unskip}     \fi
\ifx \showISSN     \undefined \def \showISSN      #1{\unskip}     \fi
\ifx \showLCCN     \undefined \def \showLCCN      #1{\unskip}     \fi
\ifx \shownote     \undefined \def \shownote      #1{#1}          \fi
\ifx \showarticletitle \undefined \def \showarticletitle #1{#1}   \fi
\ifx \showURL      \undefined \def \showURL       {\relax}        \fi
\providecommand\bibfield[2]{#2}
\providecommand\bibinfo[2]{#2}
\providecommand\natexlab[1]{#1}
\providecommand\showeprint[2][]{arXiv:#2}

\bibitem[Alliez et~al\mbox{.}(2007)]%
        {Alliez2007}
\bibfield{author}{\bibinfo{person}{Pierre Alliez}, \bibinfo{person}{David Cohen-Steiner}, \bibinfo{person}{Yiying Tong}, {and} \bibinfo{person}{Mathieu Desbrun}.} \bibinfo{year}{2007}\natexlab{}.
\newblock \showarticletitle{Voronoi-Based Variational Reconstruction of Unoriented Point Sets}. In \bibinfo{booktitle}{\emph{Proc. of SGP}}. \bibinfo{pages}{39--48}.
\newblock


\bibitem[Amenta et~al\mbox{.}(2001)]%
        {Amenta2001}
\bibfield{author}{\bibinfo{person}{Nina Amenta}, \bibinfo{person}{Sunghee Choi}, {and} \bibinfo{person}{Ravi~Krishna Kolluri}.} \bibinfo{year}{2001}\natexlab{}.
\newblock \showarticletitle{The Power Crust}. In \bibinfo{booktitle}{\emph{Proc. of ACM SMA}}. \bibinfo{pages}{249--266}.
\newblock


\bibitem[Bernardini et~al\mbox{.}(1999)]%
        {Bernardini1999}
\bibfield{author}{\bibinfo{person}{Fausto Bernardini}, \bibinfo{person}{Joshua Mittleman}, \bibinfo{person}{Holly Rushmeier}, \bibinfo{person}{Cl\'{a}udio Silva}, {and} \bibinfo{person}{Gabriel Taubin}.} \bibinfo{year}{1999}\natexlab{}.
\newblock \showarticletitle{The Ball-Pivoting Algorithm for Surface Reconstruction}.
\newblock \bibinfo{journal}{\emph{IEEE Transactions on Visualization and Computer Graphics}} \bibinfo{volume}{5}, \bibinfo{number}{4} (\bibinfo{date}{oct} \bibinfo{year}{1999}), \bibinfo{pages}{349--359}.
\newblock


\bibitem[Boykov and Kolmogorov(2004)]%
        {Boykov2004}
\bibfield{author}{\bibinfo{person}{Yuri Boykov} {and} \bibinfo{person}{Vladimir Kolmogorov}.} \bibinfo{year}{2004}\natexlab{}.
\newblock \showarticletitle{An Experimental Comparison of Min-Cut/Max-Flow Algorithms for Energy Minimization in Vision}.
\newblock \bibinfo{journal}{\emph{IEEE Trans. Pattern Anal. Mach. Intell.}} \bibinfo{volume}{26}, \bibinfo{number}{9} (\bibinfo{year}{2004}), \bibinfo{pages}{1124--1137}.
\newblock


\bibitem[Chabra et~al\mbox{.}(2020)]%
        {Chabra2020}
\bibfield{author}{\bibinfo{person}{Rohan Chabra}, \bibinfo{person}{Jan~E. Lenssen}, \bibinfo{person}{Eddy Ilg}, \bibinfo{person}{Tanner Schmidt}, \bibinfo{person}{Julian Straub}, \bibinfo{person}{Steven Lovegrove}, {and} \bibinfo{person}{Richard Newcombe}.} \bibinfo{year}{2020}\natexlab{}.
\newblock \showarticletitle{Deep Local Shapes: Learning Local SDF Priors for Detailed 3D Reconstruction}. In \bibinfo{booktitle}{\emph{Proc. of ECCV}}. \bibinfo{pages}{608--625}.
\newblock


\bibitem[Chang et~al\mbox{.}(2015)]%
        {Chang2015}
\bibfield{author}{\bibinfo{person}{Angel~X. Chang}, \bibinfo{person}{Thomas Funkhouser}, \bibinfo{person}{Leonidas Guibas}, \bibinfo{person}{Pat Hanrahan}, \bibinfo{person}{Qixing Huang}, \bibinfo{person}{Zimo Li}, \bibinfo{person}{Silvio Savarese}, \bibinfo{person}{Manolis Savva}, \bibinfo{person}{Shuran Song}, \bibinfo{person}{Hao Su}, \bibinfo{person}{Jianxiong Xiao}, \bibinfo{person}{Li Yi}, {and} \bibinfo{person}{Fisher Yu}.} \bibinfo{year}{2015}\natexlab{}.
\newblock \bibinfo{title}{ShapeNet: An Information-Rich 3D Model Repository}.
\newblock
\newblock
\showeprint[arxiv]{1512.03012}~[cs.GR]


\bibitem[Chen et~al\mbox{.}(2022a)]%
        {Chen2022}
\bibfield{author}{\bibinfo{person}{Weikai Chen}, \bibinfo{person}{Cheng Lin}, \bibinfo{person}{Weiyang Li}, {and} \bibinfo{person}{Bo Yang}.} \bibinfo{year}{2022}\natexlab{a}.
\newblock \showarticletitle{3PSDF: Three-Pole Signed Distance Function for Learning Surfaces with Arbitrary Topologies}. In \bibinfo{booktitle}{\emph{Proc. of CVPR}}. \bibinfo{pages}{18501--18510}.
\newblock


\bibitem[Chen et~al\mbox{.}(2022b)]%
        {Chen2022NDC}
\bibfield{author}{\bibinfo{person}{Zhiqin Chen}, \bibinfo{person}{Andrea Tagliasacchi}, \bibinfo{person}{Thomas Funkhouser}, {and} \bibinfo{person}{Hao Zhang}.} \bibinfo{year}{2022}\natexlab{b}.
\newblock \showarticletitle{Neural Dual Contouring}.
\newblock \bibinfo{journal}{\emph{ACM Trans. Graph.}} \bibinfo{volume}{41}, \bibinfo{number}{4}, Article \bibinfo{articleno}{104} (\bibinfo{year}{2022}), \bibinfo{numpages}{13}~pages.
\newblock


\bibitem[Chen and Zhang(2021)]%
        {Chen2021NMC}
\bibfield{author}{\bibinfo{person}{Zhiqin Chen} {and} \bibinfo{person}{Hao Zhang}.} \bibinfo{year}{2021}\natexlab{}.
\newblock \showarticletitle{Neural Marching Cubes}.
\newblock \bibinfo{journal}{\emph{ACM Trans. Graph.}} \bibinfo{volume}{40}, \bibinfo{number}{6}, Article \bibinfo{articleno}{251} (\bibinfo{year}{2021}), \bibinfo{numpages}{15}~pages.
\newblock


\bibitem[Chibane et~al\mbox{.}(2020a)]%
        {Chibane2020IFNET}
\bibfield{author}{\bibinfo{person}{Julian Chibane}, \bibinfo{person}{Thiemo Alldieck}, {and} \bibinfo{person}{Gerard Pons-Moll}.} \bibinfo{year}{2020}\natexlab{a}.
\newblock \showarticletitle{Implicit Functions in Feature Space for 3D Shape Reconstruction and Completion}. In \bibinfo{booktitle}{\emph{Proc. of CVPR}}. \bibinfo{pages}{6968--6979}.
\newblock


\bibitem[Chibane et~al\mbox{.}(2020b)]%
        {Chibane2020NDF}
\bibfield{author}{\bibinfo{person}{Julian Chibane}, \bibinfo{person}{Aymen Mir}, {and} \bibinfo{person}{Gerard Pons-Moll}.} \bibinfo{year}{2020}\natexlab{b}.
\newblock \showarticletitle{Neural Unsigned Distance Fields for Implicit Function Learning}. In \bibinfo{booktitle}{\emph{Proc. of NeurIPS}}. Article \bibinfo{articleno}{1816}, \bibinfo{numpages}{15}~pages.
\newblock


\bibitem[Choi et~al\mbox{.}(2016)]%
        {Choi2016}
\bibfield{author}{\bibinfo{person}{Sungjoon Choi}, \bibinfo{person}{Qian-Yi Zhou}, \bibinfo{person}{Stephen Miller}, {and} \bibinfo{person}{Vladlen Koltun}.} \bibinfo{year}{2016}\natexlab{}.
\newblock \bibinfo{title}{A Large Dataset of Object Scans}.
\newblock
\newblock
\showeprint[arxiv]{1602.02481}~[cs.CV]


\bibitem[Corona et~al\mbox{.}(2021)]%
        {Corona2021}
\bibfield{author}{\bibinfo{person}{Enric Corona}, \bibinfo{person}{Albert Pumarola}, \bibinfo{person}{Guillem Aleny\`{a}}, \bibinfo{person}{Gerard Pons-Moll}, {and} \bibinfo{person}{Francesc Moreno-Noguer}.} \bibinfo{year}{2021}\natexlab{}.
\newblock \showarticletitle{SMPLicit: Topology-aware Generative Model for Clothed People}. In \bibinfo{booktitle}{\emph{Proc. of CVPR}}. \bibinfo{pages}{11870--11880}.
\newblock


\bibitem[Deng et~al\mbox{.}(2022)]%
        {Deng2022}
\bibfield{author}{\bibinfo{person}{Bailin Deng}, \bibinfo{person}{Yuxin Yao}, \bibinfo{person}{Roberto~M. Dyke}, {and} \bibinfo{person}{Juyong Zhang}.} \bibinfo{year}{2022}\natexlab{}.
\newblock \showarticletitle{A Survey of Non-Rigid 3D Registration}.
\newblock \bibinfo{journal}{\emph{Computer Graphics Forum}} \bibinfo{volume}{41}, \bibinfo{number}{2} (\bibinfo{year}{2022}), \bibinfo{pages}{559--589}.
\newblock


\bibitem[Dey and Goswami(2003)]%
        {Dey2003}
\bibfield{author}{\bibinfo{person}{Tamal~K. Dey} {and} \bibinfo{person}{Samrat Goswami}.} \bibinfo{year}{2003}\natexlab{}.
\newblock \showarticletitle{Tight Cocone: A Water-Tight Surface Reconstructor}. In \bibinfo{booktitle}{\emph{Proc. of ACM SMA}}. \bibinfo{pages}{127--134}.
\newblock


\bibitem[Gropp et~al\mbox{.}(2020)]%
        {Gropp2020}
\bibfield{author}{\bibinfo{person}{Amos Gropp}, \bibinfo{person}{Lior Yariv}, \bibinfo{person}{Niv Haim}, \bibinfo{person}{Matan Atzmon}, {and} \bibinfo{person}{Yaron Lipman}.} \bibinfo{year}{2020}\natexlab{}.
\newblock \showarticletitle{Implicit Geometric Regularization for Learning Shapes}. In \bibinfo{booktitle}{\emph{Proc. of ICML}}, Vol.~\bibinfo{volume}{119}. \bibinfo{pages}{3789--3799}.
\newblock


\bibitem[Gu and Yau(2003)]%
        {Gu2003}
\bibfield{author}{\bibinfo{person}{Xianfeng Gu} {and} \bibinfo{person}{Shing-Tung Yau}.} \bibinfo{year}{2003}\natexlab{}.
\newblock \showarticletitle{Global Conformal Surface Parameterization}. In \bibinfo{booktitle}{\emph{Proc. of SGP}}. \bibinfo{pages}{127--137}.
\newblock


\bibitem[Guillard et~al\mbox{.}(2022)]%
        {Guillard2022}
\bibfield{author}{\bibinfo{person}{Benoit Guillard}, \bibinfo{person}{Federico Stella}, {and} \bibinfo{person}{Pascal Fua}.} \bibinfo{year}{2022}\natexlab{}.
\newblock \showarticletitle{MeshUDF: Fast and Differentiable Meshing of Unsigned Distance Field Networks}. In \bibinfo{booktitle}{\emph{Proc. of ECCV}}. \bibinfo{pages}{576--592}.
\newblock


\bibitem[Hatcher(2002)]%
        {Hatcher2002}
\bibfield{author}{\bibinfo{person}{Allen Hatcher}.} \bibinfo{year}{2002}\natexlab{}.
\newblock \bibinfo{booktitle}{\emph{Algebraic Topology}}.
\newblock \bibinfo{publisher}{Cambridge University Press}.
\newblock


\bibitem[Hoppe et~al\mbox{.}(1992)]%
        {Hoppe1992}
\bibfield{author}{\bibinfo{person}{Hugues Hoppe}, \bibinfo{person}{Tony DeRose}, \bibinfo{person}{Tom Duchamp}, \bibinfo{person}{John McDonald}, {and} \bibinfo{person}{Werner Stuetzle}.} \bibinfo{year}{1992}\natexlab{}.
\newblock \showarticletitle{Surface Reconstruction from Unorganized Points}. In \bibinfo{booktitle}{\emph{Proc. of ACM SIGGRAPH}}. \bibinfo{pages}{71--78}.
\newblock


\bibitem[Hou et~al\mbox{.}(2022)]%
        {Hou2022}
\bibfield{author}{\bibinfo{person}{Fei Hou}, \bibinfo{person}{Chiyu Wang}, \bibinfo{person}{Wencheng Wang}, \bibinfo{person}{Hong Qin}, \bibinfo{person}{Chen Qian}, {and} \bibinfo{person}{Ying He}.} \bibinfo{year}{2022}\natexlab{}.
\newblock \showarticletitle{Iterative Poisson Surface Reconstruction (IPSR) for Unoriented Points}.
\newblock \bibinfo{journal}{\emph{ACM Trans. Graph.}} \bibinfo{volume}{41}, \bibinfo{number}{4}, Article \bibinfo{articleno}{128} (\bibinfo{year}{2022}), \bibinfo{numpages}{13}~pages.
\newblock


\bibitem[Huang et~al\mbox{.}(2009)]%
        {Huang2009}
\bibfield{author}{\bibinfo{person}{Hui Huang}, \bibinfo{person}{Dan Li}, \bibinfo{person}{Hao Zhang}, \bibinfo{person}{Uri Ascher}, {and} \bibinfo{person}{Daniel Cohen-Or}.} \bibinfo{year}{2009}\natexlab{}.
\newblock \showarticletitle{Consolidation of Unorganized Point Clouds for Surface Reconstruction}.
\newblock \bibinfo{journal}{\emph{ACM Trans. Graph.}} \bibinfo{volume}{28}, \bibinfo{number}{5} (\bibinfo{year}{2009}), \bibinfo{pages}{1--7}.
\newblock


\bibitem[Huang et~al\mbox{.}(2019)]%
        {Huang2019}
\bibfield{author}{\bibinfo{person}{Zhiyang Huang}, \bibinfo{person}{Nathan Carr}, {and} \bibinfo{person}{Tao Ju}.} \bibinfo{year}{2019}\natexlab{}.
\newblock \showarticletitle{Variational Implicit Point Set Surfaces}.
\newblock \bibinfo{journal}{\emph{ACM Trans. Graph.}} \bibinfo{volume}{38}, \bibinfo{number}{4}, Article \bibinfo{articleno}{124} (\bibinfo{year}{2019}), \bibinfo{numpages}{13}~pages.
\newblock


\bibitem[Jiang et~al\mbox{.}(2020)]%
        {Jiang2020}
\bibfield{author}{\bibinfo{person}{Chiyu Jiang}, \bibinfo{person}{Avneesh Sud}, \bibinfo{person}{Ameesh Makadia}, \bibinfo{person}{Jingwei Huang}, \bibinfo{person}{Matthias Nießner}, {and} \bibinfo{person}{Thomas Funkhouser}.} \bibinfo{year}{2020}\natexlab{}.
\newblock \showarticletitle{Local Implicit Grid Representations for 3D Scenes}. In \bibinfo{booktitle}{\emph{Proc. of CVPR}}. \bibinfo{pages}{6001--6010}.
\newblock


\bibitem[Ju et~al\mbox{.}(2002)]%
        {Ju2002}
\bibfield{author}{\bibinfo{person}{Tao Ju}, \bibinfo{person}{Frank Losasso}, \bibinfo{person}{Scott Schaefer}, {and} \bibinfo{person}{Joe Warren}.} \bibinfo{year}{2002}\natexlab{}.
\newblock \showarticletitle{Dual Contouring of Hermite Data}.
\newblock \bibinfo{journal}{\emph{ACM Trans. Graph.}} \bibinfo{volume}{21}, \bibinfo{number}{3} (\bibinfo{year}{2002}), \bibinfo{pages}{339--346}.
\newblock


\bibitem[K\"{a}lberer et~al\mbox{.}(2007)]%
        {Kalberer2007}
\bibfield{author}{\bibinfo{person}{Felix K\"{a}lberer}, \bibinfo{person}{Matthias Nieser}, {and} \bibinfo{person}{Konrad Polthier}.} \bibinfo{year}{2007}\natexlab{}.
\newblock \showarticletitle{QuadCover - Surface Parameterization using Branched Coverings}.
\newblock \bibinfo{journal}{\emph{CGF}} \bibinfo{volume}{26}, \bibinfo{number}{3} (\bibinfo{year}{2007}), \bibinfo{pages}{375--384}.
\newblock


\bibitem[Kazhdan et~al\mbox{.}(2006)]%
        {Kazhdan2006}
\bibfield{author}{\bibinfo{person}{Michael Kazhdan}, \bibinfo{person}{Matthew Bolitho}, {and} \bibinfo{person}{Hugues Hoppe}.} \bibinfo{year}{2006}\natexlab{}.
\newblock \showarticletitle{Poisson Surface Reconstruction}. In \bibinfo{booktitle}{\emph{Proc. of SGP}}. \bibinfo{pages}{61--70}.
\newblock


\bibitem[Kazhdan and Hoppe(2013)]%
        {Kazhdan2013}
\bibfield{author}{\bibinfo{person}{Michael Kazhdan} {and} \bibinfo{person}{Hugues Hoppe}.} \bibinfo{year}{2013}\natexlab{}.
\newblock \showarticletitle{Screened Poisson Surface Reconstruction}.
\newblock \bibinfo{journal}{\emph{ACM Trans. Graph.}} \bibinfo{volume}{32}, \bibinfo{number}{3}, Article \bibinfo{articleno}{29} (\bibinfo{year}{2013}), \bibinfo{numpages}{13}~pages.
\newblock


\bibitem[Kingma and Ba(2015)]%
        {Kingma2015}
\bibfield{author}{\bibinfo{person}{Diederik~P. Kingma} {and} \bibinfo{person}{Jimmy Ba}.} \bibinfo{year}{2015}\natexlab{}.
\newblock \showarticletitle{Adam: {A} Method for Stochastic Optimization}. In \bibinfo{booktitle}{\emph{Proc. of ICLR}}.
\newblock


\bibitem[Koo et~al\mbox{.}(2005)]%
        {Koo2005}
\bibfield{author}{\bibinfo{person}{Bon~Ki Koo}, \bibinfo{person}{Young~Kyu Choi}, \bibinfo{person}{Chang~Woo Chu}, \bibinfo{person}{Jae~Chul Kim}, {and} \bibinfo{person}{Byoung~Tae Choi}.} \bibinfo{year}{2005}\natexlab{}.
\newblock \showarticletitle{Shrink-Wrapped Boundary Face Algorithm for Mesh Reconstruction from Unorganized Points}.
\newblock \bibinfo{journal}{\emph{ETRI Journal}} \bibinfo{volume}{27}, \bibinfo{number}{2} (\bibinfo{year}{2005}), \bibinfo{pages}{235--238}.
\newblock


\bibitem[Laine and Karras(2010)]%
        {Laine2010}
\bibfield{author}{\bibinfo{person}{Samuli Laine} {and} \bibinfo{person}{Tero Karras}.} \bibinfo{year}{2010}\natexlab{}.
\newblock \showarticletitle{Efficient Sparse Voxel Octrees}. In \bibinfo{booktitle}{\emph{Proc. of ACM I3D}}. \bibinfo{pages}{55--63}.
\newblock


\bibitem[Lee(2009)]%
        {Lee2009}
\bibfield{author}{\bibinfo{person}{Sang~Hun Lee}.} \bibinfo{year}{2009}\natexlab{}.
\newblock \showarticletitle{Offsetting operations on non-manifold topological models}.
\newblock \bibinfo{journal}{\emph{Computer-Aided Design}} \bibinfo{volume}{41}, \bibinfo{number}{11} (\bibinfo{year}{2009}), \bibinfo{pages}{830--846}.
\newblock


\bibitem[Lin et~al\mbox{.}(2023)]%
        {Lin2023}
\bibfield{author}{\bibinfo{person}{Siyou Lin}, \bibinfo{person}{Dong Xiao}, \bibinfo{person}{Zuoqiang Shi}, {and} \bibinfo{person}{Bin Wang}.} \bibinfo{year}{2023}\natexlab{}.
\newblock \showarticletitle{Surface Reconstruction from Point Clouds without Normals by Parametrizing the Gauss Formula}.
\newblock \bibinfo{journal}{\emph{ACM Trans. Graph.}} \bibinfo{volume}{42}, \bibinfo{number}{2}, Article \bibinfo{articleno}{14} (\bibinfo{year}{2023}), \bibinfo{numpages}{19}~pages.
\newblock


\bibitem[Ling et~al\mbox{.}(2022)]%
        {Ling2022}
\bibfield{author}{\bibinfo{person}{Selena Ling}, \bibinfo{person}{Nicholas Sharp}, {and} \bibinfo{person}{Alec Jacobson}.} \bibinfo{year}{2022}\natexlab{}.
\newblock \showarticletitle{VectorAdam for Rotation Equivariant Geometry Optimization}. In \bibinfo{booktitle}{\emph{Proc. of NeurIPS}}. \bibinfo{pages}{4111--4122}.
\newblock


\bibitem[Liu et~al\mbox{.}(2023)]%
        {Liu2023}
\bibfield{author}{\bibinfo{person}{Yu-Tao Liu}, \bibinfo{person}{Li Wang}, \bibinfo{person}{Jie Yang}, \bibinfo{person}{Weikai Chen}, \bibinfo{person}{Xiaoxu Meng}, \bibinfo{person}{Bo Yang}, {and} \bibinfo{person}{Lin Gao}.} \bibinfo{year}{2023}\natexlab{}.
\newblock \showarticletitle{NeUDF: Learning Neural Unsigned Distance Fields with Volume Rendering}. In \bibinfo{booktitle}{\emph{Proc. of CVPR}}. \bibinfo{pages}{237--247}.
\newblock


\bibitem[Long et~al\mbox{.}(2023)]%
        {Long2023}
\bibfield{author}{\bibinfo{person}{Xiaoxiao Long}, \bibinfo{person}{Cheng Lin}, \bibinfo{person}{Lingjie Liu}, \bibinfo{person}{Yuan Liu}, \bibinfo{person}{Peng Wang}, \bibinfo{person}{Christian Theobalt}, \bibinfo{person}{Taku Komura}, {and} \bibinfo{person}{Wenping Wang}.} \bibinfo{year}{2023}\natexlab{}.
\newblock \showarticletitle{NeuralUDF: Learning Unsigned Distance Fields for Multi-view Reconstruction of Surfaces with Arbitrary Topologies}. In \bibinfo{booktitle}{\emph{Proc. of CVPR}}. \bibinfo{pages}{20834--20843}.
\newblock


\bibitem[Lorensen and Cline(1987)]%
        {Lorensen1987}
\bibfield{author}{\bibinfo{person}{William~E. Lorensen} {and} \bibinfo{person}{Harvey~E. Cline}.} \bibinfo{year}{1987}\natexlab{}.
\newblock \showarticletitle{Marching Cubes: A High Resolution 3D Surface Construction Algorithm}. In \bibinfo{booktitle}{\emph{Proc. of ACM SIGGRAPH}}. \bibinfo{pages}{163--169}.
\newblock


\bibitem[Lu et~al\mbox{.}(2018)]%
        {Lu2018}
\bibfield{author}{\bibinfo{person}{Wenjia Lu}, \bibinfo{person}{Zuoqiang Shi}, \bibinfo{person}{Jian Sun}, {and} \bibinfo{person}{Bin Wang}.} \bibinfo{year}{2018}\natexlab{}.
\newblock \showarticletitle{Surface Reconstruction Based on the Modified Gauss Formula}.
\newblock \bibinfo{journal}{\emph{ACM Trans. Graph.}} \bibinfo{volume}{38}, \bibinfo{number}{1}, Article \bibinfo{articleno}{2} (\bibinfo{year}{2018}), \bibinfo{numpages}{18}~pages.
\newblock


\bibitem[Meng et~al\mbox{.}(2023)]%
        {Meng2023}
\bibfield{author}{\bibinfo{person}{Xiaoxu Meng}, \bibinfo{person}{Weikai Chen}, {and} \bibinfo{person}{Bo Yang}.} \bibinfo{year}{2023}\natexlab{}.
\newblock \showarticletitle{NeAT: Learning Neural Implicit Surfaces with Arbitrary Topologies from Multi-view Images}. In \bibinfo{booktitle}{\emph{Proc. of CVPR}}. \bibinfo{pages}{248--258}.
\newblock


\bibitem[Metzer et~al\mbox{.}(2021)]%
        {Metzer2021}
\bibfield{author}{\bibinfo{person}{Gal Metzer}, \bibinfo{person}{Rana Hanocka}, \bibinfo{person}{Denis Zorin}, \bibinfo{person}{Raja Giryes}, \bibinfo{person}{Daniele Panozzo}, {and} \bibinfo{person}{Daniel Cohen-Or}.} \bibinfo{year}{2021}\natexlab{}.
\newblock \showarticletitle{Orienting Point Clouds with Dipole Propagation}.
\newblock \bibinfo{journal}{\emph{ACM Trans. Graph.}} \bibinfo{volume}{40}, \bibinfo{number}{4}, Article \bibinfo{articleno}{165} (\bibinfo{year}{2021}), \bibinfo{numpages}{14}~pages.
\newblock


\bibitem[Mildenhall et~al\mbox{.}(2020)]%
        {Mildenhall2020}
\bibfield{author}{\bibinfo{person}{Ben Mildenhall}, \bibinfo{person}{Pratul~P. Srinivasan}, \bibinfo{person}{Matthew Tancik}, \bibinfo{person}{Jonathan~T. Barron}, \bibinfo{person}{Ravi Ramamoorthi}, {and} \bibinfo{person}{Ren Ng}.} \bibinfo{year}{2020}\natexlab{}.
\newblock \showarticletitle{NeRF: Representing Scenes as Neural Radiance Fields for View Synthesis}. In \bibinfo{booktitle}{\emph{Proc. of ECCV}}. \bibinfo{pages}{405--421}.
\newblock


\bibitem[Nieser et~al\mbox{.}(2012)]%
        {Nieser2012}
\bibfield{author}{\bibinfo{person}{Matthias Nieser}, \bibinfo{person}{Jonathan Palacios}, \bibinfo{person}{Konrad Polthier}, {and} \bibinfo{person}{Eugene Zhang}.} \bibinfo{year}{2012}\natexlab{}.
\newblock \showarticletitle{Hexagonal Global Parameterization of Arbitrary Surfaces}.
\newblock \bibinfo{journal}{\emph{IEEE TVCG}} \bibinfo{volume}{18}, \bibinfo{number}{6} (\bibinfo{year}{2012}), \bibinfo{pages}{865--878}.
\newblock


\bibitem[Ohtake et~al\mbox{.}(2003)]%
        {Ohtake2003}
\bibfield{author}{\bibinfo{person}{Yutaka Ohtake}, \bibinfo{person}{Alexander Belyaev}, \bibinfo{person}{Marc Alexa}, \bibinfo{person}{Greg Turk}, {and} \bibinfo{person}{Hans-Peter Seidel}.} \bibinfo{year}{2003}\natexlab{}.
\newblock \showarticletitle{Multi-Level Partition of Unity Implicits}.
\newblock \bibinfo{journal}{\emph{ACM Trans. Graph.}} \bibinfo{volume}{22}, \bibinfo{number}{3} (\bibinfo{year}{2003}), \bibinfo{pages}{463--470}.
\newblock


\bibitem[Palmer et~al\mbox{.}(2022)]%
        {Palmer2022}
\bibfield{author}{\bibinfo{person}{David Palmer}, \bibinfo{person}{Dmitriy Smirnov}, \bibinfo{person}{Stephanie Wang}, \bibinfo{person}{Albert Chern}, {and} \bibinfo{person}{Justin Solomon}.} \bibinfo{year}{2022}\natexlab{}.
\newblock \showarticletitle{DeepCurrents: Learning Implicit Representations of Shapes With Boundaries}. In \bibinfo{booktitle}{\emph{Proc. of CVPR}}. \bibinfo{pages}{18665--18675}.
\newblock


\bibitem[Park et~al\mbox{.}(2019)]%
        {Park2019}
\bibfield{author}{\bibinfo{person}{Jeong~Joon Park}, \bibinfo{person}{Peter Florence}, \bibinfo{person}{Julian Straub}, \bibinfo{person}{Richard Newcombe}, {and} \bibinfo{person}{Steven Lovegrove}.} \bibinfo{year}{2019}\natexlab{}.
\newblock \showarticletitle{DeepSDF: Learning Continuous Signed Distance Functions for Shape Representation}. In \bibinfo{booktitle}{\emph{Proc. of CVPR}}. \bibinfo{pages}{165--174}.
\newblock


\bibitem[Peng et~al\mbox{.}(2021)]%
        {Peng2021}
\bibfield{author}{\bibinfo{person}{Songyou Peng}, \bibinfo{person}{Chiyu~"Max" Jiang}, \bibinfo{person}{Yiyi Liao}, \bibinfo{person}{Michael Niemeyer}, \bibinfo{person}{Marc Pollefeys}, {and} \bibinfo{person}{Andreas Geiger}.} \bibinfo{year}{2021}\natexlab{}.
\newblock \showarticletitle{Shape As Points: A Differentiable Poisson Solver}. In \bibinfo{booktitle}{\emph{Proc. of NeurIPS}}. \bibinfo{pages}{13032--13044}.
\newblock


\bibitem[Ren et~al\mbox{.}(2023)]%
        {Ren2023}
\bibfield{author}{\bibinfo{person}{Siyu Ren}, \bibinfo{person}{Junhui Hou}, \bibinfo{person}{Xiaodong Chen}, \bibinfo{person}{Ying He}, {and} \bibinfo{person}{Wenping Wang}.} \bibinfo{year}{2023}\natexlab{}.
\newblock \showarticletitle{GeoUDF: Surface Reconstruction from 3D Point Clouds via Geometry-guided Distance Representation}. In \bibinfo{booktitle}{\emph{Proc. of ICCV}}. \bibinfo{pages}{14214--14224}.
\newblock


\bibitem[Schmidt et~al\mbox{.}(2019)]%
        {Schmidt2019}
\bibfield{author}{\bibinfo{person}{Patrick Schmidt}, \bibinfo{person}{Janis Born}, \bibinfo{person}{Marcel Campen}, {and} \bibinfo{person}{Leif Kobbelt}.} \bibinfo{year}{2019}\natexlab{}.
\newblock \showarticletitle{Distortion-Minimizing Injective Maps between Surfaces}.
\newblock \bibinfo{journal}{\emph{ACM Trans. Graph.}} \bibinfo{volume}{38}, \bibinfo{number}{6}, Article \bibinfo{articleno}{156} (\bibinfo{year}{2019}), \bibinfo{numpages}{15}~pages.
\newblock


\bibitem[Sharp and Crane(2020)]%
        {Sharp2020}
\bibfield{author}{\bibinfo{person}{Nicholas Sharp} {and} \bibinfo{person}{Keenan Crane}.} \bibinfo{year}{2020}\natexlab{}.
\newblock \showarticletitle{A Laplacian for Nonmanifold Triangle Meshes}.
\newblock \bibinfo{journal}{\emph{Computer Graphics Forum}} \bibinfo{volume}{39}, \bibinfo{number}{5} (\bibinfo{year}{2020}), \bibinfo{pages}{69--80}.
\newblock


\bibitem[Sitzmann et~al\mbox{.}(2020)]%
        {Sitzmann2020}
\bibfield{author}{\bibinfo{person}{Vincent Sitzmann}, \bibinfo{person}{Julien N.~P. Martel}, \bibinfo{person}{Alexander~W. Bergman}, \bibinfo{person}{David~B. Lindell}, {and} \bibinfo{person}{Gordon Wetzstein}.} \bibinfo{year}{2020}\natexlab{}.
\newblock \showarticletitle{Implicit Neural Representations with Periodic Activation Functions}. In \bibinfo{booktitle}{\emph{Proc. of NeurIPS}}. \bibinfo{pages}{7462--7473}.
\newblock


\bibitem[Venkatesh et~al\mbox{.}(2021)]%
        {Venkatesh2021}
\bibfield{author}{\bibinfo{person}{Rahul Venkatesh}, \bibinfo{person}{Tejan Karmali}, \bibinfo{person}{Sarthak Sharma}, \bibinfo{person}{Aurobrata Ghosh}, \bibinfo{person}{R.~Venkatesh Babu}, \bibinfo{person}{L\'{a}szl\'{o}~A. Jeni}, {and} \bibinfo{person}{Maneesh Singh}.} \bibinfo{year}{2021}\natexlab{}.
\newblock \showarticletitle{Deep Implicit Surface Point Prediction Networks}. In \bibinfo{booktitle}{\emph{Proc. of ICCV}}. \bibinfo{pages}{12633--12642}.
\newblock


\bibitem[Wang et~al\mbox{.}(2005)]%
        {Wang2005}
\bibfield{author}{\bibinfo{person}{Jianning Wang}, \bibinfo{person}{Manuel~M. Oliveira}, {and} \bibinfo{person}{Arie~E. Kaufman}.} \bibinfo{year}{2005}\natexlab{}.
\newblock \showarticletitle{Reconstructing manifold and non-manifold surfaces from point clouds}. In \bibinfo{booktitle}{\emph{Proc. of IEEE Visualization}}. \bibinfo{pages}{415--422}.
\newblock


\bibitem[Wang et~al\mbox{.}(2022b)]%
        {Wang2022HSDF}
\bibfield{author}{\bibinfo{person}{Li Wang}, \bibinfo{person}{Jie Yang}, \bibinfo{person}{Weikai Chen}, \bibinfo{person}{Xiaoxu Meng}, \bibinfo{person}{Bo Yang}, \bibinfo{person}{Jintao Li}, {and} \bibinfo{person}{Lin Gao}.} \bibinfo{year}{2022}\natexlab{b}.
\newblock \showarticletitle{{HSDF}: Hybrid Sign and Distance Field for Modeling Surfaces with Arbitrary Topologies}. In \bibinfo{booktitle}{\emph{Proc. of NeurIPS}}. \bibinfo{pages}{32172--32185}.
\newblock


\bibitem[Wang et~al\mbox{.}(2022a)]%
        {Wang2022IDF}
\bibfield{author}{\bibinfo{person}{Yifan Wang}, \bibinfo{person}{Lukas Rahmann}, {and} \bibinfo{person}{Olga Sorkine{-}Hornung}.} \bibinfo{year}{2022}\natexlab{a}.
\newblock \showarticletitle{Geometry-Consistent Neural Shape Representation with Implicit Displacement Fields}. In \bibinfo{booktitle}{\emph{Proc. of ICLR}}.
\newblock


\bibitem[Xu et~al\mbox{.}(2023)]%
        {Xu2023}
\bibfield{author}{\bibinfo{person}{Rui Xu}, \bibinfo{person}{Zhiyang Dou}, \bibinfo{person}{Ningna Wang}, \bibinfo{person}{Shiqing Xin}, \bibinfo{person}{Shuangmin Chen}, \bibinfo{person}{Mingyan Jiang}, \bibinfo{person}{Xiaohu Guo}, \bibinfo{person}{Wenping Wang}, {and} \bibinfo{person}{Changhe Tu}.} \bibinfo{year}{2023}\natexlab{}.
\newblock \showarticletitle{Globally Consistent Normal Orientation for Point Clouds by Regularizing the Winding-Number Field}.
\newblock \bibinfo{journal}{\emph{ACM Trans. Graph.}} \bibinfo{volume}{42}, \bibinfo{number}{4}, Article \bibinfo{articleno}{111} (\bibinfo{year}{2023}), \bibinfo{numpages}{15}~pages.
\newblock


\bibitem[Ye et~al\mbox{.}(2022)]%
        {Ye2022}
\bibfield{author}{\bibinfo{person}{Jianglong Ye}, \bibinfo{person}{Yuntao Chen}, \bibinfo{person}{Naiyan Wang}, {and} \bibinfo{person}{Xiaolong Wang}.} \bibinfo{year}{2022}\natexlab{}.
\newblock \showarticletitle{GIFS: Neural Implicit Function for General Shape Representation}. In \bibinfo{booktitle}{\emph{Proc. of CVPR}}. \bibinfo{pages}{12819--12829}.
\newblock


\bibitem[Zhao et~al\mbox{.}(2021)]%
        {Zhao2021}
\bibfield{author}{\bibinfo{person}{Fang Zhao}, \bibinfo{person}{Wenhao Wang}, \bibinfo{person}{Shengcai Liao}, {and} \bibinfo{person}{Ling Shao}.} \bibinfo{year}{2021}\natexlab{}.
\newblock \showarticletitle{Learning Anchored Unsigned Distance Functions with Gradient Direction Alignment for Single-view Garment Reconstruction}. In \bibinfo{booktitle}{\emph{Proc. of ICCV}}. \bibinfo{pages}{12654--12663}.
\newblock


\bibitem[Zhou et~al\mbox{.}(2022)]%
        {Zhou2022}
\bibfield{author}{\bibinfo{person}{Junsheng Zhou}, \bibinfo{person}{Baorui Ma}, \bibinfo{person}{Yu-Shen Liu}, \bibinfo{person}{Yi Fang}, {and} \bibinfo{person}{Zhizhong Han}.} \bibinfo{year}{2022}\natexlab{}.
\newblock \showarticletitle{Learning Consistency-Aware Unsigned Distance Functions Progressively from Raw Point Clouds}. In \bibinfo{booktitle}{\emph{Proc. of NeurIPS}}. \bibinfo{pages}{16481--16494}.
\newblock


\bibitem[Zhu et~al\mbox{.}(2020)]%
        {Zhu2020}
\bibfield{author}{\bibinfo{person}{Heming Zhu}, \bibinfo{person}{Yu Cao}, \bibinfo{person}{Hang Jin}, \bibinfo{person}{Weikai Chen}, \bibinfo{person}{Dong Du}, \bibinfo{person}{Zhangye Wang}, \bibinfo{person}{Shuguang Cui}, {and} \bibinfo{person}{Xiaoguang Han}.} \bibinfo{year}{2020}\natexlab{}.
\newblock \showarticletitle{Deep Fashion3D: A Dataset and Benchmark for 3D Garment Reconstruction from Single Images}. In \bibinfo{booktitle}{\emph{Proc. of ECCV}}. \bibinfo{pages}{512--530}.
\newblock


\end{thebibliography}

\end{document}